%% file: arxiv.tex
\documentclass{article}

\def\hidedraftcomments{}
\usepackage{iclr2027/iclr2027_conference,times}

\input{iclr2027/math_commands.tex}

\usepackage{graphicx}
\usepackage{subcaption}
\usepackage{amsmath,amssymb,amsfonts}
\usepackage{caption}
\usepackage{float}
\usepackage{wrapfig}
\makeatletter
\newcommand{\Wrapfill}{\par
    \ifnum\c@WF@wrappedlines>1
        \vspace{\dimexpr\c@WF@wrappedlines\baselineskip-\baselineskip\relax}%
        \global\c@WF@wrappedlines=0
        \WF@finale
    \fi}
\makeatother
\usepackage{booktabs}
\usepackage{array}
\usepackage{colortbl}
\usepackage[hang,flushmargin]{footmisc}
\usepackage{tcolorbox}
\usepackage{multirow}
\usepackage{tabularx}
\usepackage{tablefootnote}
\usepackage{amsthm}
\usepackage[bookmarks=true,colorlinks=true,citecolor=blue,linkcolor=blue,urlcolor=blue]{hyperref}
\usepackage{url}
\usepackage{xparse}
\usepackage{xspace}
\usepackage{pgfplots}
\usepackage[outline]{contour}
\usepackage{enumitem}
\usepgfplotslibrary{groupplots,fillbetween}
\pgfplotsset{compat=1.18}
\contourlength{0.12pt}
\DeclareMathSizes{7.2}{7.2}{5}{5}
\DeclareMathSizes{6.5}{6.5}{5}{5}
\DeclareMathSizes{6}{6}{5}{5}

\newtheorem{proposition}{Proposition}[section]
\newtheorem{corollary}[proposition]{Corollary}
\newtheorem{lemma}[proposition]{Lemma}
\newtheorem{definition}[proposition]{Definition}

\newtcolorbox[auto counter]{informalresult}[1]{%
    title={Result~\thetcbcounter: (informal; #1)},
    colback=black!2,
    colframe=black!20,
    colbacktitle=black!12,
    coltitle=black,
    fonttitle=\bfseries,
    boxrule=0.5pt,
    arc=2pt,
    outer arc=2pt,
    left=6pt,
    right=6pt,
    top=5pt,
    bottom=5pt,
    toptitle=3pt,
    bottomtitle=3pt,
    before skip=8pt,
    after skip=8pt
}

\newcommand{\methodname}{\ifmmode\text{ProAct}\else ProAct\fi\xspace}
\newcommand{\methodfullname}{Predictable and Robust Action Tokenization (\methodname)\xspace}
\newcommand{\wotp}{\ensuremath{\text{w/o\ TP}}\xspace}
\newcommand{\wotbr}{\ensuremath{\text{w/o\ TBR}}\xspace}
\newcommand{\ent}{\mathbb H} 

\graphicspath{{figures/}}

\definecolor{vqbase}{RGB}{230,85,13}
\definecolor{rvqbase}{RGB}{49,130,189}
\definecolor{nrvqbase}{RGB}{0,137,123}
\definecolor{fsqbase}{RGB}{49,163,84}
\definecolor{qincobase}{RGB}{117,107,177}
\newcommand{\tokenizerid}[2]{%
    \mbox{\textcolor{#1}{\rule{0.72em}{0.72em}}\hspace{0.35em}\texttt{#2}}%
}

\newcommand{\tablebodyfont}{\scriptsize}
\newcommand{\reconbetter}[1]{%
    #1\hspace{0.10em}%
    \textcolor{green!45!black}{%
        \raisebox{0.08ex}[0pt][0pt]{\ensuremath{\scriptstyle\downarrow}}%
    }%
}

\newif\ifshowcomments
\ifdefined\hidedraftcomments
    \showcommentsfalse
\else
    \showcommentstrue
\fi
\ifshowcomments
    \usepackage[textsize=scriptsize,textwidth=3.25cm]{todonotes}
\else
    \usepackage[disable,textsize=scriptsize]{todonotes}
\fi

\let\todonotesTodo\todo
\RenewDocumentCommand{\todo}{s O{} m}{%
    \IfBooleanTF{#1}
        {\todonotesTodo[inline,#2]{#3}}
        {\todonotesTodo[#2]{#3}}%
}

\NewDocumentCommand{\todohrc}{s O{} m}{%
    \IfBooleanTF{#1}
        {\todo*[color=yellow!15,#2]{\textbf{HRC:} #3}}
        {\todo[color=yellow!15,#2]{\textbf{HRC:} #3}}%
}
\NewDocumentCommand{\todojjt}{s O{} m}{%
    \IfBooleanTF{#1}
        {\todo*[color=blue!15,#2]{\textbf{JJT:} #3}}
        {\todo[color=blue!15,#2]{\textbf{JJT:} #3}}%
}
\NewDocumentCommand{\todomw}{s O{} m}{%
    \IfBooleanTF{#1}
        {\todo*[color=green!15,#2]{\textbf{MW} #3}}
        {\todo[color=green!15,#2]{\textbf{MW:} #3}}%
}
\NewDocumentCommand{\todokc}{s O{} m}{%
    \IfBooleanTF{#1}
        {\todo*[color=purple!15,#2]{\textbf{KC} #3}}
        {\todo[color=purple!15,#2]{\textbf{KC:} #3}}%
}

\title{Beyond Reconstruction: What Matters in\\Action Tokenization for Robot Policies?}

\author{
Haoran Chen$^{1,*}$ \quad Jingtian Ji$^{1,*}$ \quad Samuel Wheeler$^{2}$ \\
{\bfseries Kaylene Caswell Stocking$^{1}$ \quad Matthew Walter$^{1}$} \\
{\normalfont $^{1}$Toyota Technological Institute at Chicago} \\
{\normalfont $^{2}$Argonne National Laboratory} \\
{\normalfont\small $^{*}$Equal contribution.}
}
\iclrfinalcopy
\hypersetup{
    pdftitle={Beyond Reconstruction: What Matters in Action Tokenization for Robot Policies?},
    pdfauthor={Haoran Chen, Jingtian Ji, Samuel Wheeler, Kaylene Caswell Stocking, Matthew Walter}
}

\begin{document}

\maketitle
\lhead{Preprint}

\begin{abstract}
Autoregressive action-token policies such as vision-language-action models require \emph{action tokenizers} to translate discrete token sequences into precise control actions in continuous space. Many action tokenizers learn the mapping between tokens and actions via a reconstruction objective. However, as we show through extensive analysis, sufficiently accurate action reconstruction is only one part of what makes a downstream robot policy successful. It is also critical that the policy is \emph{able to predict the right tokens} for new observations, and that \emph{unseen policy token predictions still decode into reasonable actions}. These properties are downstream of tokenizer training and are not directly incentivized by a reconstruction objective alone. In this work, we introduce \textbf{Predictable and Robust Action Tokenization (ProAct)}, a tokenizer training method that strategically augments reconstruction with the goal of improving downstream predictability and robustness. ProAct is policy-agnostic and uses only action datasets for training. Across the Robomimic, LIBERO, and RoboTwin benchmarks and a diverse set of tokenizer architectures, ProAct improves rollout success by an average of 11.3 percentage points. These improvements also translate to vision-language-action policies and real-world robotic manipulation, yielding average gains of 21.8 and 36.7 percentage points, respectively. These results suggest that effective action tokenization should be designed as a policy interface—balancing fidelity, predictability, and robustness—rather than as a reconstruction problem alone.
\end{abstract}

\section{Introduction}
\label{sec:introduction}
Many robot policies predict continuous action chunks~\citep{Zhao2023ACT,Chi2023DiffusionPolicy},
whereas autoregressive vision-language models (VLMs) generate discrete token sequences.
Action tokenization bridges this gap by encoding action chunks as tokens and decoding them back to continuous control actions.
RT-1~\citep{Brohan2023RT1} discretizes robot actions, while RT-2~\citep{Zitkovich2023RT2} explicitly connects action tokens to pretrained VLMs and introduces the vision-language-action (VLA) formulation.
Action tokenization has since become standard in autoregressive VLAs~\citep{Kim2024OpenVLA,Pertsch2025FAST,Wang2025VQVLA}, in part because it preserves the native next-token interface of autoregressive VLMs.
Discrete action tokens can represent multiple valid action modes, whereas regression objectives average across modes and yield an invalid intermediate action~\citep{Florence2022ImplicitBC,Shafiullah2022BehaviorTransformer,Lee2024VQBeT}.
Moreover, action tokenization enables temporal compression~\citep{Mete2024QueST,Pertsch2025FAST} and language-model-style policy post-training~\citep{Galaxea2026G05}.

This growing adoption of action tokenization raises a more specific question: how should continuous actions be tokenized?
Independent per-dimension, per-timestep binning~\citep{Brohan2023RT1} ignores temporal redundancy within action chunks and is inefficient (Appendix~\ref{app:analytical-tokenization});
FAST~\citep{Pertsch2025FAST} exploits such temporal structure through frequency-domain compression, but relies on a fixed analytical representation~\citep{Dong2026ActionCodec} and is tied to dataset-specific frequency statistics, making it less transferable across tasks or embodiments.
Recent learned tokenizers report gains over these fixed tokenization schemes on selected benchmarks~\citep{Dong2026ActionCodec,Kang2026XTokenizer,Liu2026OAT}; however, the principles for designing effective learned action tokenizers remain underexplored.
Existing works often choose nominal bitrate, chunk size, and execution horizon empirically, while the properties of the resulting token space remain poorly understood.
More concretely, tokenizer training and policy learning involve different aspects of the same discrete token space: the tokenizer focuses on reconstructing expert actions from encoded tokens, while the downstream autoregressive (AR) policy must predict and compose those tokens from observations. As a result, reconstruction quality alone does not characterize either how predictable the induced token targets are or how policy-generated token sequences decode into actions.
These gaps expose a fundamental question: \textbf{``what matters in action tokenization for robot policies?''}


Across 22 tokenizer configurations that span key design choices~\citep{Lee2024VQBeT,Mete2024QueST,Wang2025VQVLA,Liu2025FASTer,Dong2026ActionCodec,Liu2026OAT,Kang2026XTokenizer},
we find that tokenizer reconstruction error alone is not a reliable indicator of policy success. Specifically, a sufficiently low reconstruction error is necessary for successful control, but further reductions in error do not predict rollout performance.
We further analyze how bitrate, chunk size, and the execution horizon affect downstream performance.
Beyond reconstruction, our empirical and theoretical results identify an underexplored failure mode: autoregressive policies can produce token sequences that differ from the encoded targets, yet standard training objectives do not directly control the resulting action-space error. This observation motivates \textbf{\methodfullname}, a tokenizer-training method with two components. \emph{Token Balance Regularization} makes the encoded targets easier for the downstream policy to predict. \emph{Token Perturbation} improves robustness to token mismatches. Neither requires access to the downstream policy or the observations. Together, we show that \methodname{} improves downstream rollout performance across nearly all evaluated tokenizer configurations.
Ablating each component shows that both interventions can improve rollout performance even when reconstruction and policy-validation metrics do not consistently explain the gains.
Decoded-error diagnostics support the intended robustness effect of Token Perturbation. Together, these results suggest that effective action tokenization requires balancing fidelity, predictability,
and robustness to the sequences that the downstream policy actually produces.

Our contributions are threefold:
\begin{itemize}[nosep,leftmargin=2em]
    \item We conduct a controlled study of 22 learned action tokenizers, characterizing the effects of reconstruction fidelity, nominal bitrate, chunk size, and execution horizon on downstream control. We find that sufficient reconstruction accuracy is necessary for successful rollouts, but that further improvements in reconstruction do not reliably predict better downstream performance.
    \item We show that AR policies routinely compose complete token sequences not observed in the encoded demonstrations, and develop a one-step error decomposition that distinguishes matched predictions from token mismatches, revealing that standard tokenizer and policy objectives do not control the action-space severity of mismatched predictions.
    \item Motivated by this analysis, we propose \methodname, which combines two complementary tokenizer-only training methods: Token Perturbation, which improves robustness to token mismatches, and Token Balance Regularization, which makes the encoded targets easier for the downstream policy to predict. We show that \methodname{} improves average success rate by up to $11.3$ percentage points on Robomimic, LIBERO, and RoboTwin, by $21.8$ points for a VLA policy on LIBERO, and by $36.7$ points on real-world manipulation tasks. 
\end{itemize}

\section{Related Work} \label{sec:related-work}
\paragraph{Imitation Learning and Action Chunking}

Imitation learning trains policies from expert demonstrations, commonly by behavior cloning~\citep{Ross2011DAgger}.
Action chunking extends this formulation by predicting a sequence of future actions and executing either the full chunk or a prefix before re-planning.
It is used across transformer, diffusion, and VLA policies, including ACT~\citep{Zhao2023ACT}, Diffusion Policy~\citep{Chi2023DiffusionPolicy}, and OpenVLA-OFT~\citep{Kim2025OpenVLAOFT}.
Alongside these empirical design choices, recent theory analyzes when open-loop chunk execution can mitigate compounding imitation error under stability assumptions~\citep{Zhang2025ActionChunking,Lazzati2026ActionChunking}.

\paragraph{Action-Generation Paradigms in VLA Policies}
A primary distinction among VLA policies is their output representation.
One family predicts continuous action chunks directly through an action head: $\pi_0$~\citep{Black2025Pi0}, $\pi_{0.5}$~\citep{PhysicalIntelligence2025Pi05}, and GR00T N1~\citep{NVIDIA2025GR00TN1} use flow-matching or diffusion-based objectives,
whereas OpenVLA-OFT uses an $L_1$ regression objective~\citep{Kim2025OpenVLAOFT}.
These methods map VLM features directly to continuous actions without an intervening vocabulary.
A second family predicts discrete action tokens that must be detokenized before execution; recent examples include VQ-VLA~\citep{Wang2025VQVLA}, FASTer~\citep{Liu2025FASTer}, and G0.5~\citep{Galaxea2026G05}.
This token-based formulation allows action prediction to retain the VLM's native autoregressive next-token interface.
The increased adoption of action tokenization for robot policies raises the question that we aim to address in this work: how to best train the action tokenizer that determines the VLM prediction targets.

\paragraph{Discrete Representation Learning and Tokenization}

Discrete representation learning maps continuous features to tokens using vector quantization (VQ)~\citep{Oord2017VQVAE}, residual VQ (RVQ)~\citep{Lee2022RQ}, normalized RVQ (NRVQ)~\citep{Yu2022ImprovedVQGAN,kumar2023high}, finite scalar quantization (FSQ)~\citep{Mentzer2024FSQ}, or reconstruction-conditioned codebooks such as QinCo~\citep{Huijben2024QINCo,Vallaeys2025QINCo2}.
These methods are used by tokenizers for vision~\citep{Oord2017VQVAE,Yu2022ImprovedVQGAN,Qiu2025RobustLatent} and audio~\citep{Zeghidour2022SoundStream}.

For robot control, tokenization defines the prediction targets for the downstream policy. Analytical representations include per-dimension binning~\citep{Brohan2023RT1}, FAST frequency-domain codes~\citep{Pertsch2025FAST}, and BEAST B-spline coefficients~\citep{Zhou2025BEAST}.
Learned methods include VQ-BeT~\citep{Lee2024VQBeT} and QueST~\citep{Mete2024QueST} for discrete skills, OAT for prefix decodability~\citep{Liu2026OAT}, and VQ-VLA~\citep{Wang2025VQVLA} and FASTer~\citep{Liu2025FASTer} for autoregressive VLAs.
Recent methods further incorporate vocabulary structure~\citep{Dong2026ActionCodec}, robot state~\citep{Jiang2026SAVLA}, or multimodal features~\citep{Kang2026XTokenizer}.
While such auxiliary signals can improve alignment on source tasks, they tie the tokenizer to task- or observation-specific supervision and may require realignment under distribution changes.
To avoid this, we instead train tokenizers on action chunks alone, treating auxiliary supervision as complementary.
Unlike prior systems that tune reconstruction, bitrate, and temporal horizons individually, we jointly analyze these factors and policy-generated sequences across a controlled tokenizer design space.
\section{Preliminaries}
\label{sec:preliminaries}
Action-token imitation learning first trains and freezes a tokenizer on action chunks, then trains a policy to predict the tokens that correspond to demonstration trajectory actions.
For an $H$-step action chunk $A\in\mathbb R^{H\times d_a}$ with action dimension $d_a$, a learned tokenizer consists of an encoder $E$, a quantizer $Q=(q,e)$, and a decoder $D$:
\begin{equation}
    \overbrace{
      A
      \xrightarrow{\qquad E\qquad}
      z_e
      \xrightarrow{\qquad q\qquad}
      C
    }^{\text{tokenization: }C=\mathcal T(A):=q(E(A))}
    \overbrace{
      \xrightarrow{\qquad e\qquad}
      z_q
      \xrightarrow{\qquad D\qquad}
      \widehat A
    }^{\text{detokenization: }\widehat A=\mathcal D(C):=D(e(C))}.
    \label{eq:tokenizer}
\end{equation}
Here, $q$ maps encoder embeddings to the token sequence $C$, while $e$ maps $C$ back to quantized embeddings. Across 22 configurations, we pair transformer (TF)~\citep{Liu2026OAT}, Conv2D (C2)~\citep{Wang2025VQVLA}, and convolution--transformer (CT) backbones~\citep{Liu2025FASTer} with VQ~\citep{Oord2017VQVAE}, RVQ~\citep{Lee2022RQ}, NRVQ~\citep{Yu2022ImprovedVQGAN,kumar2023high}, FSQ~\citep{Mentzer2024FSQ}, and QinCo~\citep{Huijben2024QINCo}, using shared (S) or level-specific (L) codebooks where applicable. Appendix~\ref{app:tokenizer-index} defines the complete tokenizer configurations.

Using a frozen tokenizer, each demonstration chunk yields a length-$L$ target token sequence \mbox{$C=\mathcal T(A)=(c_1,\ldots,c_L)$}. Given observation $O$, the policy predicts a sequence $\widehat C$ autoregressively according to $p_\psi(C\mid O)$ and detokenizes its prediction as $\widehat A^\pi=\mathcal D(\widehat C)$.
We train the policy with the teacher-forced negative log-likelihood (NLL)

\begin{equation}
    \mathcal L_{\mathrm{AR}}:=\mathbb E_{(O,C)}[-\log p_\psi(C\mid O)]
    =\mathbb E_{(O,C)}\left[\sum_{i=1}^{L}\ell_{\mathrm{NLL},i}\right],
    \; \ell_{\mathrm{NLL},i}:=-\log p_\psi(c_i\mid c_{<i},O).
    \label{eq:autoregressive-loss}
\end{equation}

The robot executes the first $H_{\mathrm{exec}}\leq H$ actions before replanning; the policy and rollout settings remain fixed across tokenizer configurations.

\section{What Matters for Action Tokenization?}
\label{sec:what-matters}

Understanding how action tokenization affects downstream performance is challenging, since it couples reconstruction, nominal bitrate, action horizon, and how an induced policy generates token sequences. In this section, we investigate these aspects empirically and theoretically. 

\subsection{Low Reconstruction Error Is Necessary but not Sufficient}
\label{sec:reconstruction-threshold}
\begin{figure}[!t]
    \centering
    \captionsetup{skip=-6pt}
    \input{figures/what_matters/reconstruction_error/main_validation_reconstruction_vs_success.tex}
    \caption{Expected validation reconstruction error $\mathcal E_{\mathrm{rec}}$ and FM action-noise tolerance on Robomimic. Each colored point shows a tokenizer's error $\mathcal E_{\mathrm{rec}}$ and downstream rollout success. Black curves indicate the performance of an FM policy with an increasing amount of pose noise, revealing task-dependent action-error tolerance.}
    \label{fig:reconstruction-threshold}
    \vspace{-10px}
\end{figure}
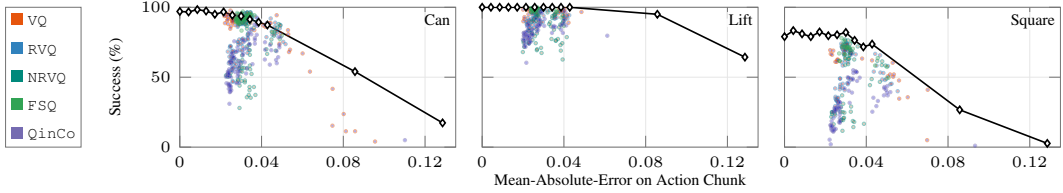
A standard reconstruction metric for an expert action chunk $A$ and its reconstruction $\widehat A$ is the mean absolute error (MAE) \(\ell(A,\widehat A)=\frac{1}{H d_a}\lVert\widehat A-A\rVert_1\).\footnote{$\ell_2$ is another standard reconstruction objective.}
We estimate the expected reconstruction error $\mathcal E_{\mathrm{rec}}=\mathbb E[\ell(A,\mathcal D(\mathcal T(A)))]$ on validation chunks and examine its relationship with downstream rollout success on three Robomimic tasks (Figure~\ref{fig:reconstruction-threshold}). High action MAE consistently results in poor policy performance, while low MAE allows for good performance only some of the time: tokenizers with similar low reconstruction error can induce substantially different downstream performance. The inability to rank tokenizers based on reconstruction error is especially clear within the RVQ, NRVQ, and QinCo subset: the Spearman correlations between reconstruction error and rollout success are non-trivially \emph{positive} on Can ($\rho=0.45$), Lift ($\rho=0.35$), and Square ($\rho=0.50$), respectively (Appendix~\ref{app:reconstruction-error-details}). Thus, once sufficient fidelity is reached, lower reconstruction error is not a reliable indicator of better rollout performance.

To calibrate when action error becomes consequential, we inject controlled noise into action chunks predicted by an observation-conditioned flow-matching (FM) policy, which serves as an expert proxy~\citep{Braun2024RFMP}.
The black curves in Figure~\ref{fig:reconstruction-threshold} show rollout success of the FM policy under increasing action perturbations, with the $x$-axis reporting the induced action-chunk MAE.
The tolerance is task-dependent: Can and Square begin to degrade at relatively small errors, while Lift remains robust to much larger perturbations.
This supports viewing low reconstruction error as a task-dependent necessary constraint rather than a metric that can reliably rank tokenizers once they reach a sufficient level of fidelity.
We provide more details in Appendix~\ref{app:reconstruction-error-details}, including the results of a direct isolation experiment in which FM-predicted action chunks are tokenized, reconstructed, and executed, yielding the same conclusion that low reconstruction error is necessary but not sufficient.

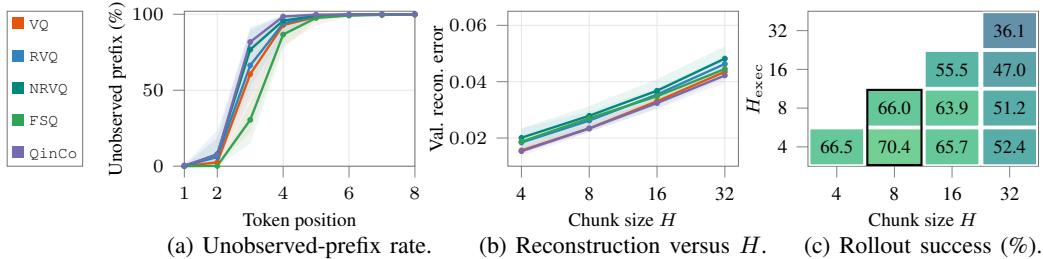
\begin{figure}[!th]
    \centering
    \captionsetup{skip=4pt}
    \def\chunkPlotWidth{3.05cm}
    \def\chunkPlotHeight{2.05cm}
    \begin{tikzpicture}
        \def\chunkPlotName{prefix}\def\chunkPlotX{2.35cm}
        \input{figures/what_matters/empirical_support/main_rollout_prefix_grouped}
        \def\chunkPlotName{reconstruction}\def\chunkPlotX{6.63cm}
        \input{figures/what_matters/bitrate_chunking/main_reconstruction_family}
        \def\chunkPlotName{success}\def\chunkPlotX{10.60cm}
        \input{figures/what_matters/bitrate_chunking/main_success_heatmap}
        \node[anchor=west,draw=black!55,line width=0.35pt,fill=white,
            minimum height=2.05cm,inner xsep=2pt,inner ysep=0pt,
            font=\fontsize{6}{6.4}\selectfont] at (0,1.025cm)
            {\vbox to 1.85cm{%
                \hbox{\tokenizerid{vqbase}{VQ}}\vfil
                \hbox{\tokenizerid{rvqbase}{RVQ}}\vfil
                \hbox{\tokenizerid{nrvqbase}{NRVQ}}\vfil
                \hbox{\tokenizerid{fsqbase}{FSQ}}\vfil
                \hbox{\tokenizerid{qincobase}{QinCo}}}};
        \node[anchor=north,inner sep=0pt,text width=4.15cm]
            at ([yshift=-0.89cm]prefix.south)
            {\subcaption{Unobserved-prefix rate.}\label{fig:unobserved-prefix-rate}};
        \node[anchor=north,inner sep=0pt,text width=4.15cm]
            at ([yshift=-0.89cm]reconstruction.south)
            {\subcaption{Reconstruction versus $H$.}\label{fig:reconerr-chunksize}};
        \node[anchor=north,inner sep=0pt,text width=3.60cm]
            at ([yshift=-0.89cm]success.south)
            {\subcaption{Rollout success (\%).}\label{fig:chunksize-executionhorizon}};
    \end{tikzpicture}\unskip
    \caption{\subref{fig:unobserved-prefix-rate} Rollout unobserved-prefix rate and \subref{fig:reconerr-chunksize} validation reconstruction error versus chunk size, where lines are quantizer-family means and shaded bands are min–max ranges over tokenizer designs; \subref{fig:chunksize-executionhorizon} success (\%) across horizon pairs on Robomimic. Plots in \subref{fig:unobserved-prefix-rate} and \subref{fig:reconerr-chunksize} share colors.}
    \vspace{-10px}
    \label{fig:bitrate-chunking}
\end{figure}

\subsection{Nominal Bitrate and Action Chunking}
\label{sec:bitrate-chunking}

A tokenizer with $L$ levels and $K$ tokens per level has a nominal bitrate (i.e., maximum representational capacity) of $L\log_2 K$ bits. Increasing this capacity generally reduces reconstruction error, but Section~\ref{sec:reconstruction-threshold} shows that beyond a sufficient level of reconstruction accuracy, further reductions in error do not imply better downstream control (Figure~\ref{fig:bitrate-reconerr}; Appendix~\ref{app:nominal-bitrate-analysis}). We therefore treat nominal bitrate as a representational capacity constraint, rather than simply making it as large as possible. Unless stated otherwise, we use $K=128$ and $L=8$ for our main experiments.

Action chunking adds two design choices for the policy: the chunk size $H$ (i.e., the policy prediction horizon), and the execution horizon $H_{\mathrm{exec}}$ (i.e., the open-loop execution length before replanning). Under stability assumptions, open-loop execution can mitigate compounding error, but its benefit diminishes beyond a modest horizon~\citep{Zhang2025ActionChunking}.
We see the same qualitative behavior experimentally, with moderate horizons providing the best rollout success (Figure~\ref{fig:chunksize-executionhorizon}). The degradation at large $H$ is severe for autoregressive policies, as longer chunks increase reconstruction error (Figure~\ref{fig:reconerr-chunksize}) above the level necessary for the task (approximately $0.04$ on Robomimic).
Increasing \(K\) or \(L\) could recover some of this lost fidelity, but it also changes the token space and therefore the downstream prediction problem; higher capacity does not imply that longer chunks become easier to learn or better for control (Appendix~\ref{app:nominal-bitrate-analysis}).
Continuous FM policies show the same performance trend in $H_{\mathrm{exec}}$ (Appendix~\ref{app:chunk-size-details}), so this temporal trade-off is not specific to tokenization. Thus, simply increasing bitrate cannot eliminate the underlying temporal trade-off. Per Figure~\ref{fig:chunksize-executionhorizon}, we primarily use $H=8$ with $H_{\mathrm{exec}}=4$ for high success and $H_{\mathrm{exec}}=8$ for fast inference.

\subsection{Beyond Empirical Training Support: A One-Step Error Analysis}

\label{sec:training-set-composition}
\label{sec:one-step-error}
While imitation learning incentivizes a policy to reproduce token sequences encoded from the demonstrations, rolling out the policy often produces token sequences that are absent from the set of encoded sequences from the training demonstrations.
Figure~\ref{fig:unobserved-prefix-rate} measures the fraction of rollout-generated prefixes that never appear in the encoded demonstrations.
Leaving the empirical training support does not necessarily imply control failure, since unseen token sequences may still decode to valid actions (Appendix~\ref{app:training-set-diagnostics}).

Typically, tokenizers are trained to minimize reconstruction error relative to expert demonstrations, while autoregressive policies are trained to predict the corresponding demonstrated tokens.
Neither objective explicitly controls the action-space error when the policy predicts a token sequence that is different from the encoded target. This motivates the following analysis:
\begin{informalresult}{Proposition~\ref{prop:one-step-error}}
    There are three types of one-step tokenizer-policy errors. When the policy predicts the encoded target action (\textbf{Match}), the error reduces to $\mathcal E_{\mathrm{rec}}$ and is directly optimized by the tokenizer reconstruction objective. When it predicts the encoding of another action close to the expert action for the same observation (\textbf{Benign Mismatch}), the error is determined by action similarity in addition to tokenizer reconstruction quality. For any other prediction (\textbf{Uncontrolled Mismatch}), the action error can be arbitrarily large relative to the reconstruction objective. Teacher forcing encourages exact matches but does not bound the severity of the action-space mismatches.
\end{informalresult}

This distinguishes two properties of token mismatches: how often token mismatches occur and how harmful they are after decoding.
Policy training mainly governs mismatch probability, whereas the tokenizer intrinsically shapes both the probability and severity of token mismatches. We make this decomposition formal in Appendix~\ref{app:one-step-error-proof}.

\section{Learning Robust and Predictable Action Tokenizers}

The analysis in Section~\ref{sec:one-step-error} motivates two complementary tokenizer-only interventions in \methodname: \emph{Token Balance Regularization} for target predictability and \emph{Token Perturbation} for mismatch robustness.
\methodname applies both during tokenizer training without access to the downstream policy, images, or proprioceptive observations, yet consistently improves downstream performance.

\subsection{Token Balance Regularization}
\label{sec:balanced-feature-learning}
Token Balance Regularization (TBR) seeks to \textbf{reduce the occurrence of mismatches} by making the tokenizer's discrete targets easier for a downstream policy to predict.
When the policy predicts the demonstrated token sequence, the decoded action error is controlled by the optimized tokenizer reconstruction error.
However, the encoded target may not be uniquely determined by the observation: demonstrations with similar observations can correspond to different valid token sequences.
Under the tokenizer-induced distribution $P(C\mid O)$, a policy loss $\mathcal L_{\mathrm{AR}}$ (Equation~\ref{eq:autoregressive-loss}) is lower-bounded by
%
%
\begin{equation}
    \mathcal L_{\mathrm{AR}}
    = \ent(C\mid O)
    + \mathbb E_O\!\left[\operatorname{KL}\!\left(P(\cdot\mid O)\,\|\,p_\psi(\cdot\mid O)\right)\right]
    \ge \ent(C\mid O).
    \label{eq:main-irreducible-policy-loss}
\end{equation}
$\ent(C\mid O)$ is the \emph{irreducible policy loss}: a tokenizer-dependent loss component in $\mathcal L_{\mathrm{AR}}$ that remains even when an unrestricted policy matches the conditional distribution of token targets.

A tokenizer partitions the action space into regions, with all actions in the same region assigned to the same token;
we refer to the interfaces between different regions as \emph{token boundaries}. The \emph{token margin} of an action is the distance to the nearest boundary.
Intuitively, given an observation, a larger margin makes nearby actions more likely to stay within the same token region, and avoids assigning them to different tokens.
As a result, nearby actions are more likely to share the same token target. We show analytically that in the case of a linear encoder model, balanced feature learning (BFL) can reduce mismatches by inducing larger token margins, as summarized in the following result.

\begin{informalresult}{Proposition~\ref{prop:irr-code-noise-sensitivity} and Corollary~\ref{cor:irr-gd-vs-ap}}
Covariance-aware preconditioning, a type of balanced feature learning, reduces disparities in learning rates across different feature directions. Assuming a linear encoder, this disparity reduction translates directly into larger token margins, which lowers the irreducible policy loss and makes it easier for a downstream policy to predict the right tokens during training. This is because the preconditioning reduces the encoder's sensitivity to perturbations in action-space.
\end{informalresult}


Therefore, we design TBR to promote more balanced feature learning (BFL) across different directions during tokenizer training.
TBR relates to methods that regulate feature scales or correlations, such as batch normalization~\citep{Ioffe2015BatchNorm} and whitening~\citep{Huang2018DBN}.
We instantiate TBR with a recent BFL method using covariance-aware preconditioning~\citep{Martens2015KFAC,Zhang2026DoPr}
for optimization updates without changing the training objective.
Appendices~\ref{app:bfl-irreducible-loss} and~\ref{app:balanced-feature-implementation} provide the details of our analysis and implementation.

\subsection{Token Perturbation}
\label{sec:token-perturbation}

Token Perturbation (TP) aims to \textbf{constrain the severity of token mismatches} by training the tokenizer to reconstruct an action from perturbed versions of its encoded token sequence.
Given an action $A$ with encoding $C=\mathcal T(A)$, we sample a perturbed sequence $C'\sim\nu(\cdot\mid C)$ and reconstruct the original action using the same training objective $\ell(A,\mathcal D(C'))$.
We implement this with a straight-through estimator to train the entire tokenizer, so the forward pass uses the perturbed token embedding while gradients still flow through the original encoder latent.
Here, $\nu(\cdot\mid C)$ denotes the conditional perturbation distribution.
One simple perturbation rule randomly selects token positions and replaces them with nearby quantizer candidates in latent space.
Reconstructing from token perturbation encourages robustness to token mismatches.
Despite being policy-agnostic proxies rather than actual policy errors, these perturbations substantially improve downstream rollout success.
We additionally study spectral-metric requantization for perturbation sampling and MaxUp~\citep{Gong2020MaxUp}; Appendix~\ref{app:token-perturbation} provides details and comparisons.

\section{Experiments}

\subsection{Experimental Setting}


We evaluate a diverse set of 22 tokenizer configurations, named ⟨\texttt{backbone}⟩-⟨\texttt{quantizer}⟩-⟨\texttt{S}/\texttt{L}⟩ (Section~\ref{sec:preliminaries}, Appendix~\ref{app:tokenizer-index}), while holding the downstream autoregressive policy configuration fixed. Our simulation study spans three manipulation benchmarks (Section~\ref{sec:exp-sim})--- Robomimic~\citep{Mandlekar2022RoboMimic}, LIBERO~\citep{Liu2023LIBERO} and RoboTwin~\citep{Mu2025RoboTwin}---covering both single-arm and bimanual manipulation tasks. Baseline tokenizers are trained with the Reconstruction Only (R.O.) objective, whereas \methodname augments tokenizer training with Token Balance Regularization and Token Perturbation. Additionally, we evaluate our method with a VLA in LIBERO (Section~\ref{sec:exp-sim}) and in real-world manipulation domains (Section~\ref{sec:real-robot-experiments}).
Appendix~\ref{app:policy-seed-study} and~\ref{app:real-robot-results} document the detailed setup, hyperparameters, and evaluation protocol. 

\input{tables/experiments/simulation_results/main_multitask}

\subsection{Simulation Results}
\label{sec:exp-sim}
Table~\ref{tab:simulation-multitask} evaluates \methodname{} across the tokenizer design space. Across Robomimic, LIBERO, and RoboTwin,
\methodname{} improves average success by $10.4$ and $11.3$ points at $H_{\mathrm{exec}}=4$ and $H_{\mathrm{exec}}=8$, respectively\footnote{Averages exclude \texttt{CT-NRVQ-S/L}, whose 0\% baseline R.O. success rate is an outlier.}. 
The gains are particularly pronounced for level-specific (L) configurations ($+12.3$/$+13.1$ points) and QinCo variants ($+20.2$/$+22.6$ points),
spanning substantially different quantization designs. Notably, these configurations often achieve lower $\mathcal E_{\mathrm{rec}}$ under R.O. (Table~\ref{tab:simulation-validation-reconstruction}) despite substantially worse rollout performance, reinforcing that better reconstruction does not necessarily translate to better downstream success.
Aggregate improvements extend across all three benchmarks using a common set of hyperparameters for most tokenizer configurations within each benchmark (Table~\ref{tab:token-balance-hyperparameters}; Table~\ref{tab:token-perturbation-hyperparameters}).
Appendix~\ref{app:simulation-results} provides per-task results and additional data-setting and checkpoint comparisons.

\begin{table}[h]
    \centering
    \captionsetup{hypcap=false}
    \vspace{6pt}
    \begin{minipage}[t]{0.38\textwidth}
        \vspace{0pt}
        \input{tables/experiments/simulation_results/main_vla_libero}
    \end{minipage}\hfill
    \begin{minipage}[t]{0.56\textwidth}
        \vspace{0pt}
        \input{tables/experiments/simulation_results/main_robomimic_aggregate_ablation}
    \end{minipage}
\end{table}

To test whether these gains transfer to a pretrained vision-language backbone, we fine-tune SmolVLM2-256M~\citep{Marafioti2025SmolVLM} with an autoregressive action expert on LIBERO, with implementation details provided in Appendix Table~\ref{tab:vla-training-hyperparameters}. \methodname{} improves success rate in 7 of 8 tokenizer pairs,
raising the average from $26.2\%$ to $48.1\%$ (Table~\ref{tab:vla-libero-results}).

\subsection{Ablations and Error Analyses}
\label{sec:exp-ablation}
We ablate the two proposed tokenizer-only interventions on Robomimic across the tokenizer design space (Table~\ref{tab:robomimic-aggregate-ablation}).\footnote{The aggregate excludes \texttt{CT-NRVQ-S/L}, whose 0\% baseline R.O. success rate is an outlier.} Both \emph{Token Balance Regularization} (\wotp{}) and \emph{Token Perturbation} (\wotbr{}) independently improve rollout success, while including both results in the best performance. Appendix~\ref{app:ablation-results} reports the complete task-level results for all tokenizer configurations.


Motivated by the one-step error analysis in Section~\ref{sec:one-step-error}, we next examine the specific effect targeted by each intervention: whether TBR makes the resulting token sequences easier for the policy to predict, and whether TP reduces the action-space severity of token mismatches.



\begin{figure}[tbp]
    \centering
    \input{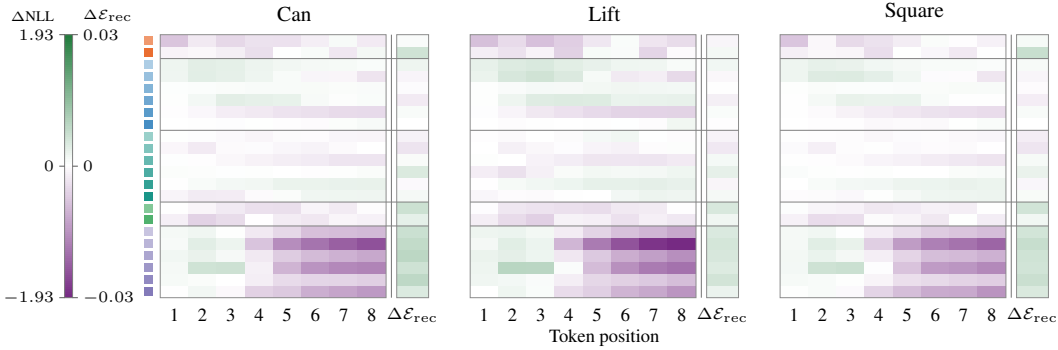}
    \captionsetup{skip=-10pt}
    \caption{Effect of Token Balance Regularization (TBR). $\Delta_{\mathrm{NLL}}$ is the difference in normalized NLL between \methodname and \wotbr at positions 1--8. $\Delta\mathcal E_{\mathrm{rec}}$ denotes the corresponding difference in validation reconstruction MAE. Each row is labeled with the same tokenizer color indicator as in Table~\ref{tab:simulation-multitask}. Purple denotes decreases and green increases.}
    \label{fig:token-position-nll-tp-ours}
    \vspace{-13px}
\end{figure}

\paragraph{Token predictability across positions.}
Token Balance Regularization is designed to make the tokenizer's discrete targets easier for a downstream policy to predict.
To test this effect directly, we measure the teacher-forced per-position NLL (Eqn.~\ref{eq:autoregressive-loss}), normalized by $\log K$, comparing \wotbr with \methodname (Figure~\ref{fig:token-position-nll-tp-ours}).
Across tokenizer variants, TBR reduces mean normalized NLL by 12.4\% on Can, 10.0\% on Lift, and 11.3\% on Square, and reduces the within-tokenizer standard deviation of NLL across positions by 18.9--21.4\%.
These gains are accompanied by a $7.2\%$ average increase in validation MAE relative to w/o TBR across tokenizers and tasks.
Figure~\ref{fig:token-position-nll-tp-ours} shows both metrics for each tokenizer.
The reduction is most pronounced at later autoregressive positions, yielding a more balanced prediction difficulty across the sequence, as seen in the smaller cross-position NLL standard deviation.
Together, these results show that TBR makes the induced token targets easier to predict and more uniform across positions, with a reconstruction trade-off. Appendix~\ref{app:prediction-error-analysis} reports aggregate diagnostics and per-tokenizer NLL curves.

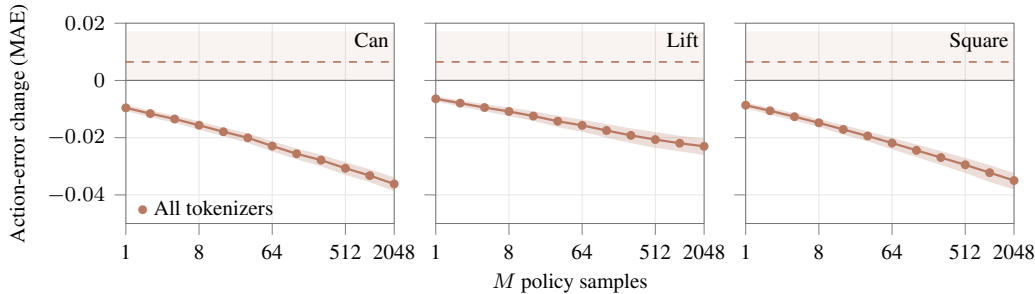
\begin{figure}[!tb]
    \centering
    \input{figures/experiments/reconstruction_error/main_decoded_robustness_pooled.tex}
    \caption{Decoded-action robustness of Token Perturbation. The curves correspond to the mean paired change in Max-$M$ mismatch error between \methodname and \wotp as the number of sampled policy sequences $M$ increases. Negative values indicate that TP reduces decoded action error; shaded bands denote the 95\% bootstrap confidence intervals. The dashed line shows the corresponding change in tokenizer reconstruction error.}
    \label{fig:decoded-robustness-pooled}
    \vspace{-14px}
\end{figure}

\paragraph{Decoded-action robustness.}
Token Perturbation aims to reduce the action-space impact of token mismatches.
For each validation observation with expert action chunk $A$, we sample $M$ policy-predicted token sequences and decode them into actions $\widehat A^{\pi,(1)},\ldots,\widehat A^{\pi,(M)}$.
Among samples whose token sequence differs from the expert encoding, we record the Max-$M$ mismatch error $\ell_{\pi}^{\max}(M)=\max_{i\in[1, M]}\ell(A,\widehat A^{\pi,(i)})$.
We then define
$\Delta_{\mathrm{mismatch}}(M)=\ell_{\methodname{}}^{\max}(M)-\ell_{\wotp{}}^{\max}(M)$
and report its average over paired tokenizer variants in Figure~\ref{fig:decoded-robustness-pooled}.
As a reference, we also report the paired difference in tokenizer reconstruction error, $\Delta\mathcal E_{\mathrm{rec}}$, between \methodname{} and \wotp{}.
This paired ablation isolates the effect of TP. As $M$ increases, $\Delta_{\mathrm{mismatch}}(M)$ becomes increasingly negative, showing that TP reduces the largest decoded errors exposed by broader policy sampling, despite a small increase in tokenizer reconstruction error.
Since policy-generated sequences rarely exactly match the expert encoding (Appendix~\ref{app:training-set-diagnostics}), this directly supports the intended effect of TP: reducing the action-space severity of token mismatches.
Appendix~\ref{app:decoded-error-analysis} reports the R.O. versus \wotbr{} comparison and discusses the effect of different policy token sampling strategies.


\subsection{Real-Robot Experiments}
\label{sec:real-robot-experiments}
\begin{wraptable}{r}{0.56\textwidth}
    \vspace{-\intextsep}
    \centering
    \begin{minipage}{\linewidth}
        \centering
        {\input{tables/experiments/real_robot/main_ur5}\par}
        \captionsetup{skip=4pt}
        \captionof{table}{UR5 results: successes/trials (success \%).}
        \label{tab:ur5-results}
    \end{minipage}\par\vspace{3pt}
    \begin{minipage}{\linewidth}
        \centering
        \captionsetup{type=figure,skip=4pt}
        \begin{tikzpicture}
            \node[inner sep=0pt,outer sep=0pt] (photo) {\includegraphics[width=0.48\linewidth]{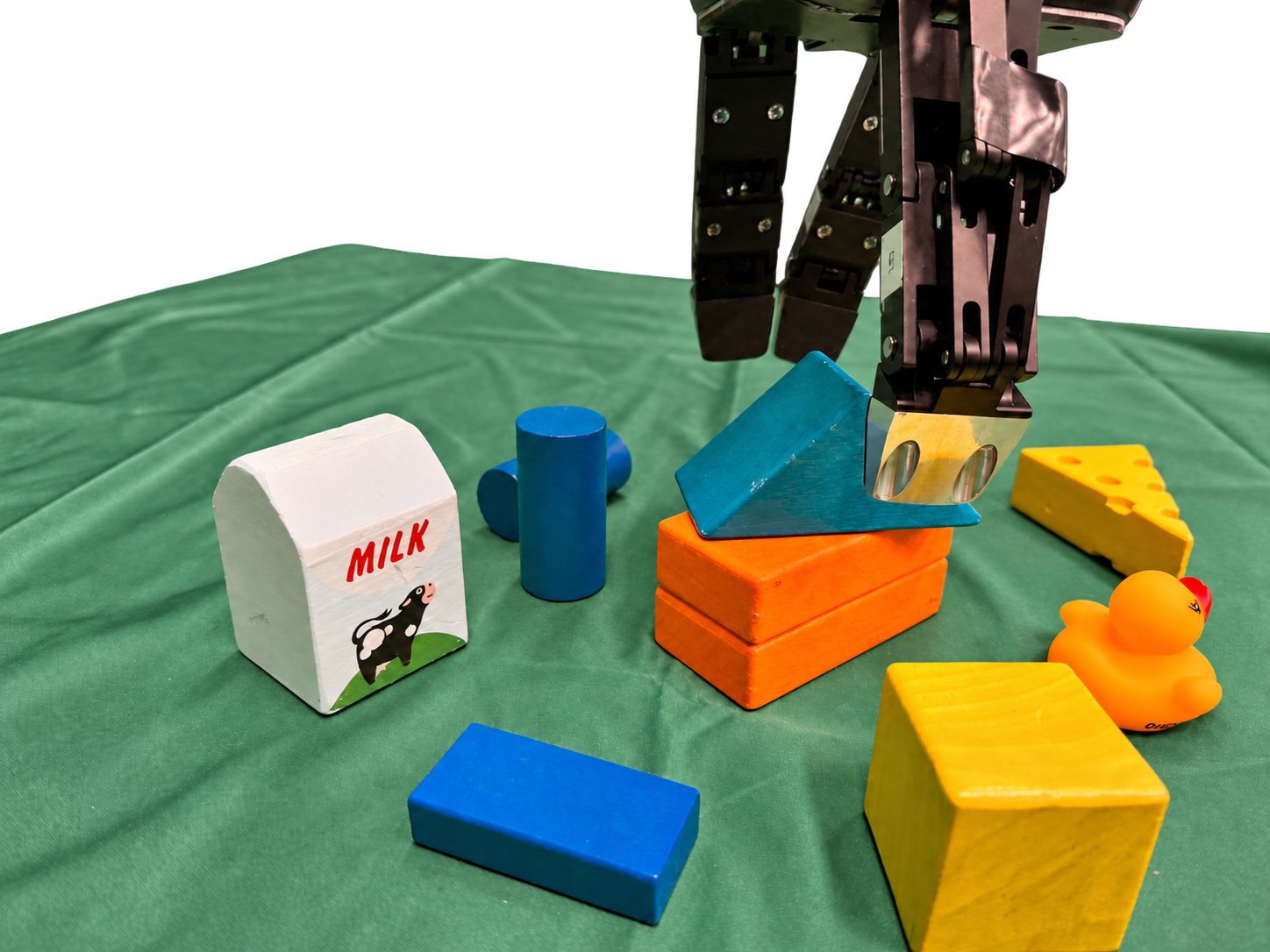}};
            \node[overlay,anchor=north west,inner sep=0pt,outer sep=0pt,align=left,text=black,font=\fontsize{8}{9}\selectfont\bfseries]
                at ([xshift=2pt,yshift=-2pt]photo.north west) {Cube\\Stacking};
        \end{tikzpicture}\hfill
        \begin{tikzpicture}
            \node[inner sep=0pt,outer sep=0pt] (photo) {\includegraphics[width=0.48\linewidth]{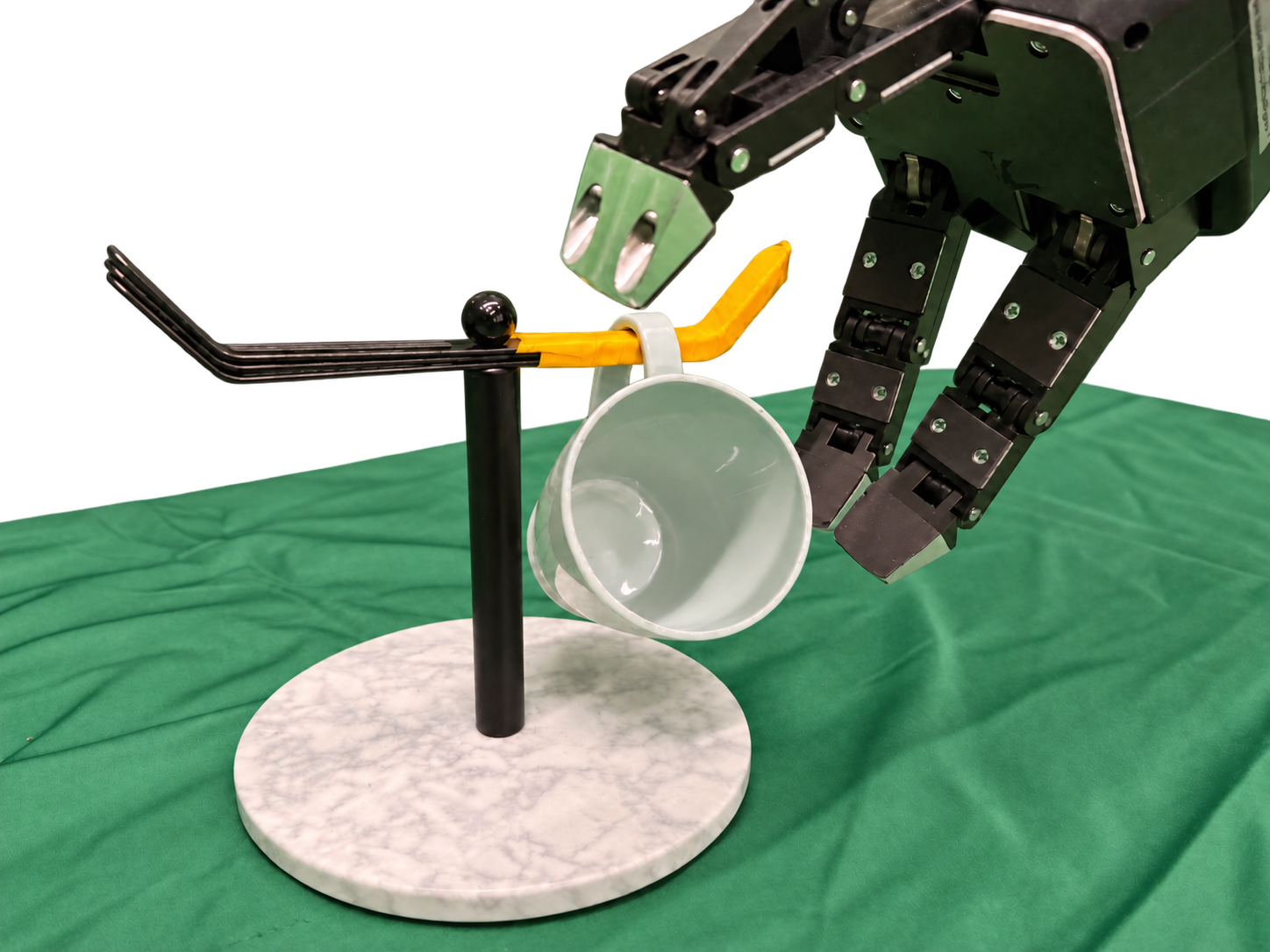}};
            \node[overlay,anchor=north west,inner sep=0pt,outer sep=0pt,align=left,text=black,font=\fontsize{8}{9}\selectfont\bfseries]
                at ([xshift=2pt,yshift=-2pt]photo.north west) {Cup\\Hanging};
        \end{tikzpicture}
        \caption{The two real-robot manipulation tasks.}
        \label{fig:ur5-tasks}
    \end{minipage}
    \vspace{-\intextsep}
\end{wraptable}
We further evaluate on two real-world UR5 manipulation tasks (Figure~\ref{fig:ur5-tasks}): \emph{Cube Stacking}, which requires placing a blue triangular block on an orange rectangular block amid cluttered distractors,
and \emph{Cup Hanging}, which requires grasping a cup and hanging it onto a cup tree with accurate positioning and alignment.
Across both tasks, \methodname{} improves success for all evaluated tokenizer configurations.
The gains are particularly pronounced for the level-specific variants on Cube Stacking, consistent with our simulation results, while Cup Hanging also shows substantial improvements across tokenizers.
Table~\ref{tab:ur5-results} reports the per-task results, and Appendix~\ref{app:real-robot-results} provides the full setup and evaluation protocol.

\par\Wrapfill

\section{Conclusions}
\label{sec:conclusions}
Effective action tokenization requires more than accurate reconstruction.
Through a controlled study across a broad tokenizer design space, we show that poor reconstruction limits downstream performance, but that among sufficiently accurate tokenizers, rollout success also varies with chunk size, token predictability, and mismatch severity.
Additionally, we provide a one-step error analysis that identifies a gap in standard tokenizer and policy objectives: they encourage accurate reconstruction and target prediction, but do not control the severity of token mismatches.
To address this gap, we introduce \methodname, which combines \emph{Token Balance Regularization} and \emph{Token Perturbation}, two policy-independent tokenizer training methods that target predictability and robustness, respectively.
\methodname improves downstream policy performance across Robomimic, LIBERO, RoboTwin, and real-robot manipulation.

Our experiments focus on autoregressive imitation-learning policies and a limited set of benchmarks, robot embodiments, and training seeds.
An important next step is to test whether these findings extend to reinforcement learning and to broader task and embodiment distributions.

\clearpage
\subsection*{AI use statement}
We used large language models and AI tools to assist with drafting and polishing manuscript sections and detailed results, analyses in the appendices, retrieving and identifying relevant literature in tokenization and robot learning, research experiment implementation, and proving and validating the mathematical claims.
This assistance included developing explanatory hypotheses, interpreting experimental results, preparing analysis and visualization scripts, and formatting tables and figures.
We did not use AI tools to generate synthetic datasets.
We are responsible for the final content of this work, including its text, claims, code, and other artifacts produced with AI assistance.

\subsection*{Ethics statement}
This work studies action tokenization for robot manipulation.
Erroneous robot actions can cause physical harm or damage, and improved performance on the evaluated benchmarks does not imply that a policy is safe to deploy.
Deployment requires appropriate human supervision, physical and software safeguards, and validation under the intended operating conditions, including tasks, environments, and embodiments beyond those studied here.
The reported improvements should therefore be interpreted within the scope of the evaluated settings rather than as safety guarantees.

\subsection*{Reproducibility statement}
We document the tokenizer configurations in Appendix~\ref{app:tokenizer-index}, rollout evaluation and uncertainty estimation in Appendix~\ref{app:rollout-evaluation}, and training hyperparameters, checkpoint conventions, and sources of randomness in Appendix~\ref{app:policy-seed-study}.
Appendices~\ref{app:balanced-feature-implementation} and~\ref{app:token-perturbation} describe the TBR and TP implementations and their hyperparameters.
Detailed simulation results and ablations are reported in Appendices~\ref{app:simulation-results} and~\ref{app:ablation-results}, while Appendix~\ref{app:real-robot-results} documents the real-robot setup and evaluation protocol.

\clearpage
\bibliography{references}
\bibliographystyle{iclr2027/iclr2027_conference}

\clearpage
\flushbottom
\appendix

\section*{Appendices}

\section{Analytical Tokenization Schemes}
\label{app:analytical-tokenization}

Our controlled study focuses on learned action tokenizers rather than fixed analytical schemes. As a representative analytical baseline, we consider per-dimension binning (Bin). Under the same autoregressive policy training protocol used throughout our study, without representation-specific augmentations or decoding heuristics,\footnote{We intentionally exclude techniques such as Masked Action Augmentation~\citep{Goyal2025VLA0} as they may mitigate limitations of the underlying token representation and make it harder to isolate the effect of the tokenizer itself.} Bin reaches $67.7\%$ / $68.6\%$ success on Robomimic at $H_{\mathrm{exec}}=4\,/\,8$ (Table~\ref{tab:analytical-rollout}), and its reconstruction MAE is $0.002533$.\footnote{This error is substantially smaller than the typical learned-tokenizer validation errors reported in Appendix~\ref{app:reconstruction-error-details}, Figure~\ref{fig:appendix-reconstruction-success-scatter}.} Bin, however, spends $56$ tokens on an $H=8$ chunk, compared with $8$ tokens for the learned tokenizers in our main experiments, and its success remains comparable with that of the learning-based tokenizers trained with the R.O. objective, while using twice the codebook size per level (Table~\ref{tab:simulation-multitask}). We therefore examine how much of the Bin code a policy actually needs. These experiments motivate our focus on learning-based tokenizers in the controlled study.

\begin{table}[H]
    \centering
    \tablebodyfont
    \begin{minipage}[t]{0.37\linewidth}
        \centering
        \caption{Configuration of the Bin tokenizer.}
        \label{tab:analytical-tokenizer-config}
        \setlength{\tabcolsep}{3pt}
        \begin{tabular}{@{}ll@{}}
            \toprule
            Representation & Per-scalar quantization \\
            Vocabulary / bins & 256 bins per scalar \\
            Normalization & Per-dim. q01/q99 to $[-1,1]$ \\
            Quantization & Uniform over $[-1,1]$ \\
            Sequence length & 56 \\
            Learned parameters & None \\
            \bottomrule
        \end{tabular}
    \end{minipage}\hfill
    \begin{minipage}[t]{0.59\linewidth}
        \centering
        \caption{Robomimic success rates (\%) of Bin and Prefix-Bin at $H_{\mathrm{exec}}=4\,/\,8$, using the checkpoint with the best teacher-forced validation accuracy and 10 seeds $\times$ 30 episodes per task.}
        \label{tab:analytical-rollout}
        \setlength{\tabcolsep}{3pt}
        \begin{tabular}{@{}lccccc@{}}
            \toprule
            Tokenizer & Tokens & Can & Lift & Square & Avg. \\
            \midrule
            Bin        & 56 & 78.7\,/\,82.0 & 96.0\,/\,98.0 & 28.3\,/\,25.7 & 67.7\,/\,68.6 \\
            Prefix-Bin & 8  & 53.7\,/\,27.3 & 83.7\,/\,55.3 & 15.7\,/\,1.7  & 51.0\,/\,28.1 \\
            \bottomrule
        \end{tabular}
    \end{minipage}
\end{table}
%
%
Table~\ref{tab:analytical-tokenizer-config} summarizes the Bin configuration. Bin independently quantizes each scalar action into 256 bins, producing a fixed 56-token sequence for an $H=8$, $d_a=7$ action chunk.

\paragraph{How much of the Bin code does the policy need?}
Bin represents an $H=8$ action chunk as $56$ tokens, one for each action dimension and timestep. If most of this sequence is redundant, a policy should not need to predict all of it. We construct \emph{Prefix-Bin}: the policy predicts only the first eight Bin tokens, after which an empirical training-set lookup completes the corresponding 56-token sequence for decoding. Prefix-Bin therefore reduces the policy prediction horizon from $56$ to $8$ tokens while retaining essentially the same reconstruction fidelity as Bin, with a negligible MAE difference of approximately $3.29\times10^{-7}$.

This lookup is almost unambiguous on the training corpus: 179,621 distinct eight-token prefixes correspond to 179,622 distinct full Bin sequences. Predicting only these eight tokens, Prefix-Bin retains $51.0\%$ success at $H_{\mathrm{exec}}=4$, about three quarters of Bin's, with a prediction horizon seven times shorter (Table~\ref{tab:analytical-rollout}). The gap widens at $H_{\mathrm{exec}}=8$ ($28.1\%$): the policy's predicted prefixes almost never appear in the training table (10 of 116,977 at rollout), so the suffix is completed from the nearest table entry, and executing more of each chunk executes more of these retrieved actions.

Bin therefore works, but inefficiently: most of its tokens are determined by a short prefix, and a policy that predicts one seventh of them retains three quarters of its success at $H_{\mathrm{exec}}=4$. A learned tokenizer can instead spend a much shorter code on the structure that varies across action chunks, which is also where TBR and TP act. We therefore focus the controlled study on learned tokenizers.

\section{Additional Preliminaries}
\label{app:additional-preliminaries}

\subsection{Tokenizer Configuration Index}
\label{app:tokenizer-index}

\subsubsection{Tokenizer Backbones}
The tokenizer backbone implements the encoder $E$ and decoder $D$ surrounding the discrete bottleneck.
We use three variants that differ in how they organize and compress temporal and action-coordinate structure.

\paragraph{Transformer (\textsc{TF}).}
The encoder embeds each action timestep, adds temporal position embeddings, appends $R$ learned register tokens, and processes the joint sequence with masked self-attention.
Each register can attend to the complete action chunk and to preceding registers, and the final register states are projected to $z_e\in\mathbb R^{R\times d_z}$.
The decoder uses one learned query per output timestep and cross-attends to the quantized registers in a single pass.
We set $R=1$ for residual quantizers and $R=L$ for the parallel VQ and FSQ families.
This construction follows the register-token action-autoencoder pattern used in recent learned action tokenizers~\citep{Liu2026OAT,Dong2026ActionCodec}.

\paragraph{Conv2D (\textsc{C2}).}
The encoder partitions the action coordinates into semantic groups and divides time into non-overlapping windows, producing a $4\times4$ or $4\times8$ patch grid depending on the action space.
Convolutional residual blocks and grid self-attention operate at successive resolutions while two downsampling stages compress both grid axes into one global latent vector.
The decoder mirrors this hierarchy with two-dimensional upsampling and unpatchifies the reconstructed grid back into an action chunk.
Although this backbone includes lightweight self-attention within the grids, its defining structure is hierarchical two-dimensional compression over time and grouped action coordinates, related to the convolutional action-tokenizer design used by VQ-VLA~\citep{Wang2025VQVLA}.

\paragraph{Convolution--Transformer (\textsc{CT}).}
The hybrid encoder divides time into windows while preserving a separate grid column for every action coordinate, with additional embeddings identifying individual coordinates and semantic action groups.
It interleaves convolutional residual blocks with two-dimensional self-attention while downsampling only the temporal axis from four rows to two and then one.
Attention pooling across the remaining action-coordinate columns produces one global latent vector; the decoder reverses the temporal hierarchy to reconstruct the action chunk.
This design combines the local hierarchy of convolution with attention over the full action grid, following the broad convolution--attention pattern explored by FASTer and QueST~\citep{Liu2025FASTer,Mete2024QueST}.

The \textsc{C2} and \textsc{CT} backbones emit one latent vector and are therefore paired with the residual quantizer families; \textsc{TF} also supports multiple register latents and is used with all five families.

\subsubsection{Quantizer Families}

The quantizer $Q=(q,e)$ maps encoder embeddings to discrete symbols through $q$ and maps those symbols back to quantized embeddings through $e$.
Let $L$ be the number of quantization layers, $\ell\in[L]$ index a layer, and $K$ be the number of possible token indices at each layer, with $k\in[K]$ indexing a candidate.
The indexing map produces $C=q(z_e)=(c_1,\ldots,c_L)\in[K]^L$, and the embedding map produces $z_q=e(C)$.
Thus each tokenizer represents one action chunk using a nominal $L\log_2K$ bits.
VQ and FSQ quantize $L$ parallel encoder latents $z_e=(z_{e,1},\ldots,z_{e,L})$, whereas RVQ, NRVQ, and QinCo generate $L$ indices sequentially from one latent vector.

\paragraph{Vector quantization (VQ).}
At layer $\ell$, VQ uses a learned codebook $\mathcal E_\ell=\{v_{\ell,k}\}_{k=1}^K$ and independently assigns latent $z_{e,\ell}$ to its nearest codeword~\citep{Oord2017VQVAE}:
\begin{equation}
    c_\ell=\underset{k\in[K]}{\arg\min}\,
    \lVert z_{e,\ell}-v_{\ell,k}\rVert_2^2,
    \qquad z_{q,\ell}=v_{\ell,c_\ell}.
    \label{eq:appendix-vq}
\end{equation}
Thus, the indices are $C=(c_1,\ldots,c_L)$ and the quantized representation is $z_q=(z_{q,1},\ldots,z_{q,L})$.

\paragraph{Residual vector quantization (RVQ).}
RVQ successively quantizes the error left by earlier layers~\citep{Lee2022RQ,Zeghidour2022SoundStream}.
At layer $\ell$, let $\mathcal E_\ell=\{v_{\ell,k}\}_{k=1}^K$ denote the learned codebook.
Starting from $r_0=z_e$, RVQ applies nearest-neighbor lookup to each residual:
\begin{equation}
\begin{aligned}
    c_\ell&=\underset{k\in[K]}{\arg\min}\,
        \lVert r_{\ell-1}-v_{\ell,k}\rVert_2^2,
    &\qquad
    r_\ell&=r_{\ell-1}-v_{\ell,c_\ell},\\
    z_q&=\sum_{\ell=1}^{L}v_{\ell,c_\ell}.
\end{aligned}
    \label{eq:appendix-rvq}
\end{equation}
Thus, the indices $C=(c_1,\ldots,c_L)$ form an ordered coarse-to-fine sequence, and the quantized representation is $z_q=\sum_{\ell=1}^{L}v_{\ell,c_\ell}$.

\paragraph{Normalized residual vector quantization (NRVQ).}
NRVQ retains the residual hierarchy of RVQ but performs assignment using normalized code directions and scales the selected direction at each layer by a learned scalar $s_\ell$~\citep{Yu2022ImprovedVQGAN,kumar2023high}.
With $r_0=z_e$, it computes
\begin{equation}
\begin{aligned}
    \bar v_{\ell,k}
        &=\frac{v_{\ell,k}}{\lVert v_{\ell,k}\rVert_2+\epsilon},
    &\qquad
    c_\ell
        &=\underset{k\in[K]}{\arg\max}\;
          r_{\ell-1}^{\mathsf T}\bar v_{\ell,k},\\
    r_\ell
        &=r_{\ell-1}-s_\ell\bar v_{\ell,c_\ell},
    &\qquad
    z_q
        &=\sum_{\ell=1}^{L}s_\ell\bar v_{\ell,c_\ell}.
\end{aligned}
    \label{eq:appendix-nrvq}
\end{equation}
The lookup normalizes the candidate directions but not the incoming residual; normalizing the residual would not change the selected index because its norm is constant across candidates.
The learned scalar $s_\ell$ controls the contribution of layer $\ell$ to the reconstructed latent.
Thus, the indices are $C=(c_1,\ldots,c_L)$ and the quantized representation is $z_q=\sum_{\ell=1}^{L}s_\ell\bar v_{\ell,c_\ell}$.

\paragraph{Finite scalar quantization (FSQ).}
FSQ quantizes each of the $L$ encoder latents using a fixed Cartesian grid~\citep{Mentzer2024FSQ}.
Let $\phi$ be a bijection from the grid points to $[K]$.
At layer $\ell$, $\Pi_\ell^{\mathrm{in}}$ projects $z_{e,\ell}$ to $d_f$ scalar coordinates, applies $\tanh$-based bounding, and rounds each coordinate to its prescribed finite level with a straight-through estimator.
The token index and quantized latent are
\begin{equation}
    c_\ell=\phi\!\left(\Pi_\ell^{\mathrm{in}}(z_{e,\ell})\right),
    \qquad
    z_{q,\ell}=\Pi_\ell^{\mathrm{out}}\!\left(\phi^{-1}(c_\ell)\right).
    \label{eq:appendix-fsq}
\end{equation}
Here $K$ is the product of the per-coordinate level counts, and $\Pi_\ell^{\mathrm{out}}$ maps the selected grid point to the decoder embedding.
Thus, the indices are $C=(c_1,\ldots,c_L)$ and the quantized representation is $z_q=(z_{q,1},\ldots,z_{q,L})$.

\paragraph{Quantization with Implicit Neural Codebooks (QinCo).}
QinCo extends residual quantization by generating the codebook at each layer conditioned on the quantized representation accumulated from previous layers~\citep{Huijben2024QINCo,Vallaeys2025QINCo2}.
Let $z_e$ denote the encoder output, $L$ the number of quantization layers, and $K$ the number of candidate code vectors at each layer.
We denote by $z_q^{(\ell)}$ the cumulative quantized representation after layer $\ell$, with $z_q^{(0)}=0$, and define the residual before layer $\ell$ as
$
r_{\ell-1}=z_e-z_q^{(\ell-1)}.
$

The first layer uses a learned static codebook
$
\mathcal E_1=\{v_{1,k}\}_{k=1}^{K},
$
where $v_{1,k}$ denotes the $k$-th code vector.
It selects the index
$
c_1=\arg\min_{k\in[K]}\lVert z_e-v_{1,k}\rVert_2^2
$
and initializes the quantized representation as
$
z_q^{(1)}=v_{1,c_1}.
$

For each later layer $\ell\geq2$, a learnable network $\mathcal V_\ell$ maps $z_q^{(\ell-1)}$ to a sample-dependent codebook, after which QinCo selects the code vector closest to the current residual:
\begin{equation}
\begin{aligned}
\mathcal V_\ell\!\left(z_q^{(\ell-1)}\right)
&=\left\{v_{\ell,k}\!\left(z_q^{(\ell-1)}\right)\right\}_{k=1}^{K},\\
c_\ell
&=\underset{k\in[K]}{\arg\min}\,
\left\lVert r_{\ell-1}-v_{\ell,k}\!\left(z_q^{(\ell-1)}\right)\right\rVert_2^2,\\
z_q^{(\ell)}
&=z_q^{(\ell-1)}+v_{\ell,c_\ell}\!\left(z_q^{(\ell-1)}\right).
\end{aligned}
\label{eq:appendix-qinco}
\end{equation}
Here, $v_{\ell,k}(z_q^{(\ell-1)})$ is the $k$-th code vector generated for layer $\ell$ under the current prefix $z_q^{(\ell-1)}$, and $c_\ell$ is the index selected at that layer.
In our implementation, each conditional code vector is formed by adding a learned static anchor to the output of a linear generator conditioned on $z_q^{(\ell-1)}$.
Therefore, unlike a static codebook, the same token index $k$ can correspond to different code vectors for different prefixes.
The resulting token sequence is $C=(c_1,\ldots,c_L)$, and the final quantized representation is $z_q=z_q^{(L)}$.

\subsubsection{Shared and Level-Specific Quantization}
\label{app:shared-level-specific}

Let $\Theta_\ell$ denote the family-specific parameter block whose sharing status defines the quantizer variant at layer $\ell$.
A \emph{shared} quantizer ties these parameters across layers, $\Theta_\ell=\Theta$, whereas a \emph{level-specific} quantizer uses independently parameterized $\Theta_\ell$.
We denote the two variants by the suffixes \texttt{S} and \texttt{L}, respectively.
Both variants produce indices $C\in[K]^L$ and therefore have the same nominal bitrate; they differ only in which quantizer parameters are reused across layers.

\paragraph{VQ.}
For VQ, $\Theta_\ell$ is the learned codebook $\mathcal E_\ell$.
Shared VQ uses one codebook $\mathcal E=\{v_k\}_{k=1}^{K}$ for every parallel latent, so the same index $k$ maps to $v_k$ at every layer.
Level-specific VQ instead learns an independent codebook $\mathcal E_\ell=\{v_{\ell,k}\}_{k=1}^{K}$ for each layer, so the vector associated with index $k$ can vary with $\ell$.

\paragraph{RVQ.}
For RVQ, $\Theta_\ell$ contains the vectors used to quantize the residual at layer $\ell$.
Shared RVQ reuses one learned table $\mathcal E=\{v_k\}_{k=1}^{K}$ throughout the residual hierarchy and learns a scalar $s_\ell$ for each layer.
Its assignment and residual update are
\begin{equation}
    c_\ell=\underset{k\in[K]}{\arg\min}\,
    \lVert r_{\ell-1}-s_\ell v_k\rVert_2^2,
    \qquad
    r_\ell=r_{\ell-1}-s_\ell v_{c_\ell}.
    \label{eq:appendix-shared-rvq}
\end{equation}
Level-specific RVQ instead learns $\mathcal E_\ell=\{v_{\ell,k}\}_{k=1}^{K}$ independently at every layer and applies the RVQ recursion in Equation~\ref{eq:appendix-rvq} directly with $v_{\ell,k}$.

\paragraph{NRVQ.}
For NRVQ, $\Theta_\ell$ is the normalized direction table.
Shared NRVQ reuses one table $\{\bar v_k\}_{k=1}^{K}$, selects $c_\ell$ using these shared directions, and subtracts $s_\ell\bar v_{c_\ell}$ from the residual.
Level-specific NRVQ instead learns an independent table $\{\bar v_{\ell,k}\}_{k=1}^{K}$ at every layer, selects $c_\ell$ using the layer-specific directions, and subtracts $s_\ell\bar v_{\ell,c_\ell}$.
Both variants retain an independent learned scale $s_\ell$ for each layer; only the normalized directions are shared or learned separately.

\paragraph{FSQ.}
For FSQ, the Cartesian grid is fixed, so $\Theta_\ell$ consists only of the learned projection pair $(\Pi_\ell^{\mathrm{in}},\Pi_\ell^{\mathrm{out}})$ surrounding the grid.
Shared FSQ ties this pair across layers,
$(\Pi_\ell^{\mathrm{in}},\Pi_\ell^{\mathrm{out}})=(\Pi^{\mathrm{in}},\Pi^{\mathrm{out}})$ for all $\ell$, whereas level-specific FSQ learns an independent projection pair for each parallel latent.

\paragraph{QinCo.}
For QinCo, $\Theta_\ell$ is the learnable generator that maps the partial quantized representation to the conditional codebook in Equation~\ref{eq:appendix-qinco}.
For each layer $\ell\geq2$, let $a_{\ell,k}$ be the learned static anchor for index $k$.
The shared variant ties one generator $g$ across layers, whereas the level-specific variant learns an independent generator $g_\ell$:
\begin{equation}
    v_{\ell,k}\!\left(z_q^{(\ell-1)}\right) =
    \begin{cases}
    a_{\ell,k}+[g(z_q^{(\ell-1)})]_k, & \text{shared},\\
    a_{\ell,k}+[g_\ell(z_q^{(\ell-1)})]_k, & \text{level-specific}.
    \end{cases}
    \label{eq:appendix-qinco-sharing}
\end{equation}
Both variants retain layer-specific anchors $\{a_{\ell,k}\}_{k=1}^{K}$, and the first-layer codebook $\mathcal E_1$ is unchanged by the sharing choice.
Thus, \texttt{S} denotes sharing the prefix-to-codebook generator, not using identical realized codebooks at every layer.

Across all five families, the suffixes \texttt{S} and \texttt{L} therefore identify the family-specific parameter block that is shared or learned independently across layers.

Table~\ref{tab:tokenizer-index} lists every tokenizer used in the controlled comparison. The structural identifier and color swatch are held fixed across all main-paper and appendix tables.

\input{tables/preliminaries/tokenizer_design/appendix_tokenizer_index}

\subsection{Rollout Evaluation Protocol}
\label{app:rollout-evaluation}

Our primary results use greedy action-token prediction.
For an evaluation unit with $n$ binary rollout outcomes, let $y_i\in\{0,1\}$ denote the outcome of episode $i$. The reported success rate is
\begin{equation}
\widehat{p}
= \frac{1}{n}\sum_{i=1}^{n} y_i.
\label{eq:evaluation-success-rate}
\end{equation}
To quantify uncertainty, we pool the $n$ binary episode outcomes and, for each bootstrap replicate $b\in\{1,\ldots,B\}$ with $B=10{,}000$, draw $n$ outcomes with replacement and compute the corresponding success rate $\widehat{p}^{*(b)}$. We estimate the bootstrap standard error and the 95\% interval half-width as
\begin{equation}
\widehat{\mathrm{SE}}_{\mathrm{boot}}
= \operatorname{sd}\!\left(
\widehat{p}^{*(1)},\ldots,\widehat{p}^{*(B)}
\right),
\qquad
\Delta_{95}
= 1.96\,\widehat{\mathrm{SE}}_{\mathrm{boot}}.
\label{eq:evaluation-bootstrap}
\end{equation}
Each individual-task or suite result is reported as
$100\widehat{p}\% \mathbin{\pm} 100\Delta_{95}$ percentage points.
Benchmark-level tables macro-average the corresponding task or suite success rates and omit confidence intervals rather than averaging individual intervals.
Execution horizons are evaluated and reported separately.

\paragraph{Robomimic.}
We use the checkpoint with the best teacher-forced validation accuracy and evaluate each task with 10 rollout seeds, with 30 episodes per seed, for a total of 300 episodes per task and execution horizon.
The rollout seeds control environment initialization and are distinct from the policy-training seeds discussed below.

\paragraph{LIBERO.}
Each of the Spatial, Object, Goal, and LIBERO-10 suites contains 10 tasks.
We use the checkpoint with the best teacher-forced validation accuracy.
We evaluate each task for 50 episodes using LIBERO's fixed set of 50 initial states.
This gives 500 episodes per suite and execution horizon.
The benchmark-level cells in Table~\ref{tab:simulation-multitask} equally macro-average the four suite success rates.


\paragraph{RoboTwin.}
We use the checkpoint with the best teacher-forced validation accuracy and evaluate each of the three tasks on 50 random initializations, for a total of 150 episodes per execution horizon.
Table~\ref{tab:rollout-evaluation} summarizes the rollout protocols across the three benchmarks.

\input{tables/preliminaries/training_evaluation/appendix_rollout_protocol}

\subsection{Training Hyperparameters and Randomness}
\label{app:policy-seed-study}

\paragraph{Tokenizer training.}
Table~\ref{tab:tokenizer-training-hyperparameters} summarizes the tokenizer-training hyperparameters used for the three simulation benchmarks.
Our primary goal is to compare downstream policy performance across tokenizer designs and configurations. Because the sweep covers a large number of configurations,
we allocate compute to broader design coverage rather than repeated tokenizer-training seeds within each configuration.
Each tokenizer configuration is therefore trained once. Consequently, our experiments characterize between-design differences but do not estimate within-design variance arising from tokenizer training;
these comparisons should be interpreted accordingly.
\input{tables/preliminaries/tokenizer_design/appendix_tokenizer_training_hyperparameters}

\paragraph{Policy training.}
Table~\ref{tab:policy-training-hyperparameters} summarizes the policy-training hyperparameters used for the three simulation benchmarks.
We estimate policy-training variability using two tokenizer instances, FSQ and NRVQ, with $K=128$ and $L=8$.
For each tokenizer, we train the same multi-task autoregressive policy with 10 policy-training seeds and evaluate every trained policy on Can, Lift, and Square using the same bank of 10 rollout seeds and 30 episodes per rollout seed.
Each run uses the checkpoint with the best teacher-forced validation accuracy.
\input{tables/preliminaries/training_evaluation/appendix_policy_training_hyperparameters}

Table~\ref{tab:policy-seed-robustness} reports the resulting variation in rollout success.
The spread is modest on Can and Lift but larger on Square: the across-training-seed standard deviation is 4.3 percentage points for FSQ and 5.8 points for NRVQ, with ranges of 14.3 and 16.7 points, respectively.

This experiment confirms that policy training exhibits non-negligible randomness, with the magnitude of variation differing across tasks and tokenizers.
At the same time, the observed variability is not so large as to obscure the broader trends that emerge repeatedly across the full tokenizer sweep.
We therefore use this study to contextualize individual pairwise differences, while relying on the large number of comparisons in the sweep to support aggregate conclusions.
Because each configuration in the full sweep is trained with a single policy seed, we interpret the sweep primarily through trends that recur across many comparisons rather than the precise magnitude of any individual pairwise difference.
\input{tables/preliminaries/training_evaluation/appendix_policy_seed_robustness}

\input{tables/preliminaries/training_evaluation/appendix_vla_training_hyperparameters}

\paragraph{VLA architecture, training, and evaluation.}
To test whether our tokenizer-side interventions also benefit a policy built on a pretrained vision-language backbone, we fine-tune SmolVLM2-256M~\citep{Marafioti2025SmolVLM} with an ActionCodec-style Llama action expert~\citep{Dong2026ActionCodec} on LIBERO.
The action expert autoregressively predicts eight tokens, which the corresponding action tokenizer decodes into an action chunk.
We evaluate eight paired tokenizer configurations (Table~\ref{tab:vla-libero-results}).
All sixteen policies use the same VLA architecture, optimization recipe, and evaluation protocol; only the action tokenizer and its resulting token targets differ.
The VLA is fine-tuned end to end without LoRA.
Although all model parameters are included in optimization, $3.3$M parameters in the final VLM decoder layer receive no gradient because the VLM supplies only its keys and values to the action expert.

Table~\ref{tab:vla-training-hyperparameters} reports the complete shared configuration. Note that we report results from the final training checkpoint after 50k gradient steps. Due to computational constraints, we were not able to train the VLA to convergence. Therefore, our results highlight the strong performance of ProAct in the limited training budget regime.

\section{Additional Action-Tokenization Analysis}
\label{app:additional-action-tokenization-analysis}

This section gives the measurements and analysis behind Section~\ref{sec:what-matters}: reconstruction error (Appendix~\ref{app:reconstruction-error-details}), nominal bitrate (Appendix~\ref{app:nominal-bitrate-analysis}), chunk size (Appendix~\ref{app:chunk-size-details}), training-support membership (Appendix~\ref{app:training-set-diagnostics}), and the one-step error decomposition (Appendix~\ref{app:one-step-error-proof}).
\paragraph{Conventions.}
A \emph{single-task} tokenizer--policy pair is trained on demonstrations from one Robomimic task and evaluated on that same task, whereas \emph{multi-task} denotes joint training on a dataset of all tasks.
Each subsection states which condition it uses, and Appendix~\ref{app:one-step-error-proof} is analytical and does not depend on the choice.
Shaded bands and horizontal spans show min--max ranges over tokenizer designs unless labeled as 95\% intervals.

\subsection{Reconstruction-Error Details}
\label{app:reconstruction-error-details}


\subsubsection{Does Reconstruction Error Rank Tokenizers?}
\label{app:recon-ranking}

\begin{figure}[h]
    \centering
    \input{figures/what_matters/reconstruction_error/appendix_reconstruction_success_by_family.tex}
    \caption{Validation reconstruction error ($\mathcal E_{\mathrm{rec}}$) and downstream rollout success for the complete single-task tokenizer--policy sweep, separated by quantizer family (rows: VQ, RVQ, NRVQ, FSQ, QinCo) and task (columns: Can, Lift, Square). Each point is one valid run; all panels share the same axes, and each column sums to the 375/242/232 Can/Lift/Square totals reported above. Bottom-right annotations report Spearman's $\rho$ for all runs / runs with validation MAE $<0.04$ (excluding tokenizers with large reconstruction error).}
    \label{fig:appendix-reconstruction-success-scatter}
\end{figure}
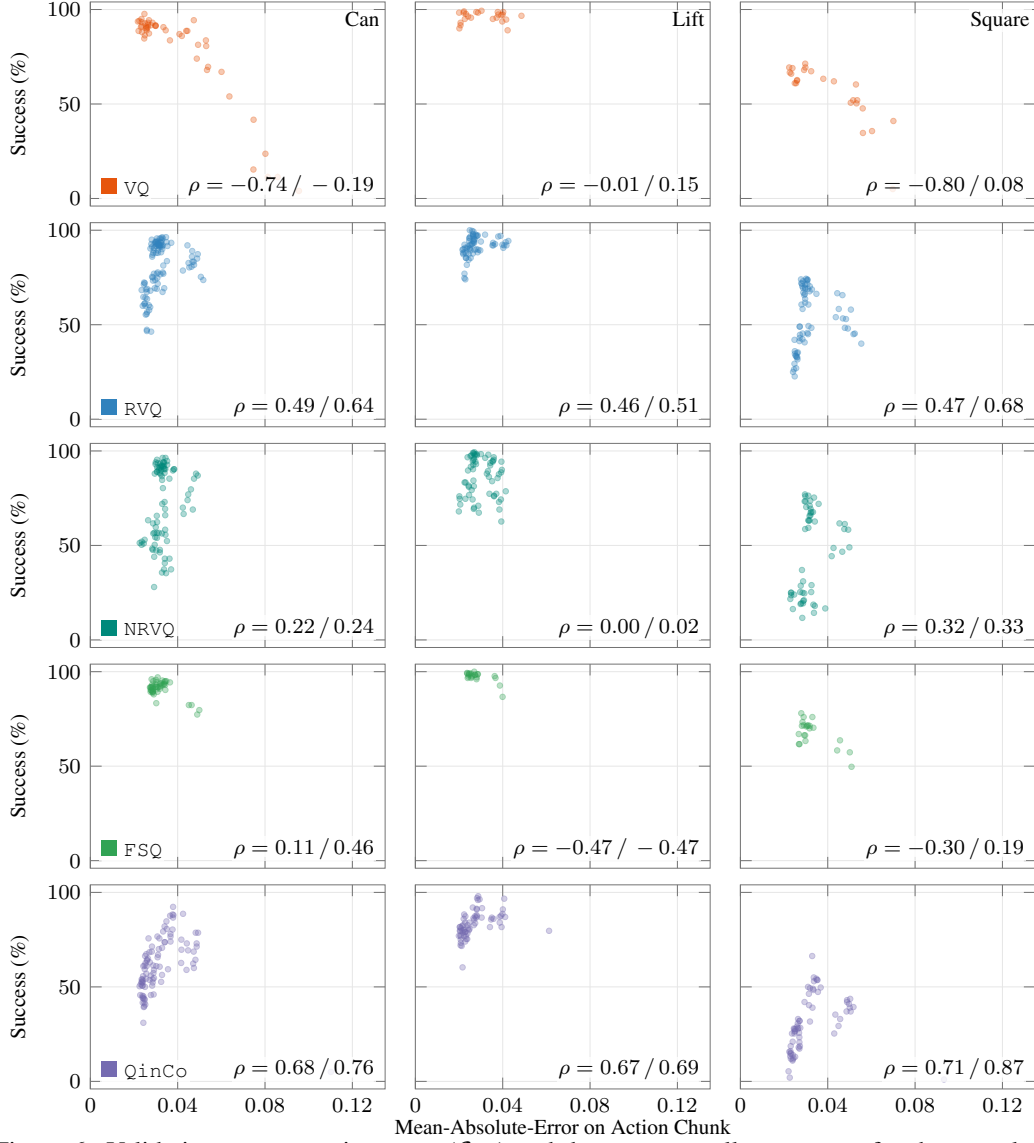

We ask when reconstruction error ceases to rank downstream performance across tokenizer--policy configurations.
The correlation study in Figure~\ref{fig:appendix-reconstruction-success-scatter} retains every valid run in the single-task sweep: 375 for Can, 242 for Lift, and 232 for Square, for 849 tokenizer--policy runs in total. The cohort spans all five quantizer families, multiple tokenizer backbones and training variants, both codebook-sharing choices where applicable, and two nominal rates. With codebook size 128, 123 runs use three levels (21 bits per chunk) and 726 use eight levels (56 bits per chunk). Every row has a unique tokenizer, a matching same-task policy with no recorded training issue, a valid best checkpoint, and an evaluation over 10 rollout seeds and 300 episodes. We therefore apply no filtering based on tokenizer family, bitrate, reconstruction error, or rollout success: high-error and low-success runs are both necessary for characterizing the relationship. Within the RVQ, NRVQ, and QinCo subset discussed in Section~\ref{sec:reconstruction-threshold}, the taskwise Spearman correlations between validation reconstruction error and rollout success are $\rho=0.45$, $0.35$, and $0.50$ for Can, Lift, and Square, respectively.

\subsubsection{Calibrating Action-Error Tolerance}
\label{app:recon-tolerance}

We calibrate action-space sensitivity using an observation-conditioned continuous FM policy with $H=H_{\mathrm{exec}}=8$ for each task separately, treating its clean prediction as the expert action reference for this diagnostic.
At every step of each predicted chunk, we independently perturb the three translation channels with $\eta_j^{\mathrm{trans}}\sim\mathcal U(-\epsilon_{\mathrm{trans}},\epsilon_{\mathrm{trans}})$ and the three rotation channels with $\eta_j^{\mathrm{rot}}\sim\mathcal U(-\epsilon_{\mathrm{rot}},\epsilon_{\mathrm{rot}})$.
The gripper is left unchanged, and the perturbed action is clipped to its valid range before execution.
Viewed as a reconstruction of the clean FM action, the perturbed action incurs a reconstruction-form error whose magnitude we control directly.
Since a variable drawn from $\mathcal U(-\epsilon,\epsilon)$ has expected absolute magnitude $\epsilon/2$, averaging the six perturbed channels and the unchanged gripper channel gives the following expected seven-dimensional MAE before clipping:
\begin{equation}
    \mathbb E\!\left[\frac{\lVert\eta\rVert_1}{7}\right]
    =\frac{3}{14}\left(\epsilon_{\mathrm{trans}}+\epsilon_{\mathrm{rot}}\right).
    \label{eq:fm-noise-calibration}
\end{equation}
This quantity measures deviation from the clean FM expert in the same raw-action units and with the same MAE form as $\ell$; it is not the tokenizer's validation reconstruction error $\mathcal E_{\mathrm{rec}}$.

We use two complementary perturbation settings.
The \emph{joint-noise} sweep sets $\epsilon_{\mathrm{trans}}=\epsilon_{\mathrm{rot}}=\epsilon$, for which Equation~\ref{eq:fm-noise-calibration} reduces to $3\epsilon/7$, and measures the overall tolerance to pose-action error.
The \emph{independent-noise} sweep varies $\epsilon_{\mathrm{trans}}$ and $\epsilon_{\mathrm{rot}}$ separately to distinguish translational from rotational sensitivity.
We first present the joint-noise calibration across Robomimic, LIBERO, and RoboTwin, followed by the independent-noise sweep on Robomimic.

\clearpage
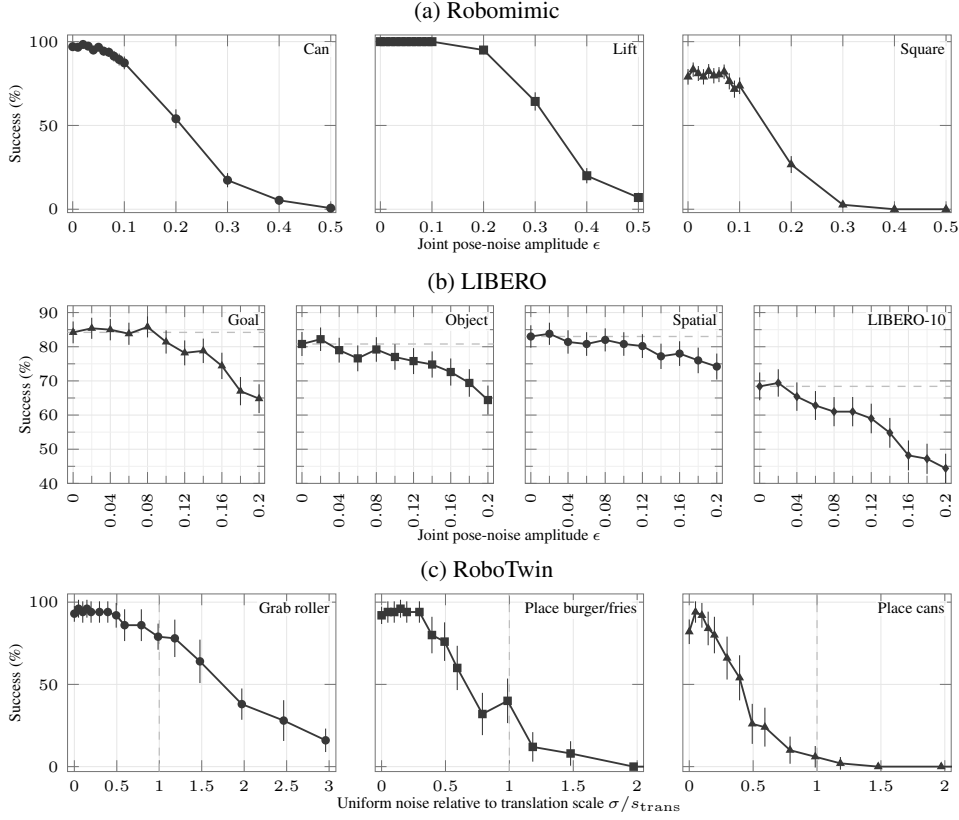
\begin{figure}[H]
    \centering
    \captionsetup[subfigure]{position=top,skip=2pt,justification=centering}
    \begin{subfigure}{\linewidth}
        \centering
        \caption{Robomimic}\label{fig:fm-noise-diagnostic}
        \input{figures/what_matters/reconstruction_error/appendix_flow_matching_noise_response.tex}
    \end{subfigure}\par\vspace{3pt}
    \begin{subfigure}{\linewidth}
        \centering
        \caption{LIBERO}\label{fig:libero-fm-noise}
        \input{figures/what_matters/reconstruction_error/appendix_libero_fm_noise_response.tex}
    \end{subfigure}\par\vspace{3pt}
    \begin{subfigure}{\linewidth}
        \centering
        \caption{RoboTwin}\label{fig:robotwin-fm-noise}
        \input{figures/what_matters/reconstruction_error/appendix_robotwin_fm_noise_response.tex}
    \end{subfigure}
    \caption{FM response to joint uniform pose-action noise across simulation benchmarks, with one panel per task or suite. Translation and rotation use equal noise amplitudes, while gripper coordinates remain unperturbed. Markers show success and cap-free bars show 95\% intervals. (a) Robomimic: 300 episodes per point; the fine sweep covers $\epsilon=0.01$--$0.10$ and the coarse sweep $\epsilon=0.20$--$0.50$. (b) LIBERO: 500 episodes per point at $\epsilon=0$--$0.20$. (c) RoboTwin: $H=64$ and $H_{\mathrm{exec}}=8$; both arms receive pose noise from $\mathcal U(-\sigma,\sigma)$. The horizontal axis normalizes $\sigma$ by the translation-channel standard deviation $s_{\mathrm{trans}}=0.00507$; dashed lines mark $\sigma=s_{\mathrm{trans}}$. Clean references use 100 episodes per task and noisy points use 50, except the three original Grab roller points at $\sigma\in\{0.005,0.010,0.015\}$, which use 100.}
    \label{fig:fm-joint-noise}
\end{figure}

In the Robomimic joint-noise sweep (Figure~\ref{fig:fm-noise-diagnostic}), success remains near the clean anchors through $\epsilon=0.07$: 93.7\% versus 97.0\% on Can, 100\% versus 100\% on Lift, and 82.0\% versus 79.0\% on Square.
At $\epsilon=0.30$, success falls to 17.3\%, 64.3\%, and 2.7\%, respectively, confirming that the tolerance scale is task dependent.

The LIBERO joint-noise sweep (Figure~\ref{fig:libero-fm-noise}) shows the same qualitative pattern across four suites: sufficiently large action noise degrades success, but the onset and rate of degradation differ by suite.

RoboTwin uses the same joint-noise design, with perturbations drawn independently from $\mathcal U(-\sigma,\sigma)$.
Its 14-dimensional dual-arm action contains 12 perturbed pose coordinates and two unchanged gripper coordinates, so the expected pre-clipping MAE is $3\sigma/7$.
Because the empirical standard deviations of the translation and rotation coordinates are $s_{\mathrm{trans}}=0.00507$ and $s_{\mathrm{rot}}=0.02957$, respectively, we report the noise level relative to the translation scale, $\sigma/s_{\mathrm{trans}}$.

The fine-grained RoboTwin sweep (Figure~\ref{fig:robotwin-fm-noise}) reveals a task-dependent plateau followed by a smooth roll-off.
At $\sigma/s_{\mathrm{trans}}\approx0.20$, success is 94\%, 94\%, and 80\% on Grab roller, Place burger/fries, and Place cans, respectively, compared with clean references of 93\%, 92\%, and 82\%.
Linear interpolation places the half-success points at $\sigma/s_{\mathrm{trans}}=1.81$, $0.69$, and $0.44$, a fourfold range across the three tasks.

Across Robomimic, LIBERO, and RoboTwin, the task-dependent onset and rate of degradation rule out a single universal action-error threshold.

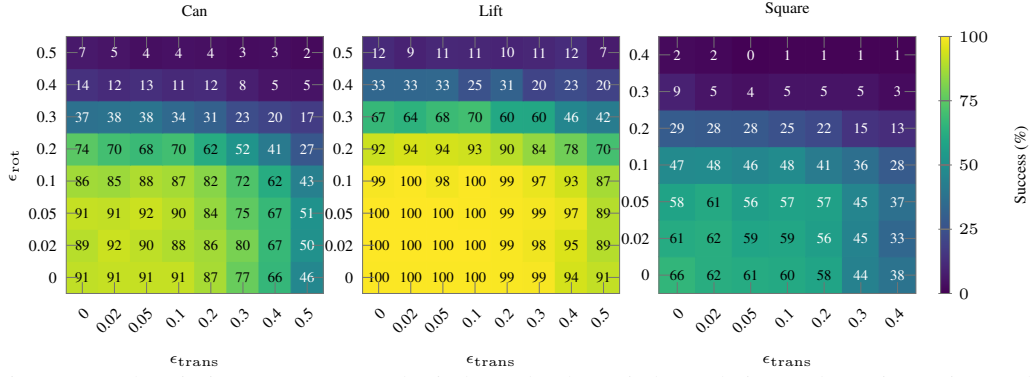
\begin{figure}[H]
    \centering
    \input{figures/what_matters/reconstruction_error/appendix_fm_noise_precision_heatmap.tex}
    \caption{Robomimic FM success under independently varied translation and rotation noise. Cells report success over 300 episodes at $H=H_{\mathrm{exec}}=8$, with the gripper unperturbed; the \textsc{Square} grid omits $\epsilon=0.5$. These runs use a different FM policy from Figure~\ref{fig:fm-noise-diagnostic}, so sensitivity patterns, but not absolute success levels, are comparable between the two diagnostics.}
    \label{fig:fm-noise-precision}
\end{figure}

By varying the two amplitudes separately, Figure~\ref{fig:fm-noise-precision} isolates their effects within each FM policy.
At comparable amplitudes, rotation noise produces the sharper degradation, though the response surface remains strongly task dependent.

\subsubsection{Isolating Tokenizer Distortion}
\label{app:recon-isolation}

The action-noise calibration establishes the magnitude of raw-action error that begins to impair control. We next ask whether reconstruction distortion introduced by the tokenizer alone can reach this consequential regime.
To isolate this effect from token prediction, we pass each action chunk $A^{\mathrm{FM}}$ produced by the FM policy through a frozen action tokenizer and detokenizer before execution, without training or invoking an AR policy.
We measure the resulting distortion using the on-policy reconstruction MAE,
$\mathbb E[\ell(A^{\mathrm{FM}},\mathcal D(\mathcal T(A^{\mathrm{FM}})))]$,
including the gripper coordinate.

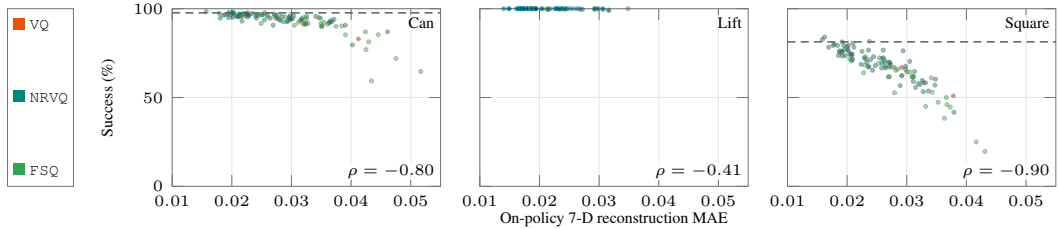
\begin{figure}[H]
    \centering
    \input{figures/what_matters/reconstruction_error/appendix_tokenized_fm_reconstruction_success.tex}
    \caption{Tokenized-FM success versus on-policy seven-dimensional reconstruction MAE. Each point is one tokenizer condition; color denotes quantizer family.
    Dashed lines mark the available un-tokenized FM references for Can and Square. Spearman correlations are shown in-panel. All results use $H=H_{\mathrm{exec}}=8$ and 300 episodes.}
    \label{fig:tokenized-fm-reconstruction}
\end{figure}

Figure~\ref{fig:tokenized-fm-reconstruction} reports all 273 tokenized conditions.
Reconstruction error is strongly negatively associated with success on Can and Square, and more weakly on Lift, where success remains near ceiling across much of the observed error range.
Because token prediction is absent, these results directly characterize how tokenizer-induced reconstruction distortion affects control.


\subsubsection{Separating Distortion from Token Prediction}
\label{app:recon-vs-prediction}

\begin{figure}[H]
    \centering
    \input{figures/what_matters/reconstruction_error/appendix_tokenized_fm_vs_policy_success.tex}
    \caption{Tokenized-FM rollout success versus tokenizer--AR policy rollout success for the same tokenizer. Each point is one tokenizer configuration: color denotes quantizer family, filled squares mark \texttt{TF-FSQ-S}, circles \texttt{TF-NRVQ-S}, and triangles \texttt{TF-NRVQ-L}. The \texttt{TF-NRVQ} points span latent dimensions 32, 64, and 128, and both families span 21--80 bits per chunk. The dashed line marks equal success. Both axes use $H=H_{\mathrm{exec}}=8$, multi-task training, and 300 episodes per condition; the panels contain the 63 tokenizers for which both evaluations are available.}
    \label{fig:tokenized-fm-vs-policy}
\end{figure}
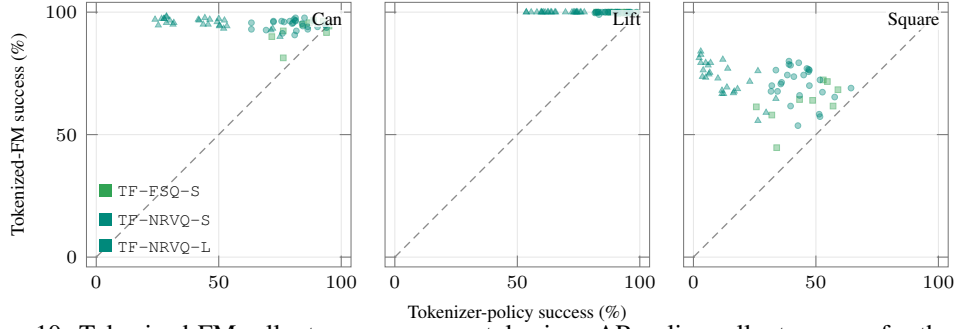

The tokenized-FM policy isolates reconstruction distortion, whereas the tokenizer--AR policy additionally has to predict the token sequence. Pairing the two evaluations on the same tokenizer therefore separates the two error sources.
Figure~\ref{fig:tokenized-fm-vs-policy} does this for the 63 tokenizers evaluated under both policies: the 9 \texttt{TF-FSQ-S} configurations of the bitrate sweep together with its 27 \texttt{TF-NRVQ-S} and 27 \texttt{TF-NRVQ-L} configurations.

Removing token prediction shifts nearly every tokenizer above the diagonal, and the two axes span very different ranges: tokenizer--AR success varies widely across these tokenizers, whereas tokenized-FM success stays within a comparatively narrow high band on each task. Reconstruction distortion alone therefore does not account for the spread in tokenizer--AR performance, which instead reflects the difficulty of predicting the token sequence.

Taken together, the four diagnostics support reconstruction as a feasibility condition: sufficiently large distortion degrades control,
yet once reconstruction error enters a sufficiently low regime, further reductions do not reliably rank downstream control performance.

\subsection{Nominal Bitrate Analysis}
\label{app:nominal-bitrate-analysis}

Higher nominal bitrate lowers reconstruction error, so it is tempting to maximize it.
Section~\ref{sec:reconstruction-threshold} showed that reconstruction only needs to be accurate enough, so higher bitrate need not improve control.
We test this at fixed chunk size: higher bitrate generally improves reconstruction, while downstream success remains non-monotonic and can even decline.

\paragraph{Nominal bitrate bounds capacity, not encoded information.}
For an $L$-level tokenizer with $K$ choices per level, $B_{\mathrm{nom}}=L\log_2K$ is the log-cardinality of the code space: the most information a chunk's token sequence could carry.
Writing $C_{1:L}$ for that sequence, the information it does carry obeys
\[
\ent(C_{1:L}) \;\leq\; L\log_2K \;=\; B_{\mathrm{nom}},
\]
This gap is common in practice, since discrete representation learning substantially under-utilizes a nominal codebook~\citep{Mentzer2024FSQ}.
Enlarging the code space enlarges the set of achievable rate--distortion points, so reconstruction should improve with $B_{\mathrm{nom}}$.
Whether that added capacity is used, and whether the code it yields is easier or harder for an autoregressive policy to predict, is not determined by $B_{\mathrm{nom}}$: excess action-token capacity can absorb high-frequency variation that complicates autoregressive learning, whereas compression can strip temporal redundancy and leave more informative prediction targets~\citep{Dong2026ActionCodec,Pertsch2025FAST}.
The rest of this subsection measures the effect on reconstruction and on downstream success separately.

\begin{figure}[!t]
    \centering
    \begingroup
    \renewcommand{\tokenizerid}[2]{%
        \mbox{\textcolor{#1}{\rule{0.72em}{0.72em}}\hspace{0.35em}\textsc{#2}}%
    }
    \input{figures/what_matters/bitrate_chunking/appendix_bitrate_success_reconstruction}
    \endgroup
    \caption{Robomimic full-decode bitrate sweep at fixed $H=H_{\mathrm{exec}}=8$. The upper panel shows tokenizer validation reconstruction MSE in raw action units, which is a property of the tokenizer and therefore shared across tasks; the lower panels show the corresponding taskwise rollout success over 300 episodes. Nominal bitrate is $L\log_2K$ for $K\in\{128,512,1024\}$ and $L\in\{3,6,8\}$. Each point is one tokenizer configuration, so no averaging or design range is involved. All tokenizers use Transformer backbones (TF); S/L denote shared and level-specific codebooks (Appendix~\ref{app:shared-level-specific}). \textsc{TF-FSQ-S} has a single configuration per bitrate.}
    \label{fig:bitrate-reconerr}
\end{figure}
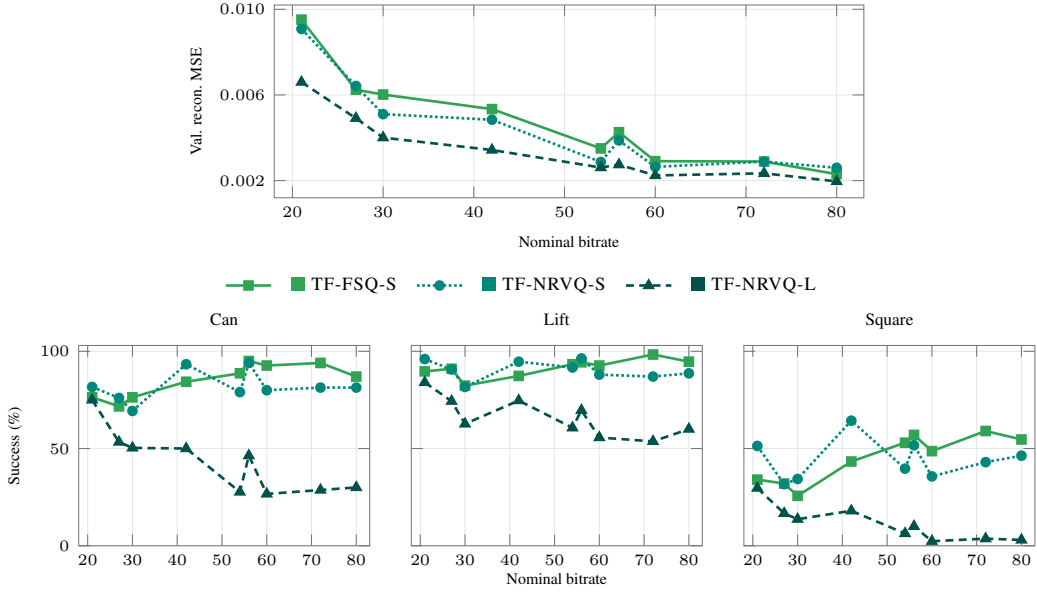

\paragraph{Setup.}
Figure~\ref{fig:bitrate-reconerr} sweeps $B_{\mathrm{nom}}$ over the full grid $K\in\{128,512,1024\}$ and $L\in\{3,6,8\}$, giving nine rates from 21 to 80 bits per chunk, for the Transformer-backbone \textsc{TF-FSQ} and \textsc{TF-NRVQ} tokenizers.
The chunk size is fixed at $H=H_{\mathrm{exec}}=8$ throughout, so bitrate is the only design factor that varies; each point is a single tokenizer configuration paired with a multi-task AR policy and evaluated over 300 rollout episodes per task.
We exclude the prefix-only partial-decode runs here, because they change the deployed bitrate without recomputing the tokenizer's offline validation reconstruction error, which would leave the two axes of Figure~\ref{fig:bitrate-reconerr} measured at different rates.
The upper panel reports validation reconstruction MSE in raw action units; this is not the MAE $\mathcal E_{\mathrm{rec}}$ of Section~\ref{sec:reconstruction-threshold}, so its magnitudes are not comparable to the tolerance scale calibrated there.
Every policy in this sweep was trained with horizontal image flipping enabled, so comparisons within the sweep hold, while absolute success levels are not comparable to the other tables in this paper.

\paragraph{Effect on reconstruction.}
Across \textsc{TF-FSQ-S}, \textsc{TF-NRVQ-S}, and \textsc{TF-NRVQ-L}, validation reconstruction MSE generally decreases as nominal bitrate increases, with only small local reversals between nearby rates.
Greater representational capacity therefore usually improves reconstruction fidelity, as expected from rate--distortion behavior.

\paragraph{Effect on downstream success.}
Downstream success does not follow the same monotonic trend.
\textsc{TF-FSQ-S} generally improves with bitrate but peaks before the largest rate, \textsc{TF-NRVQ-S} fluctuates without a consistent direction, and \textsc{TF-NRVQ-L} performs best at the lowest tested rate on every task even though it has the lowest reconstruction error of the three tokenizers at every rate.
Thus, additional bitrate can improve reconstruction without reliably improving control, and the downstream effect depends on the tokenizer design.

\paragraph{Interaction with chunk size.}
Bitrate must also be read relative to the chunk size $H$, since a fixed $B_{\mathrm{nom}}$ allocates only $B_{\mathrm{nom}}/H$ bits per action step, or $B_{\mathrm{nom}}/(Hd_a)$ per scalar action dimension.
Longer chunks consequently need more capacity to hold reconstruction fidelity fixed, which is what Appendix~\ref{app:chunk-size-details} measures.
Raising the bitrate relieves that representational bottleneck, yet it does not address the separate difficulty of predicting a longer action sequence or executing it open-loop, and the results above show that it does not reliably help even at fixed $H$.
Hence, bitrate and chunk size should not simply be maximized.
Bitrate is best treated as a \emph{conditional capacity requirement}: for a chosen $H$ it must be large enough to enter the feasible reconstruction regime of Section~\ref{sec:reconstruction-threshold}; beyond that point, $H$ and $H_{\mathrm{exec}}$ should be chosen by downstream success and inference cost.

\subsection{Chunk-Size Details}
\label{app:chunk-size-details}

Chunking introduces two distinct effects. Increasing the prediction horizon $H$ requires the policy to predict farther into the future and, for a tokenizer with fixed capacity, compresses more action steps into the same representation. Increasing the execution horizon $H_{\mathrm{exec}}$ instead reduces replanning frequency and increases the duration of open-loop execution.

We separate these effects by first sweeping $H$ and $H_{\mathrm{exec}}$ with continuous FM policies, which isolate the prediction- and execution-horizon effects without tokenization (Appendix~\ref{app:chunk-size-fm}).
We then repeat the sweep with the tokenizer--AR policy (Appendix~\ref{app:chunk-size-tokenizer}) and compare the two settings directly.
The results show that $H_{\mathrm{exec}}$ exhibits an intermediate optimum in both settings, while increasing $H$ introduces an additional cost for tokenized policies once reconstruction leaves the feasible regime identified in Section~\ref{sec:reconstruction-threshold}.
\subsubsection{Chunk Size with Flow-Matching Policies}
\label{app:chunk-size-fm}

Because FM policies predict action chunks without tokenizing them, no horizon effect measured here can be attributed to a tokenizer.
All three benchmarks show the same underlying shape in $H_{\mathrm{exec}}$: success is depressed when the policy replans almost every step, rises to an interior optimum, and falls again as execution approaches the full predicted chunk.
The swept ranges show different parts of this shape: Robomimic shows both the rise and the fall, LIBERO mainly the fall, and RoboTwin mainly the rise.

\paragraph{Evaluation protocol.}
We evaluate FM policies on Robomimic, LIBERO, and RoboTwin across the available prediction horizons $H$ and execution horizons $H_{\mathrm{exec}}$, and compare their rollout success using the same evaluation protocol as the tokenized policies.

\begin{figure}[!t]
    \centering
    \captionsetup[subfigure]{position=top,skip=2pt,justification=centering}
    \begin{subfigure}{\linewidth}
        \centering
        \caption{Robomimic}\label{fig:robomimic-fm-exec}
        \input{figures/what_matters/bitrate_chunking/appendix_robomimic_fm_exec_sweep}
    \end{subfigure}\par\vspace{3pt}
    \begin{subfigure}{\linewidth}
        \centering
        \caption{LIBERO}\label{fig:libero-fm-exec}
        \input{figures/what_matters/bitrate_chunking/appendix_libero_fm_exec_sweep}
    \end{subfigure}\par\vspace{3pt}
    \begin{subfigure}{\linewidth}
        \centering
        \caption{RoboTwin, $H=64$}\label{fig:robotwin-fm-exec}
        \input{figures/what_matters/bitrate_chunking/appendix_robotwin_fm_exec_sweep.tex}
    \end{subfigure}
    \caption{FM success across prediction horizon $H$ and execution horizon $H_{\mathrm{exec}}$, with one panel per task or suite. Bars show 95\% bootstrap intervals, and curves terminate at $H_{\mathrm{exec}}=H$. The shared color legend applies to (a--b), which use the last checkpoint. (a) Robomimic: 300 episodes per point. (b) LIBERO: 500 episodes per suite, except the hollow LIBERO-10 points at $(H,H_{\mathrm{exec}})=(32,2)$ and $(64,2)$, with 400 and 300 completed episodes, respectively. (c) RoboTwin: fixed $H=64$; evaluations at $H_{\mathrm{exec}}\in\{8,16,32,64\}$ use 100 episodes per task, and the remaining horizons use 50. Episode limits are 300, 500, and 600 environment steps for Grab roller, Place burger/fries, and Place cans, respectively.}
    \label{fig:fm-execution-horizons}
\end{figure}
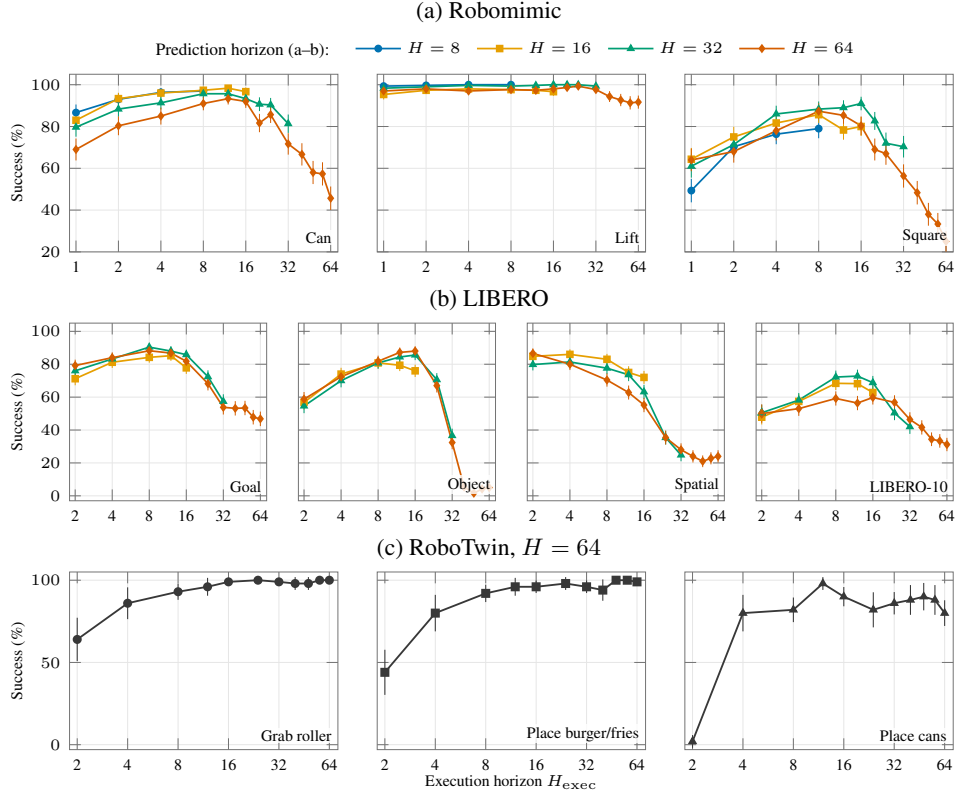

\paragraph{Robomimic.}
On Robomimic, increasing $H_{\mathrm{exec}}$ from one step consistently helps until a task- and horizon-dependent intermediate range, whereas executing most or all of a long chunk degrades success.
After macro-averaging Can, Lift, and Square, the best execution horizon is 8 for $H\in\{8,16\}$, 12 or 16 for $H=32$, and 8 or 12 for $H=64$.
For $H=64$, macro-average success falls from 92.0\% at $H_{\mathrm{exec}}\in\{8,12\}$ to 54.1\% when the entire predicted chunk is executed.
Success therefore both rises and falls with $H_{\mathrm{exec}}$, but the optimum is close to the short end, so the rise spans only a few steps.

\paragraph{LIBERO.}
LIBERO exhibits the same broad preference for moderate execution horizons, but the optimum is suite dependent.
The four-suite macro-average peaks at $H_{\mathrm{exec}}=8$ for all three prediction horizons, reaching 79.1\%, 80.2\%, and 74.9\% for $H=16$, 32, and 64, respectively.
Long open-loop execution is especially damaging for Object and Spatial; at $H=64$, the macro-average declines to 26.8\% at $H_{\mathrm{exec}}=64$.
The short-horizon end is the least completely evaluated part of this grid, since both hollow LIBERO-10 points in Figure~\ref{fig:libero-fm-exec} sit at $H_{\mathrm{exec}}=2$; we therefore do not read a replanning penalty into LIBERO on its own.


\paragraph{RoboTwin.}
RoboTwin shows the strongest benefit from increasing the execution horizon.
Increasing $H_{\mathrm{exec}}$ from 2 to 4 raises success from 64\% to 86\% on Grab Roller, from 44\% to 80\% on Place Burger and Fries, and from 2\% to 80\% on Place Cans.
Performance remains high once $H_{\mathrm{exec}}$ reaches roughly 12, with the first two tasks near ceiling and Place Cans mostly between 80\% and 90\%.
Thus, as in Robomimic, overly frequent replanning is detrimental, but the effect is substantially stronger on RoboTwin.

\paragraph{Implication for tokenizer policies.}
The tokenizer-free sweeps show that a large prediction horizon is not inherently harmful when paired with a moderate execution horizon: even at $H=64$, FM reaches 92.0\% on Robomimic and 74.9\% on LIBERO.
The degradation observed at large $H$ in the tokenized setting therefore cannot be explained by prediction horizon alone.

\subsubsection{Chunk Size with Tokenizers and Tokenizer--AR Policies}
\label{app:chunk-size-tokenizer}

\paragraph{Evaluation protocol.}
We evaluate all valid $(H,H_{\mathrm{exec}})$ pairs for $H\in\{4,8,16,32\}$ using Reconstruction-only tokenizers and matching single-task policies. To make horizon comparisons insensitive to failed tokenizer runs, each task uses the intersection of configurations valid at every $H$: 21 for Can, 22 for Lift, and 20 for Square. All success values use the best policy checkpoint, greedy decoding, and 300 episodes per tokenizer--task condition. Reconstruction error in this subsubsection is always the validation mean of $\ell$ between decoded and target chunks in raw action space rather than latent or code space. Two features of this protocol differ from the FM sweeps above, which report one policy per value and use the last checkpoint: every value here averages 20--22 tokenizer configurations, and checkpoints are selected by validation accuracy.

\begin{figure}[p]
    \centering
    \begin{subfigure}[t]{\linewidth}
        \centering
        \input{figures/what_matters/bitrate_chunking/appendix_reconstruction_by_task.tex}
        \vspace{-0.35em}
        \begin{center}
            \pgfplotslegendfromname{chunkappendixfamilylegend}
        \end{center}
        \caption{Task-specific family means and tokenizer-design ranges.}
    \end{subfigure}
    \vspace{0.25em}
    \begin{subfigure}[t]{\linewidth}
        \centering
        \input{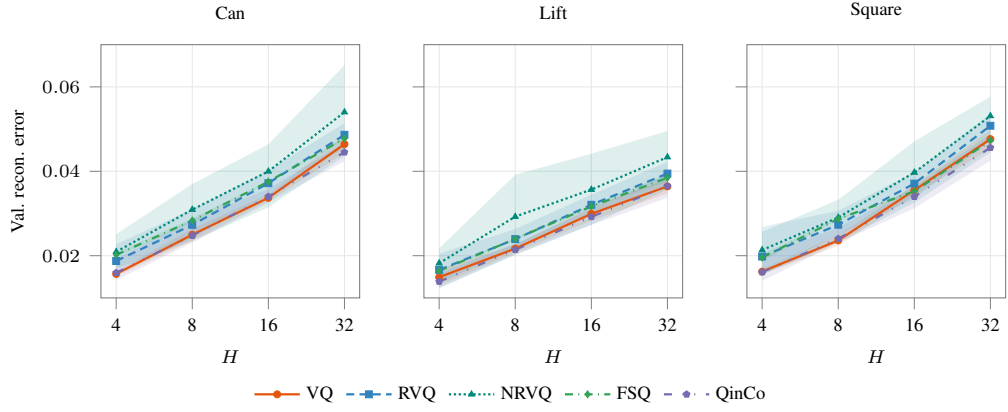}
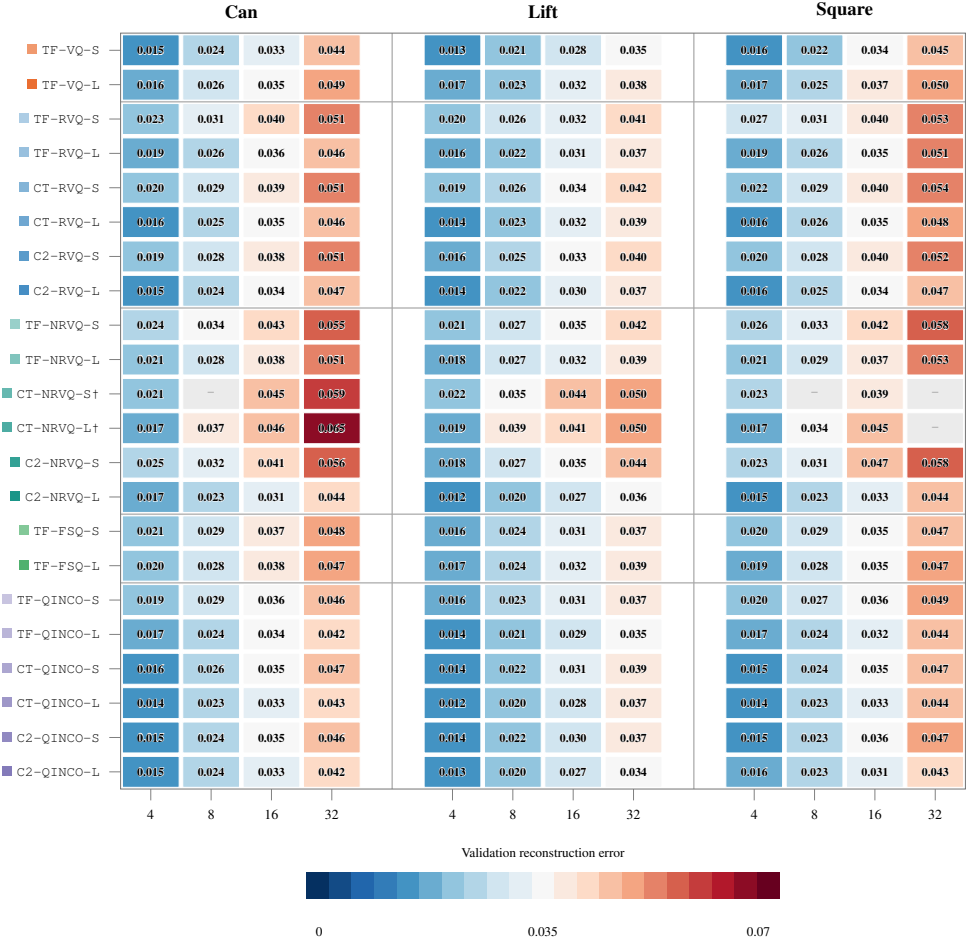
        \caption{Tokenizer-level validation reconstruction errors.}
    \end{subfigure}
    \caption{Detailed validation-reconstruction behavior as the chunk size increases. \textbf{(a)} Color identifies quantizer family, with a distinct line style and marker shape per family, and shaded regions show min--max variation over valid tokenizer designs. \textbf{(b)} Rows follow the tokenizer ordering in Appendix~\ref{app:tokenizer-index}; all task blocks share a blue-to-red scale from 0 to 0.07, with red denoting larger validation reconstruction error. A dagger marks configurations excluded from the globally matched main-text aggregation, and gray ``--'' cells denote above-threshold tokenizer runs whose validation errors are unavailable.}
    \label{fig:chunk-reconstruction-details}
\end{figure}

\paragraph{Reconstruction difficulty at larger $H$.}
We first examine whether increasing $H$ makes action chunks harder to reconstruct under a fixed token budget.
This tokenizer-specific effect could explain any additional degradation beyond the trends already observed in the tokenizer-free sweeps.
Figure~\ref{fig:chunk-reconstruction-details} reports task-specific family summaries together with the validation reconstruction error of every tokenizer configuration.
Family curves average the configurations available for each task, with min--max ranges across designs.

\clearpage

Validation reconstruction error rises systematically with chunk size. Across the multi-task tokenizer set, the mean increases from 0.0168 at $H=4$ to 0.0247, 0.0328, and 0.0437 at $H=8,16,32$, respectively. The single-task family curves show the same broad trend despite variation across tokenizer designs and tasks.
Only the largest chunk size leaves the regime that Section~\ref{sec:reconstruction-threshold} identifies as feasible: $H=32$ is the first horizon whose mean validation reconstruction error, 0.0437, exceeds the ${\approx}0.04$ Robomimic tolerance, whereas $H\in\{4,8,16\}$ all remain inside it.

\begin{figure}[!t]
    \centering
    \input{figures/what_matters/bitrate_chunking/appendix_success_colorbar.tex}\par
    \vspace{2pt}
    \def\chunkPlotHeight{3.25cm}
    \begin{tikzpicture}
        \def\chunkPlotWidth{7.65cm}
        \def\chunkPlotName{single}\def\chunkPlotX{0.65cm}
        \input{figures/what_matters/bitrate_chunking/appendix_success_by_task}
        \def\chunkPlotWidth{3.25cm}
        \def\chunkPlotName{multi}\def\chunkPlotX{9.95cm}
        \input{figures/what_matters/bitrate_chunking/appendix_success_multitask}
        \node[anchor=north,inner sep=0pt,text width=7.65cm]
            at ([yshift=-1.38cm]single.south)
            {\subcaption{Matching single-task policies.}};
        \node[anchor=north,inner sep=0pt,text width=4cm]
            at ([yshift=-1.38cm]multi.south)
            {\subcaption{Multi-task policies.}};
    \end{tikzpicture}\unskip
    \caption{Detailed downstream success across action and execution horizons. \textbf{(a)} Each row averages the fixed valid tokenizer set for one task. Columns are grouped by increasing execution horizon $H_{\mathrm{exec}}$ and, within each group, by increasing chunk size $H$; labels are shown as $H/H_{\mathrm{exec}}$ and the vertical rules separate the $H_{\mathrm{exec}}$ groups. \textbf{(b)} Each cell averages the 20 configurations valid at every multi-task horizon and then macro-averages Can, Lift, and Square. Both panels use a common Viridis scale from 0 to 100\% success; numbers inside cells are percentages.}
    \label{fig:chunk-success-details}
\end{figure}
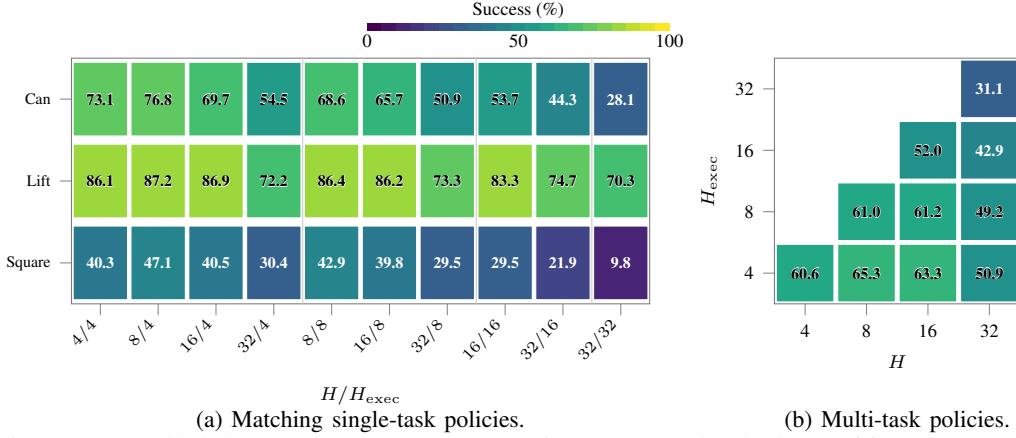

\paragraph{Downstream effects of $H$ and $H_{\mathrm{exec}}$.}
Downstream success turns over at the same chunk size. Figure~\ref{fig:chunk-success-details} separates the aggregate in Figure~\ref{fig:chunksize-executionhorizon} by task and adds the corresponding multi-task sweep.

Increasing the prediction horizon is not uniformly beneficial.
Across both execution horizons, performance is relatively stable from $H=8$ to $H=16$ but drops substantially at $H=32$: mean success decreases from 63.3\% to 50.9\% at $H_{\mathrm{exec}}=4$, and from 61.2\% to 49.2\% at $H_{\mathrm{exec}}=8$.
This degradation coincides with reconstruction error leaving the low-error regime identified in Section~\ref{sec:reconstruction-threshold}, suggesting that longer prediction horizons become harmful once they make the action chunk sufficiently difficult to reconstruct.

\subsection{Training-Support Membership Diagnostics}
\label{app:training-set-diagnostics}

We ask two questions: how often the policy's token sequences leave the empirical training support, and whether restricting decoding to observed sequences helps.

\paragraph{Setup.}
Every diagnostic in this subsection uses the \emph{single-task} condition at $H=H_{\mathrm{exec}}=8$: each of the 22 tokenizer configurations is paired with a policy trained and evaluated on the same Robomimic task, giving 66 tokenizer--task conditions, and the offline and online analyses below draw on the identical cohort.
Let $\mathcal C_{\mathrm{train}}=\{\mathcal T(A):A\in\mathcal A_{\mathrm{train}}\}$ be the finite set of encoded training chunks, which we call the empirical training support; note that $\mathcal A_{\mathrm{train}}$ is the training set of action chunks and $\mathcal C_{\mathrm{train}}$ its encoding, so membership below always refers to the latter.
To measure frequency, we record whether each generated prefix occurs among prefixes of the same length in $\mathcal C_{\mathrm{train}}$, and whether the complete unconstrained greedy sequence matches the encoded target.
To measure influence, we construct a trie from $\mathcal C_{\mathrm{train}}$ and constrain greedy decoding according to
\begin{equation}
    \widehat c_i
      =\arg\max_{\substack{v:\,\exists C\in\mathcal C_{\mathrm{train}}\\
          C_{1:i}=(\widehat C_{<i},v)}}
        \operatorname{logit}_i(v).
    \label{eq:training-set-constrained-decoding}
\end{equation}
Every selected token therefore preserves a prefix encoded from training, and the complete generated sequence belongs to $\mathcal C_{\mathrm{train}}$.
Comparing constrained against unconstrained decoding under otherwise identical conditions isolates the effect of training-support membership.

\subsubsection{Offline Analysis}
\label{app:training-set-offline}

\paragraph{Token mismatch frequency.}
Two conditions in the cohort used for the family-level validation diagnostics are degenerate: their tokenizers collapse the training set onto only two distinct token sequences, so the policy reproduces the encoded target trivially and rollout success is zero under every decoding rule.
Excluding those two, the exact target-sequence match rate is at most 3.8\%, so at least 96.2\% of greedy predictions differ from their encoded targets.
For every non-collapsed tokenizer, nearly all greedy predictions thus differ from their encoded targets.
Figure~\ref{fig:validation-training-set-details-token-position} shows where they emerge within the autoregressive sequence, and Figure~\ref{fig:validation-exact-match} resolves the exact-match rate by quantizer family and task.

\paragraph{Action-reconstruction error.}
Figure~\ref{fig:validation-training-set-details-token-position-validation-action-mae-change} reports how constraining decoding to the training support changes the validation action error, measured between the decoded prediction and the encoded target chunk.
Training-support-constrained decoding increases validation action MAE in 64 of 66 tokenizer--task conditions and ties in the remaining 2.
In the normalized raw-action $L_1$ units of $\ell$, the task-average constrained-minus-unconstrained MAE is 0.0302 on Can, 0.0227 on Lift, and 0.0335 on Square.
Restricting predictions to observed sequences worsens action accuracy in almost every condition.
The two ties are the degenerate conditions above, whose greedy predictions already lie in $\mathcal C_{\mathrm{train}}$, so the constraint is vacuous there and acts on every condition in which it can act.

\subsubsection{Online Analysis}
\label{app:training-set-online}

Figure~\ref{fig:rollout-training-set-details} repeats both measurements during rollout, where the policy conditions on its own executed actions rather than on held-out expert chunks. Its prefix curves are grouped by quantizer family for each task; shaded regions are within-family min--max ranges across tokenizer designs. The forest panels retain all 22 tokenizer configurations and their paired intervals.

\paragraph{Token mismatch frequency.}
Figure~\ref{fig:rollout-training-set-details-token-position} reports the task-specific rollout unobserved-prefix curves, which track how often the generated prefix is absent from $\mathcal C_{\mathrm{train}}$ at each position of the sequence.
As in the offline analysis, departures from the training support are not confined to a single quantizer family.

\paragraph{Rollout success.}
Constrained decoding reduces rollout success in 53 of 66 tokenizer--task conditions, improves it in 11, and ties in 2.
The mean constrained-minus-unconstrained changes are $-19.7$, $-14.7$, and $-17.6$ percentage points on Can, Lift, and Square, respectively.
The intervention therefore degrades both offline action accuracy and closed-loop performance for most configurations.
The two ties are again the degenerate conditions, whose rollout success is zero under both decoding rules, so the constraint acts on every condition in which it can act.

Together, the offline and online diagnostics show that training-support membership is neither necessary nor sufficient for a good prediction: novel token compositions are common, and forcing predictions onto observed sequences lowers offline action accuracy and, in most conditions, rollout success.

\begin{figure}[p]
    \centering
    \setlength{\parskip}{0pt}
    \renewcommand{\footnotesize}{\fontsize{8}{9}\selectfont}
    \renewcommand{\scriptsize}{\fontsize{7}{7.6}\selectfont}
    \captionsetup{font=footnotesize,skip=2pt,hypcap=false}
    \captionsetup[subfigure]{font=scriptsize,skip=0pt,justification=centering,margin={1.60cm,0.10cm}}
    \begin{subfigure}[t]{0.66\linewidth}
        \captionsetup{margin={1.30cm,0.10cm}}
        \centering
        \input{figures/what_matters/empirical_support/appendix_validation_prefix_by_task.tex}\unskip
        \caption{Token position}\label{fig:validation-training-set-details-token-position}
    \end{subfigure}\hfill
    \begin{subfigure}[t]{0.31\linewidth}
        \renewcommand{\scriptsize}{\fontsize{7.2}{7.8}\selectfont}
        \captionsetup{margin={1.20cm,0.10cm}}
        \centering
        \input{figures/what_matters/empirical_support/appendix_validation_exact_match.tex}\unskip
        \caption{Exact-match rate.}
        \label{fig:validation-exact-match}
    \end{subfigure}\par
    \begin{subfigure}{\linewidth}
        \centering
        \input{figures/what_matters/empirical_support/appendix_validation_action_error.tex}
        \caption{Validation action-MAE change.}\label{fig:validation-training-set-details-token-position-validation-action-mae-change}
    \end{subfigure}
    \caption{Held-out validation training-support diagnostics. (a) Accumulated unobserved-prefix rates: curves and bands show family means and min--max design ranges. (b) Unconstrained greedy exact-match rates: family/task means and min--max ranges over the same cohort. (c) Training-support-constrained minus unconstrained normalized raw-action $L_1$/MAE $\ell$ for every tokenizer with paired 95\% intervals; positive values indicate worse prediction. Color and line style identify family; diamonds, squares, and triangles identify Can, Lift, and Square. Tokenizer colors and order match Appendix~\ref{app:tokenizer-index}.}
    \label{fig:validation-training-set-details}
    \par\vspace{2pt}
    \begin{subfigure}{\linewidth}
        \centering
        \input{figures/what_matters/empirical_support/appendix_rollout_prefix_by_task.tex}
        \caption{Token position.}\label{fig:rollout-training-set-details-token-position}
    \end{subfigure}\par
    \begin{subfigure}{\linewidth}
        \centering
        \input{figures/what_matters/empirical_support/appendix_rollout_success_delta.tex}
        \caption{Success-rate change (pp).}\label{fig:rollout-training-set-details-success-rate-change}
    \end{subfigure}
    \caption{Rollout training-support diagnostics using greedy policies at the checkpoint with the best teacher-forced validation accuracy, $H_{\mathrm{exec}}=8$, and 300 episodes per condition. (a) Accumulated unobserved-prefix rates: family means and min--max design ranges. (b) Constrained minus unconstrained success for all 22 tokenizers with paired 95\% intervals. Family styles, task markers, and tokenizer colors/order follow Figure~\ref{fig:validation-training-set-details}.}
    \phantomsection\label{fig:rollout-training-set-details}
\end{figure}
\clearpage

\subsection{One-Step Error Decomposition}
\label{app:one-step-error-proof}

We study the expected action-chunk error of a tokenized autoregressive policy,
\begin{equation}
    J_E
      =\mathbb E_{(O,A)\sim p_{\mathrm{data}}}
         \mathbb E_{\widehat C\sim p_\psi^{\mathrm{dec}}(\cdot\mid O)}
         \!\left[\|\mathcal D(\widehat C)-A\|\right].
    \label{eq:one-step-decoded-error-def}
\end{equation}
Here $(O,A)$ is an observation--action pair from the offline expert data distribution $p_{\mathrm{data}}$: $A$ is the expert action chunk recorded for observation $O$.
We have access only to a fixed dataset of such pairs, with no additional expert queries.
The deployed policy generates $\widehat C$ from $O$, independently of the recorded $A$ conditional on $O$, and decodes it into an action chunk.
Our objective is to decompose $J_E$ and identify which contributions are controlled by reconstruction and teacher-forced training.

\paragraph{Training objectives and decoding rules.}
We use the tokenization map $\mathcal T$, detokenization map $\mathcal D$, and autoregressive model $p_\psi$ from Section~\ref{sec:preliminaries}.
An action chunk has shape $H\times d_a$, and its target encoding is $C^\star=\mathcal T(A)=(c_1^\star,\ldots,c_L^\star)$.
Errors are measured by the MAE of Section~\ref{sec:reconstruction-threshold}: we write $\|X\|:=\|X\|_1/(Hd_a)$ for the flattened chunk, so that $\ell(A,\widehat A)=\|\widehat A-A\|$.
Let $p_{\mathcal A}$ be the action marginal of $p_{\mathrm{data}}$ and define the reconstruction error $r(A)=\|\mathcal D(\mathcal T(A))-A\|$.
The reconstruction objective and teacher-forced negative log-likelihood are
\begin{equation}
\begin{aligned}
    \mathcal L_{\mathrm{tok}}
      &=\mathbb E_{A\sim p_{\mathcal A}}[r(A)],\\
    \mathcal L_{\mathrm{AR}}
      &=-\mathbb E_{(O,A)\sim p_{\mathrm{data}}}
          [\log p_\psi(C^\star\mid O)]\\
      &=-\mathbb E_{(O,A)\sim p_{\mathrm{data}}}
          \left[\sum_{i=1}^{L}
             \log p_\psi(c_i^\star\mid c_{<i}^\star,O)\right].
\end{aligned}
    \label{eq:one-step-training-objectives}
\end{equation}
All logarithms are natural.
The reconstruction objective and $J_E$ use the same action marginal; the analysis also applies when $p_{\mathrm{data}}$ is the empirical distribution of the recorded dataset.
Quantizer-specific auxiliary losses do not enter the argument.

The model distribution $p_\psi$ and deployed distribution $p_\psi^{\mathrm{dec}}$ need not coincide.
We consider three decoding rules:
\emph{sampling}, for which $p_\psi^{\mathrm{dec}}=p_\psi$;
\emph{mode-seeking}, which returns the token-wise greedy sequence or an exact sequence mode, including beam search when it returns that mode;
and \emph{truncated sampling}, which restricts each conditional distribution to a set containing its most probable token and renormalizes, as in top-$k$ or nucleus sampling.
For mode-seeking, $p_\psi^{\mathrm{dec}}$ is the point mass at the returned sequence.
The decomposition below holds for any deployed distribution; these three rules determine the constants in the mismatch-probability bound. 

\paragraph{Error decomposition.}
For each recorded pair, define
\begin{equation}
    \mathsf M=\{\widehat C=C^\star\},
    \qquad
    p_{\mathrm m}(O,A)=p_\psi^{\mathrm{dec}}(C^\star\mid O),
    \qquad
    \bar p=\mathbb E[p_{\mathrm m}(O,A)].
    \label{eq:match-weight-and-error}
\end{equation}
Unless specified otherwise, expectations and probabilities are over $(O,A)\sim p_{\mathrm{data}}$ and the deployed token draw.
Write
$\overline{\mathcal E}_{=}=\mathbb E[\mathbf 1_{\mathsf M}\|\mathcal D(\widehat C)-A\|]$
for the matched contribution and
$M_{\ne}=\mathbb E[\|\mathcal D(\widehat C)-A\|\mid\mathsf M^c]$
for the conditional mismatch error.

\begin{proposition}[One-step error decomposition]
\label{prop:one-step-error}
The expected decoded error satisfies
\begin{equation}
    J_E
      =\overline{\mathcal E}_{=}+(1-\bar p)M_{\ne},
    \qquad
    \overline{\mathcal E}_{=}\leq\mathcal L_{\mathrm{tok}}.
    \label{eq:one-step-decoded-error}
\end{equation}
For the three decoding rules above,
\begin{equation}
    1-\bar p\leq c_{\mathrm{dec}}\mathcal L_{\mathrm{AR}},
    \qquad
    c_{\mathrm{dec}}=
    \begin{cases}
        1, & \text{sampling},\\
        1/\log 2, & \text{mode-seeking},\\
        1+1/(2\log 2), & \text{truncated sampling}.
    \end{cases}
    \label{eq:one-step-mismatch-frequency}
\end{equation}
Consequently,
\begin{equation}
    J_E\leq\mathcal L_{\mathrm{tok}}
             +c_{\mathrm{dec}}\mathcal L_{\mathrm{AR}}M_{\ne}.
    \label{eq:mismatch-contribution-proof}
\end{equation}
\end{proposition}

\begin{proof}
On $\mathsf M$, the decoder receives $C^\star$ and its error is exactly $r(A)$.
Splitting the expectation over $\mathsf M$ and $\mathsf M^c$ gives the identity in Equation~\ref{eq:one-step-decoded-error}.
Moreover, $0\leq p_{\mathrm m}(O,A)\leq1$ gives
\begin{equation}
    \overline{\mathcal E}_{=}
      =\mathbb E[p_{\mathrm m}(O,A)r(A)]
      \leq\mathbb E_{A\sim p_{\mathcal A}}[r(A)]
      =\mathcal L_{\mathrm{tok}}.
    \label{eq:matched-contribution-proof}
\end{equation}
The probability bound is established in Lemma~\ref{lem:one-step-match-probability} below.
Substituting both bounds into the decomposition proves Equation~\ref{eq:mismatch-contribution-proof}.
If there are no mismatches, their contribution is zero.
\end{proof}

\paragraph{Mismatch probability.}
Teacher forcing controls the model likelihood of the target sequence.
The following lemma transfers that control to the deployed match probability.

\begin{lemma}[Match probability under decoding]
\label{lem:one-step-match-probability}
Equation~\ref{eq:one-step-mismatch-frequency} holds for each of the three decoding rules.
For sampling, the sharper bound is $1-\bar p\leq1-e^{-\mathcal L_{\mathrm{AR}}}$.
\end{lemma}

\begin{proof}
For a fixed pair $(O,A)$, write $s=p_\psi(C^\star\mid O)$.
Under sampling, $p_{\mathrm m}=s$, so Jensen's inequality yields
\begin{equation}
    \bar p=\mathbb E[s]
      \geq\exp(\mathbb E[\log s])
      =e^{-\mathcal L_{\mathrm{AR}}}
      \geq1-\mathcal L_{\mathrm{AR}}.
    \label{eq:match-mass-proof}
\end{equation}

For the other rules, observe that $s>1/2$ implies that every target-prefix conditional probability exceeds $1/2$, since their product is $s$ and each factor is at most one.
Thus every target token is the unique most probable token at its prefix, and $C^\star$ is also the unique sequence mode.
Markov's inequality gives
\begin{equation}
    \Pr(s\leq1/2)
      =\Pr(-\log s\geq\log 2)
      \leq\frac{\mathcal L_{\mathrm{AR}}}{\log 2}.
    \label{eq:one-step-target-margin}
\end{equation}
Under mode-seeking, $s>1/2$ therefore implies $p_{\mathrm m}=1$, giving
$1-\bar p\leq\Pr(s\leq1/2)\leq\mathcal L_{\mathrm{AR}}/\log 2$.

Under truncated sampling, $s>1/2$ ensures that every target token is retained along its prefix.
Renormalization can only increase its conditional probability, so $p_{\mathrm m}\geq s$ on this event.
Consequently,
\begin{equation}
\begin{aligned}
    \bar p
      &\geq\mathbb E[s\mathbf 1_{\{s>1/2\}}]
       \geq\mathbb E[s]-\tfrac12\Pr(s\leq1/2)\\
      &\geq e^{-\mathcal L_{\mathrm{AR}}}
             -\frac{\mathcal L_{\mathrm{AR}}}{2\log 2}
       \geq1-\left(1+\frac{1}{2\log 2}\right)\mathcal L_{\mathrm{AR}}.
\end{aligned}
    \label{eq:one-step-truncated-match}
\end{equation}
This proves all three cases.
\end{proof}

\paragraph{Mismatch severity.}
To examine $M_{\ne}$, define the encoded action support and the action preimage of a sequence:
\begin{equation}
\begin{aligned}
    \mathcal C_{\mathrm{seen}}
      &=\{C:\Pr_{A\sim p_{\mathcal A}}(\mathcal T(A)=C)>0\},\\
    \mathcal F(C)
      &=\{A'\in\operatorname{supp}p_{\mathcal A}:\mathcal T(A')=C\}.
\end{aligned}
    \label{eq:encoded-expert-supports}
\end{equation}
For the empirical data distribution, $\mathcal C_{\mathrm{seen}}$ is precisely the training support $\mathcal C_{\mathrm{train}}$.
The mismatch event splits into
\begin{equation}
\begin{aligned}
    \mathsf B_{\mathrm{in}}
      &=\{\widehat C\ne C^\star,\ \widehat C\in\mathcal C_{\mathrm{seen}}\},\\
    \mathsf B_{\mathrm{out}}
      &=\{\widehat C\notin\mathcal C_{\mathrm{seen}}\}.
\end{aligned}
    \label{eq:one-step-mismatch-partition}
\end{equation}
Writing $\alpha=\Pr(\mathsf B_{\mathrm{in}}\mid\mathsf M^c)$ and
$M_{\mathrm{in}},M_{\mathrm{out}}$ for their conditional decoded errors gives
\begin{equation}
    M_{\ne}=\alpha M_{\mathrm{in}}+(1-\alpha)M_{\mathrm{out}}.
    \label{eq:one-step-mismatch-severity}
\end{equation}
When $\Pr(\mathsf M^c)=0$, set $\alpha=0$ and omit the mismatch terms.

\emph{In-support mismatches.}
For $\widehat C\in\mathcal C_{\mathrm{seen}}$, choose $A'\in\mathcal F(\widehat C)$.
Then $\mathcal D(\widehat C)=\mathcal D(\mathcal T(A'))$, and the triangle inequality gives
\begin{equation}
\begin{aligned}
    \|\mathcal D(\widehat C)-A\|
      &\leq b(\widehat C,A),\\
    b(C,A)
      &:=\inf_{A'\in\mathcal F(C)}
           \left[r(A')+\|A'-A\|\right]
       \leq R_{\mathrm{rec}}+D_{\mathcal A},
\end{aligned}
    \label{eq:one-step-preimage-bound}
\end{equation}
where
\[
    R_{\mathrm{rec}}=\sup_{A\in\operatorname{supp}p_{\mathcal A}}r(A),
    \qquad
    D_{\mathcal A}
      =\sup_{A,A'\in\operatorname{supp}p_{\mathcal A}}\|A'-A\|.
\]
The last inequality follows because each $C\in\mathcal C_{\mathrm{seen}}$ has a preimage in the support, where $r(A')\leq R_{\mathrm{rec}}$.
Taking the conditional expectation yields
\[
    M_{\mathrm{in}}
      \leq\mathbb E[b(\widehat C,A)\mid\mathsf B_{\mathrm{in}}]
      \leq R_{\mathrm{rec}}+D_{\mathcal A}.
\]
The preimage bound depends on how close a corresponding action $A'$ is to the current target $A$; the diameter bound can be much looser.
Moreover, reconstruction training controls the average $r(A)$, rather than $R_{\mathrm{rec}}$.
Without a lower bound on the probability of individual actions, a small average does not imply a small worst-case error; it gives only the tail bound
\[
    \Pr_{A\sim p_{\mathcal A}}(r(A)>\varepsilon)
      \leq\mathcal L_{\mathrm{tok}}/\varepsilon.
\]

\emph{Off-support mismatches.}
On $\mathsf B_{\mathrm{out}}$, the predicted sequence has zero probability under the encoded action distribution.
The reconstruction objective therefore places no direct constraint on its decoded action, while the teacher-forced objective does not evaluate $\mathcal D$ at all.
To make this precise, suppose $\Pr(\mathsf B_{\mathrm{out}})>0$ and impose no additional range or regularity constraint on the detokenization map.
For a vector $u$ with $\|u\|=1$ and $R>0$, define
\begin{equation}
    \mathcal D_R(C)=
    \begin{cases}
        \mathcal D(C), & C\in\mathcal C_{\mathrm{seen}},\\
        \mathcal D(C)+Ru, & C\notin\mathcal C_{\mathrm{seen}}.
    \end{cases}
    \label{eq:off-support-decoder-construction}
\end{equation}
This leaves $\mathcal L_{\mathrm{tok}}$, $\mathcal L_{\mathrm{AR}}$, and the deployed token distribution unchanged.
For every off-support prediction, the reverse triangle inequality gives
$\|\mathcal D_R(\widehat C)-A\|\geq R-\|\mathcal D(\widehat C)-A\|$,
so its conditional expected error grows without bound as $R\to\infty$.
Thus no finite upper bound on $M_{\mathrm{out}}$ follows from the two loss values alone.
A bounded output range or an additional regularity condition can supply a separate bound.

This argument applies to any decoding rule that assigns positive probability to $\mathsf B_{\mathrm{out}}$.
With finite-logit softmax sampling, that probability is positive whenever $\mathcal C_{\mathrm{seen}}$ is a proper subset of the token space.
Mode-seeking or truncation may avoid the event, but do not otherwise change the absence of decoder supervision on it.

\paragraph{Combined bound.}
Combining Proposition~\ref{prop:one-step-error} with the two mismatch branches gives
\begin{equation}
    J_E
      \leq\mathcal L_{\mathrm{tok}}
        +c_{\mathrm{dec}}\mathcal L_{\mathrm{AR}}
          \left[\alpha(R_{\mathrm{rec}}+D_{\mathcal A})
                +(1-\alpha)M_{\mathrm{out}}\right].
    \label{eq:one-step-combined-bound}
\end{equation}
Reconstruction controls the matched contribution, while teacher forcing controls mismatch frequency.
The remaining error depends on what mismatched sequences decode to: even an in-support sequence may represent an action far from the current target, and off-support decoder outputs are not constrained by either objective.
The in-support bound also requires worst-case reconstruction and action-range information beyond the average loss.
Changing the decoding rule preserves this decomposition, although it can change both branch probabilities and conditional errors.
These statements concern one-step error under the offline expert data distribution.

\paragraph{Benign and uncontrolled mismatches.}
Motivated by the action-noise tolerance observed in Appendix~\ref{app:recon-tolerance}, assume each observation--action pair $(O,A)$ admits a local region $\mathcal R(O,A)$ around $A$ whose actions preserve acceptable execution at $O$.
Let $\rho\geq0$ be a local tolerance radius bound such that $\|a-A\|\leq\rho$ for every $a\in\mathcal R(O,A)$.
An in-support token mismatch is \emph{benign} when it admits a reconstruction target $A'\in\mathcal F(\widehat C)\cap\mathcal R(O,A)$ with $r(A')\leq R_{\mathrm{rec}}$.
Since $\mathcal D(\widehat C)=\mathcal D(\mathcal T(A'))$, the triangle inequality then gives
\[
    \|\mathcal D(\widehat C)-A\|
      \leq r(A')+\|A'-A\|
      \leq R_{\mathrm{rec}}+\rho.
\]
Here, benign refers to the local compatibility of the reconstruction target; the decoded action additionally incurs reconstruction error.
Retain $\alpha=\Pr(\mathsf B_{\mathrm{in}}\mid\mathsf M^c)$ from Equation~\ref{eq:one-step-mismatch-severity} and let $\beta$ be the probability of a benign in-support mismatch conditional on $\mathsf M^c$, so that $0\leq\beta\leq\alpha$.
The remaining in-support mismatches and all off-support mismatches form the \emph{uncontrolled} branch: they lack this local compatibility guarantee, which does not imply execution failure.
Splitting the in-support contribution in Equation~\ref{eq:one-step-combined-bound} gives
\begin{equation}
    J_E
      \leq\underbrace{\mathcal L_{\mathrm{tok}}}_{\text{matched}}
        +c_{\mathrm{dec}}\mathcal L_{\mathrm{AR}}\Bigl[
        \rlap{$\displaystyle
          \overbrace{\phantom{
            \beta(R_{\mathrm{rec}}+\rho)
            +(\alpha-\beta)(R_{\mathrm{rec}}+D_{\mathcal A})
          }}^{\text{in-support}}$}%
        \underbrace{\beta(R_{\mathrm{rec}}+\rho)}_{\text{benign}}
        +\underbrace{
          (\alpha-\beta)(R_{\mathrm{rec}}+D_{\mathcal A})
          +\overbrace{(1-\alpha)M_{\mathrm{out}}}^{\text{off-support}}
        }_{\text{uncontrolled}}\Bigr].
\end{equation}
%
%
The benign branch defined above identifies a sufficient condition for local compatibility, rather than exhaustively characterizing all acceptable predictions.
The empirical training support records the final tokenizer's encodings of the training chunks and need not include every sequence encountered during tokenizer training as token assignments evolve.
Such additional exposure may allow sequences outside the empirical support to decode to locally acceptable actions. These predictions therefore remain in the off-support contribution $M_{\mathrm{out}}$, whose value can be small in practice.
However, this, in general, provides no guarantee on the action error, but suggests possible mechanisms that explain the phenomena that we study in Section~\ref{sec:training-set-composition} and Appendix~\ref{app:training-set-diagnostics}.

\paragraph{Remark: connection to trajectory-level error.}
Our analysis isolates the sources of action-chunk prediction error under expert observations.
For how prediction errors accumulate during execution, we refer readers to \citet{Zhang2025ActionChunking}, who relate squared prediction error on expert demonstrations to trajectory-level error under assumptions on system stability, the policy class, and chunked execution.
Their results provide a complementary perspective on when accurate action-chunk prediction can translate into accurate trajectories.

\clearpage
\section{Additional Method Details}
\label{app:additional-method-details}

\subsection{Token Balance Regularization: Balanced Feature Learning and Irreducible Policy Loss}
\label{app:bfl-irreducible-loss}

This section motivates TBR through the geometry of the token targets and the principle of balanced feature learning (BFL). We first analyze the irreducible loss of the teacher-forced loss in policy learning (Step 1) and bound it using token margins (Step 2). Because tokenization must also support action reconstruction (Step 3), we then examine a linear model in which noise-subspace sensitivity reduces margins without changing the tokenizer’s token encoding (Step 4). Under mild assumptions, we show that reconstruction weakly penalizes this sensitivity (Step 5), and that activation preconditioning, as one concrete BFL mechanism, removes it more rapidly than gradient
descent and reduces irreducible policy log-loss under a controlled comparison (Step 6).

\subsubsection{Irreducible Policy Loss and Token Geometry}
\label{app:irr-margins}

\paragraph{Step 1: Irreducible policy log-loss.}
Let $(O,A)\sim p_{\mathrm{data}}$ be an observation--action pair as in Appendix~\ref{app:one-step-error-proof}.
Let $\tau$ be a deterministic token-assignment map, $C=\tau(A)$ the resulting token target, and $p_\psi(C\mid O)$ the policy distribution parameterized by $\psi$.
Probabilities and expectations are over $(O,A)$, $\ent$ denotes (conditional) entropy with natural logarithms, and $J_E$, $\mathcal L_{\mathrm{tok}}$, $\mathcal L_{\mathrm{AR}}$, $M_{\ne}$, $c_{\mathrm{dec}}$ are as in Proposition~\ref{prop:one-step-error}.
That proposition gives
\begin{equation}
    J_E\le\mathcal L_{\mathrm{tok}}
          +c_{\mathrm{dec}}\,\mathcal L_{\mathrm{AR}}\,M_{\ne},
    \label{eq:irr-starting-bound}
\end{equation}
where reconstruction training controls $\mathcal L_{\mathrm{tok}}$ and Token Perturbation targets $M_{\ne}$.
We next isolate the irreducible component of $\mathcal L_{\mathrm{AR}}:=\mathbb E_{(O,C)}[-\log p_\psi(C\mid O)]$, which is the sequence-level negative log-likelihood. If the training loss is averaged over the $L$ token positions, the entropy and KL terms below should also be divided by $L$.

\begin{proposition}[Irreducible part of the teacher-forced loss]
\label{prop:irr-part}
Let $P(C\mid O)$ denote the true conditional distribution of $C$ and $\operatorname{KL}$ the Kullback--Leibler divergence.
For every policy $p_\psi$,
\begin{equation}
    \mathcal L_{\mathrm{AR}}
      =\underbrace{\ent(C\mid O)}_{\text{tokenizer and data}}
       +\underbrace{\mathbb E_O\operatorname{KL}\!\big(P(C\mid O)\,\big\|\,p_\psi(C\mid O)\big)}_{\text{policy}}
      \;\ge\;\ent(C\mid O),
    \label{eq:irr-loss-split}
\end{equation}
with equality if and only if $p_\psi(\cdot\mid O)=P(\cdot\mid O)$ almost surely.
\end{proposition}

\begin{proof}
Taking expectations in $-\log p_\psi(C\mid O)=-\log P(C\mid O)+\log(P(C\mid O)/p_\psi(C\mid O))$ gives Equation~\ref{eq:irr-loss-split}.
The KL divergence is nonnegative and vanishes exactly when the two conditional distributions agree almost surely.
\end{proof}

Thus $\ent(C\mid O)$ is the \emph{irreducible log-loss} shared by all policies trained on the same token targets. Changing the tokenizer can directly affect this loss; here we focus on the first term in Equation~\ref{eq:irr-loss-split}---the conditional token entropy.

\begin{figure}[!t]
\centering
\begin{tikzpicture}[scale=0.85]
  \fill[orange!20] (1.2,-1.2) -- (1.2,1.0) -- (0.2,2.0) -- (-2.2,1.6) -- (-2.8,-0.6) -- (-1.0,-1.9) -- cycle;
  \draw[orange!70!black,thick] (1.2,-1.2) -- (1.2,1.0) -- (0.2,2.0) -- (-2.2,1.6) -- (-2.8,-0.6) -- (-1.0,-1.9) -- cycle;
  \fill[green!28] (0,0) circle (1.2);
  \draw[green!45!black,thick] (0,0) circle (1.2);
  \foreach \x/\y in {0.3/0.5,-0.35/0.15,0.55/-0.5,0.8/0.25,-0.75/-0.35,0.15/-0.9,0.0/0.95}
     \fill[black!55] (\x,\y) circle (1.1pt);
  \foreach \x/\y in {-1.6/0.6,-1.3/-1.0,0.3/1.55}
     \fill[black!55] (\x,\y) circle (1.1pt);
  \fill[red!75!black] (1.65,0.35) circle (1.4pt);
  \fill[red!75!black] (-2.75,1.75) circle (1.4pt);
  \fill (0,0) circle (1.6pt) node[below=1pt] {\small $g(O)$};
  \draw[<->] (0,0) -- (1.2,0) node[midway,above=-1pt] {\small $m(O)$};
  \node[green!35!black] at (-0.55,0.62) {\small $\mathbb B(O)$};
  \node[orange!55!black] at (-1.95,-0.2) {\small $\mathcal C(O)$};
  \node[anchor=west] at (1.45,1.2) {\small other labels};
\end{tikzpicture}
\caption{Actions recorded under one observation (dots) around their center $g(O)$. The margin ball $\mathbb B(O)$ lies in the cell $\mathcal C(O)$ of the center; samples outside $\mathcal C(O)$ (red) carry another label and lie outside $\mathbb B(O)$.}
\label{fig:irr-ball-cell}
\end{figure}
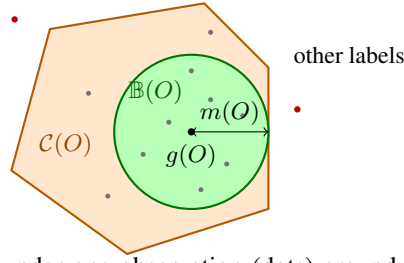

\paragraph{Step 2: Token margins bound the irreducible loss.}
Consider first a single token $c=\tau(A)\in\{1,\ldots,K\}$, where $K \geq 2$ is the codebook size.
For a sequence $C=(c_1,\ldots,c_L)$ of $L$ tokens, let $c_{<i}=(c_1,\ldots,c_{i-1})$ denote the preceding tokens.
The chain rule $\ent(C\mid O)=\sum_{i=1}^L\ent(c_i\mid O,c_{<i})$ reduces the sequence setting to the single-token case by replacing $O$ with $(O,c_{<i})$.
For each observation, choose a measurable reference action $g(O)$; the bounds below hold for any function $g$ of $O$ and the conditional mean $g(O):=\mathbb E[A\mid O]$ is one natural choice. This reference serves as a geometric witness for a token label.
We measure the margins from the reference action, around which the actions recorded at $O$ form a cloud, and call it the \emph{center} (Figure~\ref{fig:irr-ball-cell}).
We write $A$ for the random action chunk and lowercase $a$ for a fixed action chunk, following the usual convention for a random variable and its values.
Define
\begin{equation}
    \begin{aligned}
    \mathcal C(O)&=\{a:\tau(a)=\tau(g(O))\}
      &&\text{(cell of the center)},\\
    m(O)&=\inf\{\|a-g(O)\|_2:a\notin\mathcal C(O)\}
      &&\text{(margin)},\\
    \mathbb B(O)&=\{a:\|a-g(O)\|_2<m(O)\}
      &&\text{(margin ball)},\\
    p_{\ne}(O)&=\Pr[A\notin\mathcal C(O)\mid O]
      &&\text{(label inconsistency)}.
    \end{aligned}
    \label{eq:irr-objects}
\end{equation}
Thus $\mathcal C(O)$ is the token cell containing the reference action $g(O)$, and $m(O)$ is the distance from $g(O)$ to the nearest cell boundary.
The label inconsistency $p_{\ne}(O)$ is the probability that a demonstration action $A$ recorded at $O$ receives a different token from its center $g(O)$.
When $0<p_{\ne}(O)<1$, demonstrations of the same observation disagree on the token, so no policy can predict it from $O$ alone.
Let $h_2(p)=-p\log p-(1-p)\log(1-p)$ denote binary entropy.

\begin{proposition}[Irreducible loss and label inconsistency]
\label{prop:irr-whether-where}
Let $\ent^{\ne}(O):=\ent(c\mid O,A\notin\mathcal C(O))$ be the entropy among labels outside the center cell.
Then
\begin{equation}
    \ent(c\mid O)
      =\mathbb E\big[\underbrace{h_2(p_{\ne}(O))}_{\text{whether}}
        +\underbrace{p_{\ne}(O)\,\ent^{\ne}(O)}_{\text{where}}\big]
      \;\le\;\mathbb E\big[h_2(p_{\ne}(O))+p_{\ne}(O)\log(K-1)\big].
    \label{eq:irr-whether-where}
\end{equation}
Equality holds when, given $A\notin\mathcal C(O)$, the label is uniform on the other $K-1$ values, and always when $K=2$.
\end{proposition}

\begin{proof}
Fix $O=o$ and let $I=\mathbf 1\{A\notin\mathcal C(o)\}=\mathbf 1\{\tau(A)\neq \tau(g(o))\}=\mathbf 1\{c\neq \tau(g(o))\}$, which is a deterministic function of $c$.
Because $I$ is Bernoulli with mean $p_{\ne}(o)$ and $c=\tau(g(o))$ when $I=0$, the chain rule gives
$\ent(c\mid o)=\ent(I\mid o)+\ent(c\mid I,o)=h_2(p_{\ne}(o))+p_{\ne}(o)\ent^{\ne}(o)$.
Finally, $\ent^{\ne}(o)\le\log(K-1)$, with equality for a uniform distribution over the other labels; when $K=2$, equality always holds.
\end{proof}

The \emph{whether} term $h_2(p_{\ne}(O))$ is the uncertainty about whether a demonstration keeps the token of its center.
The \emph{where} term $p_{\ne}(O)\ent^{\ne}(O)$ is the uncertainty about which of the other $K-1$ tokens it receives when it does not, and it is at most $p_{\ne}(O)\log(K-1)$.
The margin controls the exit probability $p_{\ne}(O)$.

\begin{proposition}[Margin bound]
\label{prop:irr-margin}
For every $O$, $\mathbb B(O)\subseteq\mathcal C(O)$, and hence
\begin{equation}
    p_{\ne}(O)\le\Pr\big[\|A-g(O)\|_2\ge m(O)\mid O\big],
    \label{eq:irr-margin}
\end{equation}
with equality when $\mathcal C(O)$ is a ball centered at $g(O)$.
\end{proposition}

\begin{proof}
By the definition of $m(O)$, every $a$ with $\|a-g(O)\|_2<m(O)$ belongs to $\mathcal C(O)$. Consequently, $\{A \notin \mathcal C(O) \}\subseteq \{\|A-g(O)\|_2\ge m(O)\}$. Taking conditional probabilities gives Equation~\ref{eq:irr-margin}.
\end{proof}

Let $F_K(p):=h_2(p)+p\log(K-1)$ and define its nondecreasing envelope by $\overline F_K(q):=\max_{0\le p\le q}F_K(p)=F_K\!\left(\min\left\{q,1-\frac1K\right\}\right), 0\le q\le1$. Writing $q(O)=\Pr\big[\|A-g(O)\|_2\ge m(O)\mid O\big]$, we have $\ent(c \mid O)\le \mathbb E\big[F_K(p_{\ne}(O))\big]\le \mathbb E\big[\overline F_K(q(O))\big]$. For a fixed conditional action distribution and reference map $g$, this upper bound is nonincreasing as the margins grow. This implies that the smaller token margin results in a larger upper bound on the token entropies.


\paragraph{Step 3: Predictability and reconstruction.}
Tokenization should balance predictability and reconstruction. The margin bound suggests that greater separation between conditional action variation and token boundaries can improve predictability, while a sufficiently low reconstruction error is required (Section~\ref{sec:reconstruction-threshold}). A trivial case is that a tokenizer can have constant token encoding for all actions, resulting in zero token entropies while action information is totally lost.
We therefore fix $K$ and the encoded content, and ask  whether reducing sensitivity in orthogonal directions with respect to the token margin boundary can improve token predictability while preserving reconstruction.

\subsubsection{A Linear Model Motivating Balanced Feature Learning}
\label{app:irr-linear-example}

\paragraph{Assumptions and scope.} We now list the explicit distributional and optimization assumptions.
\textbf{(A1) Conditional-center subspace model.} Let $A=g(O)+\xi$, where $\xi\sim\mathcal N(0,\sigma^2I_d)$ is independent of $O$ and $\sigma>0$ is the noise scale.
Let $\mathcal S\subsetneq\mathbb R^d$ be the smallest subspace containing $g(O)$ almost surely, and let $\mathcal S^\perp$ denote its orthogonal complement. The low-dimensional assumption concerns the conditional centers, not the noisy actions: when second moments exist, $\mathbb E\big[AA^\top\big]=\mathbb E \big[g(O)g(O)^\top\big] + \sigma^2 I_d \succ 0$. Let $P_{\mathcal S}$ and $P_\perp$ be the orthogonal projections onto $\mathcal S$ and $\mathcal S^\perp$, respectively.

\textbf{(A2) Binary linear tokenizer.} Let $c=\mathbf 1\{w^\top A>\beta\}$, where $w\ne0$ is a row of the first encoder layer, $\beta\in\mathbb R$ is its threshold, and $\mathbf 1\{\cdot\}$ is the indicator function. This binary case permits an exact entropy calculation; the general bounds in Steps 1–3
already apply to arbitrary finite codebooks.

\paragraph{Step 4: Noise-subspace sensitivity reduces token margins.}
The subspace $\mathcal S$ is a property of the data alone: along $\mathcal S$ the centers vary, and along $\mathcal S^\perp$, the \emph{pure-noise directions}, actions carry only residual noise.
Under (A2), $m_w(a):=|w^\top a-\beta|/\|w\|_2$ is the distance from $a$ to the boundary, so $m(O)=m_w(g(O))$.
Let $\Phi$ be the standard normal distribution function.

\begin{proposition}[Exact irreducible loss]
\label{prop:irr-margin-over-noise}
Under (A1)--(A2),
\begin{equation}
    \Pr[c=1\mid O]=\Phi\!\Big(\frac{w^\top g(O)-\beta}{\sigma\|w\|_2}\Big),
    \qquad
    \ent(c\mid O)=\mathbb E\Big[h_2\Big(\Phi\Big(\frac{m_w(g(O))}{\sigma}\Big)\Big)\Big],
    \label{eq:irr-margin-over-noise}
\end{equation}
and the integrand is strictly decreasing in $m_w(g(O))/\sigma$.
\end{proposition}

\begin{proof}
$w^\top A=w^\top g(O)+w^\top\xi$ with $w^\top\xi\sim\mathcal N(0,\sigma^2\|w\|_2^2)$ independent of $O$, which gives the first identity.
Since $K=2$, Proposition~\ref{prop:irr-whether-where} holds with equality, $\ent(c\mid O)=\mathbb E[h_2(\Pr[c=1\mid O])]$, and $h_2(\Phi(x))=h_2(\Phi(-x))$ is strictly decreasing in $|x|$.
\end{proof}

Decompose $w=w_{\mathcal S}+u$ with $w_{\mathcal S}\in\mathcal S$, the \emph{signal-subspace} component, and $u\in\mathcal S^\perp$, the \emph{noise-subspace} component.

\begin{proposition}[Removing \emph{noise-subspace} sensitivity increases margins]
\label{prop:irr-wasted-weight}
Assume (A1)--(A2) and $w_{\mathcal S}\ne0$, and let $\alpha$ be the angle between $w$ and $w_{\mathcal S}$.
Then
\begin{equation}
    m_w(g(O))=\frac{m_{w_{\mathcal S}}(g(O))}{\kappa},
    \qquad
    \kappa:=\sqrt{1+\frac{\|u\|_2^2}{\|w_{\mathcal S}\|_2^2}}=\frac1{\cos\alpha}\ge1 .
    \label{eq:irr-kappa}
\end{equation}
Moreover, $\mathbf 1\{w^\top g(O)>\beta\}=\mathbf 1\{w_{\mathcal S}^\top g(O)>\beta\}$, so every center keeps its label, and
$\ent(c\mid O)=\mathbb E\big[h_2\big(\Phi\big(m_{w_{\mathcal S}}(g(O))/(\sigma\kappa)\big)\big)\big]$
is nondecreasing in $\|u\|_2$, strictly so if $\Pr[m_{w_{\mathcal S}}(g(O))>0]>0$.
\end{proposition}

\begin{proof}
Since $g(O)\in\mathcal S$ and $u\perp\mathcal S$, $w^\top g(O)=w_{\mathcal S}^\top g(O)$; since $w_{\mathcal S}\perp u$, $\|w\|_2=\kappa\|w_{\mathcal S}\|_2$ and $\cos\alpha=\|w_{\mathcal S}\|_2/\|w\|_2$.
The last claim follows from Proposition~\ref{prop:irr-margin-over-noise}.
\end{proof}

\begin{figure}[ht]
\centering
\begin{tikzpicture}[scale=0.78]
  \path[use as bounding box] (-3.5,-2.75) rectangle (12.95,2.35);
  \tikzset{
    irrcloudL/.style={fill=blue!22,draw=blue!45!black,line width=0.3pt},
    irrcloudR/.style={fill=orange!30,draw=orange!60!black,line width=0.3pt},
    irrspill/.style={fill=red!75}
  }
  \begin{scope}
    \draw[->,black!45] (-3.3,0) -- (4.2,0) node[right=-2pt] {\small $\mathcal S$};
    \draw[->,black!45] (-3.15,-1.6) -- (-3.15,1.8) node[above=-1pt] {\small $\mathcal S^\perp$};
    \foreach \x in {-2.3,-0.2} \path[irrcloudL] (\x,0) circle (0.85);
    \foreach \x in {1.4,3.1}  \path[irrcloudR] (\x,0) circle (0.85);
    \begin{scope}\clip (0.6,-2) rectangle (5,2);  \path[irrspill] (-0.2,0) circle (0.85);\end{scope}
    \begin{scope}\clip (-5,-2) rectangle (0.6,2); \path[irrspill] (1.4,0) circle (0.85);\end{scope}
    \foreach \x in {-2.3,-0.2,1.4,3.1} \fill (\x,0) circle (1.4pt);
    \draw[very thick,green!35!black] (0.6,-1.7) -- (0.6,1.85);
    \fill (0.6,0) circle (1.7pt);
    \node[blue!50!black] at (-1.25,1.45) {\small $c=0$};
    \node[orange!65!black] at (2.35,1.45) {\small $c=1$};
    \node at (0.45,-2.35) {\small (a) $u=0$};
  \end{scope}
  \begin{scope}[xshift=8.15cm]
    \draw[->,black!45] (-3.3,0) -- (4.2,0) node[right=-2pt] {\small $\mathcal S$};
    \draw[->,black!45] (-3.15,-1.6) -- (-3.15,1.8) node[above=-1pt] {\small $\mathcal S^\perp$};
    \foreach \x in {-2.3,-0.2} \path[irrcloudL] (\x,0) circle (0.85);
    \foreach \x in {1.4,3.1}  \path[irrcloudR] (\x,0) circle (0.85);
    \begin{scope}\clip (4.695,2.87) -- (-3.495,-2.87) -- (-0.625,-6.965) -- (7.565,-1.225) -- cycle;
      \path[irrspill] (-0.2,0) circle (0.85);\end{scope}
    \begin{scope}\clip (4.695,2.87) -- (-3.495,-2.87) -- (-6.365,1.225) -- (1.825,6.965) -- cycle;
      \path[irrspill] (1.4,0) circle (0.85);\end{scope}
    \foreach \x in {-2.3,-0.2,1.4,3.1} \fill (\x,0) circle (1.4pt);
    \draw[thick,green!35!black,densely dotted] (0.6,-1.7) -- (0.6,1.85);
    \begin{scope}\clip (-3.3,-1.75) rectangle (4.15,1.9);
      \draw[very thick,red!65!black] (-3.495,-2.87) -- (4.695,2.87);\end{scope}
    \fill (0.6,0) circle (1.7pt);
    \draw[black!65] (0.6,0.75) arc (90:35:0.75);
    \node[black!75] at (1.0,1.0) {\small $\alpha$};
    \node at (0.45,-2.35) {\small (b) $u\ne0$};
  \end{scope}
\end{tikzpicture}
\caption{The linear example in two dimensions, where $\mathcal S$ is a line. Discs are the action clouds of four observations. (a) $u=0$. (b) $u\ne0$ tilts the boundary by $\alpha$ about its cut with $\mathcal S$: no center changes side, every margin shrinks by $\kappa=1/\cos\alpha$, and more samples (red) receive a label different from that of their center.}
\label{fig:irr-tilted-boundary}
\end{figure}
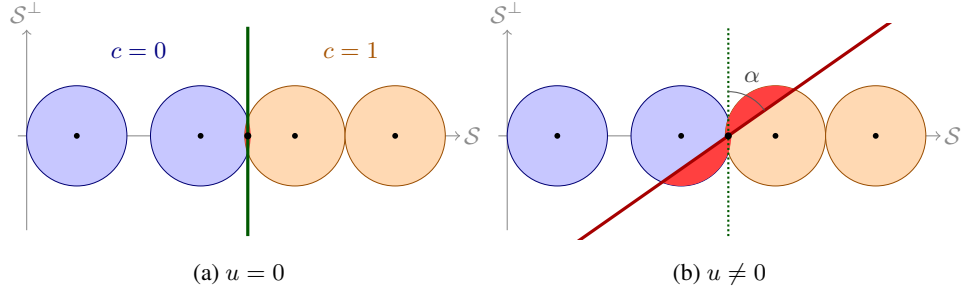

The pair $(w_{\mathcal S},\beta)$ determines where the boundary cuts $\mathcal S$, and hence what the token encodes. The component $u$ changes sensitivity in $\mathcal S^\perp$ while leaving the token encoding for the reference actions unchanged (Figure~\ref{fig:irr-tilted-boundary}).
Holding $(w_{\mathcal S},\beta)$ fixed therefore isolates the effect of noise-subspace sensitivity on token predictability.

Thus removing the noise-subspace component multiplies each center margin by $\kappa$ and reduces conditional token entropy under the condition of Proposition~\ref{prop:irr-wasted-weight}.

\paragraph{Step 5: Reconstruction weakly penalizes the wasted weight.}
Step 4 identified a component of the encoder weight that reduces token margins without changing the labels of the action-cloud centers. We now specify a reconstruction-guided model of encoder training, in which we separate the choice of target codes from the optimization that fits the encoder to those codes. The resulting quadratic objective is exact for this regression problem.

\noindent\textbf{(A3) Fixed-target code regression.}
Fix a finite set of code embeddings $z_1,\ldots,z_K\in\mathbb R^r$ and a
fixed decoder $D_\star$ whose reconstruction points satisfy
\begin{equation}
 r_j:=D_\star(z_j)\in\mathcal S,
 \qquad j=1,\ldots,K.
 \label{eq:irr-target-range}
\end{equation}
Choose the target assignment and target embedding as
\begin{equation}
 C^\dagger(A)
 :=\underset{j}{\arg\min}\,\|P_{\mathcal S}A-r_j\|_2^2,
 \qquad Z^\dagger(A):=z_{C^\dagger(A)},
 \label{eq:irr-target-assignment}
\end{equation}
During the optimization comparison, the decoder, embeddings, and target mapping $A\mapsto Z^\dagger(A)$ are fixed. The linear encoder is trained by
\begin{equation}
 \mathcal L_{\mathrm{code}}(W)
 :=\frac12\mathbb E\|WA-Z^\dagger(A)\|_2^2,
 \qquad
 W_\star:=\underset{W}{\arg\min}\,\mathcal L_{\mathrm{code}}(W).
 \label{eq:irr-code-regression}
\end{equation}
For the binary example, scalar target embeddings ($r=1$) suffice; the matrix
notation also permits the feature-wise optimization calculation below.
The range condition in Equation~\ref{eq:irr-target-range} is motivated by
optimal reconstruction for assignments based only on $P_{\mathcal S}A$.
Indeed, under (A1), $P_\perp A$ is zero-mean and independent of $P_{\mathcal S}A$.
For any fixed target assignment, the squared-error-optimal reconstruction minimizer gives $r_j \in \mathcal S$, which incurs no loss of optimality within this class of assignments. 
We therefore model encoder training by regressing $WA$ onto the corresponding embedding $Z^\dagger(A)$, using code regression as a quadratic surrogate for action reconstruction.

\begin{lemma}[Regression geometry and the noise subspace]
\label{lem:irr-principal-subspace}
Under (A1) and (A3), $C^\dagger$ minimizes squared action-reconstruction error
among assignments to the fixed reconstruction points $r_1,\ldots,r_K$, and
\begin{equation}
 \mathbb E\bigl[Z^\dagger(P_\perp A)^\top\bigr]=0.
 \label{eq:irr-target-noise-orthogonality}
\end{equation}
Moreover, with $\Sigma=\mathbb E[AA^\top]$,
\begin{equation}
 \Sigma=\mathbb E[g(O)g(O)^\top]+\sigma^2I,
 \qquad \Sigma P_\perp=\sigma^2P_\perp.
 \label{eq:irr-target-spectrum}
\end{equation}
The eigenvalues on $\mathcal S$ exceed $\sigma^2$, whereas those on
$\mathcal S^\perp$ equal $\sigma^2$. The unique regression optimum satisfies
\begin{equation}
 W_\star=\mathbb E[Z^\dagger A^\top]\Sigma^{-1},
 \qquad W_\star P_\perp=0,
 \label{eq:irr-target-regression-optimum}
\end{equation}
and the objective admits the exact decomposition
\begin{equation}
 \mathcal L_{\mathrm{code}}(W)
 =\frac12\mathbb E\|(W-W_\star)A\|_2^2
  +\mathcal L_{\mathrm{code}}(W_\star),
 \qquad
 G:=\nabla_W\mathcal L_{\mathrm{code}}(W)=(W-W_\star)\Sigma.
 \label{eq:irr-target-quadratic}
\end{equation}
\end{lemma}

\begin{proof}
For every action $a$ and reconstruction point $r_j\in\mathcal S$, orthogonality
gives
\[
 \|a-r_j\|_2^2=\|P_{\mathcal S}a-r_j\|_2^2+\|P_\perp a\|_2^2.
\]
The second term is independent of $j$, so the assignment in
Equation~\ref{eq:irr-target-assignment} is pointwise optimal for the fixed decoder
and depends only on $P_{\mathcal S}A$. Under (A1), $P_\perp A=P_\perp\xi$ is zero-mean and
independent of $P_{\mathcal S}A=g(O)+P_{\mathcal S}\xi$. Thus $Z^\dagger$ is independent of $P_\perp A$,
which proves Equation~\ref{eq:irr-target-noise-orthogonality}.

Independence of $\xi$ and $O$ gives the expression for $\Sigma$. Since
$\mathbb E[g(O)g(O)^\top]$ has range $\mathcal S$, its restriction to
$\mathcal S$ is positive definite, and it vanishes on $\mathcal S^\perp$.
This proves the spectral statements. In particular, $\Sigma\succ0$.

The least-squares normal equation is
\[
 \mathbb E[(W_\star A-Z^\dagger)A^\top]=0.
\]
It yields the stated formula for $W_\star$. Since
$\Sigma^{-1}P_\perp=\sigma^{-2}P_\perp$, we also obtain
\[
 W_\star P_\perp
 =\sigma^{-2}\mathbb E[Z^\dagger(P_\perp A)^\top]=0.
\]
Finally, expand
$WA-Z^\dagger=(W-W_\star)A+(W_\star A-Z^\dagger)$.
The normal equation makes the expected cross term zero, giving
Equation~\ref{eq:irr-target-quadratic}. Differentiation gives the gradient.
\end{proof}

For two distinct reconstruction points $r_0,r_1\in\mathcal S$, with ties assigned
to code $0$, the target assignment is itself a linear binary token:
\begin{equation}
 C^\dagger(A)
 =\mathbf 1\!\left\{(r_1-r_0)^\top A>
                  \frac{\|r_1\|_2^2-\|r_0\|_2^2}{2}\right\}.
 \label{eq:irr-target-binary}
\end{equation}
Thus no pure-noise direction is needed for an optimal assignment to these reconstruction points. The least-squares encoder need not interpolate the target embeddings.

\begin{proposition}[Different sensitivities of code loss and conditional entropy]
\label{prop:irr-code-noise-sensitivity}
Assume (A1)--(A3). For one encoder row and its target coordinate
$Z^\dagger_\ell$, write
\[
 \ell_{\mathrm{code}}(w)
 :=\frac12\mathbb E(w^\top A-Z^\dagger_\ell)^2,
 \qquad w=w_{\mathcal S}+u,
 \quad 0\ne w_{\mathcal S}\in\mathcal S,
 \quad u\in\mathcal S^\perp.
\]
Then
\begin{equation}
 \ell_{\mathrm{code}}(w)-\ell_{\mathrm{code}}(w_{\mathcal S})
 =\frac12\sigma^2\|u\|_2^2.
 \label{eq:irr-code-wasted-penalty}
\end{equation}
For the token $c_w=\mathbf 1\{w^\top A>\beta\}$, let
\[
 \kappa=\sqrt{1+\frac{\|u\|_2^2}{\|w_{\mathcal S}\|_2^2}}.
\]
Suppose the signed center distance
$s(O)=(w_{\mathcal S}^\top g(O)-\beta)/\|w_{\mathcal S}\|_2$ has a density $f$
that is continuous and positive at zero. Holding $w_{\mathcal S}$, $u$, $\beta$,
and the distribution of $g(O)$ fixed as $\sigma\to0$, we have
\begin{equation}
 \ent(c_w\mid O)=C_0\sigma\kappa+o(\sigma),
 \qquad
 C_0=2f(0)\int_0^\infty h_2(\Phi(y))\,dy>0,
 \label{eq:irr-target-entropy-asymptotic}
\end{equation}
where $C_0$ is independent of $u$. Consequently,
\begin{equation}
 \ent(c_w\mid O)-\ent(c_{w_{\mathcal S}}\mid O)
 =C_0\sigma(\kappa-1)+o(\sigma).
 \label{eq:irr-target-entropy-excess}
\end{equation}
\end{proposition}

\begin{proof}
Let $w_\star$ be the corresponding row of $W_\star$.
Lemma~\ref{lem:irr-principal-subspace} gives $w_\star\in\mathcal S$ and
\[
 \ell_{\mathrm{code}}(w)
 =\frac12(w-w_\star)^\top\Sigma(w-w_\star)+\mathrm{const}.
\]
In the decomposition $w-w_\star=(w_{\mathcal S}-w_\star)+u$, the cross term
vanishes because $\Sigma u=\sigma^2u$ and $u\perp\mathcal S$.
This proves Equation~\ref{eq:irr-code-wasted-penalty}.

The margin and entropy formulas of Step 4 give
\[
 \ent(c_w\mid O)
 =\int_{\mathbb R}h_2\!\left(\Phi\!\left(\frac{|s|}{\sigma\kappa}\right)\right)
                  f(s)\,ds
 =\sigma\kappa\int_{\mathbb R}h_2(\Phi(|y|))f(\sigma\kappa y)\,dy.
\]
To justify the limit, first restrict to $|s|<\delta$, where continuity bounds
$f$ and dominated convergence applies after the change of variables.
On $|s|\ge\delta$, the integral divided by $\sigma\kappa$ is bounded by
\[
 \frac{h_2(\Phi(\delta/(\sigma\kappa)))}{\sigma\kappa}
 \longrightarrow 0.
\]
Here the convergence follows from Gaussian tail decay. The limiting constant
is therefore $C_0$ as stated above.
Applying the same result with $u=0$ and subtracting proves
Equation~\ref{eq:irr-target-entropy-excess}.
\end{proof}

\paragraph{Implication of Step 5.}
For fixed nonzero noise-subspace weight, its excess code-regression loss is of order $\sigma^2$, whereas its excess conditional entropy is of order $\sigma$ under the stated boundary-density condition. Encoders with the same signal-subspace component and threshold can therefore have nearly equal code-fitting losses but different token predictability.

\paragraph{Step 6: Balanced feature learning in the regression model.}
The fixed-target regression problem above has the exact gradient
$G=(W-W_\star)\Sigma$. We now compare full-population gradient updates with
fixed input second moment and fixed targets. Write
\[
 \Sigma=\sum_k\gamma_kv_kv_k^\top,
 \qquad \gamma_k>0,
\]
with orthonormal eigenvectors $v_k$, and define the feature error along $v_k$ by
\begin{equation}
 \epsilon_k(W):=\|(W-W_\star)v_k\|_2,
 \qquad
 \mathcal L_{\mathrm{code}}(W)
 =\frac12\sum_k\gamma_k\epsilon_k(W)^2
   +\mathcal L_{\mathrm{code}}(W_\star).
 \label{eq:irr-target-feature-error}
\end{equation}
The objective weights each feature error by the energy of its input direction.
More precisely, the \emph{noise-subspace} weight matrix $U:=WP_\perp$ satisfies
\begin{equation}
 U=(W-W_\star)P_\perp,
 \qquad
 \|U\|_F^2=\sum_{k:\,\gamma_k=\sigma^2}\epsilon_k(W)^2,
 \label{eq:irr-target-wasted-error}
\end{equation}
by Lemma~\ref{lem:irr-principal-subspace}. Thus the \emph{noise-subspace} weight component is the regression
error on the lowest-energy input subspace. The gradient-descent and activation-preconditioning contraction calculations depend only on $G=(W-W_\star)\Sigma$, and hence apply directly to this model.

\begin{definition}[Learning-speed balance in the regression model]
\label{def:irr-balance}
If an update rule maps weights $W$ to updated weights $W^+$ and satisfies $\epsilon_k(W^+)=\theta_k\,\epsilon_k(W)$ with $\theta_k\in[0,1)$, its \emph{learning speed} along $v_k$ is $1-\theta_k$ and its \emph{balance} is
\begin{equation}
    \operatorname{bal}:=\frac{\min_k(1-\theta_k)}{\max_k(1-\theta_k)}\in(0,1].
    \label{eq:irr-balance}
\end{equation}
The rule is \emph{balanced} if $\operatorname{bal}=1$.
\end{definition}
Token Balance Regularization (TBR) refers to the procedure of applying balanced feature learning (BFL) to reduce disparity among directional learning speeds. Activation preconditioning is a concrete BFL mechanism analyzed below.

\begin{lemma}[Activation preconditioning improves balance]
\label{lem:irr-ap-balanced}
Assume (A1) and (A3).
Gradient descent, $W^+=W-\eta G$, and activation preconditioning with ridge parameter $\rho>0$, $W^+=W-\eta\,G(\Sigma+\rho I)^{-1}$, where $I$ is the identity matrix, satisfy the following relations; the superscripts $\mathrm{GD}$ and $\mathrm{AP}$ denote the two update rules:
\begin{equation}
    \theta_k^{\mathrm{GD}}=1-\eta\gamma_k,
    \quad
    \operatorname{bal}^{\mathrm{GD}}=\frac{\gamma_{\min}}{\gamma_{\max}};
    \qquad
    \theta_k^{\mathrm{AP}}=1-\frac{\eta\gamma_k}{\gamma_k+\rho},
    \quad
    \operatorname{bal}^{\mathrm{AP}}
      =\frac{\gamma_{\min}}{\gamma_{\max}}\cdot
       \frac{\gamma_{\max}+\rho}{\gamma_{\min}+\rho}
      \ge\operatorname{bal}^{\mathrm{GD}},
    \label{eq:irr-balances}
\end{equation}
and $\operatorname{bal}^{\mathrm{AP}}\to1$ as $\rho\to0$.
\end{lemma}

\begin{proof}
Under (A3), $W^+-W_\star=(W-W_\star)(I-\eta\Sigma)$ for gradient descent and $W^+-W_\star=(W-W_\star)\big(I-\eta\Sigma(\Sigma+\rho I)^{-1}\big)$ for activation preconditioning; applying both sides to $v_k$ gives $\theta_k$.
Both speeds increase with $\gamma_k$, which gives the balances.
\end{proof}

To compare contraction rates rather than balance ratios alone, match the fastest-direction speed to $\tilde \eta \in (0, 1)$, i.e. $W^+=W- \frac{\tilde \eta}{\gamma_{\max}} G$ for GD and $W^+=W- \frac{\tilde \eta (\gamma_{\max} + \rho)}{\gamma_{\max}} G(\Sigma+\rho I)^{-1}$ for AP. Then, for every $k$,
\begin{equation}
    \frac{1-\theta_k^{\mathrm{AP}}}{1-\theta_k^{\mathrm{GD}}}
      =\frac{\gamma_{\max}+\rho}{\gamma_k+\rho}\ge1 .
    \label{eq:irr-ap-faster}
\end{equation}

The factor $\Sigma$ in $G$ makes the gradient-descent speed proportional to input energy; activation preconditioning compensates for this dependence through $(\Sigma+\rho I)^{-1}$.

\begin{corollary}[A balanced update reduces irreducible loss in the linear model]
\label{cor:irr-gd-vs-ap}
Assume (A1)--(A3), $\sigma^2<\gamma_{\max}$. Let $w_\star \neq 0$ be the transpose of the corresponding row of $W_\star$, and fix $\beta$. Initialize both update rules at $w(0)=w_\star+u(0)$ with $u(0)\in\mathcal S^\perp\setminus\{0\}$, and match their fastest-direction speed to $\tilde\eta\in(0,1)$.
For an integer $T\ge1$, the weight after $T$ steps is $w(T)=w_\star+u(T)$ for both rules, with
\begin{equation}
    \|u^{\mathrm{GD}}(T)\|_2
      =\Big(1-\frac{\tilde\eta\,\sigma^2}{\gamma_{\max}}\Big)^{T}\|u(0)\|_2,
    \qquad
    \|u^{\mathrm{AP}}(T)\|_2
      =\Big(1-\frac{\tilde\eta\,\sigma^2}{\sigma^2+\rho}\cdot
             \frac{\gamma_{\max}+\rho}{\gamma_{\max}}\Big)^{T}\|u(0)\|_2 ,
    \label{eq:irr-decay}
\end{equation}
so that $\|u^{\mathrm{AP}}(T)\|_2<\|u^{\mathrm{GD}}(T)\|_2$ and, if $\Pr[m_{w_\star}(g(O))>0]>0$,
\begin{equation}
    \ent \big(c^{\mathrm{AP}}_T\mid O\big)<\ent \big(c^{\mathrm{GD}}_T\mid O\big)
    \qquad\text{for every }T\ge1 ,
    \label{eq:irr-ordering}
\end{equation}
where $c^{\mathrm{GD}}_T$ and $c^{\mathrm{AP}}_T$ are the tokens defined by $w(T)$ under the two rules.
\end{corollary}

\begin{proof}
By Lemma~\ref{lem:irr-principal-subspace}, $\mathcal S^\perp$ is an eigenspace of $\Sigma$ with eigenvalue $\sigma^2$, so both updates preserve $w_{\mathcal S}=w_\star$ and contract $u$ by the factors in Lemma~\ref{lem:irr-ap-balanced} with $\gamma_k=\sigma^2$.
Equation~\ref{eq:irr-ap-faster} and $\sigma^2 < \gamma_{\max}$ imply strictly faster contraction under activation preconditioning. Proposition~\ref{prop:irr-wasted-weight} then gives the entropy ordering under its stated strictness condition.
\end{proof}

Equation~\ref{eq:irr-ordering} directly compares the irreducible losses of two tokenizers with the same reconstruction target $w_\star$.
For $\rho\ll\gamma_{\max}$, gradient descent removes the wasted weight on the time scale $\gamma_{\max}/\sigma^2$ and activation preconditioning on the time scale $1+\rho/\sigma^2$.
Combining Propositions~\ref{prop:irr-part} and~\ref{prop:irr-wasted-weight} with Equation~\ref{eq:irr-starting-bound} gives, in the example,
\begin{equation}
    J_E\le\mathcal L_{\mathrm{tok}}
      +c_{\mathrm{dec}}\,M_{\ne}
       \Big[\,
         \mathbb E\,h_2\Big(\Phi\Big(\frac{m_{w_\star}(g(O))}{\sigma\,\kappa_T}\Big)\Big)
         +\mathbb E_O\operatorname{KL}\!\big(P(C\mid O)\,\big\|\,p_\psi(C\mid O)\big)
       \Big],
    \label{eq:irr-combined}
\end{equation}
where $\kappa_T^2=1+\theta^{2T}\|u(0)\|_2^2/\|w_\star\|_2^2$ and $\theta$ is the contraction factor of the training rule on $\mathcal S^\perp$.
Reconstruction training controls $\mathcal L_{\mathrm{tok}}$, Token Perturbation targets $M_{\ne}$, policy training controls the divergence, and TBR acts through $\theta$.

\paragraph{Extension beyond a binary token.}
The general entropy and margin bounds already apply to any finite codebook. For a linear multiclass tokenizer $\tau(a)=\arg\max_j(w_j^\top a+b_j)$, removing the noise-subspace component of every $w_j$ preserves all scores on $\mathcal S$. At a center with a unique winning label $j$, its cell margin is
\[
 \min_{k\ne j}
 \frac{(w_j-w_k)^\top g(O)+b_j-b_k}{\|w_j-w_k\|_2}.
\]
Projecting all weights onto $\mathcal S$ leaves each numerator fixed and
cannot increase any denominator, so it cannot decrease this margin, thus giving a nonincreasing entropy
upper bound.

\paragraph{A local extension to nonlinear decision boundaries.}
The results in Section~\ref{app:irr-margins} apply unchanged to nonlinear
tokenizers. For a local geometric extension, consider a binary decision
$c_\vartheta(a)=\mathbf1\{f_\vartheta(a)>\beta\}$ and a reference action
$\mu=g(o)$, where $f_\vartheta$ is twice continuously differentiable.
Define the score gap and input gradient
\[
 \Delta:=|f_\vartheta(\mu)-\beta|>0,
 \qquad v:=\nabla_a f_\vartheta(\mu).
\]
The linearized boundary has distance
$\Delta/\|v\|_2$ when $v\ne0$, giving a first-order version of the geometry in Proposition~\ref{prop:irr-wasted-weight}.

\paragraph{Discussions for multimodal conditional actions.}
The general bounds do not assume unimodality. A mode variable $M$ can instead be introduced and reference actions defined conditionally on $(O,M)$. For a finite-valued $M$,
\[
 \ent(C\mid O)=\ent(C\mid O,M)+I(C;M\mid O)
 \le \ent(C\mid O,M)+\ent(M\mid O).
\]
Modewise margin bounds control the first term and therefore also give an upper bound on the total conditional entropy. Reducing within-mode uncertainty does not alone establish a decrease in $\ent(C\mid O)$, because $I(C;M\mid O)$ can change with the tokenizer.

\paragraph{Summary.} 
Token margins provide a way to bound irreducible policy log-loss. In the linear fixed-target model, TBR through activation preconditioning implements BFL by accelerating the removal of
noise-subspace sensitivity, while preserving the clean-subspace partition in the controlled comparison. This increases margins and strictly reduces both conditional token entropy and Bayes binary token-classification error. These results identify one mechanism through which the BFL principle used by TBR can improve token predictability; the empirical analyses in Section~\ref{sec:exp-ablation} and Appendix~\ref{app:prediction-error-analysis} address its behavior in the full tokenizer.


\subsection{Token Balance Regularization: Implementation}
\label{app:balanced-feature-implementation}

\paragraph{Activation preconditioning in tokenizer training.}
We instantiate TBR with the activation-preconditioning component of DoPr~\citep{Zhang2026DoPr}, applied to the tokenizer's own optimization and to nothing else:
the downstream policy is trained afterwards on the frozen tokenizer, and the forward architecture, parameter count, and checkpoint format are unchanged.
Consider an eligible layer with weight $W\in\mathbb R^{d_{\mathrm{out}}\times d_{\mathrm{in}}}$ acting on an input activation $x\in\mathbb R^{d_{\mathrm{in}}}$ as $x\mapsto Wx$, and let $x_1,\ldots,x_n$ collect every activation that layer receives during one optimizer step.
We estimate the uncentered activation second moment and its trace-normalized ridge as
\begin{equation}
    \widehat\Sigma
      =\frac1n\sum_{t=1}^{n}x_tx_t^{\top},
    \qquad
    \rho
      =\lambda\,\frac{\operatorname{tr}\widehat\Sigma}{d_{\mathrm{in}}},
    \qquad
    \widehat\Sigma_\rho
      =\widehat\Sigma+\rho I_{d_{\mathrm{in}}},
    \label{eq:activation-second-moment}
\end{equation}
with damping $\lambda>0$. Normalizing the ridge by $\operatorname{tr}\widehat\Sigma/d_{\mathrm{in}}$ makes $\rho$ track each layer's activation scale,
so a single $\lambda$ transfers across layers and across tokenizer backbones.
Writing $G=\partial\mathcal L/\partial W$ for the raw weight gradient of the tokenizer objective, the preconditioned gradient is obtained by right-multiplication:
\begin{equation}
    \widetilde G
      =G\,\widehat\Sigma_\rho^{-1},
    \label{eq:activation-preconditioned-gradient}
\end{equation}
and the optimizer consumes $\widetilde G$ in place of $G$. Because $G=\sum_t\delta_tx_t^{\top}$ for output-side gradients $\delta_t$,
Equation~\ref{eq:activation-preconditioned-gradient} divides out the input-covariance factor that scales updates with activation energy,
equalizing progress across activation directions instead of letting a few high-variance features dominate.
This is the input-side factor of the layerwise curvature approximation used by K-FAC~\citep{Martens2015KFAC}.
We apply it inside an AdamW step, and we leave the codebook embeddings, biases, and normalization parameters to AdamW alone.
Within a training step the order is backward pass, activation preconditioning, global gradient-norm clipping, then the AdamW update.

\paragraph{Eligible layers and activation statistics.}
Preconditioning covers the linear, multi-head attention, and convolutional layers of both the tokenizer backbone and the quantizer.
For a linear layer, $x$ is its input and $d_{\mathrm{in}}$ its input width, pooled over all chunk positions in the minibatch.
For packed same-dimension multi-head attention, the query, key, and value inputs receive three separate factors,
and the corresponding row blocks of the packed input-projection gradient are preconditioned independently;
the output projection is excluded because it is reached through a functional path that does not expose its input.
For a convolution with a single group, $x$ is an extracted patch and $d_{\mathrm{in}}=C_{\mathrm{in}}k_hk_w$,
which treats the convolution as a linear map on patches.
Statistics are re-estimated from scratch at every optimizer step rather than accumulated in a running average, so $\widehat\Sigma$ always describes the current minibatch.

\paragraph{Wide layers and numerical fallbacks.}
When $d_{\mathrm{in}}\le d_{\max}$ we form $\widehat\Sigma_\rho$ explicitly and apply Equation~\ref{eq:activation-preconditioned-gradient} through its Cholesky factor. For wider layers, and for every convolution, we instead use a rank-one-plus-diagonal surrogate built from the activation mean and variance,
\begin{equation}
    \mu=\frac1n\sum_{t=1}^{n}x_t,
    \qquad
    v=\frac1n\sum_{t=1}^{n}x_t\odot x_t-\mu\odot\mu,
    \qquad
    \widehat\Sigma\approx\operatorname{diag}(v)+\mu\mu^{\top},
    \label{eq:rank-one-diagonal-surrogate}
\end{equation}
whose damped inverse follows from the Sherman--Morrison identity with $D=\operatorname{diag}(v+\rho)$,
\begin{equation}
    \widehat\Sigma_\rho^{-1}
      =D^{-1}
       -\frac{D^{-1}\mu\mu^{\top}D^{-1}}
             {1+\mu^{\top}D^{-1}\mu}.
    \label{eq:sherman-morrison-inverse}
\end{equation}
This costs $O(d_{\mathrm{in}})$ memory and time per layer instead of $O(d_{\mathrm{in}}^2)$ and $O(d_{\mathrm{in}}^3)$.
Two guards keep the step well defined. A low-rank activation batch can give $\widehat\Sigma$ small negative eigenvalues in single precision and make the Cholesky factorization fail;
we then symmetrize, take an eigendecomposition, clamp the eigenvalues at zero, and add $\rho$.
If any eligible gradient is non-finite, preconditioning is skipped for that step and the raw gradients are passed to clipping unchanged.
Table~\ref{tab:token-balance-hyperparameters} lists the damping $\lambda$ by benchmark and quantizer family and the maximum full-covariance dimension $d_{\max}$.


\input{tables/methods/token_balance_regularization/appendix_hyperparameters}
\subsection{Token Perturbation: Implementation and Alternative Sampling Studies}
\label{app:token-perturbation}

\paragraph{Perturbation sampling and training.}
During tokenizer training, TP perturbs the clean encoding $C=\mathcal T(A)$ using a conditional distribution $\nu(\cdot\mid C)$ to reconstruct $A$. Write $z_q=e(C)$, where $e$ is the token-embedding map. 
\begin{equation}
    C'\sim\nu(\cdot\mid C),
    \qquad
    z_q^{\mathrm{TP}}
      =z_q+\operatorname{sg}\!\left(e(C')-z_q\right),
    \label{eq:token-perturbation}
\end{equation}
where $\operatorname{sg}$ denotes stop-gradient. Forward decoding uses $e(C')$; the backward pass sends the identity gradient through clean $z_q$. For each batch,  we perturb each sample with Bernoulli probability $\beta$. Unperturbed samples retain clean decoder inputs. The reconstruction loss is evaluated on the mixed batch against the original actions. Codebook and commitment auxiliaries use the clean quantization pass and are not recomputed from $C'$; no separate adversarial or degradation loss is added.

\paragraph{Native-metric neighborhood sampling.}
Our main experiments use a sampler adapted from RobustTok~\citep{Qiu2025RobustLatent}. For each perturbed sample, $\nu_{\alpha,\delta}(\cdot\mid C)$ independently masks token positions with probability $\alpha$. Each masked token is sampled uniformly from the $\delta$ nearest candidates under Euclidean distance in the quantizer's latent space. For \textbf{VQ}, candidates are vectors in the applicable shared or level-specific codebook, ranked by distance to the encoder latent embedding. For \textbf{FSQ}, candidates are enumerated packed grid codes projected into the corresponding latent embedding. For \textbf{RVQ} and \textbf{NRVQ}, candidates are the level's code vectors and scaled normalized code vectors, respectively, ranked against its clean residual. \textbf{QinCo} likewise uses the clean residual, with a static codebook at level zero and prefix-conditioned codebooks thereafter. All residuals and conditioned codebooks use the clean prefix, even when earlier sampled tokens change; decoding uses the complete perturbed sequence.
Both $\alpha$ and $\delta$ are annealed linearly during training; Table~\ref{tab:token-perturbation-hyperparameters} lists the benchmark-specific settings.

\input{tables/methods/token_perturbation/appendix_hyperparameters}

\paragraph{Alternative perturbation and selection rules.}
We study two alternatives to test whether more coordinated perturbations or harder training examples improve robustness. \emph{Spectral-metric perturbation} changes how distances to candidate codes are measured: it randomly rotates the quantizer's spectral basis and assigns positive weights to the resulting directions. Candidate distance is then a weighted sum of squared differences along these directions, so a code that is farther away under ordinary Euclidean distance can become the nearest choice. The sampler requantizes every position using this modified distance. The random rotation and weighting are shared across levels, applied within each level's own basis; residual and conditioned quantizers update their state using the newly selected prefix. This construction is intended to coordinate changes across the sequence.
\emph{MaxUp} draws eight complete sequences from the native-metric neighborhood sampler and trains on the candidate with the largest reconstruction MSE. Following the worst-augmentation principle of MaxUp~\citep{Gong2020MaxUp}, this tests whether emphasizing difficult perturbations improves robustness beyond single-draw sampling.

\paragraph{Comparison results.}
We compare the alternatives on Robomimic Can. Here, TP denotes the native-metric sampling, and Alternative means either \emph{Spectral-metric perturbation} or applying \emph{MaxUp} on TP. We report best and last checkpoints separately at $H_{\mathrm{exec}}\in\{4,8\}$. 

\input{tables/methods/token_perturbation/appendix_controlled_pilots}

The performance of Spectral-metric perturbation is in general mixed relative to Token Perturbation. MaxUp remains above Reconstruction-only on average, yet is 1.37--3.90 points below the matched single-draw treatment.

\paragraph{Discussion.}
One possible explanation for the spectral variant's inconsistent gains is that changing the latent-space metric can therefore produce perturbations with different action-space consequences across quantizers. In residual and prefix-conditioned quantizers, an early replacement changes subsequent residuals or codebooks, potentially propagating its effect through the sequence. Codebook geometry may also matter: random rotations and reweighting can mix dominant and low-variance directions, changing neighborhood rankings in ways that may poorly reflect meaningful action differences. Therefore, it could disrupt useful structure rather than plausible token errors.

For MaxUp, selecting the largest reconstruction error among eight candidates emphasizes the high-error tail of the perturbation distribution. Neither candidate generation nor selection uses downstream-policy errors, so the hardest reconstruction candidate may not represent a likely policy mistake. Training on such candidates may overemphasize rare, severe changes and create a stronger trade-off between reconstruction fidelity and robustness. A severe perturbation may also produce a sequence representing a substantially different action, for which the original action is no longer an appropriate reconstruction target. Training the decoder to map such sequences back to the original action could blur distinctions between actions and compromise reconstruction fidelity. This could be particularly relevant when independently sampled replacements interact across token levels, although the mixed per-tokenizer outcomes do not establish a uniform family-level effect. These explanations remain hypotheses: distinguishing them would require examining codebook spectra, perturbation severity, and whether sampled sequences preserve action similarity and resemble policy-produced mismatches.

\clearpage

\section{Additional Experimental Results}
\label{app:additional-experimental-results}

\subsection{Primary Comparison: Task-Level Results and Robustness Checks}
\label{app:simulation-results}

This subsection expands Table~\ref{tab:simulation-multitask}. Appendix~\ref{app:simulation-reconstruction} reports validation reconstruction errors, Appendix~\ref{app:simulation-task-level} reports each task or suite separately, and Appendix~\ref{app:simulation-robustness} checks on Robomimic whether the comparison holds with the last checkpoint and with single-task policies. 

\subsubsection{Validation Reconstruction Error}
\label{app:simulation-reconstruction}

Table~\ref{tab:simulation-validation-reconstruction} reports tokenizer reconstruction MAE $\mathcal E_{\mathrm{rec}}$ on each benchmark's multi-task validation data for reconstruction-only training and \methodname{}. These errors measure reconstruction of the input action chunks in raw action units and are independent of the policy execution horizon. We report each benchmark separately because their action spaces differ.

\input{tables/experiments/simulation_results/appendix_validation_reconstruction}

\clearpage

\subsubsection{Task- and Suite-Level Results}
\label{app:simulation-task-level}

\paragraph{Robomimic.}
Table~\ref{tab:robomimic-multitask-by-task} reports Can, Lift, and Square separately for the multi-task setting at the checkpoint with the best teacher-forced validation accuracy. It compares Reconstruction-only with \methodname{} at both execution horizons and retains the confidence interval computed from each task's 300 binary rollout outcomes.

\input{tables/experiments/simulation_results/appendix_robomimic_multitask_best}

\clearpage
\paragraph{LIBERO.}
Table~\ref{tab:libero-multitask-by-suite} reports Spatial, Object, Goal, and LIBERO-10 separately for the multi-task setting. It compares Reconstruction-only with \methodname{} at both execution horizons and retains each suite's supplied confidence interval from 500 evaluation episodes.

\input{tables/experiments/simulation_results/appendix_libero_multitask_best}

\clearpage
\paragraph{RoboTwin.}
Table~\ref{tab:robotwin-multitask-by-task} reports Grab roller, Place burger/fries, and Place cans in the plastic box separately for the multi-task setting. It compares Reconstruction-only with \methodname{} at both execution horizons using 50 episodes per task and horizon.

\input{tables/experiments/simulation_results/appendix_robotwin_multitask_best}

\subsubsection{Robustness to Checkpoint and Data Setting}
\label{app:simulation-robustness}

The primary protocol fixes two reporting choices: it selects the policy checkpoint by teacher-forced validation accuracy and trains one policy jointly on all tasks of a benchmark. Neither choice is forced by the method, so we check both. Because repeating them on every simulator is computationally expensive, we restrict these auxiliary comparisons to Robomimic and keep every tokenizer configuration in each one.

\paragraph{Last checkpoint.}
Table~\ref{tab:robomimic-multitask-last} replaces the checkpoint with the best teacher-forced validation accuracy with the last one while holding multi-task training fixed. Averaging all tokenizer--task pairs, \methodname{} raises last-checkpoint success from 58.5\% to 73.9\% at $H_{\mathrm{exec}}=4$ and from 55.2\% to 71.3\% at $H_{\mathrm{exec}}=8$.

\input{tables/experiments/simulation_results/appendix_robomimic_multitask_last}

\clearpage
\paragraph{Single-task policies.}
Table~\ref{tab:robomimic-single-task} replaces multi-task training with policies trained and evaluated on one Robomimic task, again at the checkpoint with the best teacher-forced validation accuracy. Unlike the task-macro-averaged main table, these entries retain the confidence interval computed from the 300 binary rollout outcomes.

The primary conclusion also holds under single-task training: averaged equally over all tokenizer--task pairs, \methodname{} raises success from $68.2\%$ to $78.0\%$ at $H_{\mathrm{exec}}=4$ and from $63.8\%$ to $76.7\%$ at $H_{\mathrm{exec}}=8$.

\input{tables/experiments/simulation_results/appendix_robomimic_single_task}

\clearpage
\subsection{Detailed Ablation Results}
\label{app:ablation-results}

Tables~\ref{tab:robomimic-aggregate-ablation} and~\ref{tab:robomimic-family-ablation} summarize the results overall and by quantizer family. Tables~\ref{tab:robomimic-multitask-best-can}--\ref{tab:robomimic-multitask-best-square} report all four training conditions on Can, Lift, and Square, using the checkpoint with the best teacher-forced validation accuracy and both execution horizons.

\input{tables/experiments/simulation_results/appendix_robomimic_family_ablation}

\input{tables/experiments/simulation_results/appendix_robomimic_multitask_best_can}

\clearpage
\input{tables/experiments/simulation_results/appendix_robomimic_multitask_best_lift}

\clearpage
\input{tables/experiments/simulation_results/appendix_robomimic_multitask_best_square}

\clearpage
\subsection{Detailed Policy-Prediction Analysis}
\label{app:prediction-error-analysis}
\label{app:tokenizer-prediction-profiles}

We isolate TBR by comparing \methodname{} with \wotbr{} across all 22 valid tokenizer pairs. Let $\mathcal L_{\mathrm{val}}^m$ denote the teacher-forced policy validation loss at the checkpoint with the best teacher-forced validation accuracy for tokenizer treatment $m$:
\begin{equation}
    \Delta\mathcal L_{\mathrm{val}}^{\mathrm{TBR}}
      =\mathcal L_{\mathrm{val}}^{\methodname}
       -\mathcal L_{\mathrm{val}}^{\wotbr}.
    \label{eq:bfl-validation-loss-changes}
\end{equation}

\noindent
\begin{minipage}[t]{0.52\linewidth}
    \vspace{0pt}
    \centering
    \captionsetup{type=figure}
    \input{figures/experiments/prediction_error/appendix_validation_loss_by_family.tex}
    \caption{Family-mean validation-loss change after adding TBR; negative favors \methodname{}. Spans show min--max changes across tokenizers within each family.}
    \label{fig:component-error-analysis}
\end{minipage}\hfill
\begin{minipage}[t]{0.44\linewidth}
    \vspace{0pt}
    Relative to \wotbr{}, \methodname{} lowers mean validation loss by $0.655$ and raises mean token accuracy by $1.88$ percentage points. Each metric improves for 14 of 22 tokenizers. Family-mean loss decreases for VQ, NRVQ, FSQ, and QinCo, while RVQ increases slightly (Figure~\ref{fig:component-error-analysis}).

    \medskip
    Table~\ref{tab:policy-validation-metrics} reports averaged policy metrics; the curves below instead report per-position NLL normalized by $\log K$. Tokenizer treatments change the encoded targets, so these metrics describe each representation's induced prediction difficulty, rather than losses against identical labels.
\end{minipage}

\input{tables/experiments/prediction_error/appendix_validation_metrics}

\clearpage
\paragraph{Per-tokenizer token-prediction profiles.}
We show the paired normalized-NLL profiles underlying Figure~\ref{fig:token-position-nll-tp-ours}, retaining its tokenizer order.

\begin{figure}[H]
    \centering
    \input{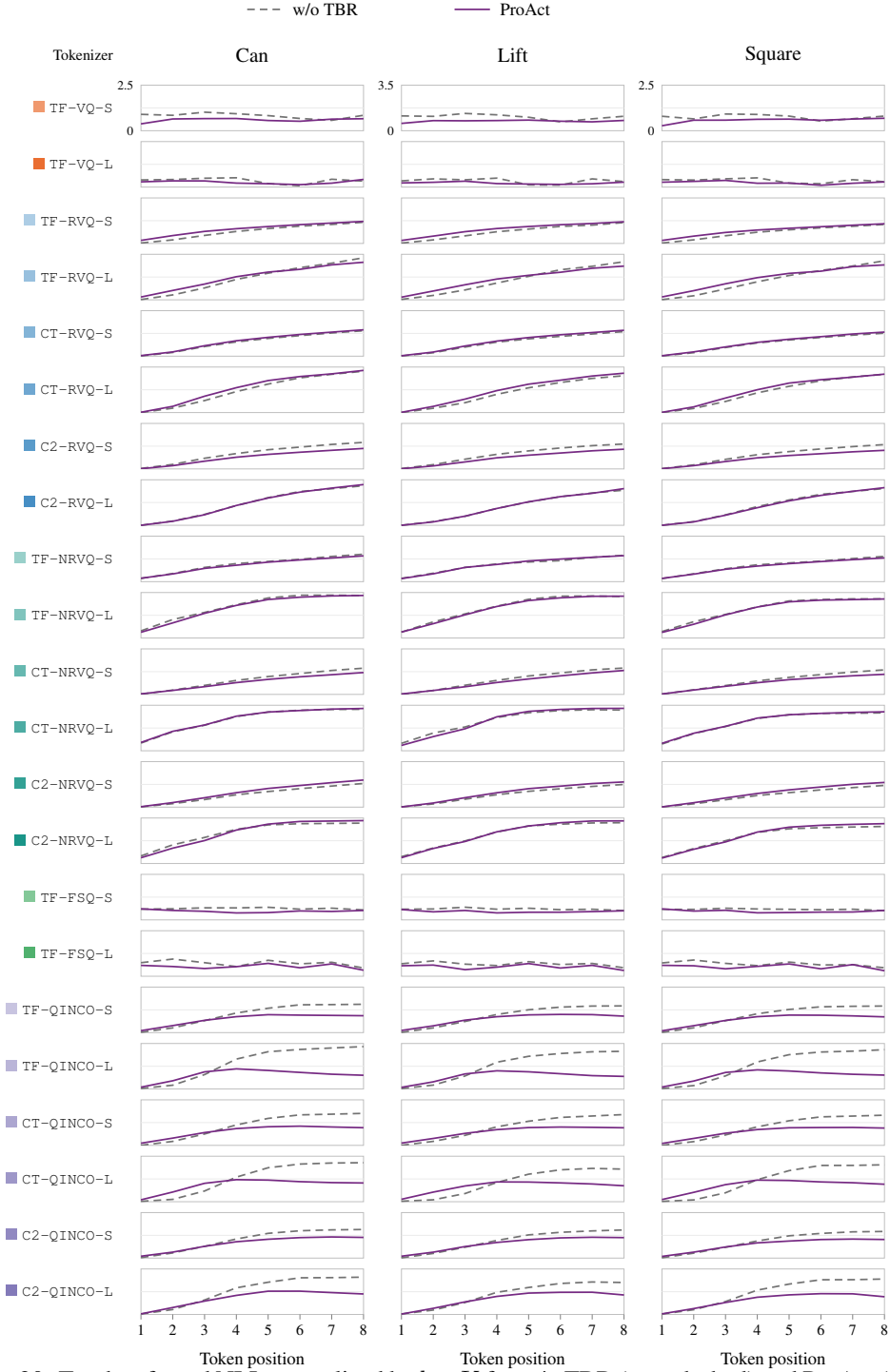}
    \caption{Teacher-forced NLL normalized by $\log K$ for \wotbr{} (gray dashed) and \methodname{} (purple solid). Each row is one tokenizer; columns are tasks. Within each task, all rows share the first row's labeled y-range.}
    \label{fig:per-tokenizer-nll}
\end{figure}

\clearpage
\subsection{Detailed Decoded-Error Analysis}
\label{app:decoded-error-analysis}

We isolate Token Perturbation through R.O. $\rightarrow$ \wotbr{} and \wotp{} $\rightarrow$ \methodname{} on validation demonstration pairs $(O,A)$. Greedy decoding is deterministic; $M=1$ draws one stochastic sequence (Appendix~\ref{app:one-step-error-proof}). For either control--treatment pair $(a,b)$, let $\widehat A_{\mathrm{greedy}}^{\pi,s}$ and $\widehat A^{\pi,s,(m)}$ denote method $s$'s greedy and $m$th sampled decoded actions. Using the normalized raw-action $L_1$ error $\ell$, define

\begin{equation}
\begin{aligned}
    \ell^{(M)}_s(A)
      &=\max_{1\leq m\leq M}
        \ell(A,\widehat A^{\pi,s,(m)}),\\[-1mm]
    \Delta\mathcal E_{\pi,\mathrm{greedy}}^{b-a}
      &=\mathbb E\!\left[
          \ell(A,\widehat A_{\mathrm{greedy}}^{\pi,b})
          -\ell(A,\widehat A_{\mathrm{greedy}}^{\pi,a})
        \right],\\[-1mm]
    \Delta\mathcal E_{\pi,\max-M}^{b-a}
      &=\mathbb E\!\left[
          \ell^{(M)}_b(A)-\ell^{(M)}_a(A)
        \right].
\end{aligned}
    \label{eq:decoded-error-changes}
\end{equation}

Here $M$ counts sequences sampled per policy, and negative changes favor TP. For Max-$M$, we retain only mismatches with each method's encoded target and require both methods to produce a mismatch on the same observation:

\begin{equation}
\begin{aligned}
\Delta\mathcal E_{\pi,\max-M}^{b-a}
  =\mathbb E\!\left[
      \max_{\substack{m\leq M:\\
          \widehat C^{b,(m)}\neq\mathcal T_b(A)}}
        \ell(A,\widehat A^{\pi,b,(m)})
      -
      \max_{\substack{m\leq M:\\
          \widehat C^{a,(m)}\neq\mathcal T_a(A)}}
        \ell(A,\widehat A^{\pi,a,(m)})
      \,\middle|\,
      \substack{
      \exists m\leq M:\widehat C^{a,(m)}\neq\mathcal T_a(A)\\
      \exists m\leq M:\widehat C^{b,(m)}\neq\mathcal T_b(A)}
    \right].
\end{aligned}
\label{eq:max-k-decoded-error}
\end{equation}

Max-$M$ probes the conditional error tail: for $n$ independent mismatch errors with conditional CDF $F_{\ne}$,

\begin{equation}
    \Pr\!\left[\max_{j\le n}E_j\le x\mid n\right]=F_{\ne}(x)^n.
    \label{eq:max-k-mismatch-error}
\end{equation}

Larger $M$ is more likely to expose severe mismatches. For R.O. $\rightarrow$ \wotbr{}, mean Greedy and Max-$2048$ changes are $-0.0046$ and $-0.0156$, improving 53 and 48 of 60 tokenizer--task conditions, respectively. For \wotp{} $\rightarrow$ \methodname{}, they are $-0.0086$ and $-0.0313$, improving 62 and 61 of 66 conditions. The larger average tail reductions are consistent with decoder robustness, but family effects vary: RVQ slightly worsens for R.O. $\rightarrow$ \wotbr{} under Max-$2048$, and VQ worsens for \wotp{} $\rightarrow$ \methodname{} under both estimators.

\begin{figure}[p]
    \centering
    \input{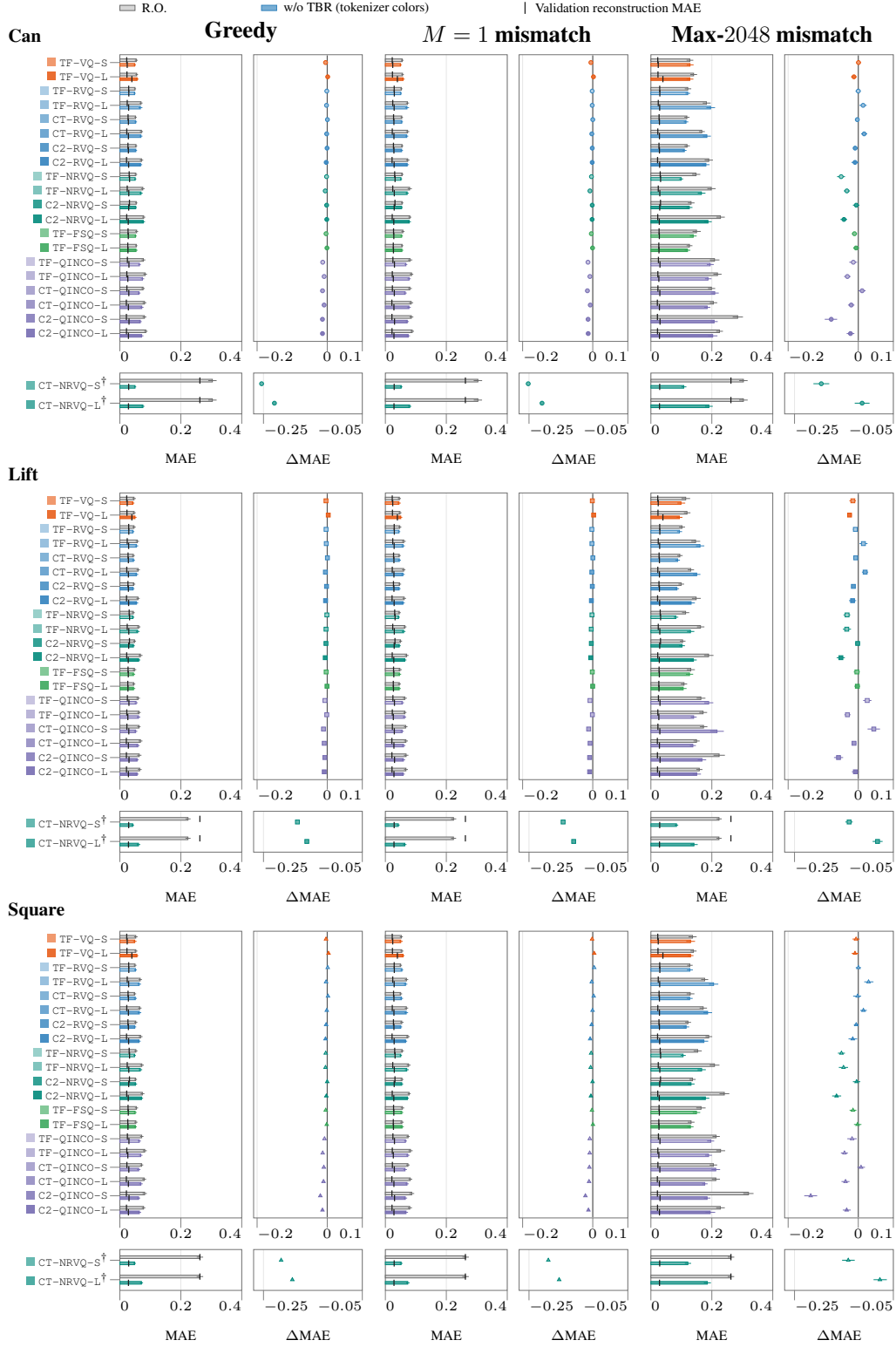}
    \captionsetup{skip=3pt}
    \caption{R.O. $\rightarrow$ \wotbr{} decoded-error details for Can, Lift, and Square (top to bottom). Each metric pairs absolute decoded MAE with its treatment-minus-control change ($\Delta$MAE). Gray and colored bars show R.O. and \wotbr{} errors with supplied method intervals; black ticks mark validation reconstruction MAE. Only Max-$2048$ changes have paired intervals. Separate daggered \textsc{CT-NRVQ-S/L} rows are excluded from the aggregate and use a distinct change-axis scale.}
    \label{fig:decoded-error-recon-tp}
    \label{fig:decoded-error-recon-tp-can}
    \label{fig:decoded-error-recon-tp-lift}
    \label{fig:decoded-error-recon-tp-square}
\end{figure}

\begin{figure}[p]
    \centering
    \input{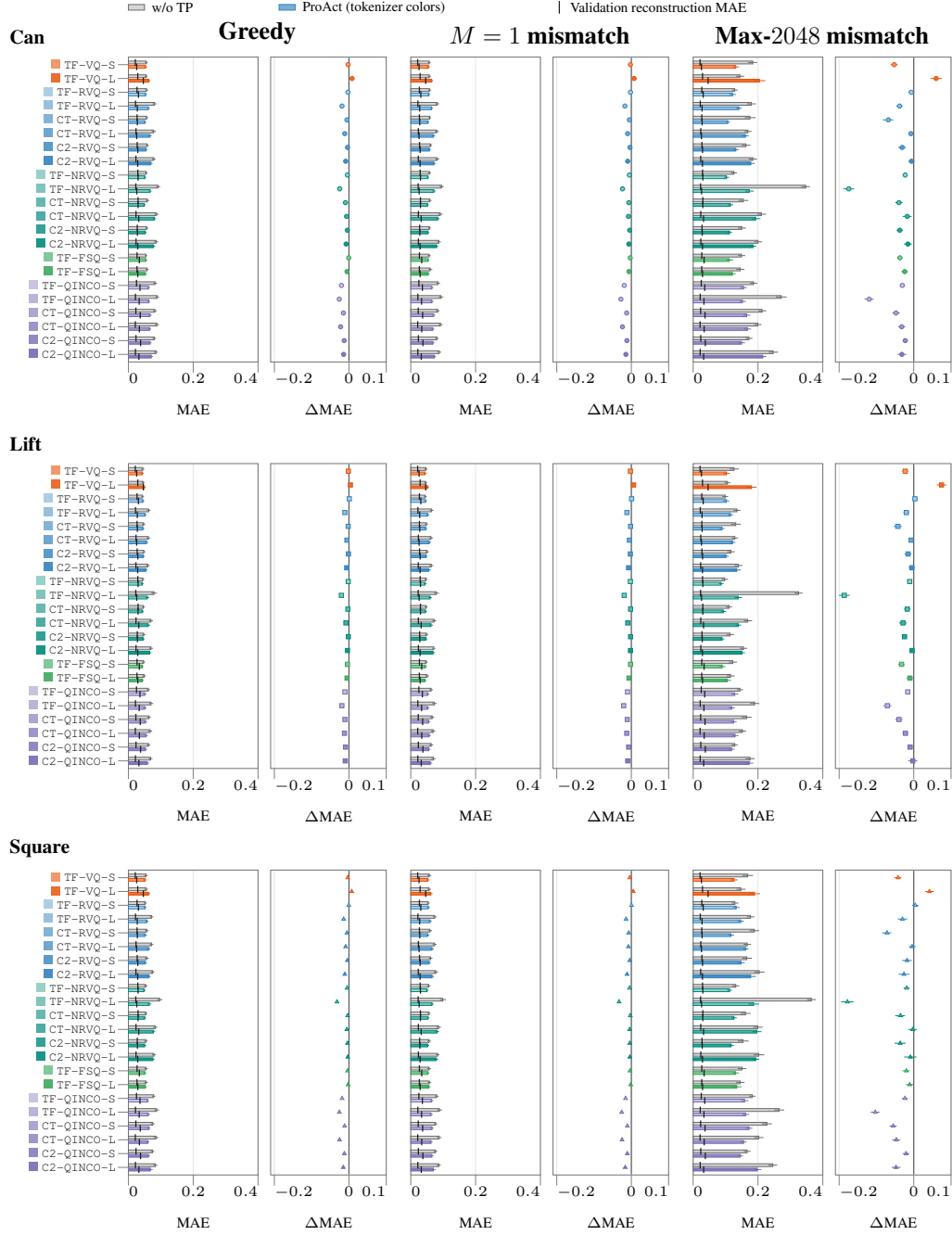}
    \captionsetup{skip=3pt}
    \caption{\wotp{} $\rightarrow$ \methodname{} decoded-error details for Can, Lift, and Square (top to bottom), retaining all 22 tokenizers. Each metric pairs absolute decoded MAE with its treatment-minus-control change ($\Delta$MAE). Gray and colored bars show \wotp{} and \methodname{} errors with supplied method intervals; black ticks mark validation reconstruction MAE. Only Max-$2048$ changes have paired intervals.}
    \label{fig:decoded-error-bfl-ours}
    \label{fig:decoded-error-bfl-ours-can}
    \label{fig:decoded-error-bfl-ours-lift}
    \label{fig:decoded-error-bfl-ours-square}
\end{figure}

Figures~\ref{fig:decoded-error-recon-tp}--\ref{fig:decoded-error-bfl-ours} give all tokenizer--task results. Absolute panels show MAE with supplied intervals and contextual reconstruction-error ticks; change panels show treatment minus control, with paired intervals only for Max-$2048$ (Greedy and $M=1$ are point estimates). Daggered \textsc{CT-NRVQ-S/L} rows are excluded only from the R.O.\ \wotbr aggregate; \wotp/\methodname includes all 22 configurations.

\begin{figure}[!t]
    \centering
    \captionsetup{skip=3pt}
    \input{figures/experiments/reconstruction_error/appendix_decoded_robustness_pooled.tex}
    \caption{Decoded-action robustness on full validation states: \wotbr{} minus R.O.. Solid curves (\texttt{CT-NRVQ-S/L} excluded) show paired Max-$M$ mismatch-error changes; darker bands are 95\% confidence intervals. Dashed lines show tokenizer validation reconstruction-error changes.}
    \label{fig:decoded-robustness-pooled-recon-tp}
\end{figure}
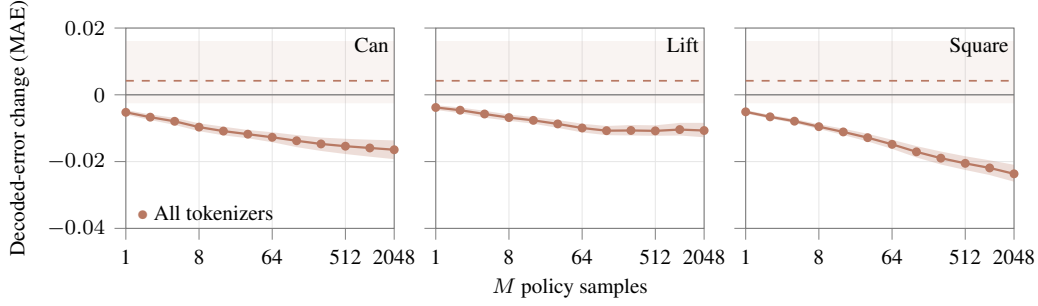

\clearpage

\clearpage

\clearpage
\subsection{Detailed Real-Robot Results}
\label{app:real-robot-results}

\paragraph{Shared setup and training.}
The physical UR5 experiments in Section~\ref{sec:real-robot-experiments} use a multi-task dataset of 780 demonstrations: 301 for Cube Stacking and 479 for Cup Hanging. For each tokenizer configuration and training objective, a single tokenizer and a single policy span both tasks; the same policy checkpoint is evaluated on each task. 

Table~\ref{tab:real-robot-hyperparameters} lists the \methodname{} tokenizer-training settings; implementation details follow Appendices~\ref{app:balanced-feature-implementation} and~\ref{app:token-perturbation}.

\begin{table}[H]
    \centering
    \captionsetup{skip=3pt}
    \caption{Real-robot \methodname{} hyperparameters. Arrows denote linear annealing with final scale $0.5$; $\beta$ remains fixed.}
    \label{tab:real-robot-hyperparameters}
    \tablebodyfont
    \setlength{\tabcolsep}{4pt}
    \renewcommand{\arraystretch}{1.06}
    \begin{tabular}{@{}llcc@{}}
        \toprule
        Component & Hyperparameter & Symbol & Value \\
        \midrule
        TBR & Activation-preconditioning damping & $\lambda$ & $0.1$ \\
            & Maximum full-covariance dimension & $d_{\max}$ & $512$ \\
        \addlinespace[3pt]
        TP & Sample-selection probability & $\beta$ & $0.1$ \\
           & Token-mask probability & $\alpha$ & $1.0\rightarrow0.5$ \\
           & Neighborhood size & $\delta$ & $10\rightarrow5$ \\
        \bottomrule
    \end{tabular}
\end{table}

\paragraph{Evaluation protocol.}
Action chunks have length $H=8$ and use the delta end effector pose representation, with nominal code capacity $K=128,\,L=8$. The policy executes four actions before replanning ($H_{\mathrm{exec}}=4$). Because the real-robot dataset is small, its held-out validation split is too small for teacher-forced validation accuracy to serve as a reliable checkpoint-selection signal; we therefore evaluate the final policy checkpoint for every condition, with 30 trials per task. Object initial poses are randomized for every trial.
Table~\ref{tab:ur5-results} reports the full per-task comparison in the main text. Across all five configurations, Cube Stacking successes increase from 28/150 (18.7\%) to 85/150 (56.7\%), and Cup Hanging successes increase from 28/150 (18.7\%) to 81/150 (54.0\%). Together, these give 56/300 (18.7\%) for Reconstruction-only and 166/300 (55.3\%) for \methodname{}. 


\end{document}

%% file: iclr2027/math_commands.tex
\usepackage{amsmath,amsfonts,bm}

\def\eqref#1{equation~\ref{#1}}

\def\1{\bm{1}}

\DeclareMathAlphabet{\mathsfit}{\encodingdefault}{\sfdefault}{m}{sl}
\SetMathAlphabet{\mathsfit}{bold}{\encodingdefault}{\sfdefault}{bx}{n}



%% file: figures/what_matters/reconstruction_error/main_validation_reconstruction_vs_success.tex
\ifdefined\appendixReconstructionScatterOnly
  \def\reconstructionPlotWidth{3.65cm}
  \def\reconstructionPlotHeight{2.65cm}
  \def\reconstructionPlotFont{\fontsize{7}{7.4}\selectfont}
  \def\reconstructionLegendHeight{2.65cm}
  \def\reconstructionLegendContentHeight{2.35cm}
  \def\reconstructionLegendShift{-1.15cm}
  \def\reconstructionXLabelOffset{-0.52cm}
\else
  \def\reconstructionPlotWidth{3.65cm}
  \def\reconstructionPlotHeight{1.85cm}
  \def\reconstructionPlotFont{\fontsize{6}{6.4}\selectfont}
  \def\reconstructionLegendHeight{1.85cm}
  \def\reconstructionLegendContentHeight{1.65cm}
  \def\reconstructionLegendShift{-1.30cm}
  \def\reconstructionXLabelOffset{-0.43cm}
\fi
\begin{tikzpicture}
\begin{groupplot}[
  group style={group size=3 by 1, horizontal sep=0.35cm,
    ylabels at=edge left, yticklabels at=edge left},
  width=\reconstructionPlotWidth, height=\reconstructionPlotHeight, scale only axis,
  xmin=0, xmax=0.135, ymin=0, ymax=100,
  xtick={0,0.04,0.08,0.12}, ytick={0,50,100},
  scaled x ticks=false,
  xticklabel style={/pgf/number format/fixed,
    /pgf/number format/precision=2, font=\reconstructionPlotFont},
  grid=major, grid style={draw=black!10, line width=0.2pt},
  axis line style={black!55, line width=0.35pt},
  tick style={black!55, line width=0.35pt},
  tick label style={font=\reconstructionPlotFont},
  label style={font=\reconstructionPlotFont},
  ylabel={Success (\%)},
]
\nextgroupplot[extra description/.code={
  \node[anchor=north east, fill=white, fill opacity=0.78,
    text opacity=1, inner sep=0.6pt,
    font=\reconstructionPlotFont]
    at (axis description cs:0.988,0.965) {Can};},]
\addplot+[only marks, mark=*, mark size=0.62pt,
  draw=vqbase, fill=vqbase, draw opacity=0.45,
  fill opacity=0.32] coordinates {(0.02177074,93.666667) (0.02209975,88.666667) (0.02244640,94.000000) (0.02316706,90.666667) (0.02343390,95.000000) (0.02354024,91.333333) (0.02354195,93.333333) (0.02366778,88.000000) (0.02469449,84.666667) (0.02477751,97.666667) (0.02496620,86.333333) (0.02505271,92.666667) (0.02553914,90.000000) (0.02558578,91.000000) (0.02578436,89.000000) (0.02586423,90.000000) (0.02629436,90.666667) (0.02635327,92.000000) (0.02640713,93.000000) (0.02684029,94.333333) (0.02687339,87.333333) (0.02776856,91.666667) (0.02848686,93.000000) (0.02982488,91.333333) (0.02987504,91.666667) (0.02991751,91.333333) (0.03347249,90.666667) (0.03450624,89.000000) (0.03645947,83.666667) (0.04080629,87.000000) (0.04191930,86.000000) (0.04390082,88.666667) (0.04444691,88.666667) (0.04739310,94.333333) (0.04876120,74.000000) (0.04934752,81.333333) (0.05293433,83.666667) (0.05306564,80.666667) (0.05342835,68.000000) (0.05393355,69.666667) (0.06008196,67.000000) (0.06368422,54.000000) (0.07468585,15.333333) (0.07475649,41.666667) (0.08018217,23.666667) (0.08125278,11.333333) (0.08577087,11.333333) (0.09553059,4.000000)};
\addplot+[only marks, mark=*, mark size=0.62pt,
  draw=rvqbase, fill=rvqbase, draw opacity=0.45,
  fill opacity=0.32] coordinates {(0.02348944,68.333333) (0.02398452,60.000000) (0.02428708,64.666667) (0.02450481,71.333333) (0.02462638,72.333333) (0.02472464,61.000000) (0.02496495,61.333333) (0.02498107,72.000000) (0.02524753,64.333333) (0.02525481,60.000000) (0.02537289,55.333333) (0.02548410,68.666667) (0.02581146,55.666667) (0.02583886,67.333333) (0.02588001,47.333333) (0.02588460,46.666667) (0.02598068,63.666667) (0.02603594,56.666667) (0.02603660,69.000000) (0.02679290,60.333333) (0.02694364,57.666667) (0.02743817,59.333333) (0.02751362,90.333333) (0.02752025,68.000000) (0.02786541,46.333333) (0.02792931,92.666667) (0.02810386,95.000000) (0.02827142,91.333333) (0.02827169,75.333333) (0.02832728,86.000000) (0.02843274,87.000000) (0.02858920,75.000000) (0.02869187,73.333333) (0.02878989,88.000000) (0.02900449,89.333333) (0.02909385,77.000000) (0.02919668,90.666667) (0.02942737,92.666667) (0.02948899,69.333333) (0.02961876,93.666667) (0.02971132,73.666667) (0.03001092,93.666667) (0.03012847,88.000000) (0.03016751,91.666667) (0.03028256,94.333333) (0.03028457,93.666667) (0.03038959,92.666667) (0.03041981,69.000000) (0.03045323,77.000000) (0.03047706,95.666667) (0.03068426,74.666667) (0.03086593,71.666667) (0.03086893,77.666667) (0.03093988,89.333333) (0.03098346,88.000000) (0.03102476,93.000000) (0.03107411,94.000000) (0.03108271,92.000000) (0.03137019,91.333333) (0.03139162,92.333333) (0.03164276,92.000000) (0.03182193,95.666667) (0.03188819,91.333333) (0.03208945,76.666667) (0.03216325,92.000000) (0.03229166,93.666667) (0.03234699,95.666667) (0.03241727,88.333333) (0.03251643,93.333333) (0.03254334,92.000000) (0.03270280,94.666667) (0.03285778,91.333333) (0.03292578,96.333333) (0.03297271,90.000000) (0.03307785,67.333333) (0.03314365,95.666667) (0.03317572,77.666667) (0.03318249,92.000000) (0.03341446,81.333333) (0.03345526,77.000000) (0.03383786,69.333333) (0.03465798,96.333333) (0.03495385,93.666667) (0.03508670,83.666667) (0.03532011,91.666667) (0.03706969,93.333333) (0.04239700,78.666667) (0.04447270,92.000000) (0.04496905,82.666667) (0.04531507,80.333333) (0.04635602,80.666667) (0.04653024,86.000000) (0.04666635,89.000000) (0.04676768,83.333333) (0.04734460,83.333333) (0.04750518,81.666667) (0.04891210,85.000000) (0.04921271,87.333333) (0.05061847,75.333333) (0.05172738,73.666667)};
\addplot+[only marks, mark=*, mark size=0.62pt,
  draw=nrvqbase, fill=nrvqbase, draw opacity=0.45,
  fill opacity=0.32] coordinates {(0.02260288,51.333333) (0.02347576,50.333333) (0.02356593,52.000000) (0.02446434,51.000000) (0.02487055,53.000000) (0.02645264,63.333333) (0.02837088,48.333333) (0.02856759,47.666667) (0.02870987,61.666667) (0.02882365,56.333333) (0.02920366,28.000000) (0.02925207,44.000000) (0.02935059,56.666667) (0.02947666,50.333333) (0.02951255,88.000000) (0.02972526,54.333333) (0.02985964,93.000000) (0.02996113,59.666667) (0.02997368,94.000000) (0.03020120,47.333333) (0.03041445,56.666667) (0.03043502,65.666667) (0.03045861,89.000000) (0.03052444,54.000000) (0.03052918,95.333333) (0.03057973,62.333333) (0.03058657,91.333333) (0.03063267,89.000000) (0.03089029,93.666667) (0.03133546,91.000000) (0.03135399,56.666667) (0.03136583,90.000000) (0.03180828,88.333333) (0.03183383,46.333333) (0.03185219,47.666667) (0.03194282,91.333333) (0.03208219,93.333333) (0.03221648,87.666667) (0.03224048,91.333333) (0.03240275,89.666667) (0.03260221,92.000000) (0.03269967,92.333333) (0.03275510,86.666667) (0.03277645,56.333333) (0.03280744,35.666667) (0.03287090,84.666667) (0.03292327,72.000000) (0.03309676,96.333333) (0.03324192,80.333333) (0.03336015,94.000000) (0.03343302,91.666667) (0.03351165,92.000000) (0.03370007,94.000000) (0.03374611,37.333333) (0.03376131,90.666667) (0.03384825,93.333333) (0.03393831,90.333333) (0.03398169,88.333333) (0.03401801,73.000000) (0.03410873,43.000000) (0.03411364,40.666667) (0.03424843,69.333333) (0.03427252,58.333333) (0.03430538,66.000000) (0.03437385,90.333333) (0.03440970,55.666667) (0.03443177,96.333333) (0.03468208,35.333333) (0.03509895,52.333333) (0.03517160,94.666667) (0.03519122,92.666667) (0.03621764,85.333333) (0.03625985,43.000000) (0.03698568,37.333333) (0.03781341,89.666667) (0.03823121,90.333333) (0.03843553,90.333333) (0.04236874,70.000000) (0.04273100,66.666667) (0.04452558,74.000000) (0.04482924,77.000000) (0.04608174,79.666667) (0.04693946,69.000000) (0.04749391,85.333333) (0.04840597,88.000000) (0.04918765,87.000000)};
\addplot+[only marks, mark=*, mark size=0.62pt,
  draw=fsqbase, fill=fsqbase, draw opacity=0.45,
  fill opacity=0.32] coordinates {(0.02729423,91.666667) (0.02754472,91.666667) (0.02759309,89.333333) (0.02765827,90.333333) (0.02784346,91.666667) (0.02793458,90.000000) (0.02805502,96.000000) (0.02805826,92.000000) (0.02821795,93.000000) (0.02851864,89.333333) (0.02855287,88.666667) (0.02857257,89.333333) (0.02863970,89.666667) (0.02866222,91.000000) (0.02866404,92.000000) (0.02883589,95.000000) (0.02904847,88.000000) (0.02990174,93.333333) (0.02990945,92.000000) (0.03002641,95.666667) (0.03020268,83.333333) (0.03031944,90.666667) (0.03032323,92.000000) (0.03036835,94.333333) (0.03078102,89.333333) (0.03084295,97.000000) (0.03086964,91.666667) (0.03176841,92.000000) (0.03190702,91.000000) (0.03195712,93.666667) (0.03233492,94.000000) (0.03238032,92.333333) (0.03292334,93.000000) (0.03297770,95.333333) (0.03367001,93.333333) (0.03395314,93.666667) (0.03402163,94.666667) (0.03429865,95.000000) (0.03430264,93.000000) (0.03435981,90.333333) (0.03481535,95.000000) (0.03649139,94.333333) (0.04511941,82.333333) (0.04638779,82.333333) (0.04895826,77.333333) (0.04991211,79.666667)};
\addplot+[only marks, mark=*, mark size=0.62pt,
  draw=qincobase, fill=qincobase, draw opacity=0.45,
  fill opacity=0.32] coordinates {(0.02256171,50.333333) (0.02283113,45.666667) (0.02306333,51.333333) (0.02347083,52.666667) (0.02351151,54.000000) (0.02360458,52.333333) (0.02378904,50.666667) (0.02384813,45.333333) (0.02389796,54.000000) (0.02390913,44.333333) (0.02391855,50.666667) (0.02397519,41.666667) (0.02398163,53.333333) (0.02408398,52.333333) (0.02415209,55.666667) (0.02420277,44.000000) (0.02426473,61.000000) (0.02428678,31.000000) (0.02430556,45.000000) (0.02445199,39.333333) (0.02457642,43.666667) (0.02468164,39.666667) (0.02473129,41.666667) (0.02486119,61.333333) (0.02492240,59.333333) (0.02496576,43.333333) (0.02504671,50.000000) (0.02507234,66.333333) (0.02518721,40.666667) (0.02529290,62.666667) (0.02551356,54.333333) (0.02563991,67.333333) (0.02564950,57.000000) (0.02579032,61.000000) (0.02590852,64.333333) (0.02603214,64.000000) (0.02639253,65.000000) (0.02665141,75.666667) (0.02703788,68.333333) (0.02711917,53.666667) (0.02726673,63.000000) (0.02769503,56.000000) (0.02771032,45.666667) (0.02816470,68.666667) (0.02823241,71.333333) (0.02832111,50.666667) (0.02854706,52.333333) (0.02859562,54.000000) (0.02878598,61.333333) (0.02891615,57.333333) (0.02899308,46.000000) (0.02915281,54.666667) (0.02915552,59.000000) (0.03021060,61.000000) (0.03072302,76.666667) (0.03080993,65.666667) (0.03100779,67.000000) (0.03136025,70.000000) (0.03163011,75.666667) (0.03176286,60.666667) (0.03243984,52.666667) (0.03284203,56.333333) (0.03284926,70.666667) (0.03334672,79.333333) (0.03400524,82.000000) (0.03403142,73.666667) (0.03410127,73.666667) (0.03485307,84.666667) (0.03502358,80.666667) (0.03556359,59.333333) (0.03652825,87.666667) (0.03671005,78.000000) (0.03671662,74.000000) (0.03672040,76.666667) (0.03765027,80.333333) (0.03767762,88.333333) (0.03788932,86.666667) (0.03789286,92.333333) (0.03806563,88.000000) (0.04159470,62.666667) (0.04168978,75.000000) (0.04173664,69.666667) (0.04246368,88.666667) (0.04404230,59.000000) (0.04418935,73.000000) (0.04476402,69.333333) (0.04709431,62.333333) (0.04733646,68.666667) (0.04743969,60.000000) (0.04813423,64.333333) (0.04825115,78.666667) (0.04869004,73.000000) (0.04871019,71.333333) (0.04934074,78.666667) (0.11014960,5.000000)};
\ifdefined\appendixReconstructionScatterOnly\else
\addplot[black, solid, line width=0.62pt, mark=diamond*,
  mark size=1.35pt, mark options={solid, draw=black, fill=white}]
  coordinates {(0.00000000,97.000000) (0.00428571,96.666667) (0.00857143,98.333333) (0.01285714,97.333333) (0.01714286,95.000000) (0.02142857,96.666667) (0.02571429,94.333333) (0.03000000,93.666667) (0.03428571,91.333333) (0.03857143,89.333333) (0.04285714,87.333333) (0.08571429,54.000000) (0.12857143,17.333333)};
\fi
\nextgroupplot[extra description/.code={
  \node[anchor=north east, fill=white, fill opacity=0.78,
    text opacity=1, inner sep=0.6pt,
    font=\reconstructionPlotFont]
    at (axis description cs:0.988,0.965) {Lift};},]
\addplot+[only marks, mark=*, mark size=0.62pt,
  draw=vqbase, fill=vqbase, draw opacity=0.45,
  fill opacity=0.32] coordinates {(0.02003136,98.333333) (0.02015980,90.000000) (0.02068917,91.666667) (0.02070285,93.000000) (0.02205708,98.333333) (0.02229138,99.000000) (0.02266154,95.000000) (0.02318682,97.666667) (0.02415088,96.333333) (0.02548922,95.666667) (0.02790197,98.666667) (0.02870759,98.333333) (0.03030386,99.333333) (0.03547833,95.000000) (0.03722124,95.666667) (0.03772391,98.666667) (0.03799803,97.333333) (0.03939824,97.000000) (0.03979339,93.666667) (0.03998317,97.333333) (0.04023505,98.666667) (0.04159456,94.666667) (0.04224774,89.000000) (0.04858917,96.666667)};
\addplot+[only marks, mark=*, mark size=0.62pt,
  draw=rvqbase, fill=rvqbase, draw opacity=0.45,
  fill opacity=0.32] coordinates {(0.02156265,89.666667) (0.02180869,88.000000) (0.02195851,87.333333) (0.02197014,90.000000) (0.02243853,74.666667) (0.02249485,77.000000) (0.02250162,92.333333) (0.02253355,88.333333) (0.02284710,90.666667) (0.02289385,74.000000) (0.02313881,85.333333) (0.02321563,87.666667) (0.02323501,85.333333) (0.02353426,90.333333) (0.02357111,81.666667) (0.02395445,91.333333) (0.02436764,84.333333) (0.02448311,87.333333) (0.02462369,94.333333) (0.02470127,96.000000) (0.02474907,90.333333) (0.02482557,95.666667) (0.02491220,100.000000) (0.02501073,92.333333) (0.02502012,90.333333) (0.02518335,96.666667) (0.02529328,89.333333) (0.02532651,85.333333) (0.02550146,92.333333) (0.02593570,99.666667) (0.02597616,92.666667) (0.02626887,93.000000) (0.02633056,87.666667) (0.02639029,97.000000) (0.02640929,98.000000) (0.02650913,97.666667) (0.02655137,94.333333) (0.02656654,99.333333) (0.02661491,96.333333) (0.02692449,94.000000) (0.02693112,95.333333) (0.02695489,93.333333) (0.02712489,95.000000) (0.02716681,89.666667) (0.02765209,97.666667) (0.02788447,97.000000) (0.02795561,90.000000) (0.02834637,97.000000) (0.02852573,88.333333) (0.02854253,97.333333) (0.02886889,89.666667) (0.02972717,89.333333) (0.03001641,94.000000) (0.03192360,97.666667) (0.03529409,92.000000) (0.03559823,93.333333) (0.03571446,92.000000) (0.03652055,92.666667) (0.03793498,96.666667) (0.03900411,97.000000) (0.03987722,92.333333) (0.04027575,90.666667) (0.04068968,92.000000) (0.04138584,93.666667) (0.04171816,92.000000) (0.04259897,94.333333)};
\addplot+[only marks, mark=*, mark size=0.62pt,
  draw=nrvqbase, fill=nrvqbase, draw opacity=0.45,
  fill opacity=0.32] coordinates {(0.01991514,68.000000) (0.02015254,76.000000) (0.02029355,74.666667) (0.02056760,71.666667) (0.02253204,83.333333) (0.02292331,74.666667) (0.02298514,83.333333) (0.02391404,96.666667) (0.02410443,89.666667) (0.02438574,81.666667) (0.02449730,81.000000) (0.02459088,76.666667) (0.02526410,97.333333) (0.02539659,91.666667) (0.02569216,97.666667) (0.02580052,79.333333) (0.02620737,74.333333) (0.02627325,94.666667) (0.02637738,99.000000) (0.02640540,90.333333) (0.02652558,92.666667) (0.02657038,95.333333) (0.02659623,84.333333) (0.02673782,98.666667) (0.02674570,69.333333) (0.02679650,81.333333) (0.02682828,96.000000) (0.02688566,70.000000) (0.02704396,99.333333) (0.02720965,97.666667) (0.02728382,98.333333) (0.02751133,94.333333) (0.02778665,93.000000) (0.02813330,97.666667) (0.02813436,98.333333) (0.02881331,71.000000) (0.02906660,67.333333) (0.03002918,98.333333) (0.03198568,96.333333) (0.03273635,88.000000) (0.03324122,89.333333) (0.03341199,98.000000) (0.03367806,86.333333) (0.03392787,77.333333) (0.03427370,92.333333) (0.03434363,93.666667) (0.03514261,84.000000) (0.03534340,94.666667) (0.03547865,94.333333) (0.03586856,95.333333) (0.03590700,96.666667) (0.03600910,76.333333) (0.03603916,76.000000) (0.03706365,77.333333) (0.03765249,85.666667) (0.03825638,87.666667) (0.03831863,73.000000) (0.03849544,69.000000) (0.03919578,74.333333) (0.03929932,88.666667) (0.03936372,62.666667) (0.03941532,94.333333) (0.03961623,90.000000) (0.04139984,78.666667)};
\addplot+[only marks, mark=*, mark size=0.62pt,
  draw=fsqbase, fill=fsqbase, draw opacity=0.45,
  fill opacity=0.32] coordinates {(0.02367191,99.333333) (0.02376146,99.000000) (0.02397210,96.666667) (0.02428537,97.333333) (0.02443883,98.333333) (0.02447251,99.000000) (0.02480140,99.000000) (0.02553767,96.666667) (0.02555467,99.000000) (0.02579706,97.333333) (0.02598581,98.000000) (0.02697431,100.000000) (0.02727093,98.333333) (0.02739137,98.333333) (0.02800736,96.000000) (0.02807896,97.000000) (0.02844621,98.666667) (0.02864792,98.666667) (0.03631472,97.666667) (0.03671036,96.666667) (0.03871506,92.666667) (0.03994464,86.666667)};
\addplot+[only marks, mark=*, mark size=0.62pt,
  draw=qincobase, fill=qincobase, draw opacity=0.45,
  fill opacity=0.32] coordinates {(0.01990622,77.000000) (0.02030039,81.666667) (0.02035363,79.666667) (0.02047066,81.666667) (0.02054162,79.666667) (0.02061678,77.333333) (0.02070485,72.000000) (0.02075229,76.666667) (0.02077532,75.000000) (0.02082252,73.333333) (0.02091343,81.000000) (0.02121928,71.666667) (0.02128435,79.666667) (0.02144174,82.333333) (0.02162034,60.333333) (0.02190195,78.666667) (0.02192267,82.666667) (0.02197417,80.000000) (0.02212134,86.000000) (0.02234299,88.333333) (0.02235025,77.000000) (0.02235642,80.333333) (0.02281601,84.000000) (0.02283674,79.000000) (0.02283972,74.000000) (0.02288601,79.000000) (0.02290697,87.333333) (0.02315066,77.000000) (0.02328739,75.333333) (0.02376116,75.333333) (0.02384939,78.333333) (0.02394402,82.666667) (0.02409785,81.000000) (0.02452767,80.000000) (0.02465730,84.000000) (0.02480572,80.666667) (0.02629533,92.000000) (0.02657327,86.333333) (0.02675882,82.666667) (0.02694648,85.666667) (0.02730063,87.333333) (0.02740274,83.666667) (0.02785628,86.666667) (0.02803704,88.000000) (0.02803858,91.333333) (0.02814725,91.000000) (0.02819923,87.333333) (0.02824701,97.000000) (0.02862924,96.000000) (0.02878260,98.000000) (0.02942093,96.333333) (0.03035654,86.666667) (0.03054457,91.666667) (0.03415819,81.666667) (0.03467647,85.666667) (0.03501029,86.666667) (0.03578804,86.000000) (0.03857145,87.000000) (0.03861829,84.000000) (0.03923600,87.666667) (0.03948309,81.666667) (0.03990473,88.666667) (0.04007799,91.000000) (0.04069104,96.666667) (0.04112524,87.000000) (0.06123085,79.666667)};
\ifdefined\appendixReconstructionScatterOnly\else
\addplot[black, solid, line width=0.62pt, mark=diamond*,
  mark size=1.35pt, mark options={solid, draw=black, fill=white}]
  coordinates {(0.00000000,100.000000) (0.00428571,100.000000) (0.00857143,100.000000) (0.01285714,100.000000) (0.01714286,100.000000) (0.02142857,100.000000) (0.02571429,100.000000) (0.03000000,100.000000) (0.03428571,100.000000) (0.03857143,100.000000) (0.04285714,100.000000) (0.08571429,95.000000) (0.12857143,64.333333)};
\fi
\nextgroupplot[extra description/.code={
  \node[anchor=north east, fill=white, fill opacity=0.78,
    text opacity=1, inner sep=0.6pt,
    font=\reconstructionPlotFont]
    at (axis description cs:0.988,0.965) {Square};},]
\addplot+[only marks, mark=*, mark size=0.62pt,
  draw=vqbase, fill=vqbase, draw opacity=0.45,
  fill opacity=0.32] coordinates {(0.02234215,69.333333) (0.02261755,66.666667) (0.02326804,66.000000) (0.02383515,69.000000) (0.02489709,61.000000) (0.02552246,61.000000) (0.02587929,62.333333) (0.02600140,62.666667) (0.02916416,68.000000) (0.02968167,71.333333) (0.02977458,69.333333) (0.03241363,67.333333) (0.03796225,63.333333) (0.04282623,62.000000) (0.05053898,50.666667) (0.05165768,52.000000) (0.05291602,60.333333) (0.05316971,50.333333) (0.05349253,52.000000) (0.05601882,47.666667) (0.05611066,34.666667) (0.06028522,35.666667) (0.06977737,5.000000) (0.07006383,41.000000)};
\addplot+[only marks, mark=*, mark size=0.62pt,
  draw=rvqbase, fill=rvqbase, draw opacity=0.45,
  fill opacity=0.32] coordinates {(0.02410176,25.000000) (0.02454357,26.333333) (0.02475948,29.333333) (0.02477814,42.000000) (0.02486900,22.666667) (0.02492261,36.000000) (0.02493796,34.666667) (0.02536970,33.666667) (0.02539196,33.333333) (0.02554216,27.000000) (0.02591750,35.000000) (0.02591750,33.000000) (0.02601925,32.666667) (0.02615064,31.666667) (0.02633995,35.333333) (0.02703546,41.333333) (0.02709607,49.000000) (0.02715069,43.000000) (0.02724176,49.000000) (0.02726233,44.666667) (0.02775210,74.000000) (0.02790210,72.000000) (0.02797028,60.666667) (0.02816819,71.666667) (0.02830390,70.333333) (0.02841399,42.333333) (0.02847353,58.333333) (0.02850331,45.333333) (0.02867648,73.000000) (0.02871386,69.333333) (0.02880806,67.000000) (0.02924408,65.666667) (0.02941125,40.666667) (0.02951895,66.333333) (0.02963292,61.666667) (0.02964262,69.666667) (0.02964720,71.333333) (0.02965302,63.666667) (0.02966402,69.000000) (0.03003563,74.333333) (0.03043450,73.333333) (0.03060491,74.000000) (0.03067433,71.666667) (0.03069128,45.666667) (0.03078830,73.666667) (0.03096367,69.000000) (0.03098330,45.000000) (0.03113065,60.666667) (0.03117106,49.333333) (0.03168628,67.666667) (0.03180016,70.333333) (0.03251520,48.333333) (0.03275800,68.666667) (0.03481970,66.333333) (0.04368111,54.000000) (0.04436558,66.666667) (0.04504055,58.333333) (0.04673477,65.666667) (0.04691772,53.333333) (0.04783748,48.333333) (0.04836843,53.000000) (0.04922578,48.000000) (0.05060019,58.000000) (0.05179549,45.000000) (0.05238445,45.333333) (0.05536420,40.000000)};
\addplot+[only marks, mark=*, mark size=0.62pt,
  draw=nrvqbase, fill=nrvqbase, draw opacity=0.45,
  fill opacity=0.32] coordinates {(0.02286470,21.666667) (0.02314518,25.000000) (0.02324052,23.666667) (0.02329838,25.000000) (0.02402932,16.333333) (0.02463671,24.000000) (0.02736518,25.333333) (0.02754839,19.000000) (0.02762529,28.666667) (0.02784950,19.333333) (0.02819663,37.000000) (0.02834720,11.666667) (0.02846678,25.000000) (0.02867295,19.333333) (0.02871473,31.000000) (0.02880284,20.666667) (0.02910943,21.000000) (0.02929383,24.666667) (0.02942019,73.333333) (0.02956861,77.000000) (0.02964590,58.666667) (0.02976197,76.000000) (0.02981571,73.333333) (0.03014475,70.333333) (0.03109993,63.333333) (0.03110230,76.333333) (0.03113680,59.333333) (0.03134534,63.000000) (0.03167364,65.666667) (0.03168733,63.333333) (0.03174090,71.000000) (0.03198401,73.666667) (0.03209322,66.000000) (0.03216525,67.333333) (0.03216546,69.333333) (0.03243450,25.333333) (0.03252098,67.666667) (0.03257504,29.000000) (0.03324804,67.666667) (0.03342229,18.666667) (0.03368782,14.333333) (0.03372723,75.333333) (0.03408827,18.000000) (0.03409332,62.666667) (0.03581841,72.000000) (0.03886157,16.666667) (0.04180280,44.333333) (0.04264782,48.666667) (0.04537988,61.666667) (0.04665446,46.666667) (0.04758598,58.666667) (0.04780250,61.333333) (0.04938277,58.333333) (0.04990637,49.000000)};
\addplot+[only marks, mark=*, mark size=0.62pt,
  draw=fsqbase, fill=fsqbase, draw opacity=0.45,
  fill opacity=0.32] coordinates {(0.02672680,67.000000) (0.02680518,61.666667) (0.02706648,61.666667) (0.02787759,78.000000) (0.02815905,71.333333) (0.02844137,73.333333) (0.02896679,76.000000) (0.02901016,71.666667) (0.02924518,66.333333) (0.02966483,66.333333) (0.02980508,63.333333) (0.03022574,71.000000) (0.03063296,71.333333) (0.03090325,71.666667) (0.03168223,70.000000) (0.03173762,71.333333) (0.03296146,76.000000) (0.03344398,70.333333) (0.04431601,58.333333) (0.04562695,63.666667) (0.05010296,57.333333) (0.05089945,49.666667)};
\addplot+[only marks, mark=*, mark size=0.62pt,
  draw=qincobase, fill=qincobase, draw opacity=0.45,
  fill opacity=0.32] coordinates {(0.02205495,5.333333) (0.02239834,15.666667) (0.02261296,2.000000) (0.02282642,13.333333) (0.02294031,16.000000) (0.02300065,15.000000) (0.02302245,14.666667) (0.02304049,18.666667) (0.02309633,12.000000) (0.02361707,25.333333) (0.02388722,11.000000) (0.02393305,11.666667) (0.02393901,21.000000) (0.02398159,14.000000) (0.02431791,12.333333) (0.02447025,28.000000) (0.02469152,27.666667) (0.02474205,25.333333) (0.02487694,26.333333) (0.02494499,26.666667) (0.02508059,13.000000) (0.02518049,27.666667) (0.02592719,29.000000) (0.02623317,27.666667) (0.02627173,30.666667) (0.02639815,33.000000) (0.02649628,28.333333) (0.02667937,18.000000) (0.02686976,25.666667) (0.02701331,23.333333) (0.02702064,32.000000) (0.02702974,17.333333) (0.02703062,20.666667) (0.02715259,32.000000) (0.02724893,18.666667) (0.02743396,28.333333) (0.02937146,42.000000) (0.03096792,50.000000) (0.03137165,46.333333) (0.03141856,40.333333) (0.03186513,31.666667) (0.03186758,49.333333) (0.03287421,66.333333) (0.03301012,39.000000) (0.03328550,49.000000) (0.03343075,48.333333) (0.03360122,55.000000) (0.03369375,53.000000) (0.03477606,54.333333) (0.03508207,53.666667) (0.03531728,53.666667) (0.03550452,47.333333) (0.03676600,49.666667) (0.04297575,25.333333) (0.04348085,35.333333) (0.04493702,29.333333) (0.04569170,33.000000) (0.04862101,37.000000) (0.04883935,43.000000) (0.04884692,42.000000) (0.04970183,41.000000) (0.05009949,39.000000) (0.05028857,43.666667) (0.05049051,37.000000) (0.05178517,39.333333) (0.09323193,1.000000)};
\ifdefined\appendixReconstructionScatterOnly\else
\addplot[black, solid, line width=0.62pt, mark=diamond*,
  mark size=1.35pt, mark options={solid, draw=black, fill=white}]
  coordinates {(0.00000000,79.000000) (0.00428571,83.333333) (0.00857143,81.000000) (0.01285714,79.000000) (0.01714286,82.333333) (0.02142857,79.666667) (0.02571429,80.333333) (0.03000000,82.000000) (0.03428571,76.333333) (0.03857143,71.666667) (0.04285714,73.666667) (0.08571429,26.666667) (0.12857143,2.666667)};
\fi
\end{groupplot}
\node[anchor=east, draw=black!55, line width=0.35pt, fill=white,
  minimum height=\reconstructionLegendHeight, inner xsep=2.0pt, inner ysep=0pt,
  font=\reconstructionPlotFont]
  at ([xshift=\reconstructionLegendShift]group c1r1.west) {\vbox to \reconstructionLegendContentHeight{
    \hbox{\tokenizerid{vqbase}{VQ}}\vfil
    \hbox{\tokenizerid{rvqbase}{RVQ}}\vfil
    \hbox{\tokenizerid{nrvqbase}{NRVQ}}\vfil
    \hbox{\tokenizerid{fsqbase}{FSQ}}\vfil
    \hbox{\tokenizerid{qincobase}{QinCo}}
  }};
\node[font=\reconstructionPlotFont, inner ysep=0pt]
  at ($(group c1r1.south west)!0.5!(group c3r1.south east)+(0,\reconstructionXLabelOffset)$)
  {Mean-Absolute-Error on Action Chunk};
\end{tikzpicture}

%% file: figures/what_matters/empirical_support/main_rollout_prefix_grouped.tex

\begin{axis}[
  width=\chunkPlotWidth,height=\chunkPlotHeight,scale only axis,
  name=\chunkPlotName,at={(\chunkPlotX,0)},anchor=south west,
  ylabel near ticks,
  label style={font=\fontsize{7.2}{7.8}\selectfont},
  tick label style={font=\fontsize{7.2}{7.8}\selectfont},
  xmin=1, xmax=8, xtick={1,2,4,6,8},
  ymin=0, ymax=102, ytick={0,50,100},
  tick align=outside, tick pos=left,
  axis line style={black!55,line width=0.35pt},
  xmajorgrids, ymajorgrids, grid style={black!10,line width=0.3pt},
  xlabel={Token position}, ylabel={Unobserved prefix (\%)},
  ylabel style={at={(yticklabel cs:0.5,2pt)},anchor=near yticklabel,inner ysep=0pt,yshift=-\pgfkeysvalueof{/pgf/inner xsep}},
  xlabel style={at={(axis description cs:0.5,-0.24)},anchor=north},
  clip=false,
]
\addplot[name path=mainVQLow,draw=none,forget plot] coordinates {(1,0.000000) (2,0.251177) (3,22.119309) (4,79.497645) (5,92.825746) (6,98.320251) (7,99.544741) (8,99.676463)};
\addplot[name path=mainVQHigh,draw=none,forget plot] coordinates {(1,0.215692) (2,5.808713) (3,81.672509) (4,98.531085) (5,99.948278) (6,99.968967) (7,99.979311) (8,99.989656)};
\addplot[draw=none,fill=vqbase!35,fill opacity=0.24,forget plot] fill between[of=mainVQLow and mainVQHigh];
\addplot[vqbase,solid,line width=0.9pt,mark=*,mark size=0.9pt,mark options={solid,fill=vqbase,draw=vqbase,line width=0.25pt}] coordinates {(1,0.094644) (2,2.504637) (3,60.597412) (4,92.898163) (5,98.163442) (6,99.457050) (7,99.801198) (8,99.891257)};
\addplot[name path=mainRVQLow,draw=none,forget plot] coordinates {(1,0.000000) (2,1.619695) (3,44.476777) (4,86.038225) (5,97.003563) (6,99.310609) (7,99.708455) (8,99.835958)};
\addplot[name path=mainRVQHigh,draw=none,forget plot] coordinates {(1,0.804290) (2,23.896935) (3,89.929269) (4,98.854101) (5,99.960843) (6,100.000000) (7,100.000000) (8,100.000000)};
\addplot[draw=none,fill=rvqbase!35,fill opacity=0.24,forget plot] fill between[of=mainRVQLow and mainRVQHigh];
\addplot[rvqbase,solid,line width=0.9pt,mark=*,mark size=0.9pt,mark options={solid,fill=rvqbase,draw=rvqbase,line width=0.25pt}] coordinates {(1,0.292006) (2,5.975444) (3,66.344264) (4,93.997212) (5,98.802413) (6,99.721923) (7,99.899180) (8,99.955173)};
\addplot[name path=mainNRVQLow,draw=none,forget plot] coordinates {(1,0.000000) (2,2.141760) (3,52.355503) (4,86.758062) (5,95.856052) (6,98.706964) (7,99.614014) (8,99.831750)};
\addplot[name path=mainNRVQHigh,draw=none,forget plot] coordinates {(1,0.145391) (2,18.045113) (3,91.409822) (4,99.677561) (5,99.971548) (6,99.992672) (7,100.000000) (8,100.000000)};
\addplot[draw=none,fill=nrvqbase!35,fill opacity=0.24,forget plot] fill between[of=mainNRVQLow and mainNRVQHigh];
\addplot[nrvqbase,solid,line width=0.9pt,mark=*,mark size=0.9pt,mark options={solid,fill=nrvqbase,draw=nrvqbase,line width=0.25pt}] coordinates {(1,0.026722) (2,7.935432) (3,76.759210) (4,95.814366) (5,99.061967) (6,99.735965) (7,99.892901) (8,99.950020)};
\addplot[name path=mainFSQLow,draw=none,forget plot] coordinates {(1,0.000000) (2,0.010786) (3,14.679367) (4,77.292056) (5,96.265371) (6,98.514539) (7,99.083439) (8,99.640934)};
\addplot[name path=mainFSQHigh,draw=none,forget plot] coordinates {(1,0.142146) (2,0.409898) (3,41.805591) (4,92.639138) (5,98.872982) (6,99.736014) (7,99.935282) (8,99.967641)};
\addplot[draw=none,fill=fsqbase!35,fill opacity=0.24,forget plot] fill between[of=mainFSQLow and mainFSQHigh];
\addplot[fsqbase,solid,line width=0.9pt,mark=*,mark size=0.9pt,mark options={solid,fill=fsqbase,draw=fsqbase,line width=0.25pt}] coordinates {(1,0.034019) (2,0.179656) (3,30.528566) (4,86.650825) (5,97.627083) (6,99.174911) (7,99.585337) (8,99.813378)};
\addplot[name path=mainQINCOLow,draw=none,forget plot] coordinates {(1,0.000000) (2,1.956637) (3,73.069056) (4,96.806993) (5,99.054584) (6,99.732430) (7,99.897013) (8,99.897013)};
\addplot[name path=mainQINCOHigh,draw=none,forget plot] coordinates {(1,1.002853) (2,19.023398) (3,89.768191) (4,99.668508) (5,99.993094) (6,99.993094) (7,100.000000) (8,100.000000)};
\addplot[draw=none,fill=qincobase!35,fill opacity=0.24,forget plot] fill between[of=mainQINCOLow and mainQINCOHigh];
\addplot[qincobase,solid,line width=0.9pt,mark=*,mark size=0.9pt,mark options={solid,fill=qincobase,draw=qincobase,line width=0.25pt}] coordinates {(1,0.471282) (2,7.403152) (3,81.774240) (4,98.525055) (5,99.748345) (6,99.932799) (7,99.977545) (8,99.982307)};
\end{axis}

%% file: figures/what_matters/bitrate_chunking/main_reconstruction_family.tex
\begin{axis}[
  width=\chunkPlotWidth,height=\chunkPlotHeight,scale only axis,
  name=\chunkPlotName,at={(\chunkPlotX,0)},anchor=south west,
  ylabel near ticks,
  xmin=0.8,xmax=4.2,xtick={1,2,3,4},xticklabels={4,8,16,32},
  ymin=0.010,ymax=0.065,ytick={0.02,0.04,0.06},
  xlabel={Chunk size $H$},ylabel={Val. recon. error},
  ylabel style={at={(yticklabel cs:0.5,2pt)},anchor=near yticklabel,inner ysep=0pt,yshift=-\pgfkeysvalueof{/pgf/inner xsep}},
  xlabel style={at={(axis description cs:0.5,-0.24)},anchor=north},
  label style={font=\fontsize{7.2}{7.8}\selectfont},
  tick label style={font=\fontsize{7.2}{7.8}\selectfont},
  scaled y ticks=false,yticklabel style={/pgf/number format/fixed,/pgf/number format/precision=2},
  tick align=outside,tick pos=left,axis line style={black!55,line width=0.35pt},
  xmajorgrids,ymajorgrids,grid style={black!10,line width=0.3pt},
  clip=false,
]
\addplot[name path=mainVQlow,draw=none,forget plot] coordinates {(1,0.01452654) (2,0.02219565) (3,0.03183509) (4,0.04147643)};
\addplot[name path=mainVQhigh,draw=none,forget plot] coordinates {(1,0.01668877) (2,0.02465525) (3,0.03426864) (4,0.04551837)};
\addplot[fill=vqbase!35,fill opacity=0.24,draw=none,forget plot] fill between[of=mainVQlow and mainVQhigh];
\addplot[name path=mainRVQlow,draw=none,forget plot] coordinates {(1,0.01488862) (2,0.02359590) (3,0.03255835) (4,0.04375479)};
\addplot[name path=mainRVQhigh,draw=none,forget plot] coordinates {(1,0.02335282) (2,0.02934322) (3,0.03768804) (4,0.04909584)};
\addplot[fill=rvqbase!35,fill opacity=0.24,draw=none,forget plot] fill between[of=mainRVQlow and mainRVQhigh];
\addplot[name path=mainNRVQlow,draw=none,forget plot] coordinates {(1,0.01492557) (2,0.02225783) (3,0.03047594) (4,0.04125788)};
\addplot[name path=mainNRVQhigh,draw=none,forget plot] coordinates {(1,0.02346225) (2,0.03132311) (3,0.04087194) (4,0.05256277)};
\addplot[fill=nrvqbase!35,fill opacity=0.24,draw=none,forget plot] fill between[of=mainNRVQlow and mainNRVQhigh];
\addplot[name path=mainFSQlow,draw=none,forget plot] coordinates {(1,0.01831350) (2,0.02665224) (3,0.03462387) (4,0.04428812)};
\addplot[name path=mainFSQhigh,draw=none,forget plot] coordinates {(1,0.01911280) (2,0.02737714) (3,0.03503872) (4,0.04472214)};
\addplot[fill=fsqbase!35,fill opacity=0.24,draw=none,forget plot] fill between[of=mainFSQlow and mainFSQhigh];
\addplot[name path=mainQinColow,draw=none,forget plot] coordinates {(1,0.01352483) (2,0.02214647) (3,0.03037344) (4,0.03954452)};
\addplot[name path=mainQinCohigh,draw=none,forget plot] coordinates {(1,0.01836000) (2,0.02602205) (3,0.03429647) (4,0.04438378)};
\addplot[fill=qincobase!35,fill opacity=0.24,draw=none,forget plot] fill between[of=mainQinColow and mainQinCohigh];
\addplot[vqbase,solid,line width=0.9pt,mark=*,mark size=0.9pt,mark options={solid,fill=vqbase,draw=vqbase,line width=0.25pt}] coordinates {(1,0.01560765) (2,0.02342545) (3,0.03305187) (4,0.04349740)};
\addplot[rvqbase,solid,line width=0.9pt,mark=*,mark size=0.9pt,mark options={solid,fill=rvqbase,draw=rvqbase,line width=0.25pt}] coordinates {(1,0.01839657) (2,0.02619353) (3,0.03544563) (4,0.04630211)};
\addplot[nrvqbase,solid,line width=0.9pt,mark=*,mark size=0.9pt,mark options={solid,fill=nrvqbase,draw=nrvqbase,line width=0.25pt}] coordinates {(1,0.02009688) (2,0.02789695) (3,0.03678982) (4,0.04820083)};
\addplot[fsqbase,solid,line width=0.9pt,mark=*,mark size=0.9pt,mark options={solid,fill=fsqbase,draw=fsqbase,line width=0.25pt}] coordinates {(1,0.01871315) (2,0.02701469) (3,0.03483130) (4,0.04450513)};
\addplot[qincobase,solid,line width=0.9pt,mark=*,mark size=0.9pt,mark options={solid,fill=qincobase,draw=qincobase,line width=0.25pt}] coordinates {(1,0.01531108) (2,0.02340869) (3,0.03240756) (4,0.04220475)};
\end{axis}

%% file: figures/what_matters/bitrate_chunking/main_success_heatmap.tex
\definecolor{chunkmainc0}{RGB}{68,1,84}
\definecolor{chunkmainc1}{RGB}{70,18,100}
\definecolor{chunkmainc2}{RGB}{72,36,117}
\definecolor{chunkmainc3}{RGB}{68,52,126}
\definecolor{chunkmainc4}{RGB}{65,68,135}
\definecolor{chunkmainc5}{RGB}{59,82,138}
\definecolor{chunkmainc6}{RGB}{53,95,141}
\definecolor{chunkmainc7}{RGB}{48,108,142}
\definecolor{chunkmainc8}{RGB}{42,120,142}
\definecolor{chunkmainc9}{RGB}{38,132,141}
\definecolor{chunkmainc10}{RGB}{33,145,140}
\definecolor{chunkmainc11}{RGB}{34,156,136}
\definecolor{chunkmainc12}{RGB}{34,168,132}
\definecolor{chunkmainc13}{RGB}{51,180,122}
\definecolor{chunkmainc14}{RGB}{68,191,112}
\definecolor{chunkmainc15}{RGB}{95,200,96}
\definecolor{chunkmainc16}{RGB}{122,209,81}
\definecolor{chunkmainc17}{RGB}{156,216,60}
\definecolor{chunkmainc18}{RGB}{189,223,38}
\definecolor{chunkmainc19}{RGB}{221,227,38}
\definecolor{chunkmainc20}{RGB}{253,231,37}
\colorlet{chunkmainc0}{chunkmainc0!75!white}
\colorlet{chunkmainc1}{chunkmainc1!75!white}
\colorlet{chunkmainc2}{chunkmainc2!75!white}
\colorlet{chunkmainc3}{chunkmainc3!75!white}
\colorlet{chunkmainc4}{chunkmainc4!75!white}
\colorlet{chunkmainc5}{chunkmainc5!75!white}
\colorlet{chunkmainc6}{chunkmainc6!75!white}
\colorlet{chunkmainc7}{chunkmainc7!75!white}
\colorlet{chunkmainc8}{chunkmainc8!75!white}
\colorlet{chunkmainc9}{chunkmainc9!75!white}
\colorlet{chunkmainc10}{chunkmainc10!75!white}
\colorlet{chunkmainc11}{chunkmainc11!75!white}
\colorlet{chunkmainc12}{chunkmainc12!75!white}
\colorlet{chunkmainc13}{chunkmainc13!75!white}
\colorlet{chunkmainc14}{chunkmainc14!75!white}
\colorlet{chunkmainc15}{chunkmainc15!75!white}
\colorlet{chunkmainc16}{chunkmainc16!75!white}
\colorlet{chunkmainc17}{chunkmainc17!75!white}
\colorlet{chunkmainc18}{chunkmainc18!75!white}
\colorlet{chunkmainc19}{chunkmainc19!75!white}
\colorlet{chunkmainc20}{chunkmainc20!75!white}
\begin{axis}[
  width=\chunkPlotWidth,height=\chunkPlotHeight,scale only axis,
  name=\chunkPlotName,at={(\chunkPlotX,0)},anchor=south west,
  ylabel near ticks,
  xmin=0.5,xmax=4.5,ymin=0.5,ymax=4.5,
  xtick={1,2,3,4},xticklabels={4,8,16,32},
  ytick={1,2,3,4},yticklabels={4,8,16,32},
  xlabel={Chunk size $H$},ylabel={$H_{\mathrm{exec}}$},
  ylabel style={at={(yticklabel cs:0.5,2pt)},anchor=near yticklabel,inner ysep=0pt,yshift=-\pgfkeysvalueof{/pgf/inner xsep}},
  xlabel style={at={(axis description cs:0.5,-0.24)},anchor=north},
  label style={font=\fontsize{7.2}{7.8}\selectfont},
  tick label style={font=\fontsize{7.2}{7.8}\selectfont},
  tick align=outside,tick pos=left,axis line style={black!55,line width=0.35pt},
  clip=false,
]
\path[fill=chunkmainc13,draw=white,line width=0.55pt] (axis cs:0.53,0.53) rectangle (axis cs:1.47,1.47);
\node[font=\fontsize{7}{7.5}\selectfont,text=black] at (axis cs:1,1) {66.5};
\path[fill=chunkmainc14,draw=white,line width=0.55pt] (axis cs:1.53,0.53) rectangle (axis cs:2.47,1.47);
\node[font=\fontsize{7}{7.5}\selectfont,text=black] at (axis cs:2,1) {70.4};
\path[fill=chunkmainc13,draw=white,line width=0.55pt] (axis cs:2.53,0.53) rectangle (axis cs:3.47,1.47);
\node[font=\fontsize{7}{7.5}\selectfont,text=black] at (axis cs:3,1) {65.7};
\path[fill=chunkmainc10,draw=white,line width=0.55pt] (axis cs:3.53,0.53) rectangle (axis cs:4.47,1.47);
\node[font=\fontsize{7}{7.5}\selectfont,text=black] at (axis cs:4,1) {52.4};
\path[fill=chunkmainc13,draw=white,line width=0.55pt] (axis cs:1.53,1.53) rectangle (axis cs:2.47,2.47);
\node[font=\fontsize{7}{7.5}\selectfont,text=black] at (axis cs:2,2) {66.0};
\path[fill=chunkmainc13,draw=white,line width=0.55pt] (axis cs:2.53,1.53) rectangle (axis cs:3.47,2.47);
\node[font=\fontsize{7}{7.5}\selectfont,text=black] at (axis cs:3,2) {63.9};
\path[fill=chunkmainc10,draw=white,line width=0.55pt] (axis cs:3.53,1.53) rectangle (axis cs:4.47,2.47);
\node[font=\fontsize{7}{7.5}\selectfont,text=black] at (axis cs:4,2) {51.2};
\path[fill=chunkmainc11,draw=white,line width=0.55pt] (axis cs:2.53,2.53) rectangle (axis cs:3.47,3.47);
\node[font=\fontsize{7}{7.5}\selectfont,text=black] at (axis cs:3,3) {55.5};
\path[fill=chunkmainc9,draw=white,line width=0.55pt] (axis cs:3.53,2.53) rectangle (axis cs:4.47,3.47);
\node[font=\fontsize{7}{7.5}\selectfont,text=black] at (axis cs:4,3) {47.0};
\path[fill=chunkmainc7,draw=white,line width=0.55pt] (axis cs:3.53,3.53) rectangle (axis cs:4.47,4.47);
\node[font=\fontsize{7}{7.5}\selectfont,text=black] at (axis cs:4,4) {36.1};
\draw[black,line width=0.8pt] (axis cs:1.53,0.53) rectangle (axis cs:2.47,2.47);
\end{axis}

%% file: tables/experiments/simulation_results/main_multitask.tex
\begin{table}[!t]
    \centering
    \caption{Simulation success rates (\%) at $H_{\mathrm{exec}}=4\,/\,8$, averaged within each benchmark. \textbf{Bold} denotes better downstream performance for the same execution horizon.}
    \label{tab:simulation-multitask}
    \tablebodyfont
    \setlength{\tabcolsep}{2.4pt}
    \renewcommand{\arraystretch}{1.02}
    \begin{tabularx}{\linewidth}{@{}l *{6}{>{\centering\arraybackslash}X}@{}}
        \toprule
        \multirow{2}{*}{Tokenizer} & \multicolumn{2}{c}{RoboMimic} & \multicolumn{2}{c}{LIBERO} & \multicolumn{2}{c}{RoboTwin} \\
        \cmidrule(lr){2-3}\cmidrule(lr){4-5}\cmidrule(l){6-7}
        & R.O. & \methodname{} & R.O. & \methodname{} & R.O. & \methodname{} \\
        \midrule
        \tokenizerid{vqbase!60}{TF-VQ-S} & 73.4\,/\,80.2 & \textbf{86.1}\,/\,\textbf{86.2} & 50.6\,/\,45.4 & \textbf{56.6}\,/\,\textbf{53.2} & \textbf{73.3}\,/\,\textbf{70.0} & 62.7\,/\,64.0 \\
        \tokenizerid{vqbase!85}{TF-VQ-L} & 75.9\,/\,71.9 & \textbf{77.6}\,/\,\textbf{72.7} & 50.0\,/\,49.4 & \textbf{58.6}\,/\,\textbf{52.7} & 32.0\,/\,41.3 & \textbf{58.7}\,/\,\textbf{70.0} \\
        \addlinespace[2pt]
        \tokenizerid{rvqbase!40}{TF-RVQ-S} & 79.2\,/\,78.4 & \textbf{84.3}\,/\,\textbf{86.2} & 68.1\,/\,64.5 & \textbf{72.1}\,/\,\textbf{69.5} & \textbf{72.0}\,/\,\textbf{69.3} & 64.0\,/\,\textbf{69.3} \\
        \tokenizerid{rvqbase!50}{TF-RVQ-L} & 59.9\,/\,55.6 & \textbf{73.1}\,/\,\textbf{68.3} & 57.5\,/\,51.8 & \textbf{67.8}\,/\,\textbf{60.2} & 23.3\,/\,22.7 & \textbf{42.7}\,/\,\textbf{38.7} \\
        \tokenizerid{rvqbase!60}{CT-RVQ-S} & \textbf{81.6}\,/\,79.6 & 80.9\,/\,\textbf{85.1} & 38.9\,/\,46.3 & \textbf{70.2}\,/\,\textbf{68.2} & 61.3\,/\,64.0 & \textbf{67.3}\,/\,\textbf{67.3} \\
        \tokenizerid{rvqbase!70}{CT-RVQ-L} & 62.0\,/\,59.2 & \textbf{66.9}\,/\,\textbf{64.8} & 48.6\,/\,46.1 & \textbf{65.8}\,/\,\textbf{58.9} & 17.3\,/\,22.7 & \textbf{33.3}\,/\,\textbf{34.0} \\
        \tokenizerid{rvqbase!80}{C2-RVQ-S} & \textbf{85.1}\,/\,83.2 & 82.0\,/\,\textbf{83.8} & \textbf{70.9}\,/\,\textbf{69.3} & 70.1\,/\,68.0 & \textbf{48.0}\,/\,\textbf{54.7} & 23.3\,/\,22.0 \\
        \tokenizerid{rvqbase!90}{C2-RVQ-L} & 59.2\,/\,57.9 & \textbf{65.9}\,/\,\textbf{61.4} & 54.4\,/\,50.0 & \textbf{64.7}\,/\,\textbf{58.4} & \textbf{30.7}\,/\,22.7 & 25.3\,/\,\textbf{30.7} \\
        \addlinespace[2pt]
        \tokenizerid{nrvqbase!40}{TF-NRVQ-S} & 80.0\,/\,80.7 & \textbf{84.3}\,/\,\textbf{84.3} & 36.2\,/\,38.7 & \textbf{41.5}\,/\,\textbf{47.5} & \textbf{62.7}\,/\,56.7 & 61.3\,/\,\textbf{58.0} \\
        \tokenizerid{nrvqbase!50}{TF-NRVQ-L} & 56.6\,/\,48.3 & \textbf{60.6}\,/\,\textbf{57.1} & 55.2\,/\,49.0 & \textbf{61.7}\,/\,\textbf{53.4} & 30.0\,/\,36.0 & \textbf{41.3}\,/\,\textbf{39.3} \\
        \tokenizerid{nrvqbase!60}{CT-NRVQ-S} & 0.0\,/\,0.0 & \textbf{81.1}\,/\,\textbf{86.6} & 0.0\,/\,0.0 & \textbf{56.7}\,/\,\textbf{63.7} & 0.0\,/\,0.0 & \textbf{48.0}\,/\,\textbf{68.0} \\
        \tokenizerid{nrvqbase!70}{CT-NRVQ-L} & 0.0\,/\,0.0 & \textbf{61.7}\,/\,\textbf{54.0} & 0.0\,/\,0.0 & \textbf{52.9}\,/\,\textbf{44.7} & 0.0\,/\,0.0 & \textbf{42.0}\,/\,\textbf{32.0} \\
        \tokenizerid{nrvqbase!80}{C2-NRVQ-S} & \textbf{80.8}\,/\,\textbf{82.9} & 78.4\,/\,\textbf{82.9} & \textbf{73.7}\,/\,\textbf{71.0} & 70.7\,/\,67.5 & \textbf{56.0}\,/\,63.3 & 55.3\,/\,\textbf{66.0} \\
        \tokenizerid{nrvqbase!90}{C2-NRVQ-L} & 56.3\,/\,48.0 & \textbf{61.4}\,/\,\textbf{52.0} & 54.0\,/\,46.3 & \textbf{54.9}\,/\,\textbf{49.5} & 40.7\,/\,\textbf{38.7} & \textbf{43.3}\,/\,37.3 \\
        \addlinespace[2pt]
        \tokenizerid{fsqbase!60}{TF-FSQ-S} & 74.7\,/\,74.8 & \textbf{80.2}\,/\,\textbf{80.8} & 30.7\,/\,32.2 & \textbf{61.0}\,/\,\textbf{58.4} & 68.7\,/\,69.3 & \textbf{72.7}\,/\,\textbf{78.0} \\
        \tokenizerid{fsqbase!85}{TF-FSQ-L} & 77.7\,/\,74.0 & \textbf{81.4}\,/\,\textbf{83.4} & 15.8\,/\,16.7 & \textbf{60.4}\,/\,\textbf{57.1} & \textbf{67.3}\,/\,68.7 & 58.7\,/\,\textbf{74.0} \\
        \addlinespace[2pt]
        \tokenizerid{qincobase!40}{TF-QINCO-S} & 48.4\,/\,40.4 & \textbf{75.8}\,/\,\textbf{73.8} & \textbf{44.9}\,/\,34.4 & 43.9\,/\,\textbf{38.9} & 14.7\,/\,14.0 & \textbf{52.7}\,/\,\textbf{48.7} \\
        \tokenizerid{qincobase!50}{TF-QINCO-L} & 44.9\,/\,32.6 & \textbf{74.0}\,/\,\textbf{72.1} & 35.2\,/\,23.8 & \textbf{37.5}\,/\,\textbf{33.8} & 17.3\,/\,18.0 & \textbf{46.0}\,/\,\textbf{46.0} \\
        \tokenizerid{qincobase!60}{CT-QINCO-S} & 58.3\,/\,46.8 & \textbf{70.9}\,/\,\textbf{63.7} & 20.0\,/\,19.0 & \textbf{35.4}\,/\,\textbf{33.1} & 18.7\,/\,19.3 & \textbf{51.3}\,/\,\textbf{49.3} \\
        \tokenizerid{qincobase!70}{CT-QINCO-L} & 52.1\,/\,42.9 & \textbf{71.2}\,/\,\textbf{64.1} & 28.4\,/\,20.6 & \textbf{42.5}\,/\,\textbf{39.5} & 23.3\,/\,19.3 & \textbf{47.3}\,/\,\textbf{44.0} \\
        \tokenizerid{qincobase!80}{C2-QINCO-S} & 50.3\,/\,41.9 & \textbf{73.7}\,/\,\textbf{71.2} & 7.3\,/\,9.4 & \textbf{38.0}\,/\,\textbf{35.9} & 42.7\,/\,27.3 & \textbf{56.7}\,/\,\textbf{47.3} \\
        \tokenizerid{qincobase!90}{C2-QINCO-L} & 50.6\,/\,40.0 & \textbf{60.8}\,/\,\textbf{52.3} & 25.4\,/\,20.2 & \textbf{63.6}\,/\,\textbf{53.6} & 20.7\,/\,15.3 & \textbf{25.3}\,/\,\textbf{24.7} \\
        \bottomrule
    \end{tabularx}
    \vspace{-10px}
\end{table}

%% file: tables/experiments/simulation_results/main_vla_libero.tex
    \centering
    \caption{VLA success rates (\%) on LIBERO at $H_{\mathrm{exec}}=8$.}
    \label{tab:vla-libero-results}
    \begingroup
    \tablebodyfont
    \setlength{\tabcolsep}{1.5pt}
    \renewcommand{\arraystretch}{1.12}
    \begin{tabular*}{\linewidth}{@{}l@{\extracolsep{\fill}}*{2}{c}@{}}
        \toprule
        Tokenizer & R.O. & \methodname{} \\
        \midrule
        \tokenizerid{vqbase!60}{TF-VQ-S} & $15.5\pm1.6$ & $\mathbf{45.4\pm2.2}$ \\
        \tokenizerid{vqbase!85}{TF-VQ-L} & $25.8\pm1.9$ & $\mathbf{48.9\pm2.2}$ \\
        \addlinespace[2pt]
        \tokenizerid{rvqbase!40}{TF-RVQ-S} & $\mathbf{53.8\pm2.2}$ & $48.1\pm2.2$ \\
        \tokenizerid{rvqbase!50}{TF-RVQ-L} & $39.2\pm2.1$ & $\mathbf{53.1\pm2.2}$ \\
        \tokenizerid{rvqbase!70}{CT-RVQ-L} & $22.7\pm1.8$ & $\mathbf{29.2\pm2.0}$ \\
        \addlinespace[2pt]
        \tokenizerid{fsqbase!60}{TF-FSQ-S} & $11.5\pm1.4$ & $\mathbf{48.8\pm2.2}$ \\
        \tokenizerid{fsqbase!85}{TF-FSQ-L} & $32.2\pm2.0$ & $\mathbf{55.1\pm2.2}$ \\
        \addlinespace[2pt]
        \tokenizerid{qincobase!90}{C2-QINCO-L} & $9.2\pm1.3$ & $\mathbf{55.9\pm2.2}$ \\
        \bottomrule
    \end{tabular*}
    \endgroup

%% file: tables/experiments/simulation_results/main_robomimic_aggregate_ablation.tex
    \centering
    \captionsetup{position=top,skip=7pt}
    \caption{Ablation on Robomimic. We report success rates (\%) at $H_{\mathrm{exec}}=4\,/\,8$. We show the averaged success by \emph{Task} and \emph{Codebook sharing}.}
    \phantomsection
    \label{tab:robomimic-aggregate-ablation}
    \begingroup
    \tablebodyfont
    \setlength{\tabcolsep}{1.5pt}
    \renewcommand{\arraystretch}{0.98}
    \begin{tabularx}{\linewidth}{@{}l *{4}{>{\centering\arraybackslash}X}@{}}
        \toprule
        Subset & R.O. & \wotbr{} & \wotp{} & \methodname{} \\
        \midrule
        \multicolumn{5}{@{}l}{\fontsize{7}{7}\selectfont\emph{Task}} \\
        Can & 76.1\,/\,70.1 & 83.5\,/\,79.2 & 85.5\,/\,77.7 & \textbf{87.0}\,/\,\textbf{83.7} \\
        Lift & 78.9\,/\,77.1 & 84.1\,/\,81.6 & \textbf{93.5}\,/\,\textbf{92.9} & 84.9\,/\,84.8 \\
        Square & 41.1\,/\,35.7 & 48.5\,/\,43.0 & 42.3\,/\,37.0 & \textbf{51.6}\,/\,\textbf{48.5} \\
        \addlinespace[2pt]
        \multicolumn{5}{@{}l}{\fontsize{7}{7}\selectfont\emph{Codebook sharing}} \\
        Shared (S) & 71.2\,/\,68.9 & 78.3\,/\,76.0 & \textbf{80.6}\,/\,77.5 & 79.7\,/\,\textbf{79.8} \\
        Level-Specific (L) & 59.5\,/\,53.0 & 65.8\,/\,59.8 & 67.0\,/\,60.9 & \textbf{69.3}\,/\,\textbf{64.8} \\
        \midrule
        \textbf{Overall} & 65.4\,/\,61.0 & 72.0\,/\,67.9 & 73.8\,/\,69.2 & \textbf{74.5}\,/\,\textbf{72.3} \\
        \bottomrule
    \end{tabularx}
    \endgroup

%% file: figures/experiments/reconstruction_error/main_decoded_robustness_pooled.tex
\begin{tikzpicture}
\definecolor{robustshared}{RGB}{184,121,99}
\definecolor{robustlevel}{RGB}{104,151,187}
\begin{groupplot}[group style={group size=3 by 1,horizontal sep=0.55cm},
  width=3.55cm,height=2.65cm,scale only axis,
  xmode=log,log basis x=2,xmin=1,xmax=2048,xtick={1,8,64,512,2048},xticklabels={1,8,64,512,2048},
  ymin=-0.0500,ymax=0.0200,ytick={-0.04,-0.02,0.00,0.02},scaled ticks=false,
  tick label style={font=\fontsize{8}{9}\selectfont,/pgf/number format/fixed,/pgf/number format/precision=2},
  label style={font=\fontsize{8}{9}\selectfont},
  tick align=outside,tick pos=left,
  axis line style={black!55,line width=0.35pt},xmajorgrids,ymajorgrids,grid style={black!10,line width=0.2pt}]
\nextgroupplot[extra description/.code={
  \node[anchor=north east,fill=none,draw=none,text opacity=1,inner sep=0.6pt,font=\fontsize{8}{9}\selectfont]
    at (axis description cs:0.988,0.965) {Can};
  \node[anchor=south west,fill=none,draw=none,text opacity=1,inner sep=0pt,align=left,font=\fontsize{8}{9}\selectfont]
    at (axis description cs:0.04,0.04) {\textcolor{robustshared}{$\bullet$} All tokenizers};
},ylabel={Action-error change (MAE)}]
\addplot[draw=none,fill=robustshared,fill opacity=0.08,forget plot] coordinates {(1,-0.0000476924) (2048,-0.0000476924) (2048,0.0172099981) (1,0.0172099981)} \closedcycle;
\addplot[robustshared,dashed,line width=0.65pt,forget plot] coordinates {(1,0.0064699895) (2048,0.0064699895)};
\addplot[name path=robust00Low,draw=none,forget plot] coordinates {(1,-0.0107014842) (2,-0.0128491896) (4,-0.0148387030) (8,-0.0172400285) (16,-0.0195127478) (32,-0.0218348768) (64,-0.0248323275) (128,-0.0275315118) (256,-0.0300361254) (512,-0.0330222920) (1024,-0.0355759010) (2048,-0.0387250186)};
\addplot[name path=robust00High,draw=none,forget plot] coordinates {(1,-0.0085356237) (2,-0.0104979323) (4,-0.0122178429) (8,-0.0142639539) (16,-0.0164540474) (32,-0.0184725737) (64,-0.0212637713) (128,-0.0237930203) (256,-0.0259170397) (512,-0.0285844591) (1024,-0.0310381546) (2048,-0.0338878176)};
\addplot[fill=robustshared,fill opacity=0.25,draw=none,forget plot] fill between[of=robust00Low and robust00High];
\addplot[robustshared,solid,line width=0.85pt,mark=*,mark size=1.25pt,mark options={solid,fill=robustshared},forget plot] coordinates {(1,-0.0095705929) (2,-0.0115836235) (4,-0.0134776621) (8,-0.0156554447) (16,-0.0179214633) (32,-0.0200506184) (64,-0.0229271205) (128,-0.0256282777) (256,-0.0278537002) (512,-0.0306831439) (1024,-0.0332519212) (2048,-0.0361804451)};
\addplot[black!55,line width=0.45pt,forget plot] coordinates {(1,0) (2048,0)};
\nextgroupplot[extra description/.code={
  \node[anchor=north east,fill=none,draw=none,text opacity=1,inner sep=0.6pt,font=\fontsize{8}{9}\selectfont]
    at (axis description cs:0.988,0.965) {Lift};
},yticklabels={}]
\addplot[draw=none,fill=robustshared,fill opacity=0.08,forget plot] coordinates {(1,-0.0000476924) (2048,-0.0000476924) (2048,0.0172099981) (1,0.0172099981)} \closedcycle;
\addplot[robustshared,dashed,line width=0.65pt,forget plot] coordinates {(1,0.0064699895) (2048,0.0064699895)};
\addplot[name path=robust10Low,draw=none,forget plot] coordinates {(1,-0.0073769669) (2,-0.0090943408) (4,-0.0107513532) (8,-0.0124474415) (16,-0.0141776944) (32,-0.0163146476) (64,-0.0179494608) (128,-0.0199959733) (256,-0.0219199873) (512,-0.0235753685) (1024,-0.0247590136) (2048,-0.0260747915)};
\addplot[name path=robust10High,draw=none,forget plot] coordinates {(1,-0.0055307727) (2,-0.0069168858) (4,-0.0082759210) (8,-0.0094477334) (16,-0.0107593913) (32,-0.0123397093) (64,-0.0137302693) (128,-0.0152372193) (256,-0.0167638746) (512,-0.0180521823) (1024,-0.0193458240) (2048,-0.0202771248)};
\addplot[fill=robustshared,fill opacity=0.25,draw=none,forget plot] fill between[of=robust10Low and robust10High];
\addplot[robustshared,solid,line width=0.85pt,mark=*,mark size=1.25pt,mark options={solid,fill=robustshared},forget plot] coordinates {(1,-0.0064209160) (2,-0.0079347437) (4,-0.0094645117) (8,-0.0108448304) (16,-0.0124463132) (32,-0.0142529035) (64,-0.0157054040) (128,-0.0174960706) (256,-0.0192030921) (512,-0.0206586382) (1024,-0.0219404303) (2048,-0.0230251277)};
\addplot[black!55,line width=0.45pt,forget plot] coordinates {(1,0) (2048,0)};
\nextgroupplot[extra description/.code={
  \node[anchor=north east,fill=none,draw=none,text opacity=1,inner sep=0.6pt,font=\fontsize{8}{9}\selectfont]
    at (axis description cs:0.988,0.965) {Square};
},yticklabels={}]
\addplot[draw=none,fill=robustshared,fill opacity=0.08,forget plot] coordinates {(1,-0.0000476924) (2048,-0.0000476924) (2048,0.0172099981) (1,0.0172099981)} \closedcycle;
\addplot[robustshared,dashed,line width=0.65pt,forget plot] coordinates {(1,0.0064699895) (2048,0.0064699895)};
\addplot[name path=robust20Low,draw=none,forget plot] coordinates {(1,-0.0098210508) (2,-0.0119436266) (4,-0.0140597692) (8,-0.0162653000) (16,-0.0189435340) (32,-0.0212987693) (64,-0.0240573787) (128,-0.0267727184) (256,-0.0295697155) (512,-0.0323016015) (1024,-0.0355013832) (2048,-0.0381107399)};
\addplot[name path=robust20High,draw=none,forget plot] coordinates {(1,-0.0077320800) (2,-0.0094954101) (4,-0.0114405852) (8,-0.0135216003) (16,-0.0155561273) (32,-0.0178110537) (64,-0.0200244251) (128,-0.0223770530) (256,-0.0246827823) (512,-0.0270786545) (1024,-0.0295844737) (2048,-0.0323172003)};
\addplot[fill=robustshared,fill opacity=0.25,draw=none,forget plot] fill between[of=robust20Low and robust20High];
\addplot[robustshared,solid,line width=0.85pt,mark=*,mark size=1.25pt,mark options={solid,fill=robustshared},forget plot] coordinates {(1,-0.0086611195) (2,-0.0106139527) (4,-0.0126684855) (8,-0.0148032581) (16,-0.0171298780) (32,-0.0194178190) (64,-0.0218674459) (128,-0.0244310213) (256,-0.0269430572) (512,-0.0294796718) (1024,-0.0322315040) (2048,-0.0349765991)};
\addplot[black!55,line width=0.45pt,forget plot] coordinates {(1,0) (2048,0)};
\end{groupplot}
\node[anchor=north,font=\fontsize{8}{9}\selectfont,inner sep=0pt] at ([yshift=-0.65cm]group c2r1.south) {$M$ policy samples};
\end{tikzpicture}

%% file: tables/experiments/real_robot/main_ur5.tex
\tablebodyfont
\setlength{\tabcolsep}{2pt}
\renewcommand{\arraystretch}{1.12}
\begin{tabularx}{\linewidth}{@{}l *{4}{>{\centering\arraybackslash}X}@{}}
    \toprule
    \multirow{2}{*}{Tokenizer} & \multicolumn{2}{c}{Cube Stacking} & \multicolumn{2}{c}{Cup hanging} \\
    \cmidrule(lr){2-3}\cmidrule(lr){4-5}
    & R.O. & \methodname{} & R.O. & \methodname{} \\
    \midrule
    \tokenizerid{fsqbase!60}{TF-FSQ-S} & 14/30 (46.7) & \textbf{17/30 (56.7)} & 10/30 (33.3) & \textbf{20/30 (66.7)} \\
    \tokenizerid{fsqbase!85}{TF-FSQ-L} & 4/30 (13.3) & \textbf{18/30 (60.0)} & 3/30 (10.0) & \textbf{18/30 (60.0)} \\
    \tokenizerid{rvqbase!40}{TF-RVQ-S} & 8/30 (26.7) & \textbf{19/30 (63.3)} & 10/30 (33.3) & \textbf{21/30 (70.0)} \\
    \tokenizerid{rvqbase!50}{TF-RVQ-L} & 1/30 (3.3) & \textbf{19/30 (63.3)} & 3/30 (10.0) & \textbf{14/30 (46.7)} \\
    \tokenizerid{rvqbase!70}{CT-RVQ-L} & 1/30 (3.3) & \textbf{12/30 (40.0)} & 2/30 (6.7) & \textbf{8/30 (26.7)} \\
    \bottomrule
\end{tabularx}

%% file: tables/preliminaries/tokenizer_design/appendix_tokenizer_index.tex
\begin{table}[H]
    \centering
    \caption{The 22 tokenizer configurations evaluated in this study. Structural IDs encode the backbone (TF: Transformer; CT: convolution--Transformer; C2: Conv2D), quantizer, and quantization-parameter sharing (S: shared; L: level-specific). The color swatch redundantly indicates the quantizer family.}
    \label{tab:tokenizer-index}
    \tablebodyfont
    \begin{tabularx}{0.88\linewidth}{@{}l X l l@{}}
        \toprule
        ID & Backbone & Quantizer & Parameter sharing \\
        \midrule
        \tokenizerid{vqbase!60}{TF-VQ-S} & Transformer & \multirow{2}{*}{VQ} & Shared \\
        \tokenizerid{vqbase!85}{TF-VQ-L} & Transformer &  & Level-specific \\
        \midrule
        \tokenizerid{rvqbase!40}{TF-RVQ-S} & Transformer & \multirow{6}{*}{RVQ} & Shared \\
        \tokenizerid{rvqbase!50}{TF-RVQ-L} & Transformer &  & Level-specific \\
        \tokenizerid{rvqbase!60}{CT-RVQ-S} & Convolution--Transformer &  & Shared \\
        \tokenizerid{rvqbase!70}{CT-RVQ-L} & Convolution--Transformer &  & Level-specific \\
        \tokenizerid{rvqbase!80}{C2-RVQ-S} & Conv2D &  & Shared \\
        \tokenizerid{rvqbase!90}{C2-RVQ-L} & Conv2D &  & Level-specific \\
        \midrule
        \tokenizerid{nrvqbase!40}{TF-NRVQ-S} & Transformer & \multirow{6}{*}{NRVQ} & Shared \\
        \tokenizerid{nrvqbase!50}{TF-NRVQ-L} & Transformer &  & Level-specific \\
        \tokenizerid{nrvqbase!60}{CT-NRVQ-S} & Convolution--Transformer &  & Shared \\
        \tokenizerid{nrvqbase!70}{CT-NRVQ-L} & Convolution--Transformer &  & Level-specific \\
        \tokenizerid{nrvqbase!80}{C2-NRVQ-S} & Conv2D &  & Shared \\
        \tokenizerid{nrvqbase!90}{C2-NRVQ-L} & Conv2D &  & Level-specific \\
        \midrule
        \tokenizerid{fsqbase!60}{TF-FSQ-S} & Transformer & \multirow{2}{*}{FSQ} & Shared \\
        \tokenizerid{fsqbase!85}{TF-FSQ-L} & Transformer &  & Level-specific \\
        \midrule
        \tokenizerid{qincobase!40}{TF-QINCO-S} & Transformer & \multirow{6}{*}{QinCo} & Shared \\
        \tokenizerid{qincobase!50}{TF-QINCO-L} & Transformer &  & Level-specific \\
        \tokenizerid{qincobase!60}{CT-QINCO-S} & Convolution--Transformer &  & Shared \\
        \tokenizerid{qincobase!70}{CT-QINCO-L} & Convolution--Transformer &  & Level-specific \\
        \tokenizerid{qincobase!80}{C2-QINCO-S} & Conv2D &  & Shared \\
        \tokenizerid{qincobase!90}{C2-QINCO-L} & Conv2D &  & Level-specific \\
        \bottomrule
    \end{tabularx}
\end{table}

%% file: tables/preliminaries/training_evaluation/appendix_rollout_protocol.tex
\begin{table}[H]
    \centering
    \caption{Primary rollout evaluation protocols across simulation benchmarks. Episode counts apply separately to each execution horizon.}
    \label{tab:rollout-evaluation}
    \label{tab:libero-rollout-evaluation}
    \label{tab:robotwin-rollout-evaluation}
    \begingroup
    \tablebodyfont
    \setlength{\tabcolsep}{4pt}
    \renewcommand{\arraystretch}{1.12}
    \begin{tabularx}{\linewidth}{@{}>{\raggedright\arraybackslash}p{0.20\linewidth}*{3}{>{\raggedright\arraybackslash}X}@{}}
        \toprule
        \textbf{Setting} & \textbf{RoboMimic} & \textbf{LIBERO} & \textbf{RoboTwin} \\
        \midrule
        Checkpoint selection & \multicolumn{3}{c}{Best teacher-forced validation accuracy} \\
        Execution horizon & \multicolumn{3}{c}{$H_{\mathrm{exec}}\in\{4,8\}$ for primary results; 8 by default} \\
        Policy sampling & \multicolumn{3}{c}{Greedy} \\
        \midrule
        Evaluation units & Individual tasks & Spatial, Object, Goal, LIBERO-10; 10 tasks per suite & 3 tasks \\
        \addlinespace[2pt]
        Episode allocation & 10 rollout seeds $\times$ 30 episodes per seed & 50 fixed initial states per task & 50 random initializations per task \\
        \addlinespace[2pt]
        Total episodes & 300 per task & 500 per suite & 150 across all 3 tasks \\
        \addlinespace[2pt]
        Episode limit & 500 environment steps & 500 environment steps for every suite & 300 / 500 / 600 environment steps (see note) \\
        \addlinespace[2pt]
        Success aggregation & Binary success averaged over 300 episodes per task & Binary success averaged over 500 episodes per suite & Binary success averaged over 50 episodes per task \\
        \bottomrule
    \end{tabularx}
    \par\vspace{3pt}
    {\raggedright\textit{Note.} RoboTwin episode limits: grabbing the roller, 300; placing the burger and fries, 500; placing cans in the plastic box, 600 environment steps.\par}
    \endgroup
\end{table}

%% file: tables/preliminaries/tokenizer_design/appendix_tokenizer_training_hyperparameters.tex
\begin{table}[H]
    \centering
    \caption{Tokenizer-training configuration for the three simulation benchmarks. A single recipe is shared by all 22 tokenizer configurations within a benchmark; rows spanning the three columns are additionally identical across benchmarks.}
    \label{tab:tokenizer-training-hyperparameters}
    \tablebodyfont
    \setlength{\tabcolsep}{6pt}
    \renewcommand{\arraystretch}{1.22}
    \begin{tabularx}{0.74\linewidth}{@{}>{\raggedright\arraybackslash}X*{3}{>{\centering\arraybackslash}p{0.145\linewidth}}@{}}
        \toprule
        \textbf{Hyperparameter} & \textbf{RoboMimic} & \textbf{LIBERO} & \textbf{RoboTwin} \\
        \midrule
        \multicolumn{4}{@{}l}{\textsc{\textbf{Data and operating point}}} \\
        \cmidrule(r{0.5em}){1-4}
        Training tasks & Can, Lift, Square & Spatial, Object, Goal, LIBERO-10 & All tasks \\
        Action dimension & $7$ & $7$ & $14$ \\
        Chunk size $H$ & \multicolumn{3}{c}{$8$} \\
        Codebook size $K$ & \multicolumn{3}{c}{$128$} \\
        Quantization levels $L$ & \multicolumn{3}{c}{$8$} \\
        Action normalization & \multicolumn{3}{c}{Per-dimension mean/standard deviation} \\
        \addlinespace[3pt]
        \cmidrule(r{0.5em}){1-4}
        \multicolumn{4}{@{}l}{\textsc{\textbf{Optimization}}} \\
        \cmidrule(r{0.5em}){1-4}
        Optimizer & \multicolumn{3}{c}{AdamW, $\beta=(0.9,\,0.95)$} \\
        Base learning rate & \multicolumn{3}{c}{$5\times10^{-5}$} \\
        Learning-rate schedule & \multicolumn{3}{c}{Constant, $100$ warmup steps} \\
        Gradient-clipping norm & \multicolumn{3}{c}{$1.0$} \\
        \addlinespace[3pt]
        \cmidrule(r{0.5em}){1-4}
        \multicolumn{4}{@{}l}{\textsc{\textbf{Training budget}}} \\
        \cmidrule(r{0.5em}){1-4}
        Training epochs & \multicolumn{3}{c}{$1000$} \\
        Batch size & \multicolumn{3}{c}{$256$} \\
        \addlinespace[3pt]
        \cmidrule(r{0.5em}){1-4}
        \multicolumn{4}{@{}l}{\textsc{\textbf{Objective and selection}}} \\
        \cmidrule(r{0.5em}){1-4}
        Commitment weight & \multicolumn{3}{c}{$0.25$ (VQ, RVQ, NRVQ, QinCo)} \\
        Checkpoint selection & \multicolumn{3}{c}{Best validation checkpoint} \\
        \bottomrule
    \end{tabularx}
\end{table}

%% file: tables/preliminaries/training_evaluation/appendix_policy_training_hyperparameters.tex
\begin{table}[H]
    \centering
    \caption{Autoregressive policy configuration for the three simulation benchmarks. A single recipe is shared by every tokenizer configuration and training objective within a benchmark; rows spanning the three columns are additionally identical across benchmarks.}
    \label{tab:policy-training-hyperparameters}
    \tablebodyfont
    \setlength{\tabcolsep}{6pt}
    \renewcommand{\arraystretch}{1.22}
    \begin{tabularx}{0.82\linewidth}{@{}>{\raggedright\arraybackslash}X*{3}{>{\centering\arraybackslash}p{0.165\linewidth}}@{}}
        \toprule
        \textbf{Hyperparameter} & \textbf{RoboMimic} & \textbf{LIBERO} & \textbf{RoboTwin} \\
        \midrule
        \multicolumn{4}{@{}l}{\textsc{\textbf{Architecture}}} \\
        \cmidrule(r{0.5em}){1-4}
        Model dimension & \multicolumn{3}{c}{$256$} \\
        Transformer layers & \multicolumn{3}{c}{$4$} \\
        Attention heads & \multicolumn{3}{c}{$4$} \\
        Feed-forward dimension & $256$ & $1024$ & $1024$ \\
        Dropout & $0.1$ & $0.0$ & $0.0$ \\
        Observation conditioning & \multicolumn{3}{c}{Cross-attention} \\
        Visual backbone & \multicolumn{3}{c}{Trainable, $512$-dim feature per view} \\
        Camera views & $2$ & $2$ & $4$ \\
        Image resolution & \multicolumn{3}{c}{$224\times224$ per view} \\
        Vocabulary size & \multicolumn{3}{c}{$128$} \\
        Action tokens per chunk & \multicolumn{3}{c}{$8$} \\
        \addlinespace[3pt]
        \cmidrule(r{0.5em}){1-4}
        \multicolumn{4}{@{}l}{\textsc{\textbf{Inputs and action space}}} \\
        \cmidrule(r{0.5em}){1-4}
        Proprioception & EEF position, quaternion, joint velocity ($14$) & EEF position, axis--angle, gripper joints ($8$) & Per arm: EEF position, $6$-D rotation, gripper ($20$) \\
        Task conditioning & One-hot task identity & Predefined task instruction & One-hot task identity \\
        Language encoder & -- & Frozen T5-base, $\leq24$ tokens & -- \\
        Action representation & \multicolumn{3}{c}{Delta end-effector pose} \\
        \addlinespace[3pt]
        \cmidrule(r{0.5em}){1-4}
        \multicolumn{4}{@{}l}{\textsc{\textbf{Optimization}}} \\
        \cmidrule(r{0.5em}){1-4}
        Optimizer & \multicolumn{3}{c}{AdamW} \\
        Base learning rate & \multicolumn{3}{c}{$1\times10^{-4}$} \\
        Learning-rate schedule & \multicolumn{3}{c}{Constant, warmup over $5\%$ of training} \\
        Weight decay & \multicolumn{3}{c}{$1\times10^{-3}$} \\
        Gradient-clipping norm & \multicolumn{3}{c}{$1.0$} \\
        EMA decay & \multicolumn{3}{c}{$0.999$} \\
        \addlinespace[3pt]
        \cmidrule(r{0.5em}){1-4}
        \multicolumn{4}{@{}l}{\textsc{\textbf{Training and selection}}} \\
        \cmidrule(r{0.5em}){1-4}
        Training epochs & $300$ & $500$ & $500$ \\
        Batch size & \multicolumn{3}{c}{$512$} \\
        Feature-space blur $\sigma_{\max}$ & $0.0$ & $2.0$ & $2.0$ \\
        Augmentation anneal & -- & $200$ epochs & $200$ epochs \\
        Checkpoint selection & \multicolumn{3}{c}{Best teacher-forced validation accuracy} \\
        \bottomrule
    \end{tabularx}
\end{table}

%% file: tables/preliminaries/training_evaluation/appendix_policy_seed_robustness.tex
\begin{table}[h]
    \centering
    \caption{Policy-training-seed robustness on RoboMimic. Each row summarizes rollout success over 10 independently trained multi-task policies paired with the same fixed tokenizer. Every policy is evaluated at $H=H_{\mathrm{exec}}=8$ on the same 300 episodes per task. Values are reported as mean $\pm$ standard deviation across policy-training seeds; all values are percentages.}
    \label{tab:policy-seed-robustness}
    \tablebodyfont
    \setlength{\tabcolsep}{5pt}
    \begin{tabular}{@{}llcc@{}}
        \toprule
        Tokenizer & Task & Mean $\pm$ SD & Min--max \\
        \midrule
        \multirow{3}{*}{FSQ} & Can    & $88.7 \pm 1.9$ & $86.0$--$91.3$ \\
                             & Lift   & $94.1 \pm 2.5$ & $88.0$--$96.3$ \\
                             & Square & $51.5 \pm 4.3$ & $44.0$--$58.3$ \\
        \addlinespace[2pt]
        \multirow{3}{*}{NRVQ} & Can    & $87.9 \pm 2.4$ & $84.3$--$91.3$ \\
                              & Lift   & $96.4 \pm 1.2$ & $94.0$--$98.0$ \\
                              & Square & $54.5 \pm 5.8$ & $44.3$--$61.0$ \\
        \bottomrule
    \end{tabular}
\end{table}

%% file: tables/preliminaries/training_evaluation/appendix_vla_training_hyperparameters.tex
\begin{table}[H]
    \centering
    \caption{Training and evaluation configuration shared by the sixteen VLA policy runs on LIBERO. The runs differ only in the action tokenizer and the resulting token targets.}
    \label{tab:vla-training-hyperparameters}
    \tablebodyfont
    \setlength{\tabcolsep}{7pt}
    \renewcommand{\arraystretch}{1.18}
    \begin{tabularx}{0.82\linewidth}{@{}>{\raggedright\arraybackslash}p{0.34\linewidth}>{\raggedright\arraybackslash}X@{}}
        \toprule
        \textbf{Hyperparameter} & \textbf{Value} \\
        \midrule
        \multicolumn{2}{@{}l}{\textsc{\textbf{Architecture and action representation}}} \\
        \cmidrule(r{0.5em}){1-2}
        VLM backbone & SmolVLM2-256M-Video-Instruct (pretrained; 228.1M parameters) \\
        Action expert & Llama, 30 layers, hidden dimension 288, feed-forward dimension 1152 (random initialization; 43.2M parameters) \\
        Total / effective trainable parameters & 271.4M / 268.0M \\
        Fine-tuning & Full-parameter \\
        Action vocabulary & $K=128$ shared tokens \\
        Action tokens per chunk & $L=8$ \\
        \addlinespace[3pt]
        \cmidrule(r{0.5em}){1-2}
        \multicolumn{2}{@{}l}{\textsc{\textbf{Inputs and preprocessing}}} \\
        \cmidrule(r{0.5em}){1-2}
        Visual input & Two camera views, tiled horizontally into one image \\
        Source frame / VLM image tile & $256\times256$ per view / 512 pixels \\
        Visual tokens & 64 total \\
        Training augmentation & Random crop (0.875), resize to $224\times224$, and color jitter $(0.2,0.2,0.2,0.05)$ \\
        Proprioception & LIBERO 8-D state rendered as text in the prompt \\
        System prompt & Enabled \\
        \addlinespace[3pt]
        \cmidrule(r{0.5em}){1-2}
        \multicolumn{2}{@{}l}{\textsc{\textbf{Optimization}}} \\
        \cmidrule(r{0.5em}){1-2}
        Optimizer & AdamW \\
        Training length & 50,000 optimizer steps \\
        Batch Size & $132$  \\
        Learning-rate schedule & 100-step linear warmup; cosine decay from $1\times10^{-4}$ to $2\times10^{-5}$ \\
        Weight decay & $1\times10^{-10}$ \\
        Gradient-clipping norm & 1.0 \\
        EMA decay & 0.999 \\
        \addlinespace[3pt]
        \cmidrule(r{0.5em}){1-2}
        \multicolumn{2}{@{}l}{\textsc{\textbf{Evaluation}}} \\
        \cmidrule(r{0.5em}){1-2}
        Checkpoint & Last checkpoint \\
        Execution horizon & $H_{\mathrm{exec}}=8$ \\
        Episodes & 500 per suite (50 for each of 10 tasks) \\
        \bottomrule
    \end{tabularx}
\end{table}

%% file: figures/what_matters/reconstruction_error/appendix_reconstruction_success_by_family.tex
\begin{tikzpicture}
\begin{groupplot}[
  group style={group size=3 by 5, horizontal sep=0.4cm, vertical sep=0.22cm,
    xlabels at=edge bottom, xticklabels at=edge bottom,
    ylabels at=edge left, yticklabels at=edge left},
  width=3.9cm, height=2.7cm, scale only axis,
  xmin=0, xmax=0.135, ymin=-4, ymax=104,
  xtick={0,0.04,0.08,0.12}, ytick={0,50,100},
  scaled x ticks=false,
  xticklabel style={/pgf/number format/fixed,
    /pgf/number format/precision=2, font=\fontsize{8}{8.6}\selectfont},
  grid=major, grid style={draw=black!10, line width=0.2pt},
  axis line style={black!55, line width=0.35pt},
  tick style={black!55, line width=0.35pt},
  tick label style={font=\fontsize{8}{8.6}\selectfont},
  label style={font=\fontsize{8}{8.6}\selectfont},
  ylabel={Success (\%)},
]
\nextgroupplot[extra description/.code={
    \node[anchor=north east, fill=white, fill opacity=0.78, text opacity=1, inner sep=0.6pt, font=\fontsize{8}{8.6}\selectfont] at (axis description cs:0.988,0.965) {Can};
    \node[anchor=south west, fill=white, fill opacity=0.78, text opacity=1, inner sep=0.6pt, font=\fontsize{8}{8.6}\selectfont] at (axis description cs:0.03,0.035) {\tokenizerid{vqbase}{VQ}};
    \node[anchor=south east, text=black, fill=white, fill opacity=0.78, text opacity=1, inner sep=0.6pt, font=\fontsize{8}{8.6}\selectfont] at (axis description cs:0.97,0.035) {$\rho=-0.74\,/\,-0.19$};
},]
\addplot[only marks, mark=*, mark size=1.0pt,
  draw=vqbase, fill=vqbase, draw opacity=0.45,
  fill opacity=0.32] coordinates {(0.02177074,93.666667) (0.02209975,88.666667) (0.02244640,94.000000) (0.02316706,90.666667) (0.02343390,95.000000) (0.02354024,91.333333) (0.02354195,93.333333) (0.02366778,88.000000) (0.02469449,84.666667) (0.02477751,97.666667) (0.02496620,86.333333) (0.02505271,92.666667) (0.02553914,90.000000) (0.02558578,91.000000) (0.02578436,89.000000) (0.02586423,90.000000) (0.02629436,90.666667) (0.02635327,92.000000) (0.02640713,93.000000) (0.02684029,94.333333) (0.02687339,87.333333) (0.02776856,91.666667) (0.02848686,93.000000) (0.02982488,91.333333) (0.02987504,91.666667) (0.02991751,91.333333) (0.03347249,90.666667) (0.03450624,89.000000) (0.03645947,83.666667) (0.04080629,87.000000) (0.04191930,86.000000) (0.04390082,88.666667) (0.04444691,88.666667) (0.04739310,94.333333) (0.04876120,74.000000) (0.04934752,81.333333) (0.05293433,83.666667) (0.05306564,80.666667) (0.05342835,68.000000) (0.05393355,69.666667) (0.06008196,67.000000) (0.06368422,54.000000) (0.07468585,15.333333) (0.07475649,41.666667) (0.08018217,23.666667) (0.08125278,11.333333) (0.08577087,11.333333) (0.09553059,4.000000)};
\nextgroupplot[extra description/.code={
    \node[anchor=north east, fill=white, fill opacity=0.78, text opacity=1, inner sep=0.6pt, font=\fontsize{8}{8.6}\selectfont] at (axis description cs:0.988,0.965) {Lift};
    \node[anchor=south east, text=black, fill=white, fill opacity=0.78, text opacity=1, inner sep=0.6pt, font=\fontsize{8}{8.6}\selectfont] at (axis description cs:0.97,0.035) {$\rho=-0.01\,/\,0.15$};
},]
\addplot[only marks, mark=*, mark size=1.0pt,
  draw=vqbase, fill=vqbase, draw opacity=0.45,
  fill opacity=0.32] coordinates {(0.02003136,98.333333) (0.02015980,90.000000) (0.02068917,91.666667) (0.02070285,93.000000) (0.02205708,98.333333) (0.02229138,99.000000) (0.02266154,95.000000) (0.02318682,97.666667) (0.02415088,96.333333) (0.02548922,95.666667) (0.02790197,98.666667) (0.02870759,98.333333) (0.03030386,99.333333) (0.03547833,95.000000) (0.03722124,95.666667) (0.03772391,98.666667) (0.03799803,97.333333) (0.03939824,97.000000) (0.03979339,93.666667) (0.03998317,97.333333) (0.04023505,98.666667) (0.04159456,94.666667) (0.04224774,89.000000) (0.04858917,96.666667)};
\nextgroupplot[extra description/.code={
    \node[anchor=north east, fill=white, fill opacity=0.78, text opacity=1, inner sep=0.6pt, font=\fontsize{8}{8.6}\selectfont] at (axis description cs:0.988,0.965) {Square};
    \node[anchor=south east, text=black, fill=white, fill opacity=0.78, text opacity=1, inner sep=0.6pt, font=\fontsize{8}{8.6}\selectfont] at (axis description cs:0.97,0.035) {$\rho=-0.80\,/\,0.08$};
},]
\addplot[only marks, mark=*, mark size=1.0pt,
  draw=vqbase, fill=vqbase, draw opacity=0.45,
  fill opacity=0.32] coordinates {(0.02234215,69.333333) (0.02261755,66.666667) (0.02326804,66.000000) (0.02383515,69.000000) (0.02489709,61.000000) (0.02552246,61.000000) (0.02587929,62.333333) (0.02600140,62.666667) (0.02916416,68.000000) (0.02968167,71.333333) (0.02977458,69.333333) (0.03241363,67.333333) (0.03796225,63.333333) (0.04282623,62.000000) (0.05053898,50.666667) (0.05165768,52.000000) (0.05291602,60.333333) (0.05316971,50.333333) (0.05349253,52.000000) (0.05601882,47.666667) (0.05611066,34.666667) (0.06028522,35.666667) (0.06977737,5.000000) (0.07006383,41.000000)};
\nextgroupplot[extra description/.code={
    \node[anchor=south west, fill=white, fill opacity=0.78, text opacity=1, inner sep=0.6pt, font=\fontsize{8}{8.6}\selectfont] at (axis description cs:0.03,0.035) {\tokenizerid{rvqbase}{RVQ}};
    \node[anchor=south east, text=black, fill=white, fill opacity=0.78, text opacity=1, inner sep=0.6pt, font=\fontsize{8}{8.6}\selectfont] at (axis description cs:0.97,0.035) {$\rho=0.49\,/\,0.64$};
},]
\addplot[only marks, mark=*, mark size=1.0pt,
  draw=rvqbase, fill=rvqbase, draw opacity=0.45,
  fill opacity=0.32] coordinates {(0.02348944,68.333333) (0.02398452,60.000000) (0.02428708,64.666667) (0.02450481,71.333333) (0.02462638,72.333333) (0.02472464,61.000000) (0.02496495,61.333333) (0.02498107,72.000000) (0.02524753,64.333333) (0.02525481,60.000000) (0.02537289,55.333333) (0.02548410,68.666667) (0.02581146,55.666667) (0.02583886,67.333333) (0.02588001,47.333333) (0.02588460,46.666667) (0.02598068,63.666667) (0.02603594,56.666667) (0.02603660,69.000000) (0.02679290,60.333333) (0.02694364,57.666667) (0.02743817,59.333333) (0.02751362,90.333333) (0.02752025,68.000000) (0.02786541,46.333333) (0.02792931,92.666667) (0.02810386,95.000000) (0.02827142,91.333333) (0.02827169,75.333333) (0.02832728,86.000000) (0.02843274,87.000000) (0.02858920,75.000000) (0.02869187,73.333333) (0.02878989,88.000000) (0.02900449,89.333333) (0.02909385,77.000000) (0.02919668,90.666667) (0.02942737,92.666667) (0.02948899,69.333333) (0.02961876,93.666667) (0.02971132,73.666667) (0.03001092,93.666667) (0.03012847,88.000000) (0.03016751,91.666667) (0.03028256,94.333333) (0.03028457,93.666667) (0.03038959,92.666667) (0.03041981,69.000000) (0.03045323,77.000000) (0.03047706,95.666667) (0.03068426,74.666667) (0.03086593,71.666667) (0.03086893,77.666667) (0.03093988,89.333333) (0.03098346,88.000000) (0.03102476,93.000000) (0.03107411,94.000000) (0.03108271,92.000000) (0.03137019,91.333333) (0.03139162,92.333333) (0.03164276,92.000000) (0.03182193,95.666667) (0.03188819,91.333333) (0.03208945,76.666667) (0.03216325,92.000000) (0.03229166,93.666667) (0.03234699,95.666667) (0.03241727,88.333333) (0.03251643,93.333333) (0.03254334,92.000000) (0.03270280,94.666667) (0.03285778,91.333333) (0.03292578,96.333333) (0.03297271,90.000000) (0.03307785,67.333333) (0.03314365,95.666667) (0.03317572,77.666667) (0.03318249,92.000000) (0.03341446,81.333333) (0.03345526,77.000000) (0.03383786,69.333333) (0.03465798,96.333333) (0.03495385,93.666667) (0.03508670,83.666667) (0.03532011,91.666667) (0.03706969,93.333333) (0.04239700,78.666667) (0.04447270,92.000000) (0.04496905,82.666667) (0.04531507,80.333333) (0.04635602,80.666667) (0.04653024,86.000000) (0.04666635,89.000000) (0.04676768,83.333333) (0.04734460,83.333333) (0.04750518,81.666667) (0.04891210,85.000000) (0.04921271,87.333333) (0.05061847,75.333333) (0.05172738,73.666667)};
\nextgroupplot[extra description/.code={
    \node[anchor=south east, text=black, fill=white, fill opacity=0.78, text opacity=1, inner sep=0.6pt, font=\fontsize{8}{8.6}\selectfont] at (axis description cs:0.97,0.035) {$\rho=0.46\,/\,0.51$};
},]
\addplot[only marks, mark=*, mark size=1.0pt,
  draw=rvqbase, fill=rvqbase, draw opacity=0.45,
  fill opacity=0.32] coordinates {(0.02156265,89.666667) (0.02180869,88.000000) (0.02195851,87.333333) (0.02197014,90.000000) (0.02243853,74.666667) (0.02249485,77.000000) (0.02250162,92.333333) (0.02253355,88.333333) (0.02284710,90.666667) (0.02289385,74.000000) (0.02313881,85.333333) (0.02321563,87.666667) (0.02323501,85.333333) (0.02353426,90.333333) (0.02357111,81.666667) (0.02395445,91.333333) (0.02436764,84.333333) (0.02448311,87.333333) (0.02462369,94.333333) (0.02470127,96.000000) (0.02474907,90.333333) (0.02482557,95.666667) (0.02491220,100.000000) (0.02501073,92.333333) (0.02502012,90.333333) (0.02518335,96.666667) (0.02529328,89.333333) (0.02532651,85.333333) (0.02550146,92.333333) (0.02593570,99.666667) (0.02597616,92.666667) (0.02626887,93.000000) (0.02633056,87.666667) (0.02639029,97.000000) (0.02640929,98.000000) (0.02650913,97.666667) (0.02655137,94.333333) (0.02656654,99.333333) (0.02661491,96.333333) (0.02692449,94.000000) (0.02693112,95.333333) (0.02695489,93.333333) (0.02712489,95.000000) (0.02716681,89.666667) (0.02765209,97.666667) (0.02788447,97.000000) (0.02795561,90.000000) (0.02834637,97.000000) (0.02852573,88.333333) (0.02854253,97.333333) (0.02886889,89.666667) (0.02972717,89.333333) (0.03001641,94.000000) (0.03192360,97.666667) (0.03529409,92.000000) (0.03559823,93.333333) (0.03571446,92.000000) (0.03652055,92.666667) (0.03793498,96.666667) (0.03900411,97.000000) (0.03987722,92.333333) (0.04027575,90.666667) (0.04068968,92.000000) (0.04138584,93.666667) (0.04171816,92.000000) (0.04259897,94.333333)};
\nextgroupplot[extra description/.code={
    \node[anchor=south east, text=black, fill=white, fill opacity=0.78, text opacity=1, inner sep=0.6pt, font=\fontsize{8}{8.6}\selectfont] at (axis description cs:0.97,0.035) {$\rho=0.47\,/\,0.68$};
},]
\addplot[only marks, mark=*, mark size=1.0pt,
  draw=rvqbase, fill=rvqbase, draw opacity=0.45,
  fill opacity=0.32] coordinates {(0.02410176,25.000000) (0.02454357,26.333333) (0.02475948,29.333333) (0.02477814,42.000000) (0.02486900,22.666667) (0.02492261,36.000000) (0.02493796,34.666667) (0.02536970,33.666667) (0.02539196,33.333333) (0.02554216,27.000000) (0.02591750,35.000000) (0.02591750,33.000000) (0.02601925,32.666667) (0.02615064,31.666667) (0.02633995,35.333333) (0.02703546,41.333333) (0.02709607,49.000000) (0.02715069,43.000000) (0.02724176,49.000000) (0.02726233,44.666667) (0.02775210,74.000000) (0.02790210,72.000000) (0.02797028,60.666667) (0.02816819,71.666667) (0.02830390,70.333333) (0.02841399,42.333333) (0.02847353,58.333333) (0.02850331,45.333333) (0.02867648,73.000000) (0.02871386,69.333333) (0.02880806,67.000000) (0.02924408,65.666667) (0.02941125,40.666667) (0.02951895,66.333333) (0.02963292,61.666667) (0.02964262,69.666667) (0.02964720,71.333333) (0.02965302,63.666667) (0.02966402,69.000000) (0.03003563,74.333333) (0.03043450,73.333333) (0.03060491,74.000000) (0.03067433,71.666667) (0.03069128,45.666667) (0.03078830,73.666667) (0.03096367,69.000000) (0.03098330,45.000000) (0.03113065,60.666667) (0.03117106,49.333333) (0.03168628,67.666667) (0.03180016,70.333333) (0.03251520,48.333333) (0.03275800,68.666667) (0.03481970,66.333333) (0.04368111,54.000000) (0.04436558,66.666667) (0.04504055,58.333333) (0.04673477,65.666667) (0.04691772,53.333333) (0.04783748,48.333333) (0.04836843,53.000000) (0.04922578,48.000000) (0.05060019,58.000000) (0.05179549,45.000000) (0.05238445,45.333333) (0.05536420,40.000000)};
\nextgroupplot[extra description/.code={
    \node[anchor=south west, fill=white, fill opacity=0.78, text opacity=1, inner sep=0.6pt, font=\fontsize{8}{8.6}\selectfont] at (axis description cs:0.03,0.035) {\tokenizerid{nrvqbase}{NRVQ}};
    \node[anchor=south east, text=black, fill=white, fill opacity=0.78, text opacity=1, inner sep=0.6pt, font=\fontsize{8}{8.6}\selectfont] at (axis description cs:0.97,0.035) {$\rho=0.22\,/\,0.24$};
},]
\addplot[only marks, mark=*, mark size=1.0pt,
  draw=nrvqbase, fill=nrvqbase, draw opacity=0.45,
  fill opacity=0.32] coordinates {(0.02260288,51.333333) (0.02347576,50.333333) (0.02356593,52.000000) (0.02446434,51.000000) (0.02487055,53.000000) (0.02645264,63.333333) (0.02837088,48.333333) (0.02856759,47.666667) (0.02870987,61.666667) (0.02882365,56.333333) (0.02920366,28.000000) (0.02925207,44.000000) (0.02935059,56.666667) (0.02947666,50.333333) (0.02951255,88.000000) (0.02972526,54.333333) (0.02985964,93.000000) (0.02996113,59.666667) (0.02997368,94.000000) (0.03020120,47.333333) (0.03041445,56.666667) (0.03043502,65.666667) (0.03045861,89.000000) (0.03052444,54.000000) (0.03052918,95.333333) (0.03057973,62.333333) (0.03058657,91.333333) (0.03063267,89.000000) (0.03089029,93.666667) (0.03133546,91.000000) (0.03135399,56.666667) (0.03136583,90.000000) (0.03180828,88.333333) (0.03183383,46.333333) (0.03185219,47.666667) (0.03194282,91.333333) (0.03208219,93.333333) (0.03221648,87.666667) (0.03224048,91.333333) (0.03240275,89.666667) (0.03260221,92.000000) (0.03269967,92.333333) (0.03275510,86.666667) (0.03277645,56.333333) (0.03280744,35.666667) (0.03287090,84.666667) (0.03292327,72.000000) (0.03309676,96.333333) (0.03324192,80.333333) (0.03336015,94.000000) (0.03343302,91.666667) (0.03351165,92.000000) (0.03370007,94.000000) (0.03374611,37.333333) (0.03376131,90.666667) (0.03384825,93.333333) (0.03393831,90.333333) (0.03398169,88.333333) (0.03401801,73.000000) (0.03410873,43.000000) (0.03411364,40.666667) (0.03424843,69.333333) (0.03427252,58.333333) (0.03430538,66.000000) (0.03437385,90.333333) (0.03440970,55.666667) (0.03443177,96.333333) (0.03468208,35.333333) (0.03509895,52.333333) (0.03517160,94.666667) (0.03519122,92.666667) (0.03621764,85.333333) (0.03625985,43.000000) (0.03698568,37.333333) (0.03781341,89.666667) (0.03823121,90.333333) (0.03843553,90.333333) (0.04236874,70.000000) (0.04273100,66.666667) (0.04452558,74.000000) (0.04482924,77.000000) (0.04608174,79.666667) (0.04693946,69.000000) (0.04749391,85.333333) (0.04840597,88.000000) (0.04918765,87.000000)};
\nextgroupplot[extra description/.code={
    \node[anchor=south east, text=black, fill=white, fill opacity=0.78, text opacity=1, inner sep=0.6pt, font=\fontsize{8}{8.6}\selectfont] at (axis description cs:0.97,0.035) {$\rho=0.00\,/\,0.02$};
},]
\addplot[only marks, mark=*, mark size=1.0pt,
  draw=nrvqbase, fill=nrvqbase, draw opacity=0.45,
  fill opacity=0.32] coordinates {(0.01991514,68.000000) (0.02015254,76.000000) (0.02029355,74.666667) (0.02056760,71.666667) (0.02253204,83.333333) (0.02292331,74.666667) (0.02298514,83.333333) (0.02391404,96.666667) (0.02410443,89.666667) (0.02438574,81.666667) (0.02449730,81.000000) (0.02459088,76.666667) (0.02526410,97.333333) (0.02539659,91.666667) (0.02569216,97.666667) (0.02580052,79.333333) (0.02620737,74.333333) (0.02627325,94.666667) (0.02637738,99.000000) (0.02640540,90.333333) (0.02652558,92.666667) (0.02657038,95.333333) (0.02659623,84.333333) (0.02673782,98.666667) (0.02674570,69.333333) (0.02679650,81.333333) (0.02682828,96.000000) (0.02688566,70.000000) (0.02704396,99.333333) (0.02720965,97.666667) (0.02728382,98.333333) (0.02751133,94.333333) (0.02778665,93.000000) (0.02813330,97.666667) (0.02813436,98.333333) (0.02881331,71.000000) (0.02906660,67.333333) (0.03002918,98.333333) (0.03198568,96.333333) (0.03273635,88.000000) (0.03324122,89.333333) (0.03341199,98.000000) (0.03367806,86.333333) (0.03392787,77.333333) (0.03427370,92.333333) (0.03434363,93.666667) (0.03514261,84.000000) (0.03534340,94.666667) (0.03547865,94.333333) (0.03586856,95.333333) (0.03590700,96.666667) (0.03600910,76.333333) (0.03603916,76.000000) (0.03706365,77.333333) (0.03765249,85.666667) (0.03825638,87.666667) (0.03831863,73.000000) (0.03849544,69.000000) (0.03919578,74.333333) (0.03929932,88.666667) (0.03936372,62.666667) (0.03941532,94.333333) (0.03961623,90.000000) (0.04139984,78.666667)};
\nextgroupplot[extra description/.code={
    \node[anchor=south east, text=black, fill=white, fill opacity=0.78, text opacity=1, inner sep=0.6pt, font=\fontsize{8}{8.6}\selectfont] at (axis description cs:0.97,0.035) {$\rho=0.32\,/\,0.33$};
},]
\addplot[only marks, mark=*, mark size=1.0pt,
  draw=nrvqbase, fill=nrvqbase, draw opacity=0.45,
  fill opacity=0.32] coordinates {(0.02286470,21.666667) (0.02314518,25.000000) (0.02324052,23.666667) (0.02329838,25.000000) (0.02402932,16.333333) (0.02463671,24.000000) (0.02736518,25.333333) (0.02754839,19.000000) (0.02762529,28.666667) (0.02784950,19.333333) (0.02819663,37.000000) (0.02834720,11.666667) (0.02846678,25.000000) (0.02867295,19.333333) (0.02871473,31.000000) (0.02880284,20.666667) (0.02910943,21.000000) (0.02929383,24.666667) (0.02942019,73.333333) (0.02956861,77.000000) (0.02964590,58.666667) (0.02976197,76.000000) (0.02981571,73.333333) (0.03014475,70.333333) (0.03109993,63.333333) (0.03110230,76.333333) (0.03113680,59.333333) (0.03134534,63.000000) (0.03167364,65.666667) (0.03168733,63.333333) (0.03174090,71.000000) (0.03198401,73.666667) (0.03209322,66.000000) (0.03216525,67.333333) (0.03216546,69.333333) (0.03243450,25.333333) (0.03252098,67.666667) (0.03257504,29.000000) (0.03324804,67.666667) (0.03342229,18.666667) (0.03368782,14.333333) (0.03372723,75.333333) (0.03408827,18.000000) (0.03409332,62.666667) (0.03581841,72.000000) (0.03886157,16.666667) (0.04180280,44.333333) (0.04264782,48.666667) (0.04537988,61.666667) (0.04665446,46.666667) (0.04758598,58.666667) (0.04780250,61.333333) (0.04938277,58.333333) (0.04990637,49.000000)};
\nextgroupplot[extra description/.code={
    \node[anchor=south west, fill=white, fill opacity=0.78, text opacity=1, inner sep=0.6pt, font=\fontsize{8}{8.6}\selectfont] at (axis description cs:0.03,0.035) {\tokenizerid{fsqbase}{FSQ}};
    \node[anchor=south east, text=black, fill=white, fill opacity=0.78, text opacity=1, inner sep=0.6pt, font=\fontsize{8}{8.6}\selectfont] at (axis description cs:0.97,0.035) {$\rho=0.11\,/\,0.46$};
},]
\addplot[only marks, mark=*, mark size=1.0pt,
  draw=fsqbase, fill=fsqbase, draw opacity=0.45,
  fill opacity=0.32] coordinates {(0.02729423,91.666667) (0.02754472,91.666667) (0.02759309,89.333333) (0.02765827,90.333333) (0.02784346,91.666667) (0.02793458,90.000000) (0.02805502,96.000000) (0.02805826,92.000000) (0.02821795,93.000000) (0.02851864,89.333333) (0.02855287,88.666667) (0.02857257,89.333333) (0.02863970,89.666667) (0.02866222,91.000000) (0.02866404,92.000000) (0.02883589,95.000000) (0.02904847,88.000000) (0.02990174,93.333333) (0.02990945,92.000000) (0.03002641,95.666667) (0.03020268,83.333333) (0.03031944,90.666667) (0.03032323,92.000000) (0.03036835,94.333333) (0.03078102,89.333333) (0.03084295,97.000000) (0.03086964,91.666667) (0.03176841,92.000000) (0.03190702,91.000000) (0.03195712,93.666667) (0.03233492,94.000000) (0.03238032,92.333333) (0.03292334,93.000000) (0.03297770,95.333333) (0.03367001,93.333333) (0.03395314,93.666667) (0.03402163,94.666667) (0.03429865,95.000000) (0.03430264,93.000000) (0.03435981,90.333333) (0.03481535,95.000000) (0.03649139,94.333333) (0.04511941,82.333333) (0.04638779,82.333333) (0.04895826,77.333333) (0.04991211,79.666667)};
\nextgroupplot[extra description/.code={
    \node[anchor=south east, text=black, fill=white, fill opacity=0.78, text opacity=1, inner sep=0.6pt, font=\fontsize{8}{8.6}\selectfont] at (axis description cs:0.97,0.035) {$\rho=-0.47\,/\,-0.47$};
},]
\addplot[only marks, mark=*, mark size=1.0pt,
  draw=fsqbase, fill=fsqbase, draw opacity=0.45,
  fill opacity=0.32] coordinates {(0.02367191,99.333333) (0.02376146,99.000000) (0.02397210,96.666667) (0.02428537,97.333333) (0.02443883,98.333333) (0.02447251,99.000000) (0.02480140,99.000000) (0.02553767,96.666667) (0.02555467,99.000000) (0.02579706,97.333333) (0.02598581,98.000000) (0.02697431,100.000000) (0.02727093,98.333333) (0.02739137,98.333333) (0.02800736,96.000000) (0.02807896,97.000000) (0.02844621,98.666667) (0.02864792,98.666667) (0.03631472,97.666667) (0.03671036,96.666667) (0.03871506,92.666667) (0.03994464,86.666667)};
\nextgroupplot[extra description/.code={
    \node[anchor=south east, text=black, fill=white, fill opacity=0.78, text opacity=1, inner sep=0.6pt, font=\fontsize{8}{8.6}\selectfont] at (axis description cs:0.97,0.035) {$\rho=-0.30\,/\,0.19$};
},]
\addplot[only marks, mark=*, mark size=1.0pt,
  draw=fsqbase, fill=fsqbase, draw opacity=0.45,
  fill opacity=0.32] coordinates {(0.02672680,67.000000) (0.02680518,61.666667) (0.02706648,61.666667) (0.02787759,78.000000) (0.02815905,71.333333) (0.02844137,73.333333) (0.02896679,76.000000) (0.02901016,71.666667) (0.02924518,66.333333) (0.02966483,66.333333) (0.02980508,63.333333) (0.03022574,71.000000) (0.03063296,71.333333) (0.03090325,71.666667) (0.03168223,70.000000) (0.03173762,71.333333) (0.03296146,76.000000) (0.03344398,70.333333) (0.04431601,58.333333) (0.04562695,63.666667) (0.05010296,57.333333) (0.05089945,49.666667)};
\nextgroupplot[extra description/.code={
    \node[anchor=south west, fill=white, fill opacity=0.78, text opacity=1, inner sep=0.6pt, font=\fontsize{8}{8.6}\selectfont] at (axis description cs:0.03,0.035) {\tokenizerid{qincobase}{QinCo}};
    \node[anchor=south east, text=black, fill=white, fill opacity=0.78, text opacity=1, inner sep=0.6pt, font=\fontsize{8}{8.6}\selectfont] at (axis description cs:0.97,0.035) {$\rho=0.68\,/\,0.76$};
},]
\addplot[only marks, mark=*, mark size=1.0pt,
  draw=qincobase, fill=qincobase, draw opacity=0.45,
  fill opacity=0.32] coordinates {(0.02256171,50.333333) (0.02283113,45.666667) (0.02306333,51.333333) (0.02347083,52.666667) (0.02351151,54.000000) (0.02360458,52.333333) (0.02378904,50.666667) (0.02384813,45.333333) (0.02389796,54.000000) (0.02390913,44.333333) (0.02391855,50.666667) (0.02397519,41.666667) (0.02398163,53.333333) (0.02408398,52.333333) (0.02415209,55.666667) (0.02420277,44.000000) (0.02426473,61.000000) (0.02428678,31.000000) (0.02430556,45.000000) (0.02445199,39.333333) (0.02457642,43.666667) (0.02468164,39.666667) (0.02473129,41.666667) (0.02486119,61.333333) (0.02492240,59.333333) (0.02496576,43.333333) (0.02504671,50.000000) (0.02507234,66.333333) (0.02518721,40.666667) (0.02529290,62.666667) (0.02551356,54.333333) (0.02563991,67.333333) (0.02564950,57.000000) (0.02579032,61.000000) (0.02590852,64.333333) (0.02603214,64.000000) (0.02639253,65.000000) (0.02665141,75.666667) (0.02703788,68.333333) (0.02711917,53.666667) (0.02726673,63.000000) (0.02769503,56.000000) (0.02771032,45.666667) (0.02816470,68.666667) (0.02823241,71.333333) (0.02832111,50.666667) (0.02854706,52.333333) (0.02859562,54.000000) (0.02878598,61.333333) (0.02891615,57.333333) (0.02899308,46.000000) (0.02915281,54.666667) (0.02915552,59.000000) (0.03021060,61.000000) (0.03072302,76.666667) (0.03080993,65.666667) (0.03100779,67.000000) (0.03136025,70.000000) (0.03163011,75.666667) (0.03176286,60.666667) (0.03243984,52.666667) (0.03284203,56.333333) (0.03284926,70.666667) (0.03334672,79.333333) (0.03400524,82.000000) (0.03403142,73.666667) (0.03410127,73.666667) (0.03485307,84.666667) (0.03502358,80.666667) (0.03556359,59.333333) (0.03652825,87.666667) (0.03671005,78.000000) (0.03671662,74.000000) (0.03672040,76.666667) (0.03765027,80.333333) (0.03767762,88.333333) (0.03788932,86.666667) (0.03789286,92.333333) (0.03806563,88.000000) (0.04159470,62.666667) (0.04168978,75.000000) (0.04173664,69.666667) (0.04246368,88.666667) (0.04404230,59.000000) (0.04418935,73.000000) (0.04476402,69.333333) (0.04709431,62.333333) (0.04733646,68.666667) (0.04743969,60.000000) (0.04813423,64.333333) (0.04825115,78.666667) (0.04869004,73.000000) (0.04871019,71.333333) (0.04934074,78.666667) (0.11014960,5.000000)};
\nextgroupplot[extra description/.code={
    \node[anchor=south east, text=black, fill=white, fill opacity=0.78, text opacity=1, inner sep=0.6pt, font=\fontsize{8}{8.6}\selectfont] at (axis description cs:0.97,0.035) {$\rho=0.67\,/\,0.69$};
},]
\addplot[only marks, mark=*, mark size=1.0pt,
  draw=qincobase, fill=qincobase, draw opacity=0.45,
  fill opacity=0.32] coordinates {(0.01990622,77.000000) (0.02030039,81.666667) (0.02035363,79.666667) (0.02047066,81.666667) (0.02054162,79.666667) (0.02061678,77.333333) (0.02070485,72.000000) (0.02075229,76.666667) (0.02077532,75.000000) (0.02082252,73.333333) (0.02091343,81.000000) (0.02121928,71.666667) (0.02128435,79.666667) (0.02144174,82.333333) (0.02162034,60.333333) (0.02190195,78.666667) (0.02192267,82.666667) (0.02197417,80.000000) (0.02212134,86.000000) (0.02234299,88.333333) (0.02235025,77.000000) (0.02235642,80.333333) (0.02281601,84.000000) (0.02283674,79.000000) (0.02283972,74.000000) (0.02288601,79.000000) (0.02290697,87.333333) (0.02315066,77.000000) (0.02328739,75.333333) (0.02376116,75.333333) (0.02384939,78.333333) (0.02394402,82.666667) (0.02409785,81.000000) (0.02452767,80.000000) (0.02465730,84.000000) (0.02480572,80.666667) (0.02629533,92.000000) (0.02657327,86.333333) (0.02675882,82.666667) (0.02694648,85.666667) (0.02730063,87.333333) (0.02740274,83.666667) (0.02785628,86.666667) (0.02803704,88.000000) (0.02803858,91.333333) (0.02814725,91.000000) (0.02819923,87.333333) (0.02824701,97.000000) (0.02862924,96.000000) (0.02878260,98.000000) (0.02942093,96.333333) (0.03035654,86.666667) (0.03054457,91.666667) (0.03415819,81.666667) (0.03467647,85.666667) (0.03501029,86.666667) (0.03578804,86.000000) (0.03857145,87.000000) (0.03861829,84.000000) (0.03923600,87.666667) (0.03948309,81.666667) (0.03990473,88.666667) (0.04007799,91.000000) (0.04069104,96.666667) (0.04112524,87.000000) (0.06123085,79.666667)};
\nextgroupplot[extra description/.code={
    \node[anchor=south east, text=black, fill=white, fill opacity=0.78, text opacity=1, inner sep=0.6pt, font=\fontsize{8}{8.6}\selectfont] at (axis description cs:0.97,0.035) {$\rho=0.71\,/\,0.87$};
},]
\addplot[only marks, mark=*, mark size=1.0pt,
  draw=qincobase, fill=qincobase, draw opacity=0.45,
  fill opacity=0.32] coordinates {(0.02205495,5.333333) (0.02239834,15.666667) (0.02261296,2.000000) (0.02282642,13.333333) (0.02294031,16.000000) (0.02300065,15.000000) (0.02302245,14.666667) (0.02304049,18.666667) (0.02309633,12.000000) (0.02361707,25.333333) (0.02388722,11.000000) (0.02393305,11.666667) (0.02393901,21.000000) (0.02398159,14.000000) (0.02431791,12.333333) (0.02447025,28.000000) (0.02469152,27.666667) (0.02474205,25.333333) (0.02487694,26.333333) (0.02494499,26.666667) (0.02508059,13.000000) (0.02518049,27.666667) (0.02592719,29.000000) (0.02623317,27.666667) (0.02627173,30.666667) (0.02639815,33.000000) (0.02649628,28.333333) (0.02667937,18.000000) (0.02686976,25.666667) (0.02701331,23.333333) (0.02702064,32.000000) (0.02702974,17.333333) (0.02703062,20.666667) (0.02715259,32.000000) (0.02724893,18.666667) (0.02743396,28.333333) (0.02937146,42.000000) (0.03096792,50.000000) (0.03137165,46.333333) (0.03141856,40.333333) (0.03186513,31.666667) (0.03186758,49.333333) (0.03287421,66.333333) (0.03301012,39.000000) (0.03328550,49.000000) (0.03343075,48.333333) (0.03360122,55.000000) (0.03369375,53.000000) (0.03477606,54.333333) (0.03508207,53.666667) (0.03531728,53.666667) (0.03550452,47.333333) (0.03676600,49.666667) (0.04297575,25.333333) (0.04348085,35.333333) (0.04493702,29.333333) (0.04569170,33.000000) (0.04862101,37.000000) (0.04883935,43.000000) (0.04884692,42.000000) (0.04970183,41.000000) (0.05009949,39.000000) (0.05028857,43.666667) (0.05049051,37.000000) (0.05178517,39.333333) (0.09323193,1.000000)};
\end{groupplot}
\node[font=\fontsize{8}{8.6}\selectfont, inner ysep=2pt]
  at ($(group c1r5.south west)!0.5!(group c3r5.south east)+(0,-0.5cm)$)
  {Mean-Absolute-Error on Action Chunk};
\end{tikzpicture}

%% file: figures/what_matters/reconstruction_error/appendix_flow_matching_noise_response.tex
\begin{tikzpicture}
\begin{groupplot}[
  group style={group size=3 by 1, horizontal sep=0.52cm,
    ylabels at=edge left, yticklabels at=edge left},
  width=3.55cm, height=2.35cm, scale only axis,
  xmin=-0.01, xmax=0.51, ymin=-2, ymax=104,
  xtick={0,0.1,0.2,0.3,0.4,0.5}, ytick={0,50,100},
  scaled x ticks=false,
  xticklabel style={/pgf/number format/fixed,
    /pgf/number format/precision=1, font=\fontsize{6}{6.4}\selectfont},
  grid=major, grid style={draw=black!10, line width=0.2pt},
  axis line style={black!55, line width=0.35pt},
  tick style={black!55, line width=0.35pt},
  tick label style={font=\fontsize{6}{6.4}\selectfont},
  label style={font=\fontsize{6}{6.4}\selectfont},
  ylabel={Success (\%)},
  ylabel style={at={(yticklabel cs:0.5,2pt)},anchor=near yticklabel,inner ysep=0pt,yshift=-\pgfkeysvalueof{/pgf/inner xsep}},
]
\nextgroupplot[extra description/.code={
  \node[anchor=north east, fill=white, fill opacity=0.82,
    text opacity=1, inner sep=0.7pt,
    font=\fontsize{6}{6.4}\selectfont]
    at (axis description cs:0.985,0.965) {Can};},]
\addplot[black!78, solid, line width=0.7pt, mark=*,
  mark size=1.35pt, mark options={solid, draw=black!78, fill=black!78},
  error bars/.cd, y dir=both, y explicit, error mark=none]
  coordinates {(0.00,97.000000) +- (0,1.927022) (0.01,96.666667) +- (0,2.019226) (0.02,98.333333) +- (0,1.440019) (0.03,97.333333) +- (0,1.817323) (0.04,95.000000) +- (0,2.471371) (0.05,96.666667) +- (0,2.006148) (0.06,94.333333) +- (0,2.613889) (0.07,93.666667) +- (0,2.795526) (0.08,91.333333) +- (0,3.159573) (0.09,89.333333) +- (0,3.501114) (0.10,87.333333) +- (0,3.801301) (0.20,54.000000) +- (0,5.613307) (0.30,17.333333) +- (0,4.226271) (0.40,5.333333) +- (0,2.543070) (0.50,0.666667) +- (0,0.923937)};
\nextgroupplot[extra description/.code={
  \node[anchor=north east, fill=white, fill opacity=0.82,
    text opacity=1, inner sep=0.7pt,
    font=\fontsize{6}{6.4}\selectfont]
    at (axis description cs:0.985,0.965) {Lift};},]
\addplot[black!78, solid, line width=0.7pt, mark=square*,
  mark size=1.35pt, mark options={solid, draw=black!78, fill=black!78},
  error bars/.cd, y dir=both, y explicit, error mark=none]
  coordinates {(0.00,100.000000) +- (0,0.000000) (0.01,100.000000) +- (0,0.000000) (0.02,100.000000) +- (0,0.000000) (0.03,100.000000) +- (0,0.000000) (0.04,100.000000) +- (0,0.000000) (0.05,100.000000) +- (0,0.000000) (0.06,100.000000) +- (0,0.000000) (0.07,100.000000) +- (0,0.000000) (0.08,100.000000) +- (0,0.000000) (0.09,100.000000) +- (0,0.000000) (0.10,100.000000) +- (0,0.000000) (0.20,95.000000) +- (0,2.451559) (0.30,64.333333) +- (0,5.445539) (0.40,20.000000) +- (0,4.537428) (0.50,7.000000) +- (0,2.885800)};
\nextgroupplot[extra description/.code={
  \node[anchor=north east, fill=white, fill opacity=0.82,
    text opacity=1, inner sep=0.7pt,
    font=\fontsize{6}{6.4}\selectfont]
    at (axis description cs:0.985,0.965) {Square};},]
\addplot[black!78, solid, line width=0.7pt, mark=triangle*,
  mark size=1.35pt, mark options={solid, draw=black!78, fill=black!78},
  error bars/.cd, y dir=both, y explicit, error mark=none]
  coordinates {(0.00,79.000000) +- (0,4.646750) (0.01,83.333333) +- (0,4.252842) (0.02,81.000000) +- (0,4.465988) (0.03,79.000000) +- (0,4.567234) (0.04,82.333333) +- (0,4.297780) (0.05,79.666667) +- (0,4.574952) (0.06,80.333333) +- (0,4.483557) (0.07,82.000000) +- (0,4.356729) (0.08,76.333333) +- (0,4.826671) (0.09,71.666667) +- (0,5.112075) (0.10,73.666667) +- (0,4.947511) (0.20,26.666667) +- (0,5.045786) (0.30,2.666667) +- (0,1.831516) (0.40,0.000000) +- (0,0.000000) (0.50,0.000000) +- (0,0.000000)};
\end{groupplot}
\node[font=\fontsize{6}{6.4}\selectfont]
  at ($(group c1r1.south west)!0.5!(group c3r1.south east)+(0,-0.43cm)$)
  {Joint pose-noise amplitude $\epsilon$};
\pgfresetboundingbox
\path[use as bounding box]
  ($(group c1r1.south west)+(-0.78cm,-0.62cm)$) rectangle
  ($(group c3r1.north east)+(0.10cm,0.05cm)$);
\end{tikzpicture}

%% file: figures/what_matters/reconstruction_error/appendix_libero_fm_noise_response.tex
\begin{tikzpicture}
\begin{groupplot}[
  group style={group size=4 by 1, horizontal sep=0.42cm,
    ylabels at=edge left, yticklabels at=edge left},
  width=2.6075cm, height=2.35cm, scale only axis,
  xmin=-0.006, xmax=0.206,
  ymin=40, ymax=92,
  xtick={0,0.04,0.08,0.12,0.16,0.20},
  minor xtick={0.02,0.06,0.10,0.14,0.18},
  ytick={40,50,60,70,80,90}, minor ytick={45,55,65,75,85},
  scaled x ticks=false,
  xticklabel style={/pgf/number format/fixed,
    /pgf/number format/precision=2, rotate=90, anchor=east, font=\fontsize{6}{6.4}\selectfont},
  grid=both, grid style={draw=black!10, line width=0.2pt},
  minor grid style={draw=black!6, line width=0.15pt},
  axis line style={black!55, line width=0.35pt},
  tick style={black!55, line width=0.35pt},
  tick label style={font=\fontsize{6}{6.4}\selectfont},
  label style={font=\fontsize{6}{6.4}\selectfont},
  ylabel={Success (\%)},
  ylabel style={at={(yticklabel cs:0.5,2pt)},anchor=near yticklabel,inner ysep=0pt,yshift=-\pgfkeysvalueof{/pgf/inner xsep}},
]
\nextgroupplot[extra description/.code={
  \node[anchor=north east, fill=white, fill opacity=0.82,
    text opacity=1, inner sep=0.7pt,
    font=\fontsize{6}{6.4}\selectfont]
    at (axis description cs:0.985,0.965) {Goal};},]
\addplot[black!25, dashed, line width=0.5pt, mark=none, forget plot] coordinates {(-0.006,84.2000) (0.206,84.2000)};
\addplot[black!78, solid, line width=0.7pt, mark=triangle*,
  mark size=1.2pt, mark options={solid, draw=black!78, fill=black!78},
  error bars/.cd, y dir=both, y explicit, error mark=none]
  coordinates {(0.00,84.2000) +- (0,3.1999) (0.02,85.4000) +- (0,3.0830) (0.04,85.0000) +- (0,3.1264) (0.06,83.8000) +- (0,3.2405) (0.08,85.8000) +- (0,3.0710) (0.10,81.4000) +- (0,3.3975) (0.12,78.2000) +- (0,3.6459) (0.14,78.8000) +- (0,3.5822) (0.16,74.4000) +- (0,3.8507) (0.18,67.0000) +- (0,4.1344) (0.20,64.8000) +- (0,4.2159)};
\nextgroupplot[extra description/.code={
  \node[anchor=north east, fill=white, fill opacity=0.82,
    text opacity=1, inner sep=0.7pt,
    font=\fontsize{6}{6.4}\selectfont]
    at (axis description cs:0.985,0.965) {Object};},]
\addplot[black!25, dashed, line width=0.5pt, mark=none, forget plot] coordinates {(-0.006,80.8000) (0.206,80.8000)};
\addplot[black!78, solid, line width=0.7pt, mark=square*,
  mark size=1.2pt, mark options={solid, draw=black!78, fill=black!78},
  error bars/.cd, y dir=both, y explicit, error mark=none]
  coordinates {(0.00,80.8000) +- (0,3.5084) (0.02,82.2000) +- (0,3.4399) (0.04,79.0000) +- (0,3.6322) (0.06,76.6000) +- (0,3.7864) (0.08,79.2000) +- (0,3.5926) (0.10,77.0000) +- (0,3.7411) (0.12,75.8000) +- (0,3.8007) (0.14,74.8000) +- (0,3.8318) (0.16,72.6000) +- (0,3.9673) (0.18,69.4000) +- (0,4.0368) (0.20,64.4000) +- (0,4.2405)};
\nextgroupplot[extra description/.code={
  \node[anchor=north east, fill=white, fill opacity=0.82,
    text opacity=1, inner sep=0.7pt,
    font=\fontsize{6}{6.4}\selectfont]
    at (axis description cs:0.985,0.965) {Spatial};},]
\addplot[black!25, dashed, line width=0.5pt, mark=none, forget plot] coordinates {(-0.006,83.0000) (0.206,83.0000)};
\addplot[black!78, solid, line width=0.7pt, mark=*,
  mark size=1.2pt, mark options={solid, draw=black!78, fill=black!78},
  error bars/.cd, y dir=both, y explicit, error mark=none]
  coordinates {(0.00,83.0000) +- (0,3.2832) (0.02,83.8000) +- (0,3.2461) (0.04,81.4000) +- (0,3.4215) (0.06,80.8000) +- (0,3.4630) (0.08,82.0000) +- (0,3.3722) (0.10,80.8000) +- (0,3.4526) (0.12,80.2000) +- (0,3.4883) (0.14,77.2000) +- (0,3.7089) (0.16,78.0000) +- (0,3.6088) (0.18,76.0000) +- (0,3.7499) (0.20,74.2000) +- (0,3.8044)};
\nextgroupplot[extra description/.code={
  \node[anchor=north east, fill=white, fill opacity=0.82,
    text opacity=1, inner sep=0.7pt,
    font=\fontsize{6}{6.4}\selectfont]
    at (axis description cs:0.985,0.965) {LIBERO-10};},]
\addplot[black!25, dashed, line width=0.5pt, mark=none, forget plot] coordinates {(-0.006,68.4000) (0.206,68.4000)};
\addplot[black!78, solid, line width=0.7pt, mark=diamond*,
  mark size=1.2pt, mark options={solid, draw=black!78, fill=black!78},
  error bars/.cd, y dir=both, y explicit, error mark=none]
  coordinates {(0.00,68.4000) +- (0,4.0737) (0.02,69.4000) +- (0,4.0022) (0.04,65.4000) +- (0,4.1569) (0.06,62.8000) +- (0,4.2276) (0.08,61.0000) +- (0,4.2704) (0.10,61.0000) +- (0,4.2903) (0.12,59.0000) +- (0,4.3335) (0.14,54.8000) +- (0,4.3674) (0.16,48.2000) +- (0,4.3745) (0.18,47.2000) +- (0,4.4093) (0.20,44.4000) +- (0,4.2955)};
\end{groupplot}
\node[font=\fontsize{6}{6.4}\selectfont]
  at ($(group c1r1.south west)!0.5!(group c4r1.south east)+(0,-0.70cm)$)
  {Joint pose-noise amplitude $\epsilon$};
\pgfresetboundingbox
\path[use as bounding box]
  ($(group c1r1.south west)+(-0.78cm,-0.85cm)$) rectangle
  ($(group c4r1.north east)+(0.10cm,0.05cm)$);
\end{tikzpicture}

%% file: figures/what_matters/reconstruction_error/appendix_robotwin_fm_noise_response.tex
\begin{tikzpicture}
\begin{groupplot}[
  group style={group size=3 by 1, horizontal sep=0.52cm,
    ylabels at=edge left, yticklabels at=edge left},
  width=3.55cm, height=2.35cm, scale only axis,
  xmin=-0.08, xmax=3.08, ymin=-3, ymax=105,
  xtick={0,0.5,1,1.5,2,2.5,3}, ytick={0,50,100},
  scaled x ticks=false,
  xticklabel style={/pgf/number format/fixed,
    /pgf/number format/precision=1, font=\fontsize{6}{6.4}\selectfont},
  grid=major, grid style={draw=black!10, line width=0.2pt},
  axis line style={black!55, line width=0.35pt},
  tick style={black!55, line width=0.35pt},
  tick label style={font=\fontsize{6}{6.4}\selectfont},
  label style={font=\fontsize{6}{6.4}\selectfont},
  ylabel={Success (\%)},
  ylabel style={at={(yticklabel cs:0.5,2pt)},anchor=near yticklabel,inner ysep=0pt,yshift=-\pgfkeysvalueof{/pgf/inner xsep}},
]
\nextgroupplot[extra description/.code={
  \node[anchor=north east, fill=white, fill opacity=0.82,
    text opacity=1, inner sep=0.7pt,
    font=\fontsize{6}{6.4}\selectfont]
    at (axis description cs:0.985,0.965) {Grab roller};},]
\addplot[black!25, dashed, line width=0.45pt] coordinates {(1,-3) (1,105)};
\addplot[black!78, solid, line width=0.7pt, mark=*,
  mark size=1.25pt, mark options={solid, draw=black!78, fill=black!78},
  error bars/.cd, y dir=both, y explicit, error mark=none]
  coordinates {(0.000,93) +- (0,4.92)
  (0.049,96) +- (0,5.43) (0.099,94) +- (0,6.49)
  (0.148,96) +- (0,5.43) (0.197,94) +- (0,6.49)
  (0.296,94) +- (0,6.49) (0.394,94) +- (0,6.49)
  (0.493,92) +- (0,7.51) (0.592,86) +- (0,9.66)
  (0.789,86) +- (0,9.66) (0.986,79) +- (0,7.94)
  (1.183,78) +- (0,11.37) (1.479,64) +- (0,13.19)
  (1.972,38) +- (0,9.51) (2.465,28) +- (0,12.39)
  (2.959,16) +- (0,7.13)};
\nextgroupplot[xmin=-0.05, xmax=2.05, xtick={0,0.5,1,1.5,2},
  extra description/.code={
  \node[anchor=north east, fill=white, fill opacity=0.82,
    text opacity=1, inner sep=0.7pt,
    font=\fontsize{6}{6.4}\selectfont]
    at (axis description cs:0.985,0.965) {Place burger/fries};},]
\addplot[black!25, dashed, line width=0.45pt] coordinates {(1,-3) (1,105)};
\addplot[black!78, solid, line width=0.7pt, mark=square*,
  mark size=1.25pt, mark options={solid, draw=black!78, fill=black!78},
  error bars/.cd, y dir=both, y explicit, error mark=none]
  coordinates {(0.000,92) +- (0,5.28)
  (0.049,94) +- (0,6.49) (0.099,94) +- (0,6.49)
  (0.148,96) +- (0,5.43) (0.197,94) +- (0,6.49)
  (0.296,94) +- (0,6.49) (0.394,80) +- (0,11.10)
  (0.493,76) +- (0,11.72) (0.592,60) +- (0,13.46)
  (0.789,32) +- (0,12.85) (0.986,40) +- (0,13.54)
  (1.183,12) +- (0,8.97) (1.479,8) +- (0,7.45)
  (1.972,0) +- (0,0) (2.465,0) +- (0,0)
  (2.959,0) +- (0,0)};
\nextgroupplot[xmin=-0.05, xmax=2.05, xtick={0,0.5,1,1.5,2},
  extra description/.code={
  \node[anchor=north east, fill=white, fill opacity=0.82,
    text opacity=1, inner sep=0.7pt,
    font=\fontsize{6}{6.4}\selectfont]
    at (axis description cs:0.985,0.965) {Place cans};},]
\addplot[black!25, dashed, line width=0.45pt] coordinates {(1,-3) (1,105)};
\addplot[black!78, solid, line width=0.7pt, mark=triangle*,
  mark size=1.25pt, mark options={solid, draw=black!78, fill=black!78},
  error bars/.cd, y dir=both, y explicit, error mark=none]
  coordinates {(0.000,82) +- (0,7.48)
  (0.049,94) +- (0,6.49) (0.099,92) +- (0,7.51)
  (0.148,84) +- (0,10.21) (0.197,80) +- (0,11.10)
  (0.296,66) +- (0,13.05) (0.394,54) +- (0,13.70)
  (0.493,26) +- (0,12.17) (0.592,24) +- (0,11.82)
  (0.789,10) +- (0,8.25) (0.986,6) +- (0,6.53)
  (1.183,2) +- (0,3.85) (1.479,0) +- (0,0)
  (1.972,0) +- (0,0) (2.465,0) +- (0,0)
  (2.959,0) +- (0,0)};
\end{groupplot}
\node[font=\fontsize{6}{6.4}\selectfont]
  at ($(group c1r1.south west)!0.5!(group c3r1.south east)+(0,-0.43cm)$)
  {Uniform noise relative to translation scale $\sigma/s_{\mathrm{trans}}$};
\pgfresetboundingbox
\path[use as bounding box]
  ($(group c1r1.south west)+(-0.78cm,-0.62cm)$) rectangle
  ($(group c3r1.north east)+(0.10cm,0.05cm)$);
\end{tikzpicture}

%% file: figures/what_matters/reconstruction_error/appendix_fm_noise_precision_heatmap.tex
\begin{tikzpicture}
\begin{groupplot}[
  group style={group size=3 by 1, horizontal sep=0.52cm},
  width=3.40cm, height=3.40cm, scale only axis,
  enlargelimits=false, axis on top,
  point meta min=0, point meta max=100,
  colormap/viridis,
  tick label style={font=\fontsize{6}{6.4}\selectfont},
  label style={font=\fontsize{6}{6.4}\selectfont},
  axis line style={black!55, line width=0.35pt},
  tick style={black!55, line width=0.35pt},
  xlabel={$\epsilon_{\mathrm{trans}}$},
  ylabel style={yshift=-3pt},
]
\nextgroupplot[title={Can}, title style={font=\fontsize{6}{6.4}\selectfont, yshift=-2pt},
  xtick={0,1,2,3,4,5,6,7}, xticklabels={0,0.02,0.05,0.1,0.2,0.3,0.4,0.5},
  ytick={0,1,2,3,4,5,6,7}, yticklabels={0,0.02,0.05,0.1,0.2,0.3,0.4,0.5},
  x tick label style={rotate=45, anchor=north east, font=\fontsize{6}{6.4}\selectfont},
  xmin=-0.5, xmax=7.5, ymin=-0.5, ymax=7.5,
  ylabel={$\epsilon_{\mathrm{rot}}$},
]
\addplot[matrix plot*, point meta=explicit, mesh/cols=8] table[meta=c] {
x y c
0 0 90.6667
1 0 91.3333
2 0 91.3333
3 0 91.3333
4 0 87.0000
5 0 77.0000
6 0 66.3333
7 0 46.3333
0 1 89.3333
1 1 92.3333
2 1 90.3333
3 1 88.3333
4 1 86.3333
5 1 79.6667
6 1 67.0000
7 1 50.3333
0 2 91.3333
1 2 91.0000
2 2 92.3333
3 2 89.6667
4 2 84.3333
5 2 75.3333
6 2 67.0000
7 2 51.0000
0 3 86.0000
1 3 85.3333
2 3 88.0000
3 3 86.6667
4 3 82.3333
5 3 72.3333
6 3 61.6667
7 3 43.3333
0 4 74.3333
1 4 70.0000
2 4 67.6667
3 4 69.6667
4 4 61.6667
5 4 51.6667
6 4 40.6667
7 4 27.0000
0 5 37.0000
1 5 38.0000
2 5 38.3333
3 5 34.0000
4 5 30.6667
5 5 23.3333
6 5 20.3333
7 5 17.3333
0 6 14.0000
1 6 12.0000
2 6 13.3333
3 6 11.3333
4 6 12.0000
5 6 8.3333
6 6 5.3333
7 6 5.0000
0 7 6.6667
1 7 5.3333
2 7 3.6667
3 7 4.3333
4 7 4.0000
5 7 3.0000
6 7 3.3333
7 7 2.3333
};
\node[font=\fontsize{5.5}{5.9}\selectfont, text=black] at (axis cs:0,0) {91};
\node[font=\fontsize{5.5}{5.9}\selectfont, text=black] at (axis cs:1,0) {91};
\node[font=\fontsize{5.5}{5.9}\selectfont, text=black] at (axis cs:2,0) {91};
\node[font=\fontsize{5.5}{5.9}\selectfont, text=black] at (axis cs:3,0) {91};
\node[font=\fontsize{5.5}{5.9}\selectfont, text=black] at (axis cs:4,0) {87};
\node[font=\fontsize{5.5}{5.9}\selectfont, text=black] at (axis cs:5,0) {77};
\node[font=\fontsize{5.5}{5.9}\selectfont, text=black] at (axis cs:6,0) {66};
\node[font=\fontsize{5.5}{5.9}\selectfont, text=white] at (axis cs:7,0) {46};
\node[font=\fontsize{5.5}{5.9}\selectfont, text=black] at (axis cs:0,1) {89};
\node[font=\fontsize{5.5}{5.9}\selectfont, text=black] at (axis cs:1,1) {92};
\node[font=\fontsize{5.5}{5.9}\selectfont, text=black] at (axis cs:2,1) {90};
\node[font=\fontsize{5.5}{5.9}\selectfont, text=black] at (axis cs:3,1) {88};
\node[font=\fontsize{5.5}{5.9}\selectfont, text=black] at (axis cs:4,1) {86};
\node[font=\fontsize{5.5}{5.9}\selectfont, text=black] at (axis cs:5,1) {80};
\node[font=\fontsize{5.5}{5.9}\selectfont, text=black] at (axis cs:6,1) {67};
\node[font=\fontsize{5.5}{5.9}\selectfont, text=white] at (axis cs:7,1) {50};
\node[font=\fontsize{5.5}{5.9}\selectfont, text=black] at (axis cs:0,2) {91};
\node[font=\fontsize{5.5}{5.9}\selectfont, text=black] at (axis cs:1,2) {91};
\node[font=\fontsize{5.5}{5.9}\selectfont, text=black] at (axis cs:2,2) {92};
\node[font=\fontsize{5.5}{5.9}\selectfont, text=black] at (axis cs:3,2) {90};
\node[font=\fontsize{5.5}{5.9}\selectfont, text=black] at (axis cs:4,2) {84};
\node[font=\fontsize{5.5}{5.9}\selectfont, text=black] at (axis cs:5,2) {75};
\node[font=\fontsize{5.5}{5.9}\selectfont, text=black] at (axis cs:6,2) {67};
\node[font=\fontsize{5.5}{5.9}\selectfont, text=white] at (axis cs:7,2) {51};
\node[font=\fontsize{5.5}{5.9}\selectfont, text=black] at (axis cs:0,3) {86};
\node[font=\fontsize{5.5}{5.9}\selectfont, text=black] at (axis cs:1,3) {85};
\node[font=\fontsize{5.5}{5.9}\selectfont, text=black] at (axis cs:2,3) {88};
\node[font=\fontsize{5.5}{5.9}\selectfont, text=black] at (axis cs:3,3) {87};
\node[font=\fontsize{5.5}{5.9}\selectfont, text=black] at (axis cs:4,3) {82};
\node[font=\fontsize{5.5}{5.9}\selectfont, text=black] at (axis cs:5,3) {72};
\node[font=\fontsize{5.5}{5.9}\selectfont, text=black] at (axis cs:6,3) {62};
\node[font=\fontsize{5.5}{5.9}\selectfont, text=white] at (axis cs:7,3) {43};
\node[font=\fontsize{5.5}{5.9}\selectfont, text=black] at (axis cs:0,4) {74};
\node[font=\fontsize{5.5}{5.9}\selectfont, text=black] at (axis cs:1,4) {70};
\node[font=\fontsize{5.5}{5.9}\selectfont, text=black] at (axis cs:2,4) {68};
\node[font=\fontsize{5.5}{5.9}\selectfont, text=black] at (axis cs:3,4) {70};
\node[font=\fontsize{5.5}{5.9}\selectfont, text=black] at (axis cs:4,4) {62};
\node[font=\fontsize{5.5}{5.9}\selectfont, text=white] at (axis cs:5,4) {52};
\node[font=\fontsize{5.5}{5.9}\selectfont, text=white] at (axis cs:6,4) {41};
\node[font=\fontsize{5.5}{5.9}\selectfont, text=white] at (axis cs:7,4) {27};
\node[font=\fontsize{5.5}{5.9}\selectfont, text=white] at (axis cs:0,5) {37};
\node[font=\fontsize{5.5}{5.9}\selectfont, text=white] at (axis cs:1,5) {38};
\node[font=\fontsize{5.5}{5.9}\selectfont, text=white] at (axis cs:2,5) {38};
\node[font=\fontsize{5.5}{5.9}\selectfont, text=white] at (axis cs:3,5) {34};
\node[font=\fontsize{5.5}{5.9}\selectfont, text=white] at (axis cs:4,5) {31};
\node[font=\fontsize{5.5}{5.9}\selectfont, text=white] at (axis cs:5,5) {23};
\node[font=\fontsize{5.5}{5.9}\selectfont, text=white] at (axis cs:6,5) {20};
\node[font=\fontsize{5.5}{5.9}\selectfont, text=white] at (axis cs:7,5) {17};
\node[font=\fontsize{5.5}{5.9}\selectfont, text=white] at (axis cs:0,6) {14};
\node[font=\fontsize{5.5}{5.9}\selectfont, text=white] at (axis cs:1,6) {12};
\node[font=\fontsize{5.5}{5.9}\selectfont, text=white] at (axis cs:2,6) {13};
\node[font=\fontsize{5.5}{5.9}\selectfont, text=white] at (axis cs:3,6) {11};
\node[font=\fontsize{5.5}{5.9}\selectfont, text=white] at (axis cs:4,6) {12};
\node[font=\fontsize{5.5}{5.9}\selectfont, text=white] at (axis cs:5,6) {8};
\node[font=\fontsize{5.5}{5.9}\selectfont, text=white] at (axis cs:6,6) {5};
\node[font=\fontsize{5.5}{5.9}\selectfont, text=white] at (axis cs:7,6) {5};
\node[font=\fontsize{5.5}{5.9}\selectfont, text=white] at (axis cs:0,7) {7};
\node[font=\fontsize{5.5}{5.9}\selectfont, text=white] at (axis cs:1,7) {5};
\node[font=\fontsize{5.5}{5.9}\selectfont, text=white] at (axis cs:2,7) {4};
\node[font=\fontsize{5.5}{5.9}\selectfont, text=white] at (axis cs:3,7) {4};
\node[font=\fontsize{5.5}{5.9}\selectfont, text=white] at (axis cs:4,7) {4};
\node[font=\fontsize{5.5}{5.9}\selectfont, text=white] at (axis cs:5,7) {3};
\node[font=\fontsize{5.5}{5.9}\selectfont, text=white] at (axis cs:6,7) {3};
\node[font=\fontsize{5.5}{5.9}\selectfont, text=white] at (axis cs:7,7) {2};
\nextgroupplot[title={Lift}, title style={font=\fontsize{6}{6.4}\selectfont, yshift=-2pt},
  xtick={0,1,2,3,4,5,6,7}, xticklabels={0,0.02,0.05,0.1,0.2,0.3,0.4,0.5},
  ytick={0,1,2,3,4,5,6,7}, yticklabels={0,0.02,0.05,0.1,0.2,0.3,0.4,0.5},
  x tick label style={rotate=45, anchor=north east, font=\fontsize{6}{6.4}\selectfont},
  xmin=-0.5, xmax=7.5, ymin=-0.5, ymax=7.5,
]
\addplot[matrix plot*, point meta=explicit, mesh/cols=8] table[meta=c] {
x y c
0 0 99.6667
1 0 99.6667
2 0 99.6667
3 0 99.6667
4 0 99.3333
5 0 99.0000
6 0 93.6667
7 0 91.0000
0 1 100.0000
1 1 100.0000
2 1 100.0000
3 1 100.0000
4 1 99.3333
5 1 98.0000
6 1 94.6667
7 1 89.0000
0 2 100.0000
1 2 100.0000
2 2 100.0000
3 2 99.6667
4 2 99.3333
5 2 98.6667
6 2 97.0000
7 2 89.0000
0 3 99.0000
1 3 99.6667
2 3 98.3333
3 3 99.6667
4 3 99.0000
5 3 97.0000
6 3 93.3333
7 3 86.6667
0 4 92.3333
1 4 94.0000
2 4 93.6667
3 4 92.6667
4 4 89.6667
5 4 84.0000
6 4 78.3333
7 4 70.3333
0 5 66.6667
1 5 64.3333
2 5 68.0000
3 5 69.6667
4 5 60.0000
5 5 59.6667
6 5 45.6667
7 5 41.6667
0 6 33.0000
1 6 33.0000
2 6 33.3333
3 6 25.0000
4 6 31.0000
5 6 20.3333
6 6 23.0000
7 6 19.6667
0 7 11.6667
1 7 9.3333
2 7 11.0000
3 7 10.6667
4 7 10.0000
5 7 10.6667
6 7 12.3333
7 7 7.3333
};
\node[font=\fontsize{5.5}{5.9}\selectfont, text=black] at (axis cs:0,0) {100};
\node[font=\fontsize{5.5}{5.9}\selectfont, text=black] at (axis cs:1,0) {100};
\node[font=\fontsize{5.5}{5.9}\selectfont, text=black] at (axis cs:2,0) {100};
\node[font=\fontsize{5.5}{5.9}\selectfont, text=black] at (axis cs:3,0) {100};
\node[font=\fontsize{5.5}{5.9}\selectfont, text=black] at (axis cs:4,0) {99};
\node[font=\fontsize{5.5}{5.9}\selectfont, text=black] at (axis cs:5,0) {99};
\node[font=\fontsize{5.5}{5.9}\selectfont, text=black] at (axis cs:6,0) {94};
\node[font=\fontsize{5.5}{5.9}\selectfont, text=black] at (axis cs:7,0) {91};
\node[font=\fontsize{5.5}{5.9}\selectfont, text=black] at (axis cs:0,1) {100};
\node[font=\fontsize{5.5}{5.9}\selectfont, text=black] at (axis cs:1,1) {100};
\node[font=\fontsize{5.5}{5.9}\selectfont, text=black] at (axis cs:2,1) {100};
\node[font=\fontsize{5.5}{5.9}\selectfont, text=black] at (axis cs:3,1) {100};
\node[font=\fontsize{5.5}{5.9}\selectfont, text=black] at (axis cs:4,1) {99};
\node[font=\fontsize{5.5}{5.9}\selectfont, text=black] at (axis cs:5,1) {98};
\node[font=\fontsize{5.5}{5.9}\selectfont, text=black] at (axis cs:6,1) {95};
\node[font=\fontsize{5.5}{5.9}\selectfont, text=black] at (axis cs:7,1) {89};
\node[font=\fontsize{5.5}{5.9}\selectfont, text=black] at (axis cs:0,2) {100};
\node[font=\fontsize{5.5}{5.9}\selectfont, text=black] at (axis cs:1,2) {100};
\node[font=\fontsize{5.5}{5.9}\selectfont, text=black] at (axis cs:2,2) {100};
\node[font=\fontsize{5.5}{5.9}\selectfont, text=black] at (axis cs:3,2) {100};
\node[font=\fontsize{5.5}{5.9}\selectfont, text=black] at (axis cs:4,2) {99};
\node[font=\fontsize{5.5}{5.9}\selectfont, text=black] at (axis cs:5,2) {99};
\node[font=\fontsize{5.5}{5.9}\selectfont, text=black] at (axis cs:6,2) {97};
\node[font=\fontsize{5.5}{5.9}\selectfont, text=black] at (axis cs:7,2) {89};
\node[font=\fontsize{5.5}{5.9}\selectfont, text=black] at (axis cs:0,3) {99};
\node[font=\fontsize{5.5}{5.9}\selectfont, text=black] at (axis cs:1,3) {100};
\node[font=\fontsize{5.5}{5.9}\selectfont, text=black] at (axis cs:2,3) {98};
\node[font=\fontsize{5.5}{5.9}\selectfont, text=black] at (axis cs:3,3) {100};
\node[font=\fontsize{5.5}{5.9}\selectfont, text=black] at (axis cs:4,3) {99};
\node[font=\fontsize{5.5}{5.9}\selectfont, text=black] at (axis cs:5,3) {97};
\node[font=\fontsize{5.5}{5.9}\selectfont, text=black] at (axis cs:6,3) {93};
\node[font=\fontsize{5.5}{5.9}\selectfont, text=black] at (axis cs:7,3) {87};
\node[font=\fontsize{5.5}{5.9}\selectfont, text=black] at (axis cs:0,4) {92};
\node[font=\fontsize{5.5}{5.9}\selectfont, text=black] at (axis cs:1,4) {94};
\node[font=\fontsize{5.5}{5.9}\selectfont, text=black] at (axis cs:2,4) {94};
\node[font=\fontsize{5.5}{5.9}\selectfont, text=black] at (axis cs:3,4) {93};
\node[font=\fontsize{5.5}{5.9}\selectfont, text=black] at (axis cs:4,4) {90};
\node[font=\fontsize{5.5}{5.9}\selectfont, text=black] at (axis cs:5,4) {84};
\node[font=\fontsize{5.5}{5.9}\selectfont, text=black] at (axis cs:6,4) {78};
\node[font=\fontsize{5.5}{5.9}\selectfont, text=black] at (axis cs:7,4) {70};
\node[font=\fontsize{5.5}{5.9}\selectfont, text=black] at (axis cs:0,5) {67};
\node[font=\fontsize{5.5}{5.9}\selectfont, text=black] at (axis cs:1,5) {64};
\node[font=\fontsize{5.5}{5.9}\selectfont, text=black] at (axis cs:2,5) {68};
\node[font=\fontsize{5.5}{5.9}\selectfont, text=black] at (axis cs:3,5) {70};
\node[font=\fontsize{5.5}{5.9}\selectfont, text=black] at (axis cs:4,5) {60};
\node[font=\fontsize{5.5}{5.9}\selectfont, text=black] at (axis cs:5,5) {60};
\node[font=\fontsize{5.5}{5.9}\selectfont, text=white] at (axis cs:6,5) {46};
\node[font=\fontsize{5.5}{5.9}\selectfont, text=white] at (axis cs:7,5) {42};
\node[font=\fontsize{5.5}{5.9}\selectfont, text=white] at (axis cs:0,6) {33};
\node[font=\fontsize{5.5}{5.9}\selectfont, text=white] at (axis cs:1,6) {33};
\node[font=\fontsize{5.5}{5.9}\selectfont, text=white] at (axis cs:2,6) {33};
\node[font=\fontsize{5.5}{5.9}\selectfont, text=white] at (axis cs:3,6) {25};
\node[font=\fontsize{5.5}{5.9}\selectfont, text=white] at (axis cs:4,6) {31};
\node[font=\fontsize{5.5}{5.9}\selectfont, text=white] at (axis cs:5,6) {20};
\node[font=\fontsize{5.5}{5.9}\selectfont, text=white] at (axis cs:6,6) {23};
\node[font=\fontsize{5.5}{5.9}\selectfont, text=white] at (axis cs:7,6) {20};
\node[font=\fontsize{5.5}{5.9}\selectfont, text=white] at (axis cs:0,7) {12};
\node[font=\fontsize{5.5}{5.9}\selectfont, text=white] at (axis cs:1,7) {9};
\node[font=\fontsize{5.5}{5.9}\selectfont, text=white] at (axis cs:2,7) {11};
\node[font=\fontsize{5.5}{5.9}\selectfont, text=white] at (axis cs:3,7) {11};
\node[font=\fontsize{5.5}{5.9}\selectfont, text=white] at (axis cs:4,7) {10};
\node[font=\fontsize{5.5}{5.9}\selectfont, text=white] at (axis cs:5,7) {11};
\node[font=\fontsize{5.5}{5.9}\selectfont, text=white] at (axis cs:6,7) {12};
\node[font=\fontsize{5.5}{5.9}\selectfont, text=white] at (axis cs:7,7) {7};
\nextgroupplot[title={Square}, title style={font=\fontsize{6}{6.4}\selectfont, yshift=-2pt},
  xtick={0,1,2,3,4,5,6}, xticklabels={0,0.02,0.05,0.1,0.2,0.3,0.4},
  ytick={0,1,2,3,4,5,6}, yticklabels={0,0.02,0.05,0.1,0.2,0.3,0.4},
  x tick label style={rotate=45, anchor=north east, font=\fontsize{6}{6.4}\selectfont},
  xmin=-0.5, xmax=6.5, ymin=-0.5, ymax=6.5,
  colorbar right, colorbar style={width=0.12cm,
    tick label style={font=\fontsize{6}{6.4}\selectfont}, ytick={0,25,50,75,100},
    ylabel={Success (\%)}, ylabel style={font=\fontsize{6}{6.4}\selectfont}},
]
\addplot[matrix plot*, point meta=explicit, mesh/cols=7] table[meta=c] {
x y c
0 0 65.6667
1 0 62.3333
2 0 61.3333
3 0 60.3333
4 0 58.3333
5 0 43.6667
6 0 38.0000
0 1 61.3333
1 1 62.3333
2 1 58.6667
3 1 59.3333
4 1 55.6667
5 1 44.6667
6 1 33.0000
0 2 58.0000
1 2 60.6667
2 2 55.6667
3 2 57.3333
4 2 56.6667
5 2 45.3333
6 2 36.6667
0 3 47.3333
1 3 48.0000
2 3 45.6667
3 3 48.0000
4 3 41.0000
5 3 36.3333
6 3 28.3333
0 4 28.6667
1 4 28.0000
2 4 27.6667
3 4 24.6667
4 4 22.0000
5 4 14.6667
6 4 12.6667
0 5 9.0000
1 5 5.3333
2 5 4.0000
3 5 5.0000
4 5 5.0000
5 5 5.0000
6 5 2.6667
0 6 2.3333
1 6 1.6667
2 6 0.3333
3 6 0.6667
4 6 0.6667
5 6 0.6667
6 6 0.6667
};
\node[font=\fontsize{5.5}{5.9}\selectfont, text=black] at (axis cs:0,0) {66};
\node[font=\fontsize{5.5}{5.9}\selectfont, text=black] at (axis cs:1,0) {62};
\node[font=\fontsize{5.5}{5.9}\selectfont, text=black] at (axis cs:2,0) {61};
\node[font=\fontsize{5.5}{5.9}\selectfont, text=black] at (axis cs:3,0) {60};
\node[font=\fontsize{5.5}{5.9}\selectfont, text=black] at (axis cs:4,0) {58};
\node[font=\fontsize{5.5}{5.9}\selectfont, text=white] at (axis cs:5,0) {44};
\node[font=\fontsize{5.5}{5.9}\selectfont, text=white] at (axis cs:6,0) {38};
\node[font=\fontsize{5.5}{5.9}\selectfont, text=black] at (axis cs:0,1) {61};
\node[font=\fontsize{5.5}{5.9}\selectfont, text=black] at (axis cs:1,1) {62};
\node[font=\fontsize{5.5}{5.9}\selectfont, text=black] at (axis cs:2,1) {59};
\node[font=\fontsize{5.5}{5.9}\selectfont, text=black] at (axis cs:3,1) {59};
\node[font=\fontsize{5.5}{5.9}\selectfont, text=white] at (axis cs:4,1) {56};
\node[font=\fontsize{5.5}{5.9}\selectfont, text=white] at (axis cs:5,1) {45};
\node[font=\fontsize{5.5}{5.9}\selectfont, text=white] at (axis cs:6,1) {33};
\node[font=\fontsize{5.5}{5.9}\selectfont, text=white] at (axis cs:0,2) {58};
\node[font=\fontsize{5.5}{5.9}\selectfont, text=black] at (axis cs:1,2) {61};
\node[font=\fontsize{5.5}{5.9}\selectfont, text=white] at (axis cs:2,2) {56};
\node[font=\fontsize{5.5}{5.9}\selectfont, text=white] at (axis cs:3,2) {57};
\node[font=\fontsize{5.5}{5.9}\selectfont, text=white] at (axis cs:4,2) {57};
\node[font=\fontsize{5.5}{5.9}\selectfont, text=white] at (axis cs:5,2) {45};
\node[font=\fontsize{5.5}{5.9}\selectfont, text=white] at (axis cs:6,2) {37};
\node[font=\fontsize{5.5}{5.9}\selectfont, text=white] at (axis cs:0,3) {47};
\node[font=\fontsize{5.5}{5.9}\selectfont, text=white] at (axis cs:1,3) {48};
\node[font=\fontsize{5.5}{5.9}\selectfont, text=white] at (axis cs:2,3) {46};
\node[font=\fontsize{5.5}{5.9}\selectfont, text=white] at (axis cs:3,3) {48};
\node[font=\fontsize{5.5}{5.9}\selectfont, text=white] at (axis cs:4,3) {41};
\node[font=\fontsize{5.5}{5.9}\selectfont, text=white] at (axis cs:5,3) {36};
\node[font=\fontsize{5.5}{5.9}\selectfont, text=white] at (axis cs:6,3) {28};
\node[font=\fontsize{5.5}{5.9}\selectfont, text=white] at (axis cs:0,4) {29};
\node[font=\fontsize{5.5}{5.9}\selectfont, text=white] at (axis cs:1,4) {28};
\node[font=\fontsize{5.5}{5.9}\selectfont, text=white] at (axis cs:2,4) {28};
\node[font=\fontsize{5.5}{5.9}\selectfont, text=white] at (axis cs:3,4) {25};
\node[font=\fontsize{5.5}{5.9}\selectfont, text=white] at (axis cs:4,4) {22};
\node[font=\fontsize{5.5}{5.9}\selectfont, text=white] at (axis cs:5,4) {15};
\node[font=\fontsize{5.5}{5.9}\selectfont, text=white] at (axis cs:6,4) {13};
\node[font=\fontsize{5.5}{5.9}\selectfont, text=white] at (axis cs:0,5) {9};
\node[font=\fontsize{5.5}{5.9}\selectfont, text=white] at (axis cs:1,5) {5};
\node[font=\fontsize{5.5}{5.9}\selectfont, text=white] at (axis cs:2,5) {4};
\node[font=\fontsize{5.5}{5.9}\selectfont, text=white] at (axis cs:3,5) {5};
\node[font=\fontsize{5.5}{5.9}\selectfont, text=white] at (axis cs:4,5) {5};
\node[font=\fontsize{5.5}{5.9}\selectfont, text=white] at (axis cs:5,5) {5};
\node[font=\fontsize{5.5}{5.9}\selectfont, text=white] at (axis cs:6,5) {3};
\node[font=\fontsize{5.5}{5.9}\selectfont, text=white] at (axis cs:0,6) {2};
\node[font=\fontsize{5.5}{5.9}\selectfont, text=white] at (axis cs:1,6) {2};
\node[font=\fontsize{5.5}{5.9}\selectfont, text=white] at (axis cs:2,6) {0};
\node[font=\fontsize{5.5}{5.9}\selectfont, text=white] at (axis cs:3,6) {1};
\node[font=\fontsize{5.5}{5.9}\selectfont, text=white] at (axis cs:4,6) {1};
\node[font=\fontsize{5.5}{5.9}\selectfont, text=white] at (axis cs:5,6) {1};
\node[font=\fontsize{5.5}{5.9}\selectfont, text=white] at (axis cs:6,6) {1};
\end{groupplot}
\end{tikzpicture}

%% file: figures/what_matters/reconstruction_error/appendix_tokenized_fm_reconstruction_success.tex
\begin{tikzpicture}
\begin{groupplot}[
  group style={group size=3 by 1, horizontal sep=0.52cm,
    ylabels at=edge left, yticklabels at=edge left},
  width=3.55cm, height=2.35cm, scale only axis,
  xmin=0.01, xmax=0.055, ymin=0, ymax=100,
  xtick={0.01,0.02,0.03,0.04,0.05}, ytick={0,50,100},
  scaled x ticks=false,
  xticklabel style={/pgf/number format/fixed,
    /pgf/number format/precision=2, font=\fontsize{6}{6.4}\selectfont},
  grid=major, grid style={draw=black!10, line width=0.2pt},
  axis line style={black!55, line width=0.35pt},
  tick style={black!55, line width=0.35pt},
  tick label style={font=\fontsize{6}{6.4}\selectfont},
  label style={font=\fontsize{6}{6.4}\selectfont},
  ylabel={Success (\%)},
]
\nextgroupplot[extra description/.code={
  \node[anchor=north east, fill=white, fill opacity=0.82,
    text opacity=1, inner sep=0.7pt,
    font=\fontsize{6}{6.4}\selectfont]
    at (axis description cs:0.985,0.965) {Can};},]
\addplot[black!65, densely dashed, line width=0.55pt, forget plot] coordinates {(0.01,97.666667) (0.055,97.666667)};
\addplot+[only marks, mark=*, mark size=0.72pt,
  draw=vqbase, fill=vqbase, draw opacity=0.58,
  fill opacity=0.40] coordinates {(0.03199674,91.333333) (0.04122258,83.000000)};
\addplot+[only marks, mark=*, mark size=0.72pt,
  draw=nrvqbase, fill=nrvqbase, draw opacity=0.58,
  fill opacity=0.40] coordinates {(0.01570123,98.333333) (0.01792156,97.333333) (0.01804294,94.666667) (0.01857954,96.333333) (0.01887604,96.666667) (0.01932934,97.000000) (0.01945665,95.000000) (0.01953249,98.000000) (0.01976423,96.666667) (0.02010338,97.666667) (0.02026901,95.333333) (0.02027015,98.000000) (0.02028331,97.333333) (0.02069944,96.000000) (0.02099061,95.666667) (0.02106729,95.333333) (0.02118381,97.000000) (0.02134470,96.666667) (0.02144365,96.333333) (0.02148793,95.666667) (0.02171690,93.000000) (0.02203969,97.333333) (0.02224414,96.000000) (0.02249265,97.000000) (0.02250581,97.000000) (0.02265901,95.333333) (0.02266011,97.666667) (0.02286223,96.333333) (0.02286941,96.333333) (0.02335450,97.333333) (0.02403964,96.000000) (0.02412730,94.666667) (0.02418157,96.333333) (0.02498224,96.666667) (0.02519177,96.666667) (0.02544817,93.666667) (0.02548574,97.666667) (0.02595524,96.333333) (0.02644457,94.666667) (0.02692819,90.666667) (0.02711449,97.000000) (0.02735982,96.333333) (0.02737918,94.666667) (0.02762570,95.000000) (0.02791416,94.000000) (0.02817527,93.333333) (0.02846770,95.666667) (0.02916053,92.666667) (0.02946215,93.000000) (0.02953045,95.000000) (0.02970570,94.333333) (0.02972121,95.000000) (0.02985291,96.666667) (0.02987183,94.000000) (0.03018088,94.333333) (0.03040048,96.000000) (0.03152205,93.333333) (0.03170664,94.333333) (0.03176084,93.666667) (0.03235063,92.000000) (0.03242557,91.333333) (0.03254105,95.666667) (0.03406520,95.666667) (0.03484172,94.666667) (0.03505636,90.000000) (0.03513045,90.333333) (0.03630055,92.000000) (0.03635669,92.333333) (0.03657256,93.333333) (0.03679003,92.666667) (0.03714599,94.000000) (0.03903189,90.666667) (0.04021075,79.666667) (0.04610499,87.000000)};
\addplot+[only marks, mark=*, mark size=0.72pt,
  draw=fsqbase, fill=fsqbase, draw opacity=0.58,
  fill opacity=0.40] coordinates {(0.02169395,95.666667) (0.02443818,96.666667) (0.02479994,96.333333) (0.02490683,91.666667) (0.02603074,95.333333) (0.02715573,96.333333) (0.02786711,96.666667) (0.02863168,92.666667) (0.02896538,95.333333) (0.03051980,90.666667) (0.03116152,92.333333) (0.03204023,95.000000) (0.03225181,91.333333) (0.03253009,94.333333) (0.03279173,91.666667) (0.03455089,95.000000) (0.03492778,91.666667) (0.03541153,92.333333) (0.03602360,91.333333) (0.03747604,90.000000) (0.03844106,90.333333) (0.03906145,85.333333) (0.04240490,87.000000) (0.04249629,77.000000) (0.04295765,81.333333) (0.04342608,59.333333) (0.04454698,85.333333) (0.04751842,72.000000) (0.05167166,64.666667)};
\node[anchor=south east, fill=white, fill opacity=0.82,
  text opacity=1, inner sep=0.6pt,
  font=\fontsize{6}{6.4}\selectfont]
  at (axis description cs:0.985,0.035) {$\rho=-0.80$};
\nextgroupplot[extra description/.code={
  \node[anchor=north east, fill=white, fill opacity=0.82,
    text opacity=1, inner sep=0.7pt,
    font=\fontsize{6}{6.4}\selectfont]
    at (axis description cs:0.985,0.965) {Lift};},]
\addplot+[only marks, mark=*, mark size=0.72pt,
  draw=nrvqbase, fill=nrvqbase, draw opacity=0.58,
  fill opacity=0.40] coordinates {(0.01399774,100.000000) (0.01482297,100.000000) (0.01494283,100.000000) (0.01612585,100.000000) (0.01632941,100.000000) (0.01646915,100.000000) (0.01679587,100.000000) (0.01679939,100.000000) (0.01681323,100.000000) (0.01690722,100.000000) (0.01707535,100.000000) (0.01735863,100.000000) (0.01738721,100.000000) (0.01745185,100.000000) (0.01794126,100.000000) (0.01799297,100.000000) (0.01802112,100.000000) (0.01846728,100.000000) (0.01855369,100.000000) (0.01862535,100.000000) (0.01870359,100.000000) (0.01877345,100.000000) (0.01891705,100.000000) (0.01919609,100.000000) (0.01932926,100.000000) (0.01939048,100.000000) (0.01947252,100.000000) (0.02028188,100.000000) (0.02053497,100.000000) (0.02133923,99.666667) (0.02147164,99.666667) (0.02178345,100.000000) (0.02254939,100.000000) (0.02275981,100.000000) (0.02300854,100.000000) (0.02302708,100.000000) (0.02334440,99.666667) (0.02342331,100.000000) (0.02376402,100.000000) (0.02381793,100.000000) (0.02409662,100.000000) (0.02445394,100.000000) (0.02450862,100.000000) (0.02495794,100.000000) (0.02553239,100.000000) (0.02616227,100.000000) (0.02664868,100.000000) (0.02720540,99.666667) (0.02923415,99.666667) (0.02935538,100.000000) (0.02981732,100.000000) (0.03055154,100.000000) (0.03071311,99.666667) (0.03163379,99.000000)};
\addplot+[only marks, mark=*, mark size=0.72pt,
  draw=fsqbase, fill=fsqbase, draw opacity=0.58,
  fill opacity=0.40] coordinates {(0.01860840,100.000000) (0.02134864,100.000000) (0.02143276,99.666667) (0.02423953,99.666667) (0.02688025,99.666667) (0.02919983,100.000000) (0.02973685,100.000000) (0.03158811,99.333333) (0.03482697,100.000000)};
\node[anchor=south east, fill=white, fill opacity=0.82,
  text opacity=1, inner sep=0.6pt,
  font=\fontsize{6}{6.4}\selectfont]
  at (axis description cs:0.985,0.035) {$\rho=-0.41$};
\nextgroupplot[extra description/.code={
  \node[anchor=north east, fill=white, fill opacity=0.82,
    text opacity=1, inner sep=0.7pt,
    font=\fontsize{6}{6.4}\selectfont]
    at (axis description cs:0.985,0.965) {Square};},]
\addplot[black!65, densely dashed, line width=0.55pt, forget plot] coordinates {(0.01,81.333333) (0.055,81.333333)};
\addplot+[only marks, mark=*, mark size=0.72pt,
  draw=vqbase, fill=vqbase, draw opacity=0.58,
  fill opacity=0.40] coordinates {(0.02914982,67.000000) (0.03781045,51.000000)};
\addplot+[only marks, mark=*, mark size=0.72pt,
  draw=nrvqbase, fill=nrvqbase, draw opacity=0.58,
  fill opacity=0.40] coordinates {(0.01582986,82.666667) (0.01621810,84.000000) (0.01691880,78.333333) (0.01764422,79.333333) (0.01784785,80.000000) (0.01838357,79.333333) (0.01854209,73.666667) (0.01868900,75.666667) (0.01877934,76.333333) (0.01902979,75.333333) (0.01915867,81.333333) (0.01920212,76.000000) (0.01921519,73.333333) (0.01947840,79.333333) (0.01948850,77.000000) (0.01968575,73.666667) (0.01993919,78.666667) (0.02010765,77.000000) (0.02015564,76.333333) (0.02025697,79.333333) (0.02046415,74.333333) (0.02064212,74.666667) (0.02072915,70.666667) (0.02078541,73.000000) (0.02085976,74.333333) (0.02090836,71.333333) (0.02096513,80.666667) (0.02114422,78.333333) (0.02231755,68.666667) (0.02251146,76.000000) (0.02288459,72.333333) (0.02313317,71.666667) (0.02331456,69.666667) (0.02355314,67.333333) (0.02358758,76.666667) (0.02363423,67.333333) (0.02389701,71.333333) (0.02425407,65.333333) (0.02444773,73.000000) (0.02472357,68.000000) (0.02577053,66.666667) (0.02583135,69.333333) (0.02583924,72.000000) (0.02606817,76.666667) (0.02607063,72.333333) (0.02625438,70.000000) (0.02642566,67.666667) (0.02644954,66.666667) (0.02693476,70.000000) (0.02709135,61.666667) (0.02710325,67.666667) (0.02713501,69.666667) (0.02726323,69.000000) (0.02782265,67.000000) (0.02889118,56.666667) (0.02918205,62.333333) (0.02939229,76.333333) (0.02979855,66.000000) (0.02981406,67.666667) (0.03003059,64.333333) (0.03050704,70.333333) (0.03109198,59.000000) (0.03109331,62.000000) (0.03130816,64.666667) (0.03148199,55.666667) (0.03217280,57.333333) (0.03258550,59.666667) (0.03275558,58.333333) (0.03277349,53.333333) (0.03315628,58.333333) (0.03363367,53.666667) (0.03472729,57.000000) (0.03482336,60.333333) (0.03791906,41.666667)};
\addplot+[only marks, mark=*, mark size=0.72pt,
  draw=fsqbase, fill=fsqbase, draw opacity=0.58,
  fill opacity=0.40] coordinates {(0.02044666,71.666667) (0.02276238,64.000000) (0.02292862,68.333333) (0.02304600,73.000000) (0.02375600,81.666667) (0.02529602,69.333333) (0.02540881,72.333333) (0.02628011,71.000000) (0.02688019,61.666667) (0.02722150,67.000000) (0.02796701,52.666667) (0.02817950,65.333333) (0.02847198,66.000000) (0.02916693,62.333333) (0.03054363,64.333333) (0.03085524,61.666667) (0.03102173,61.666667) (0.03109252,61.333333) (0.03188656,58.000000) (0.03317383,45.000000) (0.03413033,61.000000) (0.03428287,53.000000) (0.03526716,47.333333) (0.03632205,38.333333) (0.03662499,50.000000) (0.03673925,46.000000) (0.03729297,44.666667) (0.04163656,25.000000) (0.04311008,19.666667)};
\node[anchor=south east, fill=white, fill opacity=0.82,
  text opacity=1, inner sep=0.6pt,
  font=\fontsize{6}{6.4}\selectfont]
  at (axis description cs:0.985,0.035) {$\rho=-0.90$};
\end{groupplot}
\node[anchor=east, draw=black!55, line width=0.35pt, fill=white,
  minimum height=2.35cm, inner xsep=2.0pt, inner ysep=0pt,
  font=\fontsize{6}{6.4}\selectfont]
  at ([xshift=-1.30cm]group c1r1.west) {\vbox to 2.08cm{
    \hbox{\tokenizerid{vqbase}{VQ}}\vfil
    \hbox{\tokenizerid{nrvqbase}{NRVQ}}\vfil
    \hbox{\tokenizerid{fsqbase}{FSQ}}
  }};
\node[font=\fontsize{6}{6.4}\selectfont]
  at ($(group c1r1.south west)!0.5!(group c3r1.south east)+(0,-0.43cm)$)
  {On-policy 7-D reconstruction MAE};
\end{tikzpicture}

%% file: figures/what_matters/reconstruction_error/appendix_tokenized_fm_vs_policy_success.tex
\begin{tikzpicture}
\begin{groupplot}[
  group style={group size=3 by 1, horizontal sep=0.45cm,
    ylabels at=edge left, yticklabels at=edge left},
  width=3.5cm, height=3.5cm, scale only axis,
  xmin=-4, xmax=104, ymin=-4, ymax=104,
  xtick={0,50,100}, ytick={0,50,100},
  grid=major, grid style={draw=black!10, line width=0.2pt},
  axis line style={black!55, line width=0.35pt},
  tick style={black!55, line width=0.35pt},
  tick label style={font=\fontsize{7}{7.5}\selectfont},
  label style={font=\fontsize{7}{7.5}\selectfont},
  ylabel={Tokenized-FM success (\%)},
]
\nextgroupplot[extra description/.code={
    \node[anchor=north east, fill=white, fill opacity=0.78, text opacity=1, inner sep=0.6pt, font=\fontsize{7}{7.5}\selectfont] at (axis description cs:0.975,0.972) {Can};
    \node[anchor=south west, inner sep=0.6pt, font=\fontsize{6.5}{7}\selectfont] at (axis description cs:0.04,0.25) {\tokenizerid{fsqbase}{TF-FSQ-S}};
    \node[anchor=south west, inner sep=0.6pt, font=\fontsize{6.5}{7}\selectfont] at (axis description cs:0.04,0.14) {\tokenizerid{nrvqbase}{TF-NRVQ-S}};
    \node[anchor=south west, inner sep=0.6pt, font=\fontsize{6.5}{7}\selectfont] at (axis description cs:0.04,0.04) {\tokenizerid{nrvqbase}{TF-NRVQ-L}};
},]
\addplot[black!45, densely dashed, line width=0.5pt] coordinates {(0,0) (100,100)};
\addplot[only marks, mark=square*, mark size=1.05pt,
  draw=fsqbase, fill=fsqbase, draw opacity=0.55,
  fill opacity=0.38] coordinates {(71.6667,90.0000) (76.3333,81.3333) (76.3333,92.3333) (84.3333,95.0000) (87.0000,95.6667) (88.6667,96.6667) (92.6667,96.3333) (94.0000,91.6667) (95.0000,94.3333)};
\addplot[only marks, mark=*, mark size=1.05pt,
  draw=nrvqbase, fill=nrvqbase, draw opacity=0.55,
  fill opacity=0.38] coordinates {(63.3333,93.0000) (63.3333,95.6667) (69.3333,92.6667) (72.3333,96.0000) (72.3333,96.3333) (75.6667,96.3333) (76.0000,91.3333) (76.6667,96.0000) (79.0000,96.3333) (80.0000,95.3333) (80.3333,95.6667) (80.6667,97.0000) (81.0000,90.6667) (81.3333,96.3333) (81.3333,96.6667) (81.3333,97.6667) (81.6667,92.3333) (84.0000,95.3333) (84.3333,94.3333) (85.0000,97.6667) (86.3333,93.0000) (86.3333,94.0000) (88.6667,96.3333) (90.3333,94.0000) (93.3333,95.0000) (93.6667,94.6667) (94.0000,93.6667)};
\addplot[only marks, mark=triangle*, mark size=1.25pt,
  draw=nrvqbase, fill=nrvqbase, draw opacity=0.55,
  fill opacity=0.38] coordinates {(24.0000,97.0000) (25.3333,94.6667) (26.6667,97.3333) (27.6667,97.3333) (27.6667,97.3333) (28.6667,95.6667) (28.6667,98.0000) (28.6667,98.3333) (30.0000,96.6667) (31.3333,95.0000) (31.6667,95.6667) (41.6667,97.0000) (42.0000,96.6667) (43.6667,95.0000) (44.3333,97.6667) (45.6667,95.0000) (46.3333,97.0000) (49.3333,96.0000) (50.0000,97.3333) (50.3333,94.3333) (50.3333,94.6667) (52.0000,93.3333) (52.3333,96.6667) (53.3333,94.3333) (71.3333,92.6667) (72.0000,93.3333) (75.0000,90.0000)};
\nextgroupplot[extra description/.code={
    \node[anchor=north east, fill=white, fill opacity=0.78, text opacity=1, inner sep=0.6pt, font=\fontsize{7}{7.5}\selectfont] at (axis description cs:0.975,0.972) {Lift};
},]
\addplot[black!45, densely dashed, line width=0.5pt] coordinates {(0,0) (100,100)};
\addplot[only marks, mark=square*, mark size=1.05pt,
  draw=fsqbase, fill=fsqbase, draw opacity=0.55,
  fill opacity=0.38] coordinates {(82.3333,100.0000) (87.3333,100.0000) (89.6667,100.0000) (91.0000,99.3333) (92.6667,100.0000) (93.3333,99.6667) (94.3333,99.6667) (94.6667,100.0000) (98.3333,99.6667)};
\addplot[only marks, mark=*, mark size=1.05pt,
  draw=nrvqbase, fill=nrvqbase, draw opacity=0.55,
  fill opacity=0.38] coordinates {(81.6667,99.6667) (82.0000,100.0000) (82.3333,100.0000) (83.0000,100.0000) (83.0000,100.0000) (83.3333,99.0000) (84.6667,100.0000) (85.6667,100.0000) (86.0000,100.0000) (87.0000,100.0000) (87.3333,100.0000) (88.0000,100.0000) (88.6667,100.0000) (90.0000,100.0000) (90.3333,100.0000) (90.6667,99.6667) (91.0000,100.0000) (91.0000,100.0000) (91.6667,100.0000) (93.0000,99.6667) (94.6667,99.6667) (94.6667,100.0000) (95.3333,100.0000) (96.0000,100.0000) (96.3333,99.6667) (96.3333,100.0000) (98.6667,100.0000)};
\addplot[only marks, mark=triangle*, mark size=1.25pt,
  draw=nrvqbase, fill=nrvqbase, draw opacity=0.55,
  fill opacity=0.38] coordinates {(53.6667,100.0000) (53.6667,100.0000) (55.6667,100.0000) (58.0000,100.0000) (59.6667,100.0000) (60.0000,100.0000) (60.0000,100.0000) (60.6667,100.0000) (61.3333,100.0000) (62.3333,100.0000) (62.6667,100.0000) (63.6667,100.0000) (64.0000,100.0000) (65.3333,100.0000) (65.6667,100.0000) (69.6667,100.0000) (72.3333,100.0000) (73.0000,100.0000) (73.3333,100.0000) (74.0000,100.0000) (74.0000,100.0000) (74.3333,100.0000) (74.6667,100.0000) (75.6667,100.0000) (76.0000,100.0000) (77.3333,100.0000) (84.0000,99.6667)};
\nextgroupplot[extra description/.code={
    \node[anchor=north east, fill=white, fill opacity=0.78, text opacity=1, inner sep=0.6pt, font=\fontsize{7}{7.5}\selectfont] at (axis description cs:0.975,0.972) {Square};
},]
\addplot[black!45, densely dashed, line width=0.5pt] coordinates {(0,0) (100,100)};
\addplot[only marks, mark=square*, mark size=1.05pt,
  draw=fsqbase, fill=fsqbase, draw opacity=0.55,
  fill opacity=0.38] coordinates {(25.6667,61.3333) (32.0000,58.0000) (34.0000,44.6667) (43.3333,64.3333) (48.6667,64.0000) (53.0000,72.3333) (54.6667,71.6667) (57.0000,61.6667) (59.0000,68.3333)};
\addplot[only marks, mark=*, mark size=1.05pt,
  draw=nrvqbase, fill=nrvqbase, draw opacity=0.55,
  fill opacity=0.38] coordinates {(31.6667,67.6667) (32.0000,70.0000) (33.6667,76.3333) (34.3333,67.6667) (35.6667,70.6667) (36.0000,71.3333) (38.3333,74.3333) (39.0000,78.6667) (39.0000,80.0000) (39.6667,61.6667) (39.6667,78.3333) (41.0000,73.6667) (42.6667,53.6667) (43.0000,79.3333) (43.3333,66.0000) (45.0000,74.3333) (46.3333,77.0000) (47.0000,75.6667) (47.0000,76.6667) (47.3333,70.0000) (47.3333,76.3333) (51.3333,58.3333) (51.6667,57.3333) (51.6667,72.3333) (52.6667,67.3333) (57.6667,65.3333) (64.3333,69.0000)};
\addplot[only marks, mark=triangle*, mark size=1.25pt,
  draw=nrvqbase, fill=nrvqbase, draw opacity=0.55,
  fill opacity=0.38] coordinates {(2.3333,81.3333) (3.0000,79.3333) (3.0000,82.6667) (3.0000,84.0000) (3.6667,73.6667) (4.0000,76.3333) (5.0000,73.3333) (5.0000,79.3333) (6.0000,78.3333) (6.3333,74.6667) (6.6667,75.3333) (7.0000,79.3333) (10.0000,73.0000) (11.6667,66.6667) (11.6667,68.0000) (12.0000,80.6667) (12.3333,66.6667) (13.6667,69.3333) (14.0000,77.0000) (16.3333,67.0000) (16.6667,67.6667) (18.0000,68.6667) (23.0000,71.6667) (26.3333,55.6667) (26.3333,76.0000) (29.6667,59.0000) (33.6667,64.6667)};
\end{groupplot}
\node[font=\fontsize{7}{7.5}\selectfont, inner ysep=0pt]
  at ($(group c1r1.south west)!0.5!(group c3r1.south east)+(0,-0.62cm)$)
  {Tokenizer-policy success (\%)};
\end{tikzpicture}

%% file: figures/what_matters/bitrate_chunking/appendix_bitrate_success_reconstruction.tex
\begin{tikzpicture}
\begin{axis}[
  width=7.8cm, height=2.55cm, scale only axis,
  xmin=18, xmax=83, ymin=0.0012, ymax=0.0102,
  xtick={20,30,40,50,60,70,80},
  ytick={0.002,0.006,0.010},
  scaled y ticks=false,
  yticklabel style={/pgf/number format/fixed,
    /pgf/number format/fixed zerofill,/pgf/number format/precision=3,
    font=\fontsize{6}{6.4}\selectfont},
  tick label style={font=\fontsize{6}{6.4}\selectfont},
  label style={font=\fontsize{6}{6.4}\selectfont},
  xlabel={Nominal bitrate},
  ylabel={Val. recon. MSE},
  grid=major, grid style={draw=black!10, line width=0.2pt},
  axis line style={black!55, line width=0.35pt},
  tick style={black!55, line width=0.35pt},
  legend to name=bitratefamilylegend,
  legend columns=3,
  legend style={draw=none,font=\fontsize{7}{7.5}\selectfont,column sep=5pt},
]
\addplot[fsqbase,solid,line width=1.0pt,mark=square*,mark size=1.5pt,
  mark options={solid,fill=fsqbase,draw=fsqbase}] coordinates {
  (21,0.00951631) (27,0.00624222) (30,0.00601843) (42,0.00533988)
  (54,0.00350333) (56,0.00425993) (60,0.00290751) (72,0.00288973)
  (80,0.00230170)};
\addlegendentry{\tokenizerid{fsqbase}{TF-FSQ-S}}
\addplot[nrvqbase,densely dotted,line width=1.0pt,mark=*,mark size=1.5pt,
  mark options={solid,fill=nrvqbase,draw=nrvqbase}] coordinates {
  (21,0.00908255) (27,0.00642015) (30,0.00510307) (42,0.00484356)
  (54,0.00286892) (56,0.00387664) (60,0.00264104) (72,0.00288889)
  (80,0.00259926)};
\addlegendentry{\tokenizerid{nrvqbase}{TF-NRVQ-S}}
\addplot[nrvqbase!60!black,densely dashed,line width=1.0pt,mark=triangle*,mark size=1.75pt,
  mark options={solid,fill=nrvqbase!60!black,draw=nrvqbase!60!black}] coordinates {
  (21,0.00659875) (27,0.00491120) (30,0.00400723) (42,0.00342933)
  (54,0.00261033) (56,0.00274582) (60,0.00224153) (72,0.00234137)
  (80,0.00196214)};
\addlegendentry{\tokenizerid{nrvqbase!60!black}{TF-NRVQ-L}}
\end{axis}
\end{tikzpicture}

\vspace{-0.2em}
\begin{center}
  \pgfplotslegendfromname{bitratefamilylegend}
\end{center}
\vspace{-0.6em}

\begin{tikzpicture}
\begin{groupplot}[
  group style={group size=3 by 1, horizontal sep=0.55cm,
    ylabels at=edge left, yticklabels at=edge left},
  width=3.85cm, height=2.65cm, scale only axis,
  title style={font=\fontsize{6.5}{7}\selectfont, text depth=0pt, yshift=-2pt},
  xmin=18, xmax=83, ymin=0, ymax=103,
  xtick={20,30,40,50,60,70,80}, ytick={0,50,100},
  grid=major, grid style={draw=black!10, line width=0.2pt},
  axis line style={black!55, line width=0.35pt},
  tick style={black!55, line width=0.35pt},
  tick label style={font=\fontsize{6}{6.4}\selectfont},
  label style={font=\fontsize{6}{6.4}\selectfont},
  ylabel={Success (\%)},
]
\nextgroupplot[title={Can}]
\addplot[fsqbase,solid,line width=1.0pt,mark=square*,mark size=1.5pt,
  mark options={solid,fill=fsqbase,draw=fsqbase}] coordinates {
  (21,76.3333) (27,71.6667) (30,76.3333) (42,84.3333) (54,88.6667)
  (56,95.0000) (60,92.6667) (72,94.0000) (80,87.0000)};
\addplot[nrvqbase,densely dotted,line width=1.0pt,mark=*,mark size=1.5pt,
  mark options={solid,fill=nrvqbase,draw=nrvqbase}] coordinates {
  (21,81.6667) (27,76.0000) (30,69.3333) (42,93.3333) (54,79.0000)
  (56,94.0000) (60,80.0000) (72,81.3333) (80,81.3333)};
\addplot[nrvqbase!60!black,densely dashed,line width=1.0pt,mark=triangle*,mark size=1.75pt,
  mark options={solid,fill=nrvqbase!60!black,draw=nrvqbase!60!black}] coordinates {
  (21,75.0000) (27,53.3333) (30,50.3333) (42,50.0000) (54,27.6667)
  (56,46.3333) (60,26.6667) (72,28.6667) (80,30.0000)};

\nextgroupplot[title={Lift}]
\addplot[fsqbase,solid,line width=1.0pt,mark=square*,mark size=1.5pt,
  mark options={solid,fill=fsqbase,draw=fsqbase}] coordinates {
  (21,89.6667) (27,91.0000) (30,82.3333) (42,87.3333) (54,93.3333)
  (56,94.3333) (60,92.6667) (72,98.3333) (80,94.6667)};
\addplot[nrvqbase,densely dotted,line width=1.0pt,mark=*,mark size=1.5pt,
  mark options={solid,fill=nrvqbase,draw=nrvqbase}] coordinates {
  (21,96.0000) (27,90.6667) (30,81.6667) (42,94.6667) (54,91.6667)
  (56,96.3333) (60,88.0000) (72,87.0000) (80,88.6667)};
\addplot[nrvqbase!60!black,densely dashed,line width=1.0pt,mark=triangle*,mark size=1.75pt,
  mark options={solid,fill=nrvqbase!60!black,draw=nrvqbase!60!black}] coordinates {
  (21,84.0000) (27,74.3333) (30,62.6667) (42,74.6667) (54,60.6667)
  (56,69.6667) (60,55.6667) (72,53.6667) (80,60.0000)};

\nextgroupplot[title={Square}]
\addplot[fsqbase,solid,line width=1.0pt,mark=square*,mark size=1.5pt,
  mark options={solid,fill=fsqbase,draw=fsqbase}] coordinates {
  (21,34.0000) (27,32.0000) (30,25.6667) (42,43.3333) (54,53.0000)
  (56,57.0000) (60,48.6667) (72,59.0000) (80,54.6667)};
\addplot[nrvqbase,densely dotted,line width=1.0pt,mark=*,mark size=1.5pt,
  mark options={solid,fill=nrvqbase,draw=nrvqbase}] coordinates {
  (21,51.3333) (27,31.6667) (30,34.3333) (42,64.3333) (54,39.6667)
  (56,51.6667) (60,35.6667) (72,43.0000) (80,46.3333)};
\addplot[nrvqbase!60!black,densely dashed,line width=1.0pt,mark=triangle*,mark size=1.75pt,
  mark options={solid,fill=nrvqbase!60!black,draw=nrvqbase!60!black}] coordinates {
  (21,29.6667) (27,16.6667) (30,13.6667) (42,18.0000) (54,6.3333)
  (56,10.0000) (60,2.3333) (72,3.6667) (80,3.0000)};
\end{groupplot}
\node[font=\fontsize{6}{6.4}\selectfont]
  at ($(group c1r1.south west)!0.5!(group c3r1.south east)+(0,-0.43cm)$)
  {Nominal bitrate};
\end{tikzpicture}

%% file: figures/what_matters/bitrate_chunking/appendix_robomimic_fm_exec_sweep.tex
\definecolor{fmh8}{RGB}{0,114,178}
\definecolor{fmh16}{RGB}{230,159,0}
\definecolor{fmh32}{RGB}{0,158,115}
\definecolor{fmh64}{RGB}{213,94,0}
\begin{tikzpicture}
\begin{groupplot}[
  group style={group size=3 by 1, horizontal sep=0.52cm,
    ylabels at=edge left, yticklabels at=edge left},
  width=3.55cm, height=2.35cm, scale only axis,
  xmode=log, log basis x=2, log ticks with fixed point,
  xmin=0.9, xmax=72, ymin=20, ymax=105,
  xtick={1,2,4,8,16,32,64}, xticklabels={1,2,4,8,16,32,64},
  minor xtick={}, ytick={20,40,60,80,100},
  grid=major, grid style={draw=black!10, line width=0.2pt},
  axis line style={black!55, line width=0.35pt},
  tick style={black!55, line width=0.35pt},
  tick label style={font=\fontsize{6}{6.4}\selectfont},
  label style={font=\fontsize{6}{6.4}\selectfont},
  ylabel={Success (\%)},
  ylabel style={at={(yticklabel cs:0.5,2pt)},anchor=near yticklabel,inner ysep=0pt,yshift=-\pgfkeysvalueof{/pgf/inner xsep}},
  error bars/y dir=both, error bars/y explicit,
  error bars/error mark=none,
  error bars/error bar style={line width=0.3pt},
]
\nextgroupplot[
  legend to name=robomimicfmlegend,
  legend columns=4,
  legend style={draw=none, font=\fontsize{6.5}{7}\selectfont,
    /tikz/every even column/.append style={column sep=0.14cm}},
  extra description/.code={
    \node[anchor=south east, fill=white, fill opacity=0.82,
      text opacity=1, inner sep=0.7pt,
      font=\fontsize{6}{6.4}\selectfont]
      at (axis description cs:0.985,0.035) {Can};},
]
\addplot[fmh8, line width=0.65pt, mark=*, mark size=1.15pt]
  coordinates {(1,86.67) +- (0,3.86) (2,93) +- (0,2.87)
  (4,96.33) +- (0,2.12) (8,97) +- (0,1.91)};
\addlegendentry{$H=8$}
\addplot[fmh16, line width=0.65pt, mark=square*, mark size=1.1pt]
  coordinates {(1,83) +- (0,4.26) (2,93.33) +- (0,2.8)
  (4,96) +- (0,2.21) (8,97.33) +- (0,1.8)
  (12,98.33) +- (0,1.44) (16,96.67) +- (0,2.01)};
\addlegendentry{$H=16$}
\addplot[fmh32, line width=0.65pt, mark=triangle*, mark size=1.25pt]
  coordinates {(1,79.67) +- (0,4.55) (2,88.33) +- (0,3.64)
  (4,91.33) +- (0,3.18) (8,95.67) +- (0,2.29)
  (12,95.67) +- (0,2.29) (16,93.33) +- (0,2.8)
  (20,90.67) +- (0,3.29) (24,90.33) +- (0,3.34)
  (32,81.33) +- (0,4.41)};
\addlegendentry{$H=32$}
\addplot[fmh64, line width=0.65pt, mark=diamond*, mark size=1.25pt]
  coordinates {(1,69) +- (0,5.24) (2,80.33) +- (0,4.5)
  (4,85) +- (0,4.06) (8,91) +- (0,3.24)
  (12,93.33) +- (0,2.8) (16,92) +- (0,3.08)
  (20,81.67) +- (0,4.39) (24,85.67) +- (0,3.98)
  (32,71.67) +- (0,5.13) (40,66.67) +- (0,5.34)
  (48,58) +- (0,5.54) (56,57.33) +- (0,5.55)
  (64,45.67) +- (0,5.62)};
\addlegendentry{$H=64$}

\nextgroupplot[extra description/.code={
  \node[anchor=south east, fill=white, fill opacity=0.82,
    text opacity=1, inner sep=0.7pt,
    font=\fontsize{6}{6.4}\selectfont]
    at (axis description cs:0.985,0.035) {Lift};},]
\addplot[fmh8, line width=0.65pt, mark=*, mark size=1.15pt]
  coordinates {(1,99.33) +- (0,0.92) (2,99.67) +- (0,0.65)
  (4,100) +- (0,0) (8,100) +- (0,0)};
\addplot[fmh16, line width=0.65pt, mark=square*, mark size=1.1pt]
  coordinates {(1,95.33) +- (0,2.37) (2,97.33) +- (0,1.8)
  (4,98) +- (0,1.57) (8,97.67) +- (0,1.69)
  (12,97.33) +- (0,1.8) (16,96.67) +- (0,2.01)};
\addplot[fmh32, line width=0.65pt, mark=triangle*, mark size=1.25pt]
  coordinates {(1,98.33) +- (0,1.44) (2,99) +- (0,1.12)
  (4,99.67) +- (0,0.65) (8,99.33) +- (0,0.92)
  (12,99.67) +- (0,0.65) (16,100) +- (0,0)
  (20,100) +- (0,0) (24,100) +- (0,0)
  (32,99.33) +- (0,0.92)};
\addplot[fmh64, line width=0.65pt, mark=diamond*, mark size=1.25pt]
  coordinates {(1,97) +- (0,1.91) (2,98) +- (0,1.57)
  (4,97) +- (0,1.91) (8,97.67) +- (0,1.69)
  (12,97.33) +- (0,1.8) (16,98) +- (0,1.57)
  (20,98.67) +- (0,1.29) (24,99.33) +- (0,0.92)
  (32,97.67) +- (0,1.69) (40,94.33) +- (0,2.59)
  (48,92.67) +- (0,2.94) (56,91.33) +- (0,3.18)
  (64,91.67) +- (0,3.13)};

\nextgroupplot[extra description/.code={
  \node[anchor=south east, fill=white, fill opacity=0.82,
    text opacity=1, inner sep=0.7pt,
    font=\fontsize{6}{6.4}\selectfont]
    at (axis description cs:0.985,0.035) {Square};},]
\addplot[fmh8, line width=0.65pt, mark=*, mark size=1.15pt]
  coordinates {(1,49.33) +- (0,5.62) (2,70.33) +- (0,5.16)
  (4,76.33) +- (0,4.81) (8,79) +- (0,4.61)};
\addplot[fmh16, line width=0.65pt, mark=square*, mark size=1.1pt]
  coordinates {(1,64.33) +- (0,5.42) (2,75) +- (0,4.89)
  (4,81.67) +- (0,4.39) (8,85.67) +- (0,3.98)
  (12,78.33) +- (0,4.64) (16,80) +- (0,4.53)};
\addplot[fmh32, line width=0.65pt, mark=triangle*, mark size=1.25pt]
  coordinates {(1,61) +- (0,5.47) (2,71.33) +- (0,5.14)
  (4,86) +- (0,3.95) (8,88.33) +- (0,3.64)
  (12,89) +- (0,3.55) (16,91) +- (0,3.24)
  (20,82.67) +- (0,4.29) (24,72) +- (0,5.1)
  (32,70.33) +- (0,5.16)};
\addplot[fmh64, line width=0.65pt, mark=diamond*, mark size=1.25pt]
  coordinates {(1,64) +- (0,5.42) (2,68) +- (0,5.28)
  (4,78) +- (0,4.68) (8,87.33) +- (0,3.77)
  (12,85.33) +- (0,4.03) (16,80.33) +- (0,4.5)
  (20,69) +- (0,5.24) (24,67) +- (0,5.32)
  (32,56.33) +- (0,5.56) (40,48.33) +- (0,5.62)
  (48,38) +- (0,5.46) (56,33.33) +- (0,5.31)
  (64,25) +- (0,4.88)};
\end{groupplot}
\node at ($(group c1r1.north west)!0.5!(group c3r1.north east)+(0,0.34cm)$)
  {\fontsize{6.5}{7}\selectfont Prediction horizon (a--b):\quad\raisebox{-4pt}{\pgfplotslegendfromname{robomimicfmlegend}}};

\pgfresetboundingbox
\path[use as bounding box]
  ($(group c1r1.south west)+(-0.78cm,-0.33cm)$) rectangle
  ($(group c3r1.north east)+(0.10cm,0.57cm)$);
\end{tikzpicture}

%% file: figures/what_matters/bitrate_chunking/appendix_libero_fm_exec_sweep.tex
\definecolor{fmh16}{RGB}{230,159,0}
\definecolor{fmh32}{RGB}{0,158,115}
\definecolor{fmh64}{RGB}{213,94,0}
\begin{tikzpicture}
\begin{groupplot}[
  group style={group size=4 by 1, horizontal sep=0.42cm,
    ylabels at=edge left, yticklabels at=edge left},
  width=2.6075cm, height=2.35cm, scale only axis,
  xmode=log, log basis x=2, log ticks with fixed point,
  xmin=1.8, xmax=72, ymin=-3, ymax=105,
  xtick={2,4,8,16,32,64}, xticklabels={2,4,8,16,32,64},
  minor xtick={}, ytick={0,20,40,60,80,100},
  grid=major, grid style={draw=black!10, line width=0.2pt},
  axis line style={black!55, line width=0.35pt},
  tick style={black!55, line width=0.35pt},
  tick label style={font=\fontsize{6}{6.4}\selectfont},
  label style={font=\fontsize{6}{6.4}\selectfont},
  ylabel={Success (\%)},
  ylabel style={at={(yticklabel cs:0.5,2pt)},anchor=near yticklabel,inner ysep=0pt,yshift=-\pgfkeysvalueof{/pgf/inner xsep}},
  error bars/y dir=both, error bars/y explicit,
  error bars/error mark=none,
  error bars/error bar style={line width=0.3pt},
]
\nextgroupplot[
  extra description/.code={
    \node[anchor=south east, fill=white, fill opacity=0.82,
      text opacity=1, inner sep=0.7pt,
      font=\fontsize{6}{6.4}\selectfont]
      at (axis description cs:0.985,0.035) {Goal};},
]
\addplot[fmh16, line width=0.65pt, mark=square*, mark size=1.1pt]
  coordinates {(2,71.2) +- (0,3.98) (4,81.2) +- (0,3.45)
  (8,84.2) +- (0,3.22) (12,85.2) +- (0,3.13)
  (16,77.8) +- (0,3.66)};
\addplot[fmh32, line width=0.65pt, mark=triangle*, mark size=1.25pt]
  coordinates {(2,76) +- (0,3.74) (4,83.2) +- (0,3.3)
  (8,90.4) +- (0,2.6) (12,88) +- (0,2.86)
  (16,85.8) +- (0,3.08) (24,72.4) +- (0,3.92)
  (32,57.4) +- (0,4.34)};
\addplot[fmh64, line width=0.65pt, mark=diamond*, mark size=1.25pt]
  coordinates {(2,79.2) +- (0,3.58) (4,84) +- (0,3.24)
  (8,88.2) +- (0,2.84) (12,86.8) +- (0,2.98)
  (16,81.8) +- (0,3.41) (24,68.2) +- (0,4.12)
  (32,53.8) +- (0,4.36) (40,53.2) +- (0,4.37)
  (48,53.4) +- (0,4.37) (56,47.8) +- (0,4.41)
  (64,46.8) +- (0,4.41)};

\nextgroupplot[extra description/.code={
  \node[anchor=south east, fill=white, fill opacity=0.82,
    text opacity=1, inner sep=0.7pt,
    font=\fontsize{6}{6.4}\selectfont]
    at (axis description cs:0.985,0.035) {Object};},]
\addplot[fmh16, line width=0.65pt, mark=square*, mark size=1.1pt]
  coordinates {(2,56.8) +- (0,4.34) (4,74) +- (0,3.85)
  (8,80.8) +- (0,3.48) (12,79.4) +- (0,3.56)
  (16,76) +- (0,3.74)};
\addplot[fmh32, line width=0.65pt, mark=triangle*, mark size=1.25pt]
  coordinates {(2,54.6) +- (0,4.35) (4,70) +- (0,4.04)
  (8,80.8) +- (0,3.48) (12,84.4) +- (0,3.21)
  (16,85.6) +- (0,3.09) (24,70.6) +- (0,4.01)
  (32,36.6) +- (0,4.23)};
\addplot[fmh64, line width=0.65pt, mark=diamond*, mark size=1.25pt]
  coordinates {(2,58.8) +- (0,4.32) (4,72.4) +- (0,3.92)
  (8,81.8) +- (0,3.41) (12,87.2) +- (0,2.94)
  (16,88) +- (0,2.86) (24,67) +- (0,4.16)
  (32,32.4) +- (0,4.08) (40,5.4) +- (0,1.98)
  (48,1.6) +- (0,1.1) (56,4.2) +- (0,1.78)
  (64,5) +- (0,1.92)};

\nextgroupplot[extra description/.code={
  \node[anchor=south east, fill=white, fill opacity=0.82,
    text opacity=1, inner sep=0.7pt,
    font=\fontsize{6}{6.4}\selectfont]
    at (axis description cs:0.985,0.035) {Spatial};},]
\addplot[fmh16, line width=0.65pt, mark=square*, mark size=1.1pt]
  coordinates {(2,84.8) +- (0,3.17) (4,86) +- (0,3.06)
  (8,83) +- (0,3.31) (12,75) +- (0,3.8)
  (16,72) +- (0,3.95)};
\addplot[fmh32, line width=0.65pt, mark=triangle*, mark size=1.25pt]
  coordinates {(2,79.8) +- (0,3.54) (4,81.4) +- (0,3.44)
  (8,77.6) +- (0,3.67) (12,73.6) +- (0,3.87)
  (16,63.2) +- (0,4.26) (24,35.4) +- (0,4.2)
  (32,24.8) +- (0,3.78)};
\addplot[fmh64, line width=0.65pt, mark=diamond*, mark size=1.25pt]
  coordinates {(2,86.6) +- (0,3) (4,80) +- (0,3.53)
  (8,70.4) +- (0,4.02) (12,62.8) +- (0,4.26)
  (16,55.2) +- (0,4.35) (24,35.4) +- (0,4.2)
  (32,28) +- (0,3.92) (40,24) +- (0,3.74)
  (48,21) +- (0,3.56) (56,22.8) +- (0,3.67)
  (64,24) +- (0,3.74)};

\nextgroupplot[extra description/.code={
  \node[anchor=south east, fill=white, fill opacity=0.82,
    text opacity=1, inner sep=0.7pt,
    font=\fontsize{6}{6.4}\selectfont]
    at (axis description cs:0.985,0.035) {LIBERO-10};},]
\addplot[fmh16, line width=0.65pt, mark=square*, mark size=1.1pt]
  coordinates {(2,47.8) +- (0,4.41) (4,57.4) +- (0,4.34)
  (8,68.4) +- (0,4.11) (12,68.2) +- (0,4.12)
  (16,62.8) +- (0,4.26)};
\addplot[fmh32, line width=0.65pt, mark=triangle*, mark size=1.25pt]
  coordinates {(2,50.5) +- (0,4.9) (4,58.2) +- (0,4.33)
  (8,72.2) +- (0,3.94) (12,72.8) +- (0,3.91)
  (16,68.8) +- (0,4.09) (24,50.4) +- (0,4.4)
  (32,42) +- (0,4.34)};
\addplot[fmh64, line width=0.65pt, mark=diamond*, mark size=1.25pt]
  coordinates {(2,50) +- (0,5.62) (4,53) +- (0,4.38)
  (8,59.2) +- (0,4.31) (12,56.4) +- (0,4.34)
  (16,59.8) +- (0,4.29) (24,56.8) +- (0,4.34)
  (32,46.4) +- (0,4.41) (40,41.6) +- (0,4.34)
  (48,34.4) +- (0,4.16) (56,33.4) +- (0,4.12)
  (64,31.2) +- (0,4.06)};
\addplot[only marks, fmh32, mark=triangle, mark size=1.6pt,
  mark options={solid, fill=white, line width=0.45pt}]
  coordinates {(2,50.5)};
\addplot[only marks, fmh64, mark=diamond, mark size=1.6pt,
  mark options={solid, fill=white, line width=0.45pt}]
  coordinates {(2,50)};
\end{groupplot}

\pgfresetboundingbox
\path[use as bounding box]
  ($(group c1r1.south west)+(-0.78cm,-0.33cm)$) rectangle
  ($(group c4r1.north east)+(0.10cm,0.05cm)$);
\end{tikzpicture}

%% file: figures/what_matters/bitrate_chunking/appendix_robotwin_fm_exec_sweep.tex
\begin{tikzpicture}
\begin{groupplot}[
  group style={group size=3 by 1, horizontal sep=0.52cm,
    ylabels at=edge left, yticklabels at=edge left},
  width=3.55cm, height=2.35cm, scale only axis,
  xmode=log, log basis x=2, log ticks with fixed point,
  xmin=1.8, xmax=72, ymin=-3, ymax=105,
  xtick={2,4,8,16,32,64}, xticklabels={2,4,8,16,32,64},
  minor xtick={}, ytick={0,50,100},
  grid=major, grid style={draw=black!10, line width=0.2pt},
  axis line style={black!55, line width=0.35pt},
  tick style={black!55, line width=0.35pt},
  tick label style={font=\fontsize{6}{6.4}\selectfont},
  label style={font=\fontsize{6}{6.4}\selectfont},
  ylabel={Success (\%)},
  ylabel style={at={(yticklabel cs:0.5,2pt)},anchor=near yticklabel,inner ysep=0pt,yshift=-\pgfkeysvalueof{/pgf/inner xsep}},
]
\nextgroupplot[extra description/.code={
  \node[anchor=south east, fill=white, fill opacity=0.82,
    text opacity=1, inner sep=0.7pt,
    font=\fontsize{6}{6.4}\selectfont]
    at (axis description cs:0.985,0.035) {Grab roller};},]
\addplot[black!78, solid, line width=0.7pt, mark=*,
  mark size=1.35pt, mark options={solid, draw=black!78, fill=black!78},
  error bars/.cd, y dir=both, y explicit, error mark=none]
  coordinates {(2,64) +- (0,13.19) (4,86) +- (0,9.66)
  (8,93) +- (0,4.92) (12,96) +- (0,5.43)
  (16,99) +- (0,1.96) (24,100) +- (0,0)
  (32,99) +- (0,1.96) (40,98) +- (0,3.90)
  (48,98) +- (0,3.90) (56,100) +- (0,0)
  (64,100) +- (0,0)};
\nextgroupplot[extra description/.code={
  \node[anchor=south east, fill=white, fill opacity=0.82,
    text opacity=1, inner sep=0.7pt,
    font=\fontsize{6}{6.4}\selectfont]
    at (axis description cs:0.985,0.035) {Place burger/fries};},]
\addplot[black!78, solid, line width=0.7pt, mark=square*,
  mark size=1.35pt, mark options={solid, draw=black!78, fill=black!78},
  error bars/.cd, y dir=both, y explicit, error mark=none]
  coordinates {(2,44) +- (0,13.71) (4,80) +- (0,11.10)
  (8,92) +- (0,5.28) (12,96) +- (0,5.43)
  (16,96) +- (0,3.80) (24,98) +- (0,3.90)
  (32,96) +- (0,3.80) (40,94) +- (0,6.49)
  (48,100) +- (0,0) (56,100) +- (0,0)
  (64,99) +- (0,1.96)};
\nextgroupplot[extra description/.code={
  \node[anchor=south east, fill=white, fill opacity=0.82,
    text opacity=1, inner sep=0.7pt,
    font=\fontsize{6}{6.4}\selectfont]
    at (axis description cs:0.985,0.035) {Place cans};},]
\addplot[black!78, solid, line width=0.7pt, mark=triangle*,
  mark size=1.35pt, mark options={solid, draw=black!78, fill=black!78},
  error bars/.cd, y dir=both, y explicit, error mark=none]
  coordinates {(2,2) +- (0,3.85) (4,80) +- (0,11.10)
  (8,82) +- (0,7.48) (12,98) +- (0,3.90)
  (16,90) +- (0,5.86) (24,82) +- (0,10.67)
  (32,86) +- (0,6.80) (40,88) +- (0,9.05)
  (48,90) +- (0,8.33) (56,88) +- (0,9.05)
  (64,80) +- (0,7.81)};
\end{groupplot}
\node[font=\fontsize{6}{6.4}\selectfont]
  at ($(group c1r1.south west)!0.5!(group c3r1.south east)+(0,-0.43cm)$)
  {Execution horizon $H_{\mathrm{exec}}$};
\pgfresetboundingbox
\path[use as bounding box]
  ($(group c1r1.south west)+(-0.78cm,-0.62cm)$) rectangle
  ($(group c3r1.north east)+(0.10cm,0.05cm)$);
\end{tikzpicture}

%% file: figures/what_matters/bitrate_chunking/appendix_reconstruction_by_task.tex
\begin{tikzpicture}
\begin{groupplot}[
  group style={group size=3 by 1,horizontal sep=0.85cm},
  width=0.245\linewidth,height=3.35cm,scale only axis,
  xmin=0.8,xmax=4.2,xtick={1,2,3,4},xticklabels={4,8,16,32},
  ymin=0.010,ymax=0.070,ytick={0.02,0.04,0.06},
  xlabel={$H$},label style={font=\scriptsize},tick label style={font=\scriptsize},
  scaled y ticks=false,yticklabel style={/pgf/number format/fixed,/pgf/number format/precision=2},
  tick align=outside,tick pos=left,axis line style={black!55,line width=0.35pt},
  xmajorgrids,ymajorgrids,grid style={black!10,line width=0.3pt},
  clip=false,
]
\nextgroupplot[title={Can},title style={font=\scriptsize},ylabel={Val. recon. error},legend to name=chunkappendixfamilylegend,legend columns=5,legend style={draw=none,font=\scriptsize,/tikz/every even column/.append style={column sep=0.25em}}]
\addplot[name path=canVQlow,draw=none,forget plot] coordinates {(1,0.01504725) (2,0.02354195) (3,0.03293284) (4,0.04432960)};
\addplot[name path=canVQhigh,draw=none,forget plot] coordinates {(1,0.01631313) (2,0.02640713) (3,0.03450727) (4,0.04853114)};
\addplot[fill=vqbase,fill opacity=0.13,draw=none,forget plot] fill between[of=canVQlow and canVQhigh];
\addplot[name path=canRVQlow,draw=none,forget plot] coordinates {(1,0.01546017) (2,0.02428708) (3,0.03433384) (4,0.04557735)};
\addplot[name path=canRVQhigh,draw=none,forget plot] coordinates {(1,0.02272052) (2,0.03102476) (3,0.04029677) (4,0.05141996)};
\addplot[fill=rvqbase,fill opacity=0.13,draw=none,forget plot] fill between[of=canRVQlow and canRVQhigh];
\addplot[name path=canNRVQlow,draw=none,forget plot] coordinates {(1,0.01697364) (2,0.02347576) (3,0.03148814) (4,0.04358678)};
\addplot[name path=canNRVQhigh,draw=none,forget plot] coordinates {(1,0.02510576) (2,0.03698568) (3,0.04640709) (4,0.06515185)};
\addplot[fill=nrvqbase,fill opacity=0.13,draw=none,forget plot] fill between[of=canNRVQlow and canNRVQhigh];
\addplot[name path=canFSQlow,draw=none,forget plot] coordinates {(1,0.01971088) (2,0.02784346) (3,0.03739149) (4,0.04737675)};
\addplot[name path=canFSQhigh,draw=none,forget plot] coordinates {(1,0.02066899) (2,0.02883589) (3,0.03770537) (4,0.04813852)};
\addplot[fill=fsqbase,fill opacity=0.13,draw=none,forget plot] fill between[of=canFSQlow and canFSQhigh];
\addplot[name path=canQinColow,draw=none,forget plot] coordinates {(1,0.01421225) (2,0.02306333) (3,0.03255710) (4,0.04239710)};
\addplot[name path=canQinCohigh,draw=none,forget plot] coordinates {(1,0.01929189) (2,0.02854706) (3,0.03582842) (4,0.04682034)};
\addplot[fill=qincobase,fill opacity=0.13,draw=none,forget plot] fill between[of=canQinColow and canQinCohigh];
\addplot[vqbase,solid,line width=0.85pt,mark=*,mark size=1.3pt,mark options={solid,fill=vqbase,draw=vqbase,line width=0.25pt}] coordinates {(1,0.01568019) (2,0.02497454) (3,0.03372006) (4,0.04643037)};
\addlegendentry{VQ}
\addplot[rvqbase,densely dashed,line width=0.85pt,mark=square*,mark size=1.3pt,mark options={solid,fill=rvqbase,draw=rvqbase,line width=0.25pt}] coordinates {(1,0.01874115) (2,0.02731475) (3,0.03717286) (4,0.04864726)};
\addlegendentry{RVQ}
\addplot[nrvqbase,densely dotted,line width=0.85pt,mark=triangle*,mark size=1.3pt,mark options={solid,fill=nrvqbase,draw=nrvqbase,line width=0.25pt}] coordinates {(1,0.02091797) (2,0.03091209) (3,0.03997839) (4,0.05401707)};
\addlegendentry{NRVQ}
\addplot[fsqbase,dashdotted,line width=0.85pt,mark=diamond*,mark size=1.3pt,mark options={solid,fill=fsqbase,draw=fsqbase,line width=0.25pt}] coordinates {(1,0.02018994) (2,0.02833967) (3,0.03754843) (4,0.04775763)};
\addlegendentry{FSQ}
\addplot[qincobase,loosely dashdotdotted,line width=0.85pt,mark=pentagon*,mark size=1.3pt,mark options={solid,fill=qincobase,draw=qincobase,line width=0.25pt}] coordinates {(1,0.01595128) (2,0.02477409) (3,0.03401628) (4,0.04451772)};
\addlegendentry{QinCo}
\nextgroupplot[title={Lift},title style={font=\scriptsize},yticklabels={}]
\addplot[name path=liftVQlow,draw=none,forget plot] coordinates {(1,0.01294632) (2,0.02070285) (3,0.02830588) (4,0.03470653)};
\addplot[name path=liftVQhigh,draw=none,forget plot] coordinates {(1,0.01687876) (2,0.02266154) (3,0.03152362) (4,0.03814945)};
\addplot[fill=vqbase,fill opacity=0.13,draw=none,forget plot] fill between[of=liftVQlow and liftVQhigh];
\addplot[name path=liftRVQlow,draw=none,forget plot] coordinates {(1,0.01351539) (2,0.02156265) (3,0.02980145) (4,0.03683278)};
\addplot[name path=liftRVQhigh,draw=none,forget plot] coordinates {(1,0.02049104) (2,0.02633056) (3,0.03446158) (4,0.04225016)};
\addplot[fill=rvqbase,fill opacity=0.13,draw=none,forget plot] fill between[of=liftRVQlow and liftRVQhigh];
\addplot[name path=liftNRVQlow,draw=none,forget plot] coordinates {(1,0.01231371) (2,0.02015254) (3,0.02733079) (4,0.03568859)};
\addplot[name path=liftNRVQhigh,draw=none,forget plot] coordinates {(1,0.02175615) (2,0.03919578) (3,0.04417420) (4,0.04954093)};
\addplot[fill=nrvqbase,fill opacity=0.13,draw=none,forget plot] fill between[of=liftNRVQlow and liftNRVQhigh];
\addplot[name path=liftFSQlow,draw=none,forget plot] coordinates {(1,0.01628317) (2,0.02367191) (3,0.03133726) (4,0.03748524)};
\addplot[name path=liftFSQhigh,draw=none,forget plot] coordinates {(1,0.01651182) (2,0.02428537) (3,0.03201223) (4,0.03947091)};
\addplot[fill=fsqbase,fill opacity=0.13,draw=none,forget plot] fill between[of=liftFSQlow and liftFSQhigh];
\addplot[name path=liftQinColow,draw=none,forget plot] coordinates {(1,0.01237095) (2,0.02030039) (3,0.02727261) (4,0.03373348)};
\addplot[name path=liftQinCohigh,draw=none,forget plot] coordinates {(1,0.01589643) (2,0.02283972) (3,0.03105195) (4,0.03923297)};
\addplot[fill=qincobase,fill opacity=0.13,draw=none,forget plot] fill between[of=liftQinColow and liftQinCohigh];
\addplot[vqbase,solid,line width=0.85pt,mark=*,mark size=1.3pt,mark options={solid,fill=vqbase,draw=vqbase,line width=0.25pt}] coordinates {(1,0.01491254) (2,0.02168220) (3,0.02991475) (4,0.03642799)};
\addplot[rvqbase,densely dashed,line width=0.85pt,mark=square*,mark size=1.3pt,mark options={solid,fill=rvqbase,draw=rvqbase,line width=0.25pt}] coordinates {(1,0.01664564) (2,0.02392379) (3,0.03206842) (4,0.03946372)};
\addplot[nrvqbase,densely dotted,line width=0.85pt,mark=triangle*,mark size=1.3pt,mark options={solid,fill=nrvqbase,draw=nrvqbase,line width=0.25pt}] coordinates {(1,0.01821678) (2,0.02925414) (3,0.03566054) (4,0.04335308)};
\addplot[fsqbase,dashdotted,line width=0.85pt,mark=diamond*,mark size=1.3pt,mark options={solid,fill=fsqbase,draw=fsqbase,line width=0.25pt}] coordinates {(1,0.01639749) (2,0.02397864) (3,0.03167474) (4,0.03847807)};
\addplot[qincobase,loosely dashdotdotted,line width=0.85pt,mark=pentagon*,mark size=1.3pt,mark options={solid,fill=qincobase,draw=qincobase,line width=0.25pt}] coordinates {(1,0.01387204) (2,0.02141929) (3,0.02921749) (4,0.03649621)};
\nextgroupplot[title={Square},title style={font=\scriptsize},yticklabels={}]
\addplot[name path=squareVQlow,draw=none,forget plot] coordinates {(1,0.01558606) (2,0.02234215) (3,0.03426656) (4,0.04539316)};
\addplot[name path=squareVQhigh,draw=none,forget plot] coordinates {(1,0.01687441) (2,0.02489709) (3,0.03677504) (4,0.04987451)};
\addplot[fill=vqbase,fill opacity=0.13,draw=none,forget plot] fill between[of=squareVQlow and squareVQhigh];
\addplot[name path=squareRVQlow,draw=none,forget plot] coordinates {(1,0.01556109) (2,0.02493796) (3,0.03353977) (4,0.04724480)};
\addplot[name path=squareRVQhigh,draw=none,forget plot] coordinates {(1,0.02684691) (2,0.03067433) (3,0.03977302) (4,0.05361740)};
\addplot[fill=rvqbase,fill opacity=0.13,draw=none,forget plot] fill between[of=squareRVQlow and squareRVQhigh];
\addplot[name path=squareNRVQlow,draw=none,forget plot] coordinates {(1,0.01548937) (2,0.02314518) (3,0.03260889) (4,0.04449829)};
\addplot[name path=squareNRVQhigh,draw=none,forget plot] coordinates {(1,0.02569039) (2,0.03324804) (3,0.04710395) (4,0.05771313)};
\addplot[fill=nrvqbase,fill opacity=0.13,draw=none,forget plot] fill between[of=squareNRVQlow and squareNRVQhigh];
\addplot[name path=squareFSQlow,draw=none,forget plot] coordinates {(1,0.01871781) (2,0.02844137) (3,0.03514285) (4,0.04724059)};
\addplot[name path=squareFSQhigh,draw=none,forget plot] coordinates {(1,0.02038624) (2,0.02901016) (3,0.03539858) (4,0.04731877)};
\addplot[fill=fsqbase,fill opacity=0.13,draw=none,forget plot] fill between[of=squareFSQlow and squareFSQhigh];
\addplot[name path=squareQinColow,draw=none,forget plot] coordinates {(1,0.01399130) (2,0.02302245) (3,0.03123848) (4,0.04250297)};
\addplot[name path=squareQinCohigh,draw=none,forget plot] coordinates {(1,0.01989167) (2,0.02667937) (3,0.03644334) (4,0.04879211)};
\addplot[fill=qincobase,fill opacity=0.13,draw=none,forget plot] fill between[of=squareQinColow and squareQinCohigh];
\addplot[vqbase,solid,line width=0.85pt,mark=*,mark size=1.3pt,mark options={solid,fill=vqbase,draw=vqbase,line width=0.25pt}] coordinates {(1,0.01623023) (2,0.02361962) (3,0.03552080) (4,0.04763383)};
\addplot[rvqbase,densely dashed,line width=0.85pt,mark=square*,mark size=1.3pt,mark options={solid,fill=rvqbase,draw=rvqbase,line width=0.25pt}] coordinates {(1,0.01980292) (2,0.02734205) (3,0.03709562) (4,0.05079535)};
\addplot[nrvqbase,densely dotted,line width=0.85pt,mark=triangle*,mark size=1.3pt,mark options={solid,fill=nrvqbase,draw=nrvqbase,line width=0.25pt}] coordinates {(1,0.02134020) (2,0.02905074) (3,0.03972038) (4,0.05310308)};
\addplot[fsqbase,dashdotted,line width=0.85pt,mark=diamond*,mark size=1.3pt,mark options={solid,fill=fsqbase,draw=fsqbase,line width=0.25pt}] coordinates {(1,0.01955203) (2,0.02872577) (3,0.03527071) (4,0.04727968)};
\addplot[qincobase,loosely dashdotdotted,line width=0.85pt,mark=pentagon*,mark size=1.3pt,mark options={solid,fill=qincobase,draw=qincobase,line width=0.25pt}] coordinates {(1,0.01610992) (2,0.02403268) (3,0.03398891) (4,0.04560033)};
\end{groupplot}
\end{tikzpicture}

%% file: figures/what_matters/bitrate_chunking/appendix_success_colorbar.tex
\definecolor{chunkscalec0}{RGB}{68,1,84}
\definecolor{chunkscalec1}{RGB}{70,18,100}
\definecolor{chunkscalec2}{RGB}{72,36,117}
\definecolor{chunkscalec3}{RGB}{68,52,126}
\definecolor{chunkscalec4}{RGB}{65,68,135}
\definecolor{chunkscalec5}{RGB}{59,82,138}
\definecolor{chunkscalec6}{RGB}{53,95,141}
\definecolor{chunkscalec7}{RGB}{48,108,142}
\definecolor{chunkscalec8}{RGB}{42,120,142}
\definecolor{chunkscalec9}{RGB}{38,132,141}
\definecolor{chunkscalec10}{RGB}{33,145,140}
\definecolor{chunkscalec11}{RGB}{34,156,136}
\definecolor{chunkscalec12}{RGB}{34,168,132}
\definecolor{chunkscalec13}{RGB}{51,180,122}
\definecolor{chunkscalec14}{RGB}{68,191,112}
\definecolor{chunkscalec15}{RGB}{95,200,96}
\definecolor{chunkscalec16}{RGB}{122,209,81}
\definecolor{chunkscalec17}{RGB}{156,216,60}
\definecolor{chunkscalec18}{RGB}{189,223,38}
\definecolor{chunkscalec19}{RGB}{221,227,38}
\definecolor{chunkscalec20}{RGB}{253,231,37}
\begin{tikzpicture}[x=1cm,y=1cm]
\fill[chunkscalec0] (0.00000,0) rectangle (0.19048,0.10);
\fill[chunkscalec1] (0.19048,0) rectangle (0.38095,0.10);
\fill[chunkscalec2] (0.38095,0) rectangle (0.57143,0.10);
\fill[chunkscalec3] (0.57143,0) rectangle (0.76190,0.10);
\fill[chunkscalec4] (0.76190,0) rectangle (0.95238,0.10);
\fill[chunkscalec5] (0.95238,0) rectangle (1.14286,0.10);
\fill[chunkscalec6] (1.14286,0) rectangle (1.33333,0.10);
\fill[chunkscalec7] (1.33333,0) rectangle (1.52381,0.10);
\fill[chunkscalec8] (1.52381,0) rectangle (1.71429,0.10);
\fill[chunkscalec9] (1.71429,0) rectangle (1.90476,0.10);
\fill[chunkscalec10] (1.90476,0) rectangle (2.09524,0.10);
\fill[chunkscalec11] (2.09524,0) rectangle (2.28571,0.10);
\fill[chunkscalec12] (2.28571,0) rectangle (2.47619,0.10);
\fill[chunkscalec13] (2.47619,0) rectangle (2.66667,0.10);
\fill[chunkscalec14] (2.66667,0) rectangle (2.85714,0.10);
\fill[chunkscalec15] (2.85714,0) rectangle (3.04762,0.10);
\fill[chunkscalec16] (3.04762,0) rectangle (3.23810,0.10);
\fill[chunkscalec17] (3.23810,0) rectangle (3.42857,0.10);
\fill[chunkscalec18] (3.42857,0) rectangle (3.61905,0.10);
\fill[chunkscalec19] (3.61905,0) rectangle (3.80952,0.10);
\fill[chunkscalec20] (3.80952,0) rectangle (4.00000,0.10);
\node[anchor=south,font=\scriptsize,inner sep=1pt] at (2,0.13) {Success (\%)};
\node[anchor=north,font=\scriptsize,inner sep=1pt] at (0,0) {0};
\node[anchor=north,font=\scriptsize,inner sep=1pt] at (2,0) {50};
\node[anchor=north,font=\scriptsize,inner sep=1pt] at (4,0) {100};
\end{tikzpicture}

%% file: figures/what_matters/bitrate_chunking/appendix_success_by_task.tex
\definecolor{chunktaskc0}{RGB}{68,1,84}
\definecolor{chunktaskc1}{RGB}{70,18,100}
\definecolor{chunktaskc2}{RGB}{72,36,117}
\definecolor{chunktaskc3}{RGB}{68,52,126}
\definecolor{chunktaskc4}{RGB}{65,68,135}
\definecolor{chunktaskc5}{RGB}{59,82,138}
\definecolor{chunktaskc6}{RGB}{53,95,141}
\definecolor{chunktaskc7}{RGB}{48,108,142}
\definecolor{chunktaskc8}{RGB}{42,120,142}
\definecolor{chunktaskc9}{RGB}{38,132,141}
\definecolor{chunktaskc10}{RGB}{33,145,140}
\definecolor{chunktaskc11}{RGB}{34,156,136}
\definecolor{chunktaskc12}{RGB}{34,168,132}
\definecolor{chunktaskc13}{RGB}{51,180,122}
\definecolor{chunktaskc14}{RGB}{68,191,112}
\definecolor{chunktaskc15}{RGB}{95,200,96}
\definecolor{chunktaskc16}{RGB}{122,209,81}
\definecolor{chunktaskc17}{RGB}{156,216,60}
\definecolor{chunktaskc18}{RGB}{189,223,38}
\definecolor{chunktaskc19}{RGB}{221,227,38}
\definecolor{chunktaskc20}{RGB}{253,231,37}
\begin{axis}[
  width=\chunkPlotWidth,height=\chunkPlotHeight,scale only axis,
  name=\chunkPlotName,at={(\chunkPlotX,0)},anchor=south west,
  xmin=0.5,xmax=10.5,ymin=0.5,ymax=3.5,
  xtick={1,...,10},xticklabels={$4/4$,$8/4$,$16/4$,$32/4$,$8/8$,$16/8$,$32/8$,$16/16$,$32/16$,$32/32$},
  x tick label style={rotate=48,anchor=east,font=\fontsize{6}{6.5}\selectfont},
  ytick={1,2,3},yticklabels={Square,Lift,Can},
  yticklabel style={font=\fontsize{6}{6.5}\selectfont},
  xlabel={$H/H_{\mathrm{exec}}$},label style={font=\scriptsize},
  tick align=outside,tick pos=left,axis line style={black!55,line width=0.35pt},
  clip=false,
]
\path[fill=chunktaskc15,draw=white,line width=0.4pt] (axis cs:0.53,2.55) rectangle (axis cs:1.47,3.45);
\node[font=\fontsize{6}{6.5}\selectfont\bfseries,text=black] at (axis cs:1,3) {\contour{white}{73.1}};
\path[fill=chunktaskc15,draw=white,line width=0.4pt] (axis cs:1.53,2.55) rectangle (axis cs:2.47,3.45);
\node[font=\fontsize{6}{6.5}\selectfont\bfseries,text=black] at (axis cs:2,3) {\contour{white}{76.8}};
\path[fill=chunktaskc14,draw=white,line width=0.4pt] (axis cs:2.53,2.55) rectangle (axis cs:3.47,3.45);
\node[font=\fontsize{6}{6.5}\selectfont\bfseries,text=black] at (axis cs:3,3) {\contour{white}{69.7}};
\path[fill=chunktaskc11,draw=white,line width=0.4pt] (axis cs:3.53,2.55) rectangle (axis cs:4.47,3.45);
\node[font=\fontsize{6}{6.5}\selectfont\bfseries,text=black] at (axis cs:4,3) {\contour{white}{54.5}};
\path[fill=chunktaskc14,draw=white,line width=0.4pt] (axis cs:4.53,2.55) rectangle (axis cs:5.47,3.45);
\node[font=\fontsize{6}{6.5}\selectfont\bfseries,text=black] at (axis cs:5,3) {\contour{white}{68.6}};
\path[fill=chunktaskc13,draw=white,line width=0.4pt] (axis cs:5.53,2.55) rectangle (axis cs:6.47,3.45);
\node[font=\fontsize{6}{6.5}\selectfont\bfseries,text=black] at (axis cs:6,3) {\contour{white}{65.7}};
\path[fill=chunktaskc10,draw=white,line width=0.4pt] (axis cs:6.53,2.55) rectangle (axis cs:7.47,3.45);
\node[font=\fontsize{6}{6.5}\selectfont\bfseries,text=black] at (axis cs:7,3) {\contour{white}{50.9}};
\path[fill=chunktaskc11,draw=white,line width=0.4pt] (axis cs:7.53,2.55) rectangle (axis cs:8.47,3.45);
\node[font=\fontsize{6}{6.5}\selectfont\bfseries,text=black] at (axis cs:8,3) {\contour{white}{53.7}};
\path[fill=chunktaskc9,draw=white,line width=0.4pt] (axis cs:8.53,2.55) rectangle (axis cs:9.47,3.45);
\node[font=\fontsize{6}{6.5}\selectfont\bfseries,text=white] at (axis cs:9,3) {44.3};
\path[fill=chunktaskc6,draw=white,line width=0.4pt] (axis cs:9.53,2.55) rectangle (axis cs:10.47,3.45);
\node[font=\fontsize{6}{6.5}\selectfont\bfseries,text=white] at (axis cs:10,3) {28.1};
\path[fill=chunktaskc17,draw=white,line width=0.4pt] (axis cs:0.53,1.55) rectangle (axis cs:1.47,2.45);
\node[font=\fontsize{6}{6.5}\selectfont\bfseries,text=black] at (axis cs:1,2) {\contour{white}{86.1}};
\path[fill=chunktaskc17,draw=white,line width=0.4pt] (axis cs:1.53,1.55) rectangle (axis cs:2.47,2.45);
\node[font=\fontsize{6}{6.5}\selectfont\bfseries,text=black] at (axis cs:2,2) {\contour{white}{87.2}};
\path[fill=chunktaskc17,draw=white,line width=0.4pt] (axis cs:2.53,1.55) rectangle (axis cs:3.47,2.45);
\node[font=\fontsize{6}{6.5}\selectfont\bfseries,text=black] at (axis cs:3,2) {\contour{white}{86.9}};
\path[fill=chunktaskc14,draw=white,line width=0.4pt] (axis cs:3.53,1.55) rectangle (axis cs:4.47,2.45);
\node[font=\fontsize{6}{6.5}\selectfont\bfseries,text=black] at (axis cs:4,2) {\contour{white}{72.2}};
\path[fill=chunktaskc17,draw=white,line width=0.4pt] (axis cs:4.53,1.55) rectangle (axis cs:5.47,2.45);
\node[font=\fontsize{6}{6.5}\selectfont\bfseries,text=black] at (axis cs:5,2) {\contour{white}{86.4}};
\path[fill=chunktaskc17,draw=white,line width=0.4pt] (axis cs:5.53,1.55) rectangle (axis cs:6.47,2.45);
\node[font=\fontsize{6}{6.5}\selectfont\bfseries,text=black] at (axis cs:6,2) {\contour{white}{86.2}};
\path[fill=chunktaskc15,draw=white,line width=0.4pt] (axis cs:6.53,1.55) rectangle (axis cs:7.47,2.45);
\node[font=\fontsize{6}{6.5}\selectfont\bfseries,text=black] at (axis cs:7,2) {\contour{white}{73.3}};
\path[fill=chunktaskc17,draw=white,line width=0.4pt] (axis cs:7.53,1.55) rectangle (axis cs:8.47,2.45);
\node[font=\fontsize{6}{6.5}\selectfont\bfseries,text=black] at (axis cs:8,2) {\contour{white}{83.3}};
\path[fill=chunktaskc15,draw=white,line width=0.4pt] (axis cs:8.53,1.55) rectangle (axis cs:9.47,2.45);
\node[font=\fontsize{6}{6.5}\selectfont\bfseries,text=black] at (axis cs:9,2) {\contour{white}{74.7}};
\path[fill=chunktaskc14,draw=white,line width=0.4pt] (axis cs:9.53,1.55) rectangle (axis cs:10.47,2.45);
\node[font=\fontsize{6}{6.5}\selectfont\bfseries,text=black] at (axis cs:10,2) {\contour{white}{70.3}};
\path[fill=chunktaskc8,draw=white,line width=0.4pt] (axis cs:0.53,0.55) rectangle (axis cs:1.47,1.45);
\node[font=\fontsize{6}{6.5}\selectfont\bfseries,text=white] at (axis cs:1,1) {40.3};
\path[fill=chunktaskc9,draw=white,line width=0.4pt] (axis cs:1.53,0.55) rectangle (axis cs:2.47,1.45);
\node[font=\fontsize{6}{6.5}\selectfont\bfseries,text=white] at (axis cs:2,1) {47.1};
\path[fill=chunktaskc8,draw=white,line width=0.4pt] (axis cs:2.53,0.55) rectangle (axis cs:3.47,1.45);
\node[font=\fontsize{6}{6.5}\selectfont\bfseries,text=white] at (axis cs:3,1) {40.5};
\path[fill=chunktaskc6,draw=white,line width=0.4pt] (axis cs:3.53,0.55) rectangle (axis cs:4.47,1.45);
\node[font=\fontsize{6}{6.5}\selectfont\bfseries,text=white] at (axis cs:4,1) {30.4};
\path[fill=chunktaskc9,draw=white,line width=0.4pt] (axis cs:4.53,0.55) rectangle (axis cs:5.47,1.45);
\node[font=\fontsize{6}{6.5}\selectfont\bfseries,text=white] at (axis cs:5,1) {42.9};
\path[fill=chunktaskc8,draw=white,line width=0.4pt] (axis cs:5.53,0.55) rectangle (axis cs:6.47,1.45);
\node[font=\fontsize{6}{6.5}\selectfont\bfseries,text=white] at (axis cs:6,1) {39.8};
\path[fill=chunktaskc6,draw=white,line width=0.4pt] (axis cs:6.53,0.55) rectangle (axis cs:7.47,1.45);
\node[font=\fontsize{6}{6.5}\selectfont\bfseries,text=white] at (axis cs:7,1) {29.5};
\path[fill=chunktaskc6,draw=white,line width=0.4pt] (axis cs:7.53,0.55) rectangle (axis cs:8.47,1.45);
\node[font=\fontsize{6}{6.5}\selectfont\bfseries,text=white] at (axis cs:8,1) {29.5};
\path[fill=chunktaskc4,draw=white,line width=0.4pt] (axis cs:8.53,0.55) rectangle (axis cs:9.47,1.45);
\node[font=\fontsize{6}{6.5}\selectfont\bfseries,text=white] at (axis cs:9,1) {21.9};
\path[fill=chunktaskc2,draw=white,line width=0.4pt] (axis cs:9.53,0.55) rectangle (axis cs:10.47,1.45);
\node[font=\fontsize{6}{6.5}\selectfont\bfseries,text=white] at (axis cs:10,1) {9.8};
\addplot[black!22,line width=0.35pt,forget plot] coordinates {(4.5,0.5) (4.5,3.5)};
\addplot[black!22,line width=0.35pt,forget plot] coordinates {(7.5,0.5) (7.5,3.5)};
\addplot[black!22,line width=0.35pt,forget plot] coordinates {(9.5,0.5) (9.5,3.5)};
\end{axis}

%% file: figures/what_matters/bitrate_chunking/appendix_success_multitask.tex
\definecolor{chunkmultic0}{RGB}{68,1,84}
\definecolor{chunkmultic1}{RGB}{70,18,100}
\definecolor{chunkmultic2}{RGB}{72,36,117}
\definecolor{chunkmultic3}{RGB}{68,52,126}
\definecolor{chunkmultic4}{RGB}{65,68,135}
\definecolor{chunkmultic5}{RGB}{59,82,138}
\definecolor{chunkmultic6}{RGB}{53,95,141}
\definecolor{chunkmultic7}{RGB}{48,108,142}
\definecolor{chunkmultic8}{RGB}{42,120,142}
\definecolor{chunkmultic9}{RGB}{38,132,141}
\definecolor{chunkmultic10}{RGB}{33,145,140}
\definecolor{chunkmultic11}{RGB}{34,156,136}
\definecolor{chunkmultic12}{RGB}{34,168,132}
\definecolor{chunkmultic13}{RGB}{51,180,122}
\definecolor{chunkmultic14}{RGB}{68,191,112}
\definecolor{chunkmultic15}{RGB}{95,200,96}
\definecolor{chunkmultic16}{RGB}{122,209,81}
\definecolor{chunkmultic17}{RGB}{156,216,60}
\definecolor{chunkmultic18}{RGB}{189,223,38}
\definecolor{chunkmultic19}{RGB}{221,227,38}
\definecolor{chunkmultic20}{RGB}{253,231,37}
\begin{axis}[
  width=\chunkPlotWidth,height=\chunkPlotHeight,scale only axis,
  name=\chunkPlotName,at={(\chunkPlotX,0)},anchor=south west,
  xmin=0.5,xmax=4.5,ymin=0.5,ymax=4.5,
  xtick={1,2,3,4},xticklabels={4,8,16,32},
  ytick={1,2,3,4},yticklabels={4,8,16,32},
  xlabel={$H$},ylabel={$H_{\mathrm{exec}}$},
  label style={font=\scriptsize},tick label style={font=\scriptsize},
  tick align=outside,tick pos=left,axis line style={black!55,line width=0.35pt},
  clip=false,
]
\path[fill=chunkmultic12,draw=white,line width=0.45pt] (axis cs:0.53,0.53) rectangle (axis cs:1.47,1.47);
\node[font=\fontsize{6}{6.5}\selectfont\bfseries,text=black] at (axis cs:1,1) {\contour{white}{60.6}};
\path[fill=chunkmultic13,draw=white,line width=0.45pt] (axis cs:1.53,0.53) rectangle (axis cs:2.47,1.47);
\node[font=\fontsize{6}{6.5}\selectfont\bfseries,text=black] at (axis cs:2,1) {\contour{white}{65.3}};
\path[fill=chunkmultic13,draw=white,line width=0.45pt] (axis cs:2.53,0.53) rectangle (axis cs:3.47,1.47);
\node[font=\fontsize{6}{6.5}\selectfont\bfseries,text=black] at (axis cs:3,1) {\contour{white}{63.3}};
\path[fill=chunkmultic10,draw=white,line width=0.45pt] (axis cs:3.53,0.53) rectangle (axis cs:4.47,1.47);
\node[font=\fontsize{6}{6.5}\selectfont\bfseries,text=black] at (axis cs:4,1) {\contour{white}{50.9}};
\path[fill=chunkmultic12,draw=white,line width=0.45pt] (axis cs:1.53,1.53) rectangle (axis cs:2.47,2.47);
\node[font=\fontsize{6}{6.5}\selectfont\bfseries,text=black] at (axis cs:2,2) {\contour{white}{61.0}};
\path[fill=chunkmultic12,draw=white,line width=0.45pt] (axis cs:2.53,1.53) rectangle (axis cs:3.47,2.47);
\node[font=\fontsize{6}{6.5}\selectfont\bfseries,text=black] at (axis cs:3,2) {\contour{white}{61.2}};
\path[fill=chunkmultic10,draw=white,line width=0.45pt] (axis cs:3.53,1.53) rectangle (axis cs:4.47,2.47);
\node[font=\fontsize{6}{6.5}\selectfont\bfseries,text=black] at (axis cs:4,2) {\contour{white}{49.2}};
\path[fill=chunkmultic10,draw=white,line width=0.45pt] (axis cs:2.53,2.53) rectangle (axis cs:3.47,3.47);
\node[font=\fontsize{6}{6.5}\selectfont\bfseries,text=black] at (axis cs:3,3) {\contour{white}{52.0}};
\path[fill=chunkmultic9,draw=white,line width=0.45pt] (axis cs:3.53,2.53) rectangle (axis cs:4.47,3.47);
\node[font=\fontsize{6}{6.5}\selectfont\bfseries,text=white] at (axis cs:4,3) {42.9};
\path[fill=chunkmultic6,draw=white,line width=0.45pt] (axis cs:3.53,3.53) rectangle (axis cs:4.47,4.47);
\node[font=\fontsize{6}{6.5}\selectfont\bfseries,text=white] at (axis cs:4,4) {31.1};
\end{axis}

%% file: figures/what_matters/empirical_support/appendix_validation_prefix_by_task.tex

\begin{tikzpicture}
\begin{groupplot}[
  group style={group size=3 by 1,horizontal sep=0.35cm},
  width=2.35cm, height=2.35cm, scale only axis,
  title style={font=\fontsize{7.8}{8.5}\selectfont\bfseries,yshift=-1pt},
  label style={font=\fontsize{7.0}{7.6}\selectfont},
  tick label style={font=\fontsize{7.2}{7.7}\selectfont},
  xmin=1, xmax=8, xtick={1,2,4,6,8},
  ymin=0, ymax=102, ytick={0,50,100},
  tick align=outside, tick pos=left,
  axis line style={black!55,line width=0.35pt},
  xmajorgrids, ymajorgrids, grid style={black!10,line width=0.3pt},
  clip=false,
]
\nextgroupplot[title={Can},ylabel={Unobserved prefix (\%)},ylabel style={at={(yticklabel cs:0.5,2pt)},anchor=near yticklabel},legend columns=1,legend image code/.code={\draw[mark repeat=2,mark phase=2,#1] plot coordinates {(0cm,0cm) (0.21cm,0cm) (0.42cm,0cm)};},legend style={at={(0.98,0.04)},anchor=south east,draw=none,fill=white,fill opacity=0.85,text opacity=1,inner sep=1pt,column sep=0.5pt,row sep=1pt,nodes={inner xsep=0.5pt,inner ysep=0pt},font=\fontsize{6.5}{7}\selectfont}]
\addplot[name path=canVQLow,draw=none,forget plot] coordinates {(1,0.000000) (2,0.566038) (3,29.622642) (4,81.132075) (5,88.301887) (6,93.584906) (7,95.660377) (8,96.226415)};
\addplot[name path=canVQHigh,draw=none,forget plot] coordinates {(1,0.377358) (2,4.716981) (3,75.094340) (4,91.886792) (5,93.962264) (6,95.283019) (7,96.037736) (8,96.603774)};
\addplot[draw=none,fill=vqbase!35,fill opacity=0.24,forget plot] fill between[of=canVQLow and canVQHigh];
\addplot[vqbase,solid,line width=0.85pt,mark=*,mark size=1.3pt,mark options={solid,fill=vqbase,draw=vqbase,line width=0.25pt}] coordinates {(1,0.188679) (2,2.641509) (3,52.358491) (4,86.509434) (5,91.132075) (6,94.433962) (7,95.849057) (8,96.415094)};
\addlegendentry{VQ}
\addplot[name path=canRVQLow,draw=none,forget plot] coordinates {(1,0.000000) (2,0.754717) (3,49.056604) (4,82.452830) (5,90.188679) (6,92.452830) (7,92.830189) (8,93.018868)};
\addplot[name path=canRVQHigh,draw=none,forget plot] coordinates {(1,0.188679) (2,6.792453) (3,74.905660) (4,91.320755) (5,94.150943) (6,95.660377) (7,96.415094) (8,96.792453)};
\addplot[draw=none,fill=rvqbase!35,fill opacity=0.24,forget plot] fill between[of=canRVQLow and canRVQHigh];
\addplot[rvqbase,densely dashed,line width=0.85pt,mark=*,mark size=1.3pt,mark options={solid,fill=rvqbase,draw=rvqbase,line width=0.25pt}] coordinates {(1,0.031447) (2,3.018868) (3,57.955975) (4,86.572327) (5,92.389937) (6,94.308176) (7,94.874214) (8,95.283019)};
\addlegendentry{RVQ}
\addplot[name path=canNRVQLow,draw=none,forget plot] coordinates {(1,0.000000) (2,3.018868) (3,54.339623) (4,83.773585) (5,89.622642) (6,92.075472) (7,93.207547) (8,93.773585)};
\addplot[name path=canNRVQHigh,draw=none,forget plot] coordinates {(1,0.000000) (2,11.320755) (3,81.132075) (4,95.660377) (5,96.792453) (6,97.358491) (7,97.547170) (8,97.735849)};
\addplot[draw=none,fill=nrvqbase!35,fill opacity=0.24,forget plot] fill between[of=canNRVQLow and canNRVQHigh];
\addplot[nrvqbase,densely dotted,line width=0.85pt,mark=*,mark size=1.3pt,mark options={solid,fill=nrvqbase,draw=nrvqbase,line width=0.25pt}] coordinates {(1,0.000000) (2,6.377358) (3,71.132075) (4,90.377358) (5,93.584906) (6,94.641509) (7,95.132075) (8,95.547170)};
\addlegendentry{NRVQ}
\addplot[name path=canFSQLow,draw=none,forget plot] coordinates {(1,0.000000) (2,0.000000) (3,16.603774) (4,74.339623) (5,91.886792) (6,93.396226) (7,94.339623) (8,95.094340)};
\addplot[name path=canFSQHigh,draw=none,forget plot] coordinates {(1,0.000000) (2,0.188679) (3,41.698113) (4,85.283019) (5,92.075472) (6,94.716981) (7,96.037736) (8,96.792453)};
\addplot[draw=none,fill=fsqbase!35,fill opacity=0.24,forget plot] fill between[of=canFSQLow and canFSQHigh];
\addplot[fsqbase,dashdotted,line width=0.85pt,mark=*,mark size=1.3pt,mark options={solid,fill=fsqbase,draw=fsqbase,line width=0.25pt}] coordinates {(1,0.000000) (2,0.094340) (3,29.150943) (4,79.811321) (5,91.981132) (6,94.056604) (7,95.188679) (8,95.943396)};
\addlegendentry{FSQ}
\addplot[name path=canQINCOLow,draw=none,forget plot] coordinates {(1,0.000000) (2,1.698113) (3,62.641509) (4,89.245283) (5,92.075472) (6,93.584906) (7,94.716981) (8,95.471698)};
\addplot[name path=canQINCOHigh,draw=none,forget plot] coordinates {(1,0.000000) (2,6.792453) (3,77.169811) (4,92.830189) (5,95.283019) (6,96.603774) (7,97.358491) (8,97.735849)};
\addplot[draw=none,fill=qincobase!35,fill opacity=0.24,forget plot] fill between[of=canQINCOLow and canQINCOHigh];
\addplot[qincobase,loosely dashdotdotted,line width=0.85pt,mark=*,mark size=1.3pt,mark options={solid,fill=qincobase,draw=qincobase,line width=0.25pt}] coordinates {(1,0.000000) (2,4.056604) (3,69.874214) (4,90.974843) (5,94.025157) (6,95.377358) (7,96.257862) (8,96.666667)};
\addlegendentry{QinCo}
\nextgroupplot[title={Lift},yticklabels={}]
\addplot[name path=liftVQLow,draw=none,forget plot] coordinates {(1,0.000000) (2,2.772277) (3,72.079208) (4,91.089109) (5,95.445545) (6,96.831683) (7,98.019802) (8,98.019802)};
\addplot[name path=liftVQHigh,draw=none,forget plot] coordinates {(1,0.198020) (2,3.168317) (3,77.029703) (4,95.445545) (5,97.425743) (6,98.613861) (7,98.613861) (8,100.000000)};
\addplot[draw=none,fill=vqbase!35,fill opacity=0.24,forget plot] fill between[of=liftVQLow and liftVQHigh];
\addplot[vqbase,solid,line width=0.85pt,mark=*,mark size=1.3pt,mark options={solid,fill=vqbase,draw=vqbase,line width=0.25pt}] coordinates {(1,0.099010) (2,2.970297) (3,74.554455) (4,93.267327) (5,96.435644) (6,97.722772) (7,98.316832) (8,99.009901)};
\addplot[name path=liftRVQLow,draw=none,forget plot] coordinates {(1,0.000000) (2,2.574257) (3,63.168317) (4,88.514851) (5,94.455446) (6,97.623762) (7,98.811881) (8,99.603960)};
\addplot[name path=liftRVQHigh,draw=none,forget plot] coordinates {(1,0.000000) (2,17.821782) (3,81.386139) (4,95.841584) (5,98.811881) (6,99.801980) (7,100.000000) (8,100.000000)};
\addplot[draw=none,fill=rvqbase!35,fill opacity=0.24,forget plot] fill between[of=liftRVQLow and liftRVQHigh];
\addplot[rvqbase,densely dashed,line width=0.85pt,mark=*,mark size=1.3pt,mark options={solid,fill=rvqbase,draw=rvqbase,line width=0.25pt}] coordinates {(1,0.000000) (2,7.491749) (3,73.300330) (4,91.650165) (5,96.699670) (6,98.811881) (7,99.570957) (8,99.834983)};
\addplot[name path=liftNRVQLow,draw=none,forget plot] coordinates {(1,0.000000) (2,4.950495) (3,60.792079) (4,90.693069) (5,96.633663) (6,98.019802) (7,99.207921) (8,99.603960)};
\addplot[name path=liftNRVQHigh,draw=none,forget plot] coordinates {(1,0.000000) (2,18.019802) (3,87.524752) (4,97.029703) (5,99.603960) (6,100.000000) (7,100.000000) (8,100.000000)};
\addplot[draw=none,fill=nrvqbase!35,fill opacity=0.24,forget plot] fill between[of=liftNRVQLow and liftNRVQHigh];
\addplot[nrvqbase,densely dotted,line width=0.85pt,mark=*,mark size=1.3pt,mark options={solid,fill=nrvqbase,draw=nrvqbase,line width=0.25pt}] coordinates {(1,0.000000) (2,11.155116) (3,76.072607) (4,93.465347) (5,97.986799) (6,99.108911) (7,99.669967) (8,99.900990)};
\addplot[name path=liftFSQLow,draw=none,forget plot] coordinates {(1,0.000000) (2,0.000000) (3,38.811881) (4,88.712871) (5,95.445545) (6,96.831683) (7,98.217822) (8,98.811881)};
\addplot[name path=liftFSQHigh,draw=none,forget plot] coordinates {(1,0.000000) (2,0.198020) (3,41.782178) (4,89.504950) (5,96.435644) (6,97.227723) (7,98.613861) (8,99.009901)};
\addplot[draw=none,fill=fsqbase!35,fill opacity=0.24,forget plot] fill between[of=liftFSQLow and liftFSQHigh];
\addplot[fsqbase,dashdotted,line width=0.85pt,mark=*,mark size=1.3pt,mark options={solid,fill=fsqbase,draw=fsqbase,line width=0.25pt}] coordinates {(1,0.000000) (2,0.099010) (3,40.297030) (4,89.108911) (5,95.940594) (6,97.029703) (7,98.415842) (8,98.910891)};
\addplot[name path=liftQINCOLow,draw=none,forget plot] coordinates {(1,0.000000) (2,4.554455) (3,75.247525) (4,91.485149) (5,96.237624) (6,98.019802) (7,99.405941) (8,99.405941)};
\addplot[name path=liftQINCOHigh,draw=none,forget plot] coordinates {(1,0.198020) (2,16.039604) (3,81.782178) (4,94.257426) (5,97.623762) (6,99.207921) (7,99.801980) (8,100.000000)};
\addplot[draw=none,fill=qincobase!35,fill opacity=0.24,forget plot] fill between[of=liftQINCOLow and liftQINCOHigh];
\addplot[qincobase,loosely dashdotdotted,line width=0.85pt,mark=*,mark size=1.3pt,mark options={solid,fill=qincobase,draw=qincobase,line width=0.25pt}] coordinates {(1,0.033003) (2,7.887789) (3,78.811881) (4,92.574257) (5,96.831683) (6,98.613861) (7,99.504950) (8,99.834983)};
\nextgroupplot[title={Square},yticklabels={}]
\addplot[name path=squareVQLow,draw=none,forget plot] coordinates {(1,0.000000) (2,0.000000) (3,33.513514) (4,82.972973) (5,92.432432) (6,93.783784) (7,94.864865) (8,94.864865)};
\addplot[name path=squareVQHigh,draw=none,forget plot] coordinates {(1,0.000000) (2,3.513514) (3,65.405405) (4,91.891892) (5,95.675676) (6,96.486486) (7,97.297297) (8,97.837838)};
\addplot[draw=none,fill=vqbase!35,fill opacity=0.24,forget plot] fill between[of=squareVQLow and squareVQHigh];
\addplot[vqbase,solid,line width=0.85pt,mark=*,mark size=1.3pt,mark options={solid,fill=vqbase,draw=vqbase,line width=0.25pt}] coordinates {(1,0.000000) (2,1.756757) (3,49.459459) (4,87.432432) (5,94.054054) (6,95.135135) (7,96.081081) (8,96.351351)};
\addplot[name path=squareRVQLow,draw=none,forget plot] coordinates {(1,0.000000) (2,0.540541) (3,44.864865) (4,80.540541) (5,90.000000) (6,92.162162) (7,93.243243) (8,94.054054)};
\addplot[name path=squareRVQHigh,draw=none,forget plot] coordinates {(1,0.000000) (2,5.405405) (3,71.351351) (4,90.540541) (5,95.405405) (6,96.756757) (7,97.567568) (8,97.567568)};
\addplot[draw=none,fill=rvqbase!35,fill opacity=0.24,forget plot] fill between[of=squareRVQLow and squareRVQHigh];
\addplot[rvqbase,densely dashed,line width=0.85pt,mark=*,mark size=1.3pt,mark options={solid,fill=rvqbase,draw=rvqbase,line width=0.25pt}] coordinates {(1,0.000000) (2,2.612613) (3,56.621622) (4,85.675676) (5,92.522523) (6,94.234234) (7,95.315315) (8,96.126126)};
\addplot[name path=squareNRVQLow,draw=none,forget plot] coordinates {(1,0.000000) (2,1.891892) (3,51.621622) (4,82.432432) (5,86.756757) (6,90.270270) (7,91.351351) (8,92.432432)};
\addplot[name path=squareNRVQHigh,draw=none,forget plot] coordinates {(1,0.270270) (2,8.108108) (3,80.000000) (4,92.702703) (5,95.675676) (6,96.216216) (7,97.567568) (8,97.837838)};
\addplot[draw=none,fill=nrvqbase!35,fill opacity=0.24,forget plot] fill between[of=squareNRVQLow and squareNRVQHigh];
\addplot[nrvqbase,densely dotted,line width=0.85pt,mark=*,mark size=1.3pt,mark options={solid,fill=nrvqbase,draw=nrvqbase,line width=0.25pt}] coordinates {(1,0.054054) (2,4.918919) (3,69.081081) (4,88.810811) (5,92.486486) (6,93.945946) (7,95.135135) (8,95.945946)};
\addplot[name path=squareFSQLow,draw=none,forget plot] coordinates {(1,0.000000) (2,0.000000) (3,25.405405) (4,80.000000) (5,92.972973) (6,95.135135) (7,95.405405) (8,95.675676)};
\addplot[name path=squareFSQHigh,draw=none,forget plot] coordinates {(1,0.000000) (2,0.270270) (3,26.756757) (4,85.405405) (5,93.513514) (6,95.405405) (7,95.405405) (8,96.216216)};
\addplot[draw=none,fill=fsqbase!35,fill opacity=0.24,forget plot] fill between[of=squareFSQLow and squareFSQHigh];
\addplot[fsqbase,dashdotted,line width=0.85pt,mark=*,mark size=1.3pt,mark options={solid,fill=fsqbase,draw=fsqbase,line width=0.25pt}] coordinates {(1,0.000000) (2,0.135135) (3,26.081081) (4,82.702703) (5,93.243243) (6,95.270270) (7,95.405405) (8,95.945946)};
\addplot[name path=squareQINCOLow,draw=none,forget plot] coordinates {(1,0.000000) (2,0.540541) (3,65.945946) (4,91.351351) (5,94.594595) (6,95.675676) (7,96.486486) (8,97.027027)};
\addplot[name path=squareQINCOHigh,draw=none,forget plot] coordinates {(1,0.270270) (2,9.729730) (3,82.972973) (4,92.162162) (5,95.675676) (6,96.756757) (7,97.027027) (8,97.567568)};
\addplot[draw=none,fill=qincobase!35,fill opacity=0.24,forget plot] fill between[of=squareQINCOLow and squareQINCOHigh];
\addplot[qincobase,loosely dashdotdotted,line width=0.85pt,mark=*,mark size=1.3pt,mark options={solid,fill=qincobase,draw=qincobase,line width=0.25pt}] coordinates {(1,0.090090) (2,3.873874) (3,72.297297) (4,91.936937) (5,95.000000) (6,96.306306) (7,96.846847) (8,97.387387)};
\end{groupplot}
\pgfresetboundingbox
\path[use as bounding box]
  ($(group c1r1.south west)+(-1.30cm,-0.48cm)$) rectangle
  ($(group c3r1.north east)+(0.10cm,0.82cm)$);
\end{tikzpicture}

%% file: figures/what_matters/empirical_support/appendix_validation_exact_match.tex

\begin{tikzpicture}
\begin{axis}[
  name=exactmatch, width=3.0cm, height=2.35cm, scale only axis,
  label style={font=\fontsize{7.2}{7.8}\selectfont},
  tick label style={font=\fontsize{7.0}{7.6}\selectfont},
  yticklabel style={anchor=east,inner sep=1.0pt,font=\fontsize{6.1}{6.6}\selectfont},
  xmin=0, xmax=4.2, xtick={0,2,4},
  xticklabels={$0\%$,$2\%$,$4\%$},
  ymin=0.45, ymax=5.55, ytick={1,...,5},
  yticklabels={\tokenizerid{qincobase}{QinCo},\tokenizerid{fsqbase}{FSQ},\tokenizerid{nrvqbase}{NRVQ},\tokenizerid{rvqbase}{RVQ},\tokenizerid{vqbase}{VQ}},
  tick align=outside, tick pos=left,
  axis line style={black!55,line width=0.35pt},
  xmajorgrids, grid style={black!10,line width=0.3pt},
  legend columns=1,
  legend style={at={(0.98,0.04)},anchor=south east,draw=none,
    fill=white,fill opacity=0.85,text opacity=1,
    inner sep=1pt,column sep=1pt,row sep=-0.5pt,
    font=\fontsize{5}{5.5}\selectfont},
  clip=false,
]
\addplot+[only marks,forget plot,mark=diamond*,mark size=1.75pt,mark options={solid,fill=vqbase,draw=vqbase,line width=0.55pt},error bars/.cd,x dir=both,x explicit,error bar style={draw=vqbase!72,line width=0.7pt},error mark=none] coordinates {(2.169811,5.200) += (0.283019,0) -= (0.283019,0)};
\addplot+[only marks,forget plot,mark=square*,mark size=1.75pt,mark options={solid,fill=vqbase,draw=vqbase,line width=0.55pt},error bars/.cd,x dir=both,x explicit,error bar style={draw=vqbase!72,line width=0.7pt},error mark=none] coordinates {(0.099010,5.000) += (0.099010,0) -= (0.099010,0)};
\addplot+[only marks,forget plot,mark=triangle*,mark size=1.75pt,mark options={solid,fill=vqbase,draw=vqbase,line width=0.55pt},error bars/.cd,x dir=both,x explicit,error bar style={draw=vqbase!72,line width=0.7pt},error mark=none] coordinates {(1.756757,4.800) += (0.405405,0) -= (0.405405,0)};
\addplot+[only marks,forget plot,mark=diamond*,mark size=1.75pt,mark options={solid,fill=rvqbase,draw=rvqbase,line width=0.55pt},error bars/.cd,x dir=both,x explicit,error bar style={draw=rvqbase!72,line width=0.7pt},error mark=none] coordinates {(2.358491,4.200) += (1.415094,0) -= (0.849057,0)};
\addplot+[only marks,forget plot,mark=square*,mark size=1.75pt,mark options={solid,fill=rvqbase,draw=rvqbase,line width=0.55pt},error bars/.cd,x dir=both,x explicit,error bar style={draw=rvqbase!72,line width=0.7pt},error mark=none] coordinates {(0.033003,4.000) += (0.165017,0) -= (0.033003,0)};
\addplot+[only marks,forget plot,mark=triangle*,mark size=1.75pt,mark options={solid,fill=rvqbase,draw=rvqbase,line width=0.55pt},error bars/.cd,x dir=both,x explicit,error bar style={draw=rvqbase!72,line width=0.7pt},error mark=none] coordinates {(1.306306,3.800) += (0.585586,0) -= (0.495495,0)};
\addplot+[only marks,forget plot,mark=diamond*,mark size=1.75pt,mark options={solid,fill=nrvqbase,draw=nrvqbase,line width=0.55pt},error bars/.cd,x dir=both,x explicit,error bar style={draw=nrvqbase!72,line width=0.7pt},error mark=none] coordinates {(2.188679,3.200) += (0.830189,0) -= (1.056604,0)};
\addplot+[only marks,forget plot,mark=square*,mark size=1.75pt,mark options={solid,fill=nrvqbase,draw=nrvqbase,line width=0.55pt},error bars/.cd,x dir=both,x explicit,error bar style={draw=nrvqbase!72,line width=0.7pt},error mark=none] coordinates {(0.000000,3.000) += (0.000000,0) -= (0.000000,0)};
\addplot+[only marks,forget plot,mark=triangle*,mark size=1.75pt,mark options={solid,fill=nrvqbase,draw=nrvqbase,line width=0.55pt},error bars/.cd,x dir=both,x explicit,error bar style={draw=nrvqbase!72,line width=0.7pt},error mark=none] coordinates {(1.567568,2.800) += (0.864865,0) -= (0.486486,0)};
\addplot+[only marks,forget plot,mark=diamond*,mark size=1.75pt,mark options={solid,fill=fsqbase,draw=fsqbase,line width=0.55pt},error bars/.cd,x dir=both,x explicit,error bar style={draw=fsqbase!72,line width=0.7pt},error mark=none] coordinates {(2.169811,2.200) += (0.283019,0) -= (0.283019,0)};
\addplot+[only marks,forget plot,mark=square*,mark size=1.75pt,mark options={solid,fill=fsqbase,draw=fsqbase,line width=0.55pt},error bars/.cd,x dir=both,x explicit,error bar style={draw=fsqbase!72,line width=0.7pt},error mark=none] coordinates {(0.099010,2.000) += (0.099010,0) -= (0.099010,0)};
\addplot+[only marks,forget plot,mark=triangle*,mark size=1.75pt,mark options={solid,fill=fsqbase,draw=fsqbase,line width=0.55pt},error bars/.cd,x dir=both,x explicit,error bar style={draw=fsqbase!72,line width=0.7pt},error mark=none] coordinates {(1.486486,1.800) += (0.135135,0) -= (0.135135,0)};
\addplot+[only marks,forget plot,mark=diamond*,mark size=1.75pt,mark options={solid,fill=qincobase,draw=qincobase,line width=0.55pt},error bars/.cd,x dir=both,x explicit,error bar style={draw=qincobase!72,line width=0.7pt},error mark=none] coordinates {(1.666667,1.200) += (0.597484,0) -= (0.723270,0)};
\addplot+[only marks,forget plot,mark=square*,mark size=1.75pt,mark options={solid,fill=qincobase,draw=qincobase,line width=0.55pt},error bars/.cd,x dir=both,x explicit,error bar style={draw=qincobase!72,line width=0.7pt},error mark=none] coordinates {(0.000000,1.000) += (0.000000,0) -= (0.000000,0)};
\addplot+[only marks,forget plot,mark=triangle*,mark size=1.75pt,mark options={solid,fill=qincobase,draw=qincobase,line width=0.55pt},error bars/.cd,x dir=both,x explicit,error bar style={draw=qincobase!72,line width=0.7pt},error mark=none] coordinates {(1.216216,0.800) += (0.405405,0) -= (0.135135,0)};
\addlegendimage{only marks,mark=diamond*,mark size=1.25pt,color=black,mark options={solid,fill=white,draw=black,line width=0.55pt}}
\addlegendentry{Can}
\addlegendimage{only marks,mark=square*,mark size=1.25pt,color=black,mark options={solid,fill=white,draw=black,line width=0.55pt}}
\addlegendentry{Lift}
\addlegendimage{only marks,mark=triangle*,mark size=1.25pt,color=black,mark options={solid,fill=white,draw=black,line width=0.55pt}}
\addlegendentry{Square}
\end{axis}
\pgfresetboundingbox
\path[use as bounding box] ($(exactmatch.south west)+(-1.20cm,-0.48cm)$)
  rectangle ($(exactmatch.north east)+(0.10cm,0.82cm)$);
\end{tikzpicture}

%% file: figures/what_matters/empirical_support/appendix_validation_action_error.tex

\begin{tikzpicture}
\begin{groupplot}[
  group style={group size=3 by 1,horizontal sep=0.42cm},
  width=3.55cm, height=4.6cm, scale only axis,
  title style={font=\fontsize{7.8}{8.5}\selectfont\bfseries,yshift=-1pt},
  label style={font=\fontsize{7.0}{7.6}\selectfont},
  tick label style={font=\fontsize{7.2}{7.7}\selectfont},
  xmin=-0.01, xmax=0.23, xtick={0,0.1,0.2},
  ymin=0.45, ymax=22.55, ytick={1,...,22},
  ytick style={draw=none},
  tick align=outside, tick pos=left,
  axis line style={black!55,line width=0.35pt},
  xmajorgrids, grid style={black!10,line width=0.3pt},
  clip=false,
]
\nextgroupplot[title={Can},yticklabels={\tokenizerid{qincobase!90}{C2-QINCO-L},\tokenizerid{qincobase!80}{C2-QINCO-S},\tokenizerid{qincobase!70}{CT-QINCO-L},\tokenizerid{qincobase!60}{CT-QINCO-S},\tokenizerid{qincobase!50}{TF-QINCO-L},\tokenizerid{qincobase!40}{TF-QINCO-S},\tokenizerid{fsqbase!85}{TF-FSQ-L},\tokenizerid{fsqbase!60}{TF-FSQ-S},\tokenizerid{nrvqbase!90}{C2-NRVQ-L},\tokenizerid{nrvqbase!80}{C2-NRVQ-S},\tokenizerid{nrvqbase!70}{CT-NRVQ-L},\tokenizerid{nrvqbase!60}{CT-NRVQ-S},\tokenizerid{nrvqbase!50}{TF-NRVQ-L},\tokenizerid{nrvqbase!40}{TF-NRVQ-S},\tokenizerid{rvqbase!90}{C2-RVQ-L},\tokenizerid{rvqbase!80}{C2-RVQ-S},\tokenizerid{rvqbase!70}{CT-RVQ-L},\tokenizerid{rvqbase!60}{CT-RVQ-S},\tokenizerid{rvqbase!50}{TF-RVQ-L},\tokenizerid{rvqbase!40}{TF-RVQ-S},\tokenizerid{vqbase!85}{TF-VQ-L},\tokenizerid{vqbase!60}{TF-VQ-S}},yticklabel style={anchor=east,inner sep=1.0pt,font=\fontsize{5.6}{6.0}\selectfont},]
\addplot[black!60,densely dashed,line width=0.45pt,forget plot] coordinates {(0,0.5) (0,22.5)};
\addplot[black!18,line width=0.3pt,forget plot] coordinates {(-0.01,20.5) (0.23,20.5)};
\addplot[black!18,line width=0.3pt,forget plot] coordinates {(-0.01,14.5) (0.23,14.5)};
\addplot[black!18,line width=0.3pt,forget plot] coordinates {(-0.01,8.5) (0.23,8.5)};
\addplot[black!18,line width=0.3pt,forget plot] coordinates {(-0.01,6.5) (0.23,6.5)};
\addplot+[only marks,mark=diamond*,mark size=1.65pt,mark options={fill=vqbase!60,draw=vqbase,line width=0.35pt},error bars/.cd,x dir=both,x explicit,error bar style={draw=vqbase,line width=0.35pt},error mark=none] coordinates {(0.12089356,22) += (0.01797151,0) -= (0.01782252,0)};
\addplot+[only marks,mark=diamond*,mark size=1.65pt,mark options={fill=vqbase!85,draw=vqbase,line width=0.35pt},error bars/.cd,x dir=both,x explicit,error bar style={draw=vqbase,line width=0.35pt},error mark=none] coordinates {(0.18665717,21) += (0.02245538,0) -= (0.02153323,0)};
\addplot+[only marks,mark=diamond*,mark size=1.65pt,mark options={fill=rvqbase!40,draw=rvqbase,line width=0.35pt},error bars/.cd,x dir=both,x explicit,error bar style={draw=rvqbase,line width=0.35pt},error mark=none] coordinates {(0.01528987,20) += (0.00346235,0) -= (0.00324665,0)};
\addplot+[only marks,mark=diamond*,mark size=1.65pt,mark options={fill=rvqbase!50,draw=rvqbase,line width=0.35pt},error bars/.cd,x dir=both,x explicit,error bar style={draw=rvqbase,line width=0.35pt},error mark=none] coordinates {(0.00816447,19) += (0.00428308,0) -= (0.00434955,0)};
\addplot+[only marks,mark=diamond*,mark size=1.65pt,mark options={fill=rvqbase!60,draw=rvqbase,line width=0.35pt},error bars/.cd,x dir=both,x explicit,error bar style={draw=rvqbase,line width=0.35pt},error mark=none] coordinates {(0.01775290,18) += (0.00366511,0) -= (0.00354658,0)};
\addplot+[only marks,mark=diamond*,mark size=1.65pt,mark options={fill=rvqbase!70,draw=rvqbase,line width=0.35pt},error bars/.cd,x dir=both,x explicit,error bar style={draw=rvqbase,line width=0.35pt},error mark=none] coordinates {(0.00656147,17) += (0.00447010,0) -= (0.00423589,0)};
\addplot+[only marks,mark=diamond*,mark size=1.65pt,mark options={fill=rvqbase!80,draw=rvqbase,line width=0.35pt},error bars/.cd,x dir=both,x explicit,error bar style={draw=rvqbase,line width=0.35pt},error mark=none] coordinates {(0.01511366,16) += (0.00441083,0) -= (0.00394598,0)};
\addplot+[only marks,mark=diamond*,mark size=1.65pt,mark options={fill=rvqbase!90,draw=rvqbase,line width=0.35pt},error bars/.cd,x dir=both,x explicit,error bar style={draw=rvqbase,line width=0.35pt},error mark=none] coordinates {(0.00361030,15) += (0.00468201,0) -= (0.00438754,0)};
\addplot+[only marks,mark=diamond*,mark size=1.65pt,mark options={fill=nrvqbase!40,draw=nrvqbase,line width=0.35pt},error bars/.cd,x dir=both,x explicit,error bar style={draw=nrvqbase,line width=0.35pt},error mark=none] coordinates {(0.01350079,14) += (0.00283359,0) -= (0.00286085,0)};
\addplot+[only marks,mark=diamond*,mark size=1.65pt,mark options={fill=nrvqbase!50,draw=nrvqbase,line width=0.35pt},error bars/.cd,x dir=both,x explicit,error bar style={draw=nrvqbase,line width=0.35pt},error mark=none] coordinates {(0.00578207,13) += (0.00296312,0) -= (0.00295158,0)};
\addplot+[only marks,mark=diamond*,mark size=1.65pt,mark options={fill=nrvqbase!60,draw=nrvqbase,line width=0.35pt},error bars/.cd,x dir=both,x explicit,error bar style={draw=nrvqbase,line width=0.35pt},error mark=none] coordinates {(0.00000000,12) += (0.00000000,0) -= (0.00000000,0)};
\addplot+[only marks,mark=diamond*,mark size=1.65pt,mark options={fill=nrvqbase!70,draw=nrvqbase,line width=0.35pt},error bars/.cd,x dir=both,x explicit,error bar style={draw=nrvqbase,line width=0.35pt},error mark=none] coordinates {(0.00374551,11) += (0.00335561,0) -= (0.00324011,0)};
\addplot+[only marks,mark=diamond*,mark size=1.65pt,mark options={fill=nrvqbase!80,draw=nrvqbase,line width=0.35pt},error bars/.cd,x dir=both,x explicit,error bar style={draw=nrvqbase,line width=0.35pt},error mark=none] coordinates {(0.01788998,10) += (0.00391856,0) -= (0.00364444,0)};
\addplot+[only marks,mark=diamond*,mark size=1.65pt,mark options={fill=nrvqbase!90,draw=nrvqbase,line width=0.35pt},error bars/.cd,x dir=both,x explicit,error bar style={draw=nrvqbase,line width=0.35pt},error mark=none] coordinates {(0.00255511,9) += (0.00440400,0) -= (0.00440331,0)};
\addplot+[only marks,mark=diamond*,mark size=1.65pt,mark options={fill=fsqbase!60,draw=fsqbase,line width=0.35pt},error bars/.cd,x dir=both,x explicit,error bar style={draw=fsqbase,line width=0.35pt},error mark=none] coordinates {(0.05680200,8) += (0.00659491,0) -= (0.00665929,0)};
\addplot+[only marks,mark=diamond*,mark size=1.65pt,mark options={fill=fsqbase!85,draw=fsqbase,line width=0.35pt},error bars/.cd,x dir=both,x explicit,error bar style={draw=fsqbase,line width=0.35pt},error mark=none] coordinates {(0.15733817,7) += (0.01769965,0) -= (0.01793929,0)};
\addplot+[only marks,mark=diamond*,mark size=1.65pt,mark options={fill=qincobase!40,draw=qincobase,line width=0.35pt},error bars/.cd,x dir=both,x explicit,error bar style={draw=qincobase,line width=0.35pt},error mark=none] coordinates {(0.00168682,6) += (0.00316570,0) -= (0.00332166,0)};
\addplot+[only marks,mark=diamond*,mark size=1.65pt,mark options={fill=qincobase!50,draw=qincobase,line width=0.35pt},error bars/.cd,x dir=both,x explicit,error bar style={draw=qincobase,line width=0.35pt},error mark=none] coordinates {(0.00365155,5) += (0.00425249,0) -= (0.00419721,0)};
\addplot+[only marks,mark=diamond*,mark size=1.65pt,mark options={fill=qincobase!60,draw=qincobase,line width=0.35pt},error bars/.cd,x dir=both,x explicit,error bar style={draw=qincobase,line width=0.35pt},error mark=none] coordinates {(0.00803430,4) += (0.00379570,0) -= (0.00391973,0)};
\addplot+[only marks,mark=diamond*,mark size=1.65pt,mark options={fill=qincobase!70,draw=qincobase,line width=0.35pt},error bars/.cd,x dir=both,x explicit,error bar style={draw=qincobase,line width=0.35pt},error mark=none] coordinates {(0.00455601,3) += (0.00500645,0) -= (0.00485250,0)};
\addplot+[only marks,mark=diamond*,mark size=1.65pt,mark options={fill=qincobase!80,draw=qincobase,line width=0.35pt},error bars/.cd,x dir=both,x explicit,error bar style={draw=qincobase,line width=0.35pt},error mark=none] coordinates {(0.00941867,2) += (0.00570699,0) -= (0.00555904,0)};
\addplot+[only marks,mark=diamond*,mark size=1.65pt,mark options={fill=qincobase!90,draw=qincobase,line width=0.35pt},error bars/.cd,x dir=both,x explicit,error bar style={draw=qincobase,line width=0.35pt},error mark=none] coordinates {(0.00521477,1) += (0.00460027,0) -= (0.00477067,0)};
\nextgroupplot[title={Lift},yticklabels={},]
\addplot[black!60,densely dashed,line width=0.45pt,forget plot] coordinates {(0,0.5) (0,22.5)};
\addplot[black!18,line width=0.3pt,forget plot] coordinates {(-0.01,20.5) (0.23,20.5)};
\addplot[black!18,line width=0.3pt,forget plot] coordinates {(-0.01,14.5) (0.23,14.5)};
\addplot[black!18,line width=0.3pt,forget plot] coordinates {(-0.01,8.5) (0.23,8.5)};
\addplot[black!18,line width=0.3pt,forget plot] coordinates {(-0.01,6.5) (0.23,6.5)};
\addplot+[only marks,mark=square*,mark size=1.65pt,mark options={fill=vqbase!60,draw=vqbase,line width=0.35pt},error bars/.cd,x dir=both,x explicit,error bar style={draw=vqbase,line width=0.35pt},error mark=none] coordinates {(0.10350752,22) += (0.01859938,0) -= (0.01778784,0)};
\addplot+[only marks,mark=square*,mark size=1.65pt,mark options={fill=vqbase!85,draw=vqbase,line width=0.35pt},error bars/.cd,x dir=both,x explicit,error bar style={draw=vqbase,line width=0.35pt},error mark=none] coordinates {(0.10754112,21) += (0.01418827,0) -= (0.01383060,0)};
\addplot+[only marks,mark=square*,mark size=1.65pt,mark options={fill=rvqbase!40,draw=rvqbase,line width=0.35pt},error bars/.cd,x dir=both,x explicit,error bar style={draw=rvqbase,line width=0.35pt},error mark=none] coordinates {(0.01266623,20) += (0.00259663,0) -= (0.00248446,0)};
\addplot+[only marks,mark=square*,mark size=1.65pt,mark options={fill=rvqbase!50,draw=rvqbase,line width=0.35pt},error bars/.cd,x dir=both,x explicit,error bar style={draw=rvqbase,line width=0.35pt},error mark=none] coordinates {(0.00775997,19) += (0.00357493,0) -= (0.00347662,0)};
\addplot+[only marks,mark=square*,mark size=1.65pt,mark options={fill=rvqbase!60,draw=rvqbase,line width=0.35pt},error bars/.cd,x dir=both,x explicit,error bar style={draw=rvqbase,line width=0.35pt},error mark=none] coordinates {(0.01192724,18) += (0.00396316,0) -= (0.00389370,0)};
\addplot+[only marks,mark=square*,mark size=1.65pt,mark options={fill=rvqbase!70,draw=rvqbase,line width=0.35pt},error bars/.cd,x dir=both,x explicit,error bar style={draw=rvqbase,line width=0.35pt},error mark=none] coordinates {(0.00196838,17) += (0.00423831,0) -= (0.00422489,0)};
\addplot+[only marks,mark=square*,mark size=1.65pt,mark options={fill=rvqbase!80,draw=rvqbase,line width=0.35pt},error bars/.cd,x dir=both,x explicit,error bar style={draw=rvqbase,line width=0.35pt},error mark=none] coordinates {(0.01289620,16) += (0.00367234,0) -= (0.00363509,0)};
\addplot+[only marks,mark=square*,mark size=1.65pt,mark options={fill=rvqbase!90,draw=rvqbase,line width=0.35pt},error bars/.cd,x dir=both,x explicit,error bar style={draw=rvqbase,line width=0.35pt},error mark=none] coordinates {(0.00780938,15) += (0.00383627,0) -= (0.00384265,0)};
\addplot+[only marks,mark=square*,mark size=1.65pt,mark options={fill=nrvqbase!40,draw=nrvqbase,line width=0.35pt},error bars/.cd,x dir=both,x explicit,error bar style={draw=nrvqbase,line width=0.35pt},error mark=none] coordinates {(0.00796118,14) += (0.00305827,0) -= (0.00344519,0)};
\addplot+[only marks,mark=square*,mark size=1.65pt,mark options={fill=nrvqbase!50,draw=nrvqbase,line width=0.35pt},error bars/.cd,x dir=both,x explicit,error bar style={draw=nrvqbase,line width=0.35pt},error mark=none] coordinates {(0.00521109,13) += (0.00458890,0) -= (0.00431088,0)};
\addplot+[only marks,mark=square*,mark size=1.65pt,mark options={fill=nrvqbase!60,draw=nrvqbase,line width=0.35pt},error bars/.cd,x dir=both,x explicit,error bar style={draw=nrvqbase,line width=0.35pt},error mark=none] coordinates {(0.00965718,12) += (0.00320941,0) -= (0.00304141,0)};
\addplot+[only marks,mark=square*,mark size=1.65pt,mark options={fill=nrvqbase!70,draw=nrvqbase,line width=0.35pt},error bars/.cd,x dir=both,x explicit,error bar style={draw=nrvqbase,line width=0.35pt},error mark=none] coordinates {(0.00472083,11) += (0.00307546,0) -= (0.00289962,0)};
\addplot+[only marks,mark=square*,mark size=1.65pt,mark options={fill=nrvqbase!80,draw=nrvqbase,line width=0.35pt},error bars/.cd,x dir=both,x explicit,error bar style={draw=nrvqbase,line width=0.35pt},error mark=none] coordinates {(0.01281038,10) += (0.00278476,0) -= (0.00260045,0)};
\addplot+[only marks,mark=square*,mark size=1.65pt,mark options={fill=nrvqbase!90,draw=nrvqbase,line width=0.35pt},error bars/.cd,x dir=both,x explicit,error bar style={draw=nrvqbase,line width=0.35pt},error mark=none] coordinates {(0.00746532,9) += (0.00338587,0) -= (0.00315426,0)};
\addplot+[only marks,mark=square*,mark size=1.65pt,mark options={fill=fsqbase!60,draw=fsqbase,line width=0.35pt},error bars/.cd,x dir=both,x explicit,error bar style={draw=fsqbase,line width=0.35pt},error mark=none] coordinates {(0.11152642,8) += (0.01405423,0) -= (0.01378349,0)};
\addplot+[only marks,mark=square*,mark size=1.65pt,mark options={fill=fsqbase!85,draw=fsqbase,line width=0.35pt},error bars/.cd,x dir=both,x explicit,error bar style={draw=fsqbase,line width=0.35pt},error mark=none] coordinates {(0.04738941,7) += (0.00753951,0) -= (0.00710674,0)};
\addplot+[only marks,mark=square*,mark size=1.65pt,mark options={fill=qincobase!40,draw=qincobase,line width=0.35pt},error bars/.cd,x dir=both,x explicit,error bar style={draw=qincobase,line width=0.35pt},error mark=none] coordinates {(0.00589725,6) += (0.00480954,0) -= (0.00444440,0)};
\addplot+[only marks,mark=square*,mark size=1.65pt,mark options={fill=qincobase!50,draw=qincobase,line width=0.35pt},error bars/.cd,x dir=both,x explicit,error bar style={draw=qincobase,line width=0.35pt},error mark=none] coordinates {(0.00226011,5) += (0.00293534,0) -= (0.00292561,0)};
\addplot+[only marks,mark=square*,mark size=1.65pt,mark options={fill=qincobase!60,draw=qincobase,line width=0.35pt},error bars/.cd,x dir=both,x explicit,error bar style={draw=qincobase,line width=0.35pt},error mark=none] coordinates {(0.00564811,4) += (0.00447833,0) -= (0.00449315,0)};
\addplot+[only marks,mark=square*,mark size=1.65pt,mark options={fill=qincobase!70,draw=qincobase,line width=0.35pt},error bars/.cd,x dir=both,x explicit,error bar style={draw=qincobase,line width=0.35pt},error mark=none] coordinates {(0.00519373,3) += (0.00474503,0) -= (0.00420043,0)};
\addplot+[only marks,mark=square*,mark size=1.65pt,mark options={fill=qincobase!80,draw=qincobase,line width=0.35pt},error bars/.cd,x dir=both,x explicit,error bar style={draw=qincobase,line width=0.35pt},error mark=none] coordinates {(0.00521763,2) += (0.00514377,0) -= (0.00530871,0)};
\addplot+[only marks,mark=square*,mark size=1.65pt,mark options={fill=qincobase!90,draw=qincobase,line width=0.35pt},error bars/.cd,x dir=both,x explicit,error bar style={draw=qincobase,line width=0.35pt},error mark=none] coordinates {(0.00223750,1) += (0.00294090,0) -= (0.00306694,0)};
\nextgroupplot[title={Square},yticklabels={},]
\addplot[black!60,densely dashed,line width=0.45pt,forget plot] coordinates {(0,0.5) (0,22.5)};
\addplot[black!18,line width=0.3pt,forget plot] coordinates {(-0.01,20.5) (0.23,20.5)};
\addplot[black!18,line width=0.3pt,forget plot] coordinates {(-0.01,14.5) (0.23,14.5)};
\addplot[black!18,line width=0.3pt,forget plot] coordinates {(-0.01,8.5) (0.23,8.5)};
\addplot[black!18,line width=0.3pt,forget plot] coordinates {(-0.01,6.5) (0.23,6.5)};
\addplot+[only marks,mark=triangle*,mark size=1.65pt,mark options={fill=vqbase!60,draw=vqbase,line width=0.35pt},error bars/.cd,x dir=both,x explicit,error bar style={draw=vqbase,line width=0.35pt},error mark=none] coordinates {(0.20100483,22) += (0.01942963,0) -= (0.01944763,0)};
\addplot+[only marks,mark=triangle*,mark size=1.65pt,mark options={fill=vqbase!85,draw=vqbase,line width=0.35pt},error bars/.cd,x dir=both,x explicit,error bar style={draw=vqbase,line width=0.35pt},error mark=none] coordinates {(0.04730622,21) += (0.00780342,0) -= (0.00748630,0)};
\addplot+[only marks,mark=triangle*,mark size=1.65pt,mark options={fill=rvqbase!40,draw=rvqbase,line width=0.35pt},error bars/.cd,x dir=both,x explicit,error bar style={draw=rvqbase,line width=0.35pt},error mark=none] coordinates {(0.01728758,20) += (0.00352150,0) -= (0.00348986,0)};
\addplot+[only marks,mark=triangle*,mark size=1.65pt,mark options={fill=rvqbase!50,draw=rvqbase,line width=0.35pt},error bars/.cd,x dir=both,x explicit,error bar style={draw=rvqbase,line width=0.35pt},error mark=none] coordinates {(0.01115146,19) += (0.00524296,0) -= (0.00482594,0)};
\addplot+[only marks,mark=triangle*,mark size=1.65pt,mark options={fill=rvqbase!60,draw=rvqbase,line width=0.35pt},error bars/.cd,x dir=both,x explicit,error bar style={draw=rvqbase,line width=0.35pt},error mark=none] coordinates {(0.01751271,18) += (0.00479940,0) -= (0.00450719,0)};
\addplot+[only marks,mark=triangle*,mark size=1.65pt,mark options={fill=rvqbase!70,draw=rvqbase,line width=0.35pt},error bars/.cd,x dir=both,x explicit,error bar style={draw=rvqbase,line width=0.35pt},error mark=none] coordinates {(0.01449031,17) += (0.00520473,0) -= (0.00447380,0)};
\addplot+[only marks,mark=triangle*,mark size=1.65pt,mark options={fill=rvqbase!80,draw=rvqbase,line width=0.35pt},error bars/.cd,x dir=both,x explicit,error bar style={draw=rvqbase,line width=0.35pt},error mark=none] coordinates {(0.01730728,16) += (0.00401803,0) -= (0.00396603,0)};
\addplot+[only marks,mark=triangle*,mark size=1.65pt,mark options={fill=rvqbase!90,draw=rvqbase,line width=0.35pt},error bars/.cd,x dir=both,x explicit,error bar style={draw=rvqbase,line width=0.35pt},error mark=none] coordinates {(0.00894485,15) += (0.00683391,0) -= (0.00694617,0)};
\addplot+[only marks,mark=triangle*,mark size=1.65pt,mark options={fill=nrvqbase!40,draw=nrvqbase,line width=0.35pt},error bars/.cd,x dir=both,x explicit,error bar style={draw=nrvqbase,line width=0.35pt},error mark=none] coordinates {(0.01664089,14) += (0.00408469,0) -= (0.00378367,0)};
\addplot+[only marks,mark=triangle*,mark size=1.65pt,mark options={fill=nrvqbase!50,draw=nrvqbase,line width=0.35pt},error bars/.cd,x dir=both,x explicit,error bar style={draw=nrvqbase,line width=0.35pt},error mark=none] coordinates {(0.00847036,13) += (0.00527816,0) -= (0.00470975,0)};
\addplot+[only marks,mark=triangle*,mark size=1.65pt,mark options={fill=nrvqbase!60,draw=nrvqbase,line width=0.35pt},error bars/.cd,x dir=both,x explicit,error bar style={draw=nrvqbase,line width=0.35pt},error mark=none] coordinates {(0.00000000,12) += (0.00000000,0) -= (0.00000000,0)};
\addplot+[only marks,mark=triangle*,mark size=1.65pt,mark options={fill=nrvqbase!70,draw=nrvqbase,line width=0.35pt},error bars/.cd,x dir=both,x explicit,error bar style={draw=nrvqbase,line width=0.35pt},error mark=none] coordinates {(0.00283347,11) += (0.00380745,0) -= (0.00372365,0)};
\addplot+[only marks,mark=triangle*,mark size=1.65pt,mark options={fill=nrvqbase!80,draw=nrvqbase,line width=0.35pt},error bars/.cd,x dir=both,x explicit,error bar style={draw=nrvqbase,line width=0.35pt},error mark=none] coordinates {(0.01806125,10) += (0.00405236,0) -= (0.00392424,0)};
\addplot+[only marks,mark=triangle*,mark size=1.65pt,mark options={fill=nrvqbase!90,draw=nrvqbase,line width=0.35pt},error bars/.cd,x dir=both,x explicit,error bar style={draw=nrvqbase,line width=0.35pt},error mark=none] coordinates {(0.00490515,9) += (0.00457553,0) -= (0.00440024,0)};
\addplot+[only marks,mark=triangle*,mark size=1.65pt,mark options={fill=fsqbase!60,draw=fsqbase,line width=0.35pt},error bars/.cd,x dir=both,x explicit,error bar style={draw=fsqbase,line width=0.35pt},error mark=none] coordinates {(0.13765722,8) += (0.02497529,0) -= (0.02380783,0)};
\addplot+[only marks,mark=triangle*,mark size=1.65pt,mark options={fill=fsqbase!85,draw=fsqbase,line width=0.35pt},error bars/.cd,x dir=both,x explicit,error bar style={draw=fsqbase,line width=0.35pt},error mark=none] coordinates {(0.16133438,7) += (0.01929491,0) -= (0.02002095,0)};
\addplot+[only marks,mark=triangle*,mark size=1.65pt,mark options={fill=qincobase!40,draw=qincobase,line width=0.35pt},error bars/.cd,x dir=both,x explicit,error bar style={draw=qincobase,line width=0.35pt},error mark=none] coordinates {(0.00674854,6) += (0.00408090,0) -= (0.00423701,0)};
\addplot+[only marks,mark=triangle*,mark size=1.65pt,mark options={fill=qincobase!50,draw=qincobase,line width=0.35pt},error bars/.cd,x dir=both,x explicit,error bar style={draw=qincobase,line width=0.35pt},error mark=none] coordinates {(0.00659722,5) += (0.00416950,0) -= (0.00438690,0)};
\addplot+[only marks,mark=triangle*,mark size=1.65pt,mark options={fill=qincobase!60,draw=qincobase,line width=0.35pt},error bars/.cd,x dir=both,x explicit,error bar style={draw=qincobase,line width=0.35pt},error mark=none] coordinates {(0.00539813,4) += (0.00421876,0) -= (0.00407967,0)};
\addplot+[only marks,mark=triangle*,mark size=1.65pt,mark options={fill=qincobase!70,draw=qincobase,line width=0.35pt},error bars/.cd,x dir=both,x explicit,error bar style={draw=qincobase,line width=0.35pt},error mark=none] coordinates {(0.00647626,3) += (0.00446911,0) -= (0.00450949,0)};
\addplot+[only marks,mark=triangle*,mark size=1.65pt,mark options={fill=qincobase!80,draw=qincobase,line width=0.35pt},error bars/.cd,x dir=both,x explicit,error bar style={draw=qincobase,line width=0.35pt},error mark=none] coordinates {(0.01733378,2) += (0.00966383,0) -= (0.00919810,0)};
\addplot+[only marks,mark=triangle*,mark size=1.65pt,mark options={fill=qincobase!90,draw=qincobase,line width=0.35pt},error bars/.cd,x dir=both,x explicit,error bar style={draw=qincobase,line width=0.35pt},error mark=none] coordinates {(0.00904223,1) += (0.00666700,0) -= (0.00601376,0)};
\end{groupplot}
\pgfresetboundingbox
\path[use as bounding box]
  ($(group c1r1.south west)+(-1.60cm,-0.48cm)$) rectangle
  ($(group c3r1.north east)+(0.10cm,0.55cm)$);
\end{tikzpicture}

%% file: figures/what_matters/empirical_support/appendix_rollout_prefix_by_task.tex

\begin{tikzpicture}
\begin{groupplot}[
  group style={group size=3 by 1,horizontal sep=0.42cm},
  width=3.55cm, height=2.35cm, scale only axis,
  title style={font=\fontsize{7.8}{8.5}\selectfont\bfseries,yshift=-1pt},
  label style={font=\fontsize{7.0}{7.6}\selectfont},
  tick label style={font=\fontsize{7.2}{7.7}\selectfont},
  xmin=1, xmax=8, xtick={1,2,4,6,8},
  ymin=0, ymax=102, ytick={0,50,100},
  tick align=outside, tick pos=left,
  axis line style={black!55,line width=0.35pt},
  xmajorgrids, ymajorgrids, grid style={black!10,line width=0.3pt},
  clip=false,
]
\nextgroupplot[title={Can},ylabel={Unobserved prefix (\%)},ylabel style={at={(yticklabel cs:0.5,2pt)},anchor=near yticklabel},legend columns=1,legend image code/.code={\draw[mark repeat=2,mark phase=2,#1] plot coordinates {(0cm,0cm) (0.21cm,0cm) (0.42cm,0cm)};},legend style={at={(0.98,0.04)},anchor=south east,draw=none,fill=white,fill opacity=0.85,text opacity=1,inner sep=1pt,column sep=0.5pt,row sep=1pt,nodes={inner xsep=0.5pt,inner ysep=0pt},font=\fontsize{6.5}{7}\selectfont}]
\addplot[name path=canVQLow,draw=none,forget plot] coordinates {(1,0.000000) (2,0.251177) (3,22.119309) (4,79.497645) (5,92.825746) (6,98.320251) (7,99.544741) (8,99.858713)};
\addplot[name path=canVQHigh,draw=none,forget plot] coordinates {(1,0.156986) (2,3.792481) (3,76.489903) (4,96.568708) (5,99.491052) (6,99.885076) (7,99.934329) (8,99.950747)};
\addplot[draw=none,fill=vqbase!35,fill opacity=0.24,forget plot] fill between[of=canVQLow and canVQHigh];
\addplot[vqbase,solid,line width=0.85pt,mark=*,mark size=1.3pt,mark options={solid,fill=vqbase,draw=vqbase,line width=0.25pt}] coordinates {(1,0.078493) (2,2.021829) (3,49.304606) (4,88.033177) (5,96.158399) (6,99.102664) (7,99.739535) (8,99.904730)};
\addlegendentry{VQ}
\addplot[name path=canRVQLow,draw=none,forget plot] coordinates {(1,0.000000) (2,1.619695) (3,45.578231) (4,86.038225) (5,97.003563) (6,99.319728) (7,99.708455) (8,99.835958)};
\addplot[name path=canRVQHigh,draw=none,forget plot] coordinates {(1,0.696469) (2,14.256068) (3,86.528918) (4,98.854101) (5,99.873680) (6,99.990977) (7,100.000000) (8,100.000000)};
\addplot[draw=none,fill=rvqbase!35,fill opacity=0.24,forget plot] fill between[of=canRVQLow and canRVQHigh];
\addplot[rvqbase,densely dashed,line width=0.85pt,mark=*,mark size=1.3pt,mark options={solid,fill=rvqbase,draw=rvqbase,line width=0.25pt}] coordinates {(1,0.304531) (2,4.664506) (3,62.096758) (4,92.938910) (5,98.583012) (6,99.662084) (7,99.857580) (8,99.937480)};
\addlegendentry{RVQ}
\addplot[name path=canNRVQLow,draw=none,forget plot] coordinates {(1,0.000000) (2,2.281794) (3,54.463312) (4,86.758062) (5,95.856052) (6,98.706964) (7,99.641689) (8,99.844212)};
\addplot[name path=canNRVQHigh,draw=none,forget plot] coordinates {(1,0.015957) (2,10.819426) (3,88.533658) (4,99.353085) (5,99.962498) (6,99.991348) (7,100.000000) (8,100.000000)};
\addplot[draw=none,fill=nrvqbase!35,fill opacity=0.24,forget plot] fill between[of=canNRVQLow and canNRVQHigh];
\addplot[nrvqbase,densely dotted,line width=0.85pt,mark=*,mark size=1.3pt,mark options={solid,fill=nrvqbase,draw=nrvqbase,line width=0.25pt}] coordinates {(1,0.003191) (2,6.526741) (3,75.622580) (4,95.530028) (5,98.863075) (6,99.684804) (7,99.912381) (8,99.965651)};
\addlegendentry{NRVQ}
\addplot[name path=canFSQLow,draw=none,forget plot] coordinates {(1,0.000000) (2,0.034019) (3,14.679367) (4,77.292056) (5,96.265371) (6,98.937301) (7,99.499013) (8,99.802642)};
\addplot[name path=canFSQHigh,draw=none,forget plot] coordinates {(1,0.030363) (2,0.409898) (3,37.118567) (4,84.211325) (5,96.666100) (6,99.166525) (7,99.710835) (8,99.897942)};
\addplot[draw=none,fill=fsqbase!35,fill opacity=0.24,forget plot] fill between[of=canFSQLow and canFSQHigh];
\addplot[fsqbase,dashdotted,line width=0.85pt,mark=*,mark size=1.3pt,mark options={solid,fill=fsqbase,draw=fsqbase,line width=0.25pt}] coordinates {(1,0.015181) (2,0.221959) (3,25.898967) (4,80.751691) (5,96.465735) (6,99.051913) (7,99.604924) (8,99.850292)};
\addlegendentry{FSQ}
\addplot[name path=canQINCOLow,draw=none,forget plot] coordinates {(1,0.357626) (2,3.653586) (3,73.400348) (4,98.156817) (5,99.561147) (6,99.902477) (7,99.980645) (8,99.980645)};
\addplot[name path=canQINCOHigh,draw=none,forget plot] coordinates {(1,1.002853) (2,9.503532) (3,83.741411) (4,99.074955) (5,99.909959) (6,99.991355) (7,100.000000) (8,100.000000)};
\addplot[draw=none,fill=qincobase!35,fill opacity=0.24,forget plot] fill between[of=canQINCOLow and canQINCOHigh];
\addplot[qincobase,loosely dashdotdotted,line width=0.85pt,mark=*,mark size=1.3pt,mark options={solid,fill=qincobase,draw=qincobase,line width=0.25pt}] coordinates {(1,0.532261) (2,6.754269) (3,79.524751) (4,98.645340) (5,99.823500) (6,99.962839) (7,99.996774) (8,99.996774)};
\addlegendentry{QinCo}
\nextgroupplot[title={Lift},yticklabels={}]
\addplot[name path=liftVQLow,draw=none,forget plot] coordinates {(1,0.000000) (2,2.911836) (3,72.499326) (4,94.203289) (5,98.004853) (6,98.948504) (7,99.595578) (8,99.676463)};
\addplot[name path=liftVQHigh,draw=none,forget plot] coordinates {(1,0.215692) (2,5.808713) (3,81.672509) (4,98.347521) (5,99.649474) (6,99.749624) (7,99.799700) (8,99.899850)};
\addplot[draw=none,fill=vqbase!35,fill opacity=0.24,forget plot] fill between[of=liftVQLow and liftVQHigh];
\addplot[vqbase,solid,line width=0.85pt,mark=*,mark size=1.3pt,mark options={solid,fill=vqbase,draw=vqbase,line width=0.25pt}] coordinates {(1,0.107846) (2,4.360275) (3,77.085917) (4,96.275405) (5,98.827164) (6,99.349064) (7,99.697639) (8,99.788156)};
\addplot[name path=liftRVQLow,draw=none,forget plot] coordinates {(1,0.000000) (2,3.657602) (3,57.679050) (4,91.325163) (5,97.587131) (6,99.310609) (7,99.712754) (8,99.846802)};
\addplot[name path=liftRVQHigh,draw=none,forget plot] coordinates {(1,0.804290) (2,23.896935) (3,89.929269) (4,98.534860) (5,99.814752) (6,99.966319) (7,99.983159) (8,100.000000)};
\addplot[draw=none,fill=rvqbase!35,fill opacity=0.24,forget plot] fill between[of=liftRVQLow and liftRVQHigh];
\addplot[rvqbase,densely dashed,line width=0.85pt,mark=*,mark size=1.3pt,mark options={solid,fill=rvqbase,draw=rvqbase,line width=0.25pt}] coordinates {(1,0.337804) (2,9.592987) (3,76.134522) (4,95.459011) (5,98.798228) (6,99.710114) (7,99.902592) (8,99.957891)};
\addplot[name path=liftNRVQLow,draw=none,forget plot] coordinates {(1,0.000000) (2,3.156385) (3,57.163353) (4,92.439663) (5,98.109916) (6,99.389357) (7,99.825531) (8,99.912765)};
\addplot[name path=liftNRVQHigh,draw=none,forget plot] coordinates {(1,0.145391) (2,18.045113) (3,91.335603) (4,99.022536) (5,99.877817) (6,99.949552) (7,99.962164) (8,99.973305)};
\addplot[draw=none,fill=nrvqbase!35,fill opacity=0.24,forget plot] fill between[of=liftNRVQLow and liftNRVQHigh];
\addplot[nrvqbase,densely dotted,line width=0.85pt,mark=*,mark size=1.3pt,mark options={solid,fill=nrvqbase,draw=nrvqbase,line width=0.25pt}] coordinates {(1,0.052593) (2,10.952925) (3,77.821383) (4,95.709438) (5,99.078129) (6,99.737265) (7,99.884597) (8,99.942204)};
\addplot[name path=liftFSQLow,draw=none,forget plot] coordinates {(1,0.000000) (2,0.063211) (3,36.883692) (4,89.728192) (5,97.724399) (6,98.514539) (7,99.083439) (8,99.640934)};
\addplot[name path=liftFSQHigh,draw=none,forget plot] coordinates {(1,0.031606) (2,0.153886) (3,41.805591) (4,92.639138) (5,98.358553) (6,99.051039) (7,99.435753) (8,99.652339)};
\addplot[draw=none,fill=fsqbase!35,fill opacity=0.24,forget plot] fill between[of=liftFSQLow and liftFSQHigh];
\addplot[fsqbase,dashdotted,line width=0.85pt,mark=*,mark size=1.3pt,mark options={solid,fill=fsqbase,draw=fsqbase,line width=0.25pt}] coordinates {(1,0.015803) (2,0.108548) (3,39.344641) (4,91.183665) (5,98.041476) (6,98.782789) (7,99.259596) (8,99.646636)};
\addplot[name path=liftQINCOLow,draw=none,forget plot] coordinates {(1,0.167252) (2,5.034287) (3,83.270930) (4,96.806993) (5,99.054584) (6,99.732430) (7,99.897013) (8,99.897013)};
\addplot[name path=liftQINCOHigh,draw=none,forget plot] coordinates {(1,0.713521) (2,19.023398) (3,86.048539) (4,98.451260) (5,99.832979) (6,99.969633) (7,99.970934) (8,99.984816)};
\addplot[draw=none,fill=qincobase!35,fill opacity=0.24,forget plot] fill between[of=liftQINCOLow and liftQINCOHigh];
\addplot[qincobase,loosely dashdotdotted,line width=0.85pt,mark=*,mark size=1.3pt,mark options={solid,fill=qincobase,draw=qincobase,line width=0.25pt}] coordinates {(1,0.474688) (2,9.778450) (3,84.430993) (4,97.675960) (5,99.469184) (6,99.855022) (7,99.945558) (8,99.956160)};
\nextgroupplot[title={Square},yticklabels={}]
\addplot[name path=squareVQLow,draw=none,forget plot] coordinates {(1,0.000000) (2,0.474022) (3,36.230133) (4,90.240729) (5,99.061251) (6,99.869876) (7,99.953527) (8,99.972116)};
\addplot[name path=squareVQHigh,draw=none,forget plot] coordinates {(1,0.195185) (2,1.789593) (3,74.573291) (4,98.531085) (5,99.948278) (6,99.968967) (7,99.979311) (8,99.989656)};
\addplot[draw=none,fill=vqbase!35,fill opacity=0.24,forget plot] fill between[of=squareVQLow and squareVQHigh];
\addplot[vqbase,solid,line width=0.85pt,mark=*,mark size=1.3pt,mark options={solid,fill=vqbase,draw=vqbase,line width=0.25pt}] coordinates {(1,0.097593) (2,1.131808) (3,55.401712) (4,94.385907) (5,99.504764) (6,99.919421) (7,99.966419) (8,99.980886)};
\addplot[name path=squareRVQLow,draw=none,forget plot] coordinates {(1,0.019839) (2,1.723559) (3,44.476777) (4,86.916620) (5,97.459429) (6,99.406827) (7,99.820929) (8,99.910723)};
\addplot[name path=squareRVQHigh,draw=none,forget plot] coordinates {(1,0.492445) (2,7.345916) (3,80.617120) (4,98.621662) (5,99.960843) (6,100.000000) (7,100.000000) (8,100.000000)};
\addplot[draw=none,fill=rvqbase!35,fill opacity=0.24,forget plot] fill between[of=squareRVQLow and squareRVQHigh];
\addplot[rvqbase,densely dashed,line width=0.85pt,mark=*,mark size=1.3pt,mark options={solid,fill=rvqbase,draw=rvqbase,line width=0.25pt}] coordinates {(1,0.233684) (2,3.668840) (3,60.801513) (4,93.593716) (5,99.026000) (6,99.793571) (7,99.937368) (8,99.970150)};
\addplot[name path=squareNRVQLow,draw=none,forget plot] coordinates {(1,0.000000) (2,2.141760) (3,52.355503) (4,89.241884) (5,97.486144) (6,99.237926) (7,99.614014) (8,99.831750)};
\addplot[name path=squareNRVQHigh,draw=none,forget plot] coordinates {(1,0.096043) (2,8.560684) (3,91.409822) (4,99.677561) (5,99.971548) (6,99.992672) (7,99.992672) (8,100.000000)};
\addplot[draw=none,fill=nrvqbase!35,fill opacity=0.24,forget plot] fill between[of=squareNRVQLow and squareNRVQHigh];
\addplot[nrvqbase,densely dotted,line width=0.85pt,mark=*,mark size=1.3pt,mark options={solid,fill=nrvqbase,draw=nrvqbase,line width=0.25pt}] coordinates {(1,0.019209) (2,5.723130) (3,76.621231) (4,96.224617) (5,99.241463) (6,99.785564) (7,99.883386) (8,99.943766)};
\addplot[name path=squareFSQLow,draw=none,forget plot] coordinates {(1,0.000000) (2,0.010786) (3,22.467911) (4,83.496926) (5,97.875094) (6,99.644051) (7,99.847700) (8,99.918773)};
\addplot[name path=squareFSQHigh,draw=none,forget plot] coordinates {(1,0.142146) (2,0.406133) (3,30.216266) (4,92.537313) (5,98.872982) (6,99.736014) (7,99.935282) (8,99.967641)};
\addplot[draw=none,fill=fsqbase!35,fill opacity=0.24,forget plot] fill between[of=squareFSQLow and squareFSQHigh];
\addplot[fsqbase,dashdotted,line width=0.85pt,mark=*,mark size=1.3pt,mark options={solid,fill=fsqbase,draw=fsqbase,line width=0.25pt}] coordinates {(1,0.071073) (2,0.208459) (3,26.342088) (4,88.017120) (5,98.374038) (6,99.690033) (7,99.891491) (8,99.943207)};
\addplot[name path=squareQINCOLow,draw=none,forget plot] coordinates {(1,0.000000) (2,1.956637) (3,73.069056) (4,98.640175) (5,99.894236) (6,99.954673) (7,99.969782) (8,99.984891)};
\addplot[name path=squareQINCOHigh,draw=none,forget plot] coordinates {(1,0.672023) (2,9.965684) (3,89.768191) (4,99.668508) (5,99.993094) (6,99.993094) (7,100.000000) (8,100.000000)};
\addplot[draw=none,fill=qincobase!35,fill opacity=0.24,forget plot] fill between[of=squareQINCOLow and squareQINCOHigh];
\addplot[qincobase,loosely dashdotdotted,line width=0.85pt,mark=*,mark size=1.3pt,mark options={solid,fill=qincobase,draw=qincobase,line width=0.25pt}] coordinates {(1,0.406898) (2,5.676736) (3,81.366976) (4,99.253865) (5,99.952352) (6,99.980536) (7,99.990303) (8,99.993988)};
\end{groupplot}
\pgfresetboundingbox
\path[use as bounding box]
  ($(group c1r1.south west)+(-1.60cm,-0.48cm)$) rectangle
  ($(group c3r1.north east)+(0.10cm,0.82cm)$);
\end{tikzpicture}

%% file: figures/what_matters/empirical_support/appendix_rollout_success_delta.tex

\begin{tikzpicture}
\begin{groupplot}[
  group style={group size=3 by 1,horizontal sep=0.42cm},
  width=3.55cm, height=4.6cm, scale only axis,
  title style={font=\fontsize{7.8}{8.5}\selectfont\bfseries,yshift=-1pt},
  label style={font=\fontsize{7.0}{7.6}\selectfont},
  tick label style={font=\fontsize{7.2}{7.7}\selectfont},
  xmin=-100, xmax=20, xtick={-100,-50,0},
  ymin=0.45, ymax=22.55, ytick={1,...,22},
  ytick style={draw=none},
  tick align=outside, tick pos=left,
  axis line style={black!55,line width=0.35pt},
  xmajorgrids, grid style={black!10,line width=0.3pt},
  clip=false,
]
\nextgroupplot[title={Can},yticklabels={\tokenizerid{qincobase!90}{C2-QINCO-L},\tokenizerid{qincobase!80}{C2-QINCO-S},\tokenizerid{qincobase!70}{CT-QINCO-L},\tokenizerid{qincobase!60}{CT-QINCO-S},\tokenizerid{qincobase!50}{TF-QINCO-L},\tokenizerid{qincobase!40}{TF-QINCO-S},\tokenizerid{fsqbase!85}{TF-FSQ-L},\tokenizerid{fsqbase!60}{TF-FSQ-S},\tokenizerid{nrvqbase!90}{C2-NRVQ-L},\tokenizerid{nrvqbase!80}{C2-NRVQ-S},\tokenizerid{nrvqbase!70}{CT-NRVQ-L},\tokenizerid{nrvqbase!60}{CT-NRVQ-S},\tokenizerid{nrvqbase!50}{TF-NRVQ-L},\tokenizerid{nrvqbase!40}{TF-NRVQ-S},\tokenizerid{rvqbase!90}{C2-RVQ-L},\tokenizerid{rvqbase!80}{C2-RVQ-S},\tokenizerid{rvqbase!70}{CT-RVQ-L},\tokenizerid{rvqbase!60}{CT-RVQ-S},\tokenizerid{rvqbase!50}{TF-RVQ-L},\tokenizerid{rvqbase!40}{TF-RVQ-S},\tokenizerid{vqbase!85}{TF-VQ-L},\tokenizerid{vqbase!60}{TF-VQ-S}},yticklabel style={anchor=east,inner sep=1.0pt,font=\fontsize{5.6}{6.0}\selectfont},]
\addplot[black!60,densely dashed,line width=0.45pt,forget plot] coordinates {(0,0.5) (0,22.5)};
\addplot[black!18,line width=0.3pt,forget plot] coordinates {(-100,20.5) (20,20.5)};
\addplot[black!18,line width=0.3pt,forget plot] coordinates {(-100,14.5) (20,14.5)};
\addplot[black!18,line width=0.3pt,forget plot] coordinates {(-100,8.5) (20,8.5)};
\addplot[black!18,line width=0.3pt,forget plot] coordinates {(-100,6.5) (20,6.5)};
\addplot+[only marks,mark=diamond*,mark size=1.65pt,mark options={fill=vqbase!60,draw=vqbase,line width=0.35pt},error bars/.cd,x dir=both,x explicit,error bar style={draw=vqbase,line width=0.35pt},error mark=none] coordinates {(-87.000000,22) += (4.000000,0) -= (3.666667,0)};
\addplot+[only marks,mark=diamond*,mark size=1.65pt,mark options={fill=vqbase!85,draw=vqbase,line width=0.35pt},error bars/.cd,x dir=both,x explicit,error bar style={draw=vqbase,line width=0.35pt},error mark=none] coordinates {(-92.333333,21) += (3.000000,0) -= (3.000000,0)};
\addplot+[only marks,mark=diamond*,mark size=1.65pt,mark options={fill=rvqbase!40,draw=rvqbase,line width=0.35pt},error bars/.cd,x dir=both,x explicit,error bar style={draw=rvqbase,line width=0.35pt},error mark=none] coordinates {(-15.000000,20) += (5.000000,0) -= (5.000000,0)};
\addplot+[only marks,mark=diamond*,mark size=1.65pt,mark options={fill=rvqbase!50,draw=rvqbase,line width=0.35pt},error bars/.cd,x dir=both,x explicit,error bar style={draw=rvqbase,line width=0.35pt},error mark=none] coordinates {(-1.333333,19) += (7.333333,0) -= (7.333333,0)};
\addplot+[only marks,mark=diamond*,mark size=1.65pt,mark options={fill=rvqbase!60,draw=rvqbase,line width=0.35pt},error bars/.cd,x dir=both,x explicit,error bar style={draw=rvqbase,line width=0.35pt},error mark=none] coordinates {(-15.333333,18) += (5.333333,0) -= (5.333333,0)};
\addplot+[only marks,mark=diamond*,mark size=1.65pt,mark options={fill=rvqbase!70,draw=rvqbase,line width=0.35pt},error bars/.cd,x dir=both,x explicit,error bar style={draw=rvqbase,line width=0.35pt},error mark=none] coordinates {(-7.666667,17) += (7.333333,0) -= (7.333333,0)};
\addplot+[only marks,mark=diamond*,mark size=1.65pt,mark options={fill=rvqbase!80,draw=rvqbase,line width=0.35pt},error bars/.cd,x dir=both,x explicit,error bar style={draw=rvqbase,line width=0.35pt},error mark=none] coordinates {(-18.333333,16) += (5.333333,0) -= (5.333333,0)};
\addplot+[only marks,mark=diamond*,mark size=1.65pt,mark options={fill=rvqbase!90,draw=rvqbase,line width=0.35pt},error bars/.cd,x dir=both,x explicit,error bar style={draw=rvqbase,line width=0.35pt},error mark=none] coordinates {(-4.666667,15) += (7.333333,0) -= (7.333333,0)};
\addplot+[only marks,mark=diamond*,mark size=1.65pt,mark options={fill=nrvqbase!40,draw=nrvqbase,line width=0.35pt},error bars/.cd,x dir=both,x explicit,error bar style={draw=nrvqbase,line width=0.35pt},error mark=none] coordinates {(-6.333333,14) += (5.333333,0) -= (5.333333,0)};
\addplot+[only marks,mark=diamond*,mark size=1.65pt,mark options={fill=nrvqbase!50,draw=nrvqbase,line width=0.35pt},error bars/.cd,x dir=both,x explicit,error bar style={draw=nrvqbase,line width=0.35pt},error mark=none] coordinates {(-5.333333,13) += (7.666667,0) -= (7.666667,0)};
\addplot+[only marks,mark=diamond*,mark size=1.65pt,mark options={fill=nrvqbase!60,draw=nrvqbase,line width=0.35pt},error bars/.cd,x dir=both,x explicit,error bar style={draw=nrvqbase,line width=0.35pt},error mark=none] coordinates {(0.000000,12) += (0.000000,0) -= (0.000000,0)};
\addplot+[only marks,mark=diamond*,mark size=1.65pt,mark options={fill=nrvqbase!70,draw=nrvqbase,line width=0.35pt},error bars/.cd,x dir=both,x explicit,error bar style={draw=nrvqbase,line width=0.35pt},error mark=none] coordinates {(-3.666667,11) += (7.333333,0) -= (7.000000,0)};
\addplot+[only marks,mark=diamond*,mark size=1.65pt,mark options={fill=nrvqbase!80,draw=nrvqbase,line width=0.35pt},error bars/.cd,x dir=both,x explicit,error bar style={draw=nrvqbase,line width=0.35pt},error mark=none] coordinates {(-16.666667,10) += (5.333333,0) -= (5.666667,0)};
\addplot+[only marks,mark=diamond*,mark size=1.65pt,mark options={fill=nrvqbase!90,draw=nrvqbase,line width=0.35pt},error bars/.cd,x dir=both,x explicit,error bar style={draw=nrvqbase,line width=0.35pt},error mark=none] coordinates {(-2.333333,9) += (7.666667,0) -= (7.666667,0)};
\addplot+[only marks,mark=diamond*,mark size=1.65pt,mark options={fill=fsqbase!60,draw=fsqbase,line width=0.35pt},error bars/.cd,x dir=both,x explicit,error bar style={draw=fsqbase,line width=0.35pt},error mark=none] coordinates {(-73.333333,8) += (5.333333,0) -= (5.333333,0)};
\addplot+[only marks,mark=diamond*,mark size=1.65pt,mark options={fill=fsqbase!85,draw=fsqbase,line width=0.35pt},error bars/.cd,x dir=both,x explicit,error bar style={draw=fsqbase,line width=0.35pt},error mark=none] coordinates {(-90.333333,7) += (3.666667,0) -= (3.000000,0)};
\addplot+[only marks,mark=diamond*,mark size=1.65pt,mark options={fill=qincobase!40,draw=qincobase,line width=0.35pt},error bars/.cd,x dir=both,x explicit,error bar style={draw=qincobase,line width=0.35pt},error mark=none] coordinates {(-3.666667,6) += (7.675000,0) -= (7.666667,0)};
\addplot+[only marks,mark=diamond*,mark size=1.65pt,mark options={fill=qincobase!50,draw=qincobase,line width=0.35pt},error bars/.cd,x dir=both,x explicit,error bar style={draw=qincobase,line width=0.35pt},error mark=none] coordinates {(2.333333,5) += (7.333333,0) -= (7.333333,0)};
\addplot+[only marks,mark=diamond*,mark size=1.65pt,mark options={fill=qincobase!60,draw=qincobase,line width=0.35pt},error bars/.cd,x dir=both,x explicit,error bar style={draw=qincobase,line width=0.35pt},error mark=none] coordinates {(-0.666667,4) += (7.000000,0) -= (6.666667,0)};
\addplot+[only marks,mark=diamond*,mark size=1.65pt,mark options={fill=qincobase!70,draw=qincobase,line width=0.35pt},error bars/.cd,x dir=both,x explicit,error bar style={draw=qincobase,line width=0.35pt},error mark=none] coordinates {(9.333333,3) += (7.666667,0) -= (7.666667,0)};
\addplot+[only marks,mark=diamond*,mark size=1.65pt,mark options={fill=qincobase!80,draw=qincobase,line width=0.35pt},error bars/.cd,x dir=both,x explicit,error bar style={draw=qincobase,line width=0.35pt},error mark=none] coordinates {(-0.666667,2) += (8.000000,0) -= (7.666667,0)};
\addplot+[only marks,mark=diamond*,mark size=1.65pt,mark options={fill=qincobase!90,draw=qincobase,line width=0.35pt},error bars/.cd,x dir=both,x explicit,error bar style={draw=qincobase,line width=0.35pt},error mark=none] coordinates {(0.333333,1) += (7.333333,0) -= (7.333333,0)};
\nextgroupplot[title={Lift},yticklabels={},]
\addplot[black!60,densely dashed,line width=0.45pt,forget plot] coordinates {(0,0.5) (0,22.5)};
\addplot[black!18,line width=0.3pt,forget plot] coordinates {(-100,20.5) (20,20.5)};
\addplot[black!18,line width=0.3pt,forget plot] coordinates {(-100,14.5) (20,14.5)};
\addplot[black!18,line width=0.3pt,forget plot] coordinates {(-100,8.5) (20,8.5)};
\addplot[black!18,line width=0.3pt,forget plot] coordinates {(-100,6.5) (20,6.5)};
\addplot+[only marks,mark=square*,mark size=1.65pt,mark options={fill=vqbase!60,draw=vqbase,line width=0.35pt},error bars/.cd,x dir=both,x explicit,error bar style={draw=vqbase,line width=0.35pt},error mark=none] coordinates {(-73.000000,22) += (5.333333,0) -= (5.000000,0)};
\addplot+[only marks,mark=square*,mark size=1.65pt,mark options={fill=vqbase!85,draw=vqbase,line width=0.35pt},error bars/.cd,x dir=both,x explicit,error bar style={draw=vqbase,line width=0.35pt},error mark=none] coordinates {(-78.666667,21) += (4.666667,0) -= (4.666667,0)};
\addplot+[only marks,mark=square*,mark size=1.65pt,mark options={fill=rvqbase!40,draw=rvqbase,line width=0.35pt},error bars/.cd,x dir=both,x explicit,error bar style={draw=rvqbase,line width=0.35pt},error mark=none] coordinates {(-4.333333,20) += (4.666667,0) -= (5.000000,0)};
\addplot+[only marks,mark=square*,mark size=1.65pt,mark options={fill=rvqbase!50,draw=rvqbase,line width=0.35pt},error bars/.cd,x dir=both,x explicit,error bar style={draw=rvqbase,line width=0.35pt},error mark=none] coordinates {(-8.666667,19) += (6.000000,0) -= (6.000000,0)};
\addplot+[only marks,mark=square*,mark size=1.65pt,mark options={fill=rvqbase!60,draw=rvqbase,line width=0.35pt},error bars/.cd,x dir=both,x explicit,error bar style={draw=rvqbase,line width=0.35pt},error mark=none] coordinates {(-3.333333,18) += (4.333333,0) -= (4.000000,0)};
\addplot+[only marks,mark=square*,mark size=1.65pt,mark options={fill=rvqbase!70,draw=rvqbase,line width=0.35pt},error bars/.cd,x dir=both,x explicit,error bar style={draw=rvqbase,line width=0.35pt},error mark=none] coordinates {(-5.000000,17) += (5.333333,0) -= (5.333333,0)};
\addplot+[only marks,mark=square*,mark size=1.65pt,mark options={fill=rvqbase!80,draw=rvqbase,line width=0.35pt},error bars/.cd,x dir=both,x explicit,error bar style={draw=rvqbase,line width=0.35pt},error mark=none] coordinates {(-2.000000,16) += (3.666667,0) -= (3.666667,0)};
\addplot+[only marks,mark=square*,mark size=1.65pt,mark options={fill=rvqbase!90,draw=rvqbase,line width=0.35pt},error bars/.cd,x dir=both,x explicit,error bar style={draw=rvqbase,line width=0.35pt},error mark=none] coordinates {(2.000000,15) += (4.666667,0) -= (4.666667,0)};
\addplot+[only marks,mark=square*,mark size=1.65pt,mark options={fill=nrvqbase!40,draw=nrvqbase,line width=0.35pt},error bars/.cd,x dir=both,x explicit,error bar style={draw=nrvqbase,line width=0.35pt},error mark=none] coordinates {(-3.000000,14) += (2.666667,0) -= (2.666667,0)};
\addplot+[only marks,mark=square*,mark size=1.65pt,mark options={fill=nrvqbase!50,draw=nrvqbase,line width=0.35pt},error bars/.cd,x dir=both,x explicit,error bar style={draw=nrvqbase,line width=0.35pt},error mark=none] coordinates {(-3.666667,13) += (6.666667,0) -= (7.000000,0)};
\addplot+[only marks,mark=square*,mark size=1.65pt,mark options={fill=nrvqbase!60,draw=nrvqbase,line width=0.35pt},error bars/.cd,x dir=both,x explicit,error bar style={draw=nrvqbase,line width=0.35pt},error mark=none] coordinates {(0.666667,12) += (4.000000,0) -= (4.333333,0)};
\addplot+[only marks,mark=square*,mark size=1.65pt,mark options={fill=nrvqbase!70,draw=nrvqbase,line width=0.35pt},error bars/.cd,x dir=both,x explicit,error bar style={draw=nrvqbase,line width=0.35pt},error mark=none] coordinates {(-1.666667,11) += (7.333333,0) -= (7.000000,0)};
\addplot+[only marks,mark=square*,mark size=1.65pt,mark options={fill=nrvqbase!80,draw=nrvqbase,line width=0.35pt},error bars/.cd,x dir=both,x explicit,error bar style={draw=nrvqbase,line width=0.35pt},error mark=none] coordinates {(-2.666667,10) += (3.333333,0) -= (3.000000,0)};
\addplot+[only marks,mark=square*,mark size=1.65pt,mark options={fill=nrvqbase!90,draw=nrvqbase,line width=0.35pt},error bars/.cd,x dir=both,x explicit,error bar style={draw=nrvqbase,line width=0.35pt},error mark=none] coordinates {(1.666667,9) += (6.333333,0) -= (6.333333,0)};
\addplot+[only marks,mark=square*,mark size=1.65pt,mark options={fill=fsqbase!60,draw=fsqbase,line width=0.35pt},error bars/.cd,x dir=both,x explicit,error bar style={draw=fsqbase,line width=0.35pt},error mark=none] coordinates {(-70.000000,8) += (5.666667,0) -= (5.666667,0)};
\addplot+[only marks,mark=square*,mark size=1.65pt,mark options={fill=fsqbase!85,draw=fsqbase,line width=0.35pt},error bars/.cd,x dir=both,x explicit,error bar style={draw=fsqbase,line width=0.35pt},error mark=none] coordinates {(-51.666667,7) += (5.666667,0) -= (5.666667,0)};
\addplot+[only marks,mark=square*,mark size=1.65pt,mark options={fill=qincobase!40,draw=qincobase,line width=0.35pt},error bars/.cd,x dir=both,x explicit,error bar style={draw=qincobase,line width=0.35pt},error mark=none] coordinates {(-5.666667,6) += (6.666667,0) -= (6.666667,0)};
\addplot+[only marks,mark=square*,mark size=1.65pt,mark options={fill=qincobase!50,draw=qincobase,line width=0.35pt},error bars/.cd,x dir=both,x explicit,error bar style={draw=qincobase,line width=0.35pt},error mark=none] coordinates {(-2.333333,5) += (6.666667,0) -= (7.000000,0)};
\addplot+[only marks,mark=square*,mark size=1.65pt,mark options={fill=qincobase!60,draw=qincobase,line width=0.35pt},error bars/.cd,x dir=both,x explicit,error bar style={draw=qincobase,line width=0.35pt},error mark=none] coordinates {(1.000000,4) += (5.000000,0) -= (5.000000,0)};
\addplot+[only marks,mark=square*,mark size=1.65pt,mark options={fill=qincobase!70,draw=qincobase,line width=0.35pt},error bars/.cd,x dir=both,x explicit,error bar style={draw=qincobase,line width=0.35pt},error mark=none] coordinates {(-6.000000,3) += (6.000000,0) -= (6.000000,0)};
\addplot+[only marks,mark=square*,mark size=1.65pt,mark options={fill=qincobase!80,draw=qincobase,line width=0.35pt},error bars/.cd,x dir=both,x explicit,error bar style={draw=qincobase,line width=0.35pt},error mark=none] coordinates {(-4.666667,2) += (5.333333,0) -= (5.333333,0)};
\addplot+[only marks,mark=square*,mark size=1.65pt,mark options={fill=qincobase!90,draw=qincobase,line width=0.35pt},error bars/.cd,x dir=both,x explicit,error bar style={draw=qincobase,line width=0.35pt},error mark=none] coordinates {(-2.666667,1) += (5.666667,0) -= (5.666667,0)};
\nextgroupplot[title={Square},yticklabels={},]
\addplot[black!60,densely dashed,line width=0.45pt,forget plot] coordinates {(0,0.5) (0,22.5)};
\addplot[black!18,line width=0.3pt,forget plot] coordinates {(-100,20.5) (20,20.5)};
\addplot[black!18,line width=0.3pt,forget plot] coordinates {(-100,14.5) (20,14.5)};
\addplot[black!18,line width=0.3pt,forget plot] coordinates {(-100,8.5) (20,8.5)};
\addplot[black!18,line width=0.3pt,forget plot] coordinates {(-100,6.5) (20,6.5)};
\addplot+[only marks,mark=triangle*,mark size=1.65pt,mark options={fill=vqbase!60,draw=vqbase,line width=0.35pt},error bars/.cd,x dir=both,x explicit,error bar style={draw=vqbase,line width=0.35pt},error mark=none] coordinates {(-67.333333,22) += (5.333333,0) -= (5.333333,0)};
\addplot+[only marks,mark=triangle*,mark size=1.65pt,mark options={fill=vqbase!85,draw=vqbase,line width=0.35pt},error bars/.cd,x dir=both,x explicit,error bar style={draw=vqbase,line width=0.35pt},error mark=none] coordinates {(-55.666667,21) += (5.666667,0) -= (5.666667,0)};
\addplot+[only marks,mark=triangle*,mark size=1.65pt,mark options={fill=rvqbase!40,draw=rvqbase,line width=0.35pt},error bars/.cd,x dir=both,x explicit,error bar style={draw=rvqbase,line width=0.35pt},error mark=none] coordinates {(-23.666667,20) += (7.333333,0) -= (7.333333,0)};
\addplot+[only marks,mark=triangle*,mark size=1.65pt,mark options={fill=rvqbase!50,draw=rvqbase,line width=0.35pt},error bars/.cd,x dir=both,x explicit,error bar style={draw=rvqbase,line width=0.35pt},error mark=none] coordinates {(-3.333333,19) += (6.666667,0) -= (6.666667,0)};
\addplot+[only marks,mark=triangle*,mark size=1.65pt,mark options={fill=rvqbase!60,draw=rvqbase,line width=0.35pt},error bars/.cd,x dir=both,x explicit,error bar style={draw=rvqbase,line width=0.35pt},error mark=none] coordinates {(-13.666667,18) += (7.333333,0) -= (7.333333,0)};
\addplot+[only marks,mark=triangle*,mark size=1.65pt,mark options={fill=rvqbase!70,draw=rvqbase,line width=0.35pt},error bars/.cd,x dir=both,x explicit,error bar style={draw=rvqbase,line width=0.35pt},error mark=none] coordinates {(-8.000000,17) += (7.333333,0) -= (7.333333,0)};
\addplot+[only marks,mark=triangle*,mark size=1.65pt,mark options={fill=rvqbase!80,draw=rvqbase,line width=0.35pt},error bars/.cd,x dir=both,x explicit,error bar style={draw=rvqbase,line width=0.35pt},error mark=none] coordinates {(-20.000000,16) += (7.333333,0) -= (7.000000,0)};
\addplot+[only marks,mark=triangle*,mark size=1.65pt,mark options={fill=rvqbase!90,draw=rvqbase,line width=0.35pt},error bars/.cd,x dir=both,x explicit,error bar style={draw=rvqbase,line width=0.35pt},error mark=none] coordinates {(-7.333333,15) += (7.333333,0) -= (7.333333,0)};
\addplot+[only marks,mark=triangle*,mark size=1.65pt,mark options={fill=nrvqbase!40,draw=nrvqbase,line width=0.35pt},error bars/.cd,x dir=both,x explicit,error bar style={draw=nrvqbase,line width=0.35pt},error mark=none] coordinates {(-17.000000,14) += (7.666667,0) -= (7.333333,0)};
\addplot+[only marks,mark=triangle*,mark size=1.65pt,mark options={fill=nrvqbase!50,draw=nrvqbase,line width=0.35pt},error bars/.cd,x dir=both,x explicit,error bar style={draw=nrvqbase,line width=0.35pt},error mark=none] coordinates {(1.666667,13) += (6.000000,0) -= (6.333333,0)};
\addplot+[only marks,mark=triangle*,mark size=1.65pt,mark options={fill=nrvqbase!60,draw=nrvqbase,line width=0.35pt},error bars/.cd,x dir=both,x explicit,error bar style={draw=nrvqbase,line width=0.35pt},error mark=none] coordinates {(0.000000,12) += (0.000000,0) -= (0.000000,0)};
\addplot+[only marks,mark=triangle*,mark size=1.65pt,mark options={fill=nrvqbase!70,draw=nrvqbase,line width=0.35pt},error bars/.cd,x dir=both,x explicit,error bar style={draw=nrvqbase,line width=0.35pt},error mark=none] coordinates {(1.333333,11) += (5.000000,0) -= (5.333333,0)};
\addplot+[only marks,mark=triangle*,mark size=1.65pt,mark options={fill=nrvqbase!80,draw=nrvqbase,line width=0.35pt},error bars/.cd,x dir=both,x explicit,error bar style={draw=nrvqbase,line width=0.35pt},error mark=none] coordinates {(-24.000000,10) += (7.333333,0) -= (7.333333,0)};
\addplot+[only marks,mark=triangle*,mark size=1.65pt,mark options={fill=nrvqbase!90,draw=nrvqbase,line width=0.35pt},error bars/.cd,x dir=both,x explicit,error bar style={draw=nrvqbase,line width=0.35pt},error mark=none] coordinates {(-3.333333,9) += (6.333333,0) -= (6.333333,0)};
\addplot+[only marks,mark=triangle*,mark size=1.65pt,mark options={fill=fsqbase!60,draw=fsqbase,line width=0.35pt},error bars/.cd,x dir=both,x explicit,error bar style={draw=fsqbase,line width=0.35pt},error mark=none] coordinates {(-71.666667,8) += (5.000000,0) -= (5.000000,0)};
\addplot+[only marks,mark=triangle*,mark size=1.65pt,mark options={fill=fsqbase!85,draw=fsqbase,line width=0.35pt},error bars/.cd,x dir=both,x explicit,error bar style={draw=fsqbase,line width=0.35pt},error mark=none] coordinates {(-68.000000,7) += (5.333333,0) -= (5.333333,0)};
\addplot+[only marks,mark=triangle*,mark size=1.65pt,mark options={fill=qincobase!40,draw=qincobase,line width=0.35pt},error bars/.cd,x dir=both,x explicit,error bar style={draw=qincobase,line width=0.35pt},error mark=none] coordinates {(-5.666667,6) += (6.000000,0) -= (6.000000,0)};
\addplot+[only marks,mark=triangle*,mark size=1.65pt,mark options={fill=qincobase!50,draw=qincobase,line width=0.35pt},error bars/.cd,x dir=both,x explicit,error bar style={draw=qincobase,line width=0.35pt},error mark=none] coordinates {(5.000000,5) += (4.666667,0) -= (4.666667,0)};
\addplot+[only marks,mark=triangle*,mark size=1.65pt,mark options={fill=qincobase!60,draw=qincobase,line width=0.35pt},error bars/.cd,x dir=both,x explicit,error bar style={draw=qincobase,line width=0.35pt},error mark=none] coordinates {(-1.333333,4) += (7.000000,0) -= (6.675000,0)};
\addplot+[only marks,mark=triangle*,mark size=1.65pt,mark options={fill=qincobase!70,draw=qincobase,line width=0.35pt},error bars/.cd,x dir=both,x explicit,error bar style={draw=qincobase,line width=0.35pt},error mark=none] coordinates {(3.666667,3) += (5.333333,0) -= (5.333333,0)};
\addplot+[only marks,mark=triangle*,mark size=1.65pt,mark options={fill=qincobase!80,draw=qincobase,line width=0.35pt},error bars/.cd,x dir=both,x explicit,error bar style={draw=qincobase,line width=0.35pt},error mark=none] coordinates {(-2.666667,2) += (4.666667,0) -= (4.333333,0)};
\addplot+[only marks,mark=triangle*,mark size=1.65pt,mark options={fill=qincobase!90,draw=qincobase,line width=0.35pt},error bars/.cd,x dir=both,x explicit,error bar style={draw=qincobase,line width=0.35pt},error mark=none] coordinates {(-6.000000,1) += (6.000000,0) -= (6.000000,0)};
\end{groupplot}
\pgfresetboundingbox
\path[use as bounding box]
  ($(group c1r1.south west)+(-1.60cm,-0.48cm)$) rectangle
  ($(group c3r1.north east)+(0.10cm,0.55cm)$);
\end{tikzpicture}

%% file: tables/methods/token_balance_regularization/appendix_hyperparameters.tex
\begin{table}[H]
    \centering
    \caption{Token Balance Regularization hyperparameters used for TBR-enabled tokenizer training in each simulation benchmark. The LIBERO VQ values apply to both global and per-level configurations from the replacement runs.}
    \label{tab:token-balance-hyperparameters}
    \tablebodyfont
    \setlength{\tabcolsep}{4.0pt}
    \renewcommand{\arraystretch}{1.06}
    \begin{tabular}{@{}llccc@{}}
        \toprule
        Hyperparameter & Symbol & RoboMimic & LIBERO & RoboTwin \\
        \midrule
        Damping, VQ & $\lambda$ & $10^{-1}$ & $10^{-1}$ & $10^{-1}$ \\
        Damping, RVQ/NRVQ/FSQ/QinCo & $\lambda$ & $10^{-4}$ & $10^{-4}$ & $10^{-3}$ \\
        Maximum full-covariance dimension & $d_{\max}$ & $512$ & $512$ & $512$ \\
        \bottomrule
    \end{tabular}
\end{table}

%% file: tables/methods/token_perturbation/appendix_hyperparameters.tex
\begin{table}[H]
    \centering
    \caption{Token Perturbation hyperparameters used in each simulation benchmark. The same setting is shared by every tokenizer configuration within a benchmark.}
    \label{tab:token-perturbation-hyperparameters}
    \tablebodyfont
    \setlength{\tabcolsep}{4.0pt}
    \renewcommand{\arraystretch}{1.06}
    \begin{tabular}{@{}llccc@{}}
        \toprule
        Hyperparameter & Symbol & RoboMimic & LIBERO & RoboTwin \\
        \midrule
        Row selection probability & $\beta$ & $0.1$ & $0.1$ & $0.1$ \\
        Mask probability & $\alpha$ & $1.0\rightarrow0.5$ & $1.0\rightarrow0.5$ & $0.5\rightarrow0.125$ \\
        Neighborhood size & $\delta$ & $15\rightarrow8$ & $15\rightarrow8$ & $4\rightarrow1$ \\
        \bottomrule
    \end{tabular}
\end{table}

%% file: tables/methods/token_perturbation/appendix_controlled_pilots.tex
\begin{table}[H]
    \centering
    \caption{Controlled Token Perturbation studies on RoboMimic Can. Success rates and deltas are percentage points, averaged over the tokenizer configurations in each study. $\Delta_{\mathrm{R}}$ compares the alternative with Reconstruction-only, while $\Delta_{\mathrm{TP}}$ compares it with Token Perturbation. W/L/T counts compare the alternative against the corresponding reference configuration by configuration.}
    \label{tab:token-perturbation-controlled}
    \tablebodyfont
    \setlength{\tabcolsep}{3.0pt}
    \renewcommand{\arraystretch}{1.06}
    \begin{tabular}{@{}llrrrrcrc@{}}
        \toprule
        Ckpt. & $H_{\mathrm{exec}}$ & Recon. & TP & Alternative & $\Delta_{\mathrm{R}}$ & W/L/T & $\Delta_{\mathrm{TP}}$ & W/L/T \\
        \midrule
        \multicolumn{9}{@{}l}{\textit{Spectral-metric perturbation (8 configurations)}}\\
        Best & 4 & 77.08 & 82.46 & 74.37 & $-2.71$ & 4/4/0 & $-8.08$ & 3/4/1 \\
        Best & 8 & 68.12 & 78.88 & 70.17 & $+2.04$ & 4/4/0 & $-8.71$ & 2/5/1 \\
        Last & 4 & 75.79 & 82.17 & 73.58 & $-2.21$ & 5/3/0 & $-8.58$ & 4/4/0 \\
        Last & 8 & 68.38 & 76.79 & 69.54 & $+1.17$ & 6/2/0 & $-7.25$ & 3/5/0 \\
        \addlinespace[2pt]
        \multicolumn{9}{@{}l}{\textit{MaxUp $K=8$ (10 Transformer configurations)}}\\
        Best & 4 & 81.13 & 86.03 & 83.30 & $+2.17$ & 6/4/0 & $-2.73$ & 4/6/0 \\
        Best & 8 & 74.70 & 83.07 & 79.17 & $+4.47$ & 7/3/0 & $-3.90$ & 4/6/0 \\
        Last & 4 & 80.53 & 86.27 & 84.00 & $+3.47$ & 6/4/0 & $-2.27$ & 5/4/1 \\
        Last & 8 & 73.23 & 79.80 & 78.43 & $+5.20$ & 7/3/0 & $-1.37$ & 5/5/0 \\
        \bottomrule
    \end{tabular}
\end{table}

%% file: tables/experiments/simulation_results/appendix_validation_reconstruction.tex
\begin{table}[H]
    \centering
    \caption{Multi-task validation reconstruction MAE $\mathcal E_{\mathrm{rec}}$ in raw action units ($\downarrow$). R.O. denotes reconstruction-only training. \textbf{Bold} marks the lower value within each benchmark and tokenizer pair; ties at the displayed precision are both bold.}
    \label{tab:simulation-validation-reconstruction}
    \begingroup
    \tablebodyfont
    \setlength{\tabcolsep}{3pt}
    \renewcommand{\arraystretch}{1.05}
    \begin{tabularx}{\linewidth}{@{}l *{6}{>{\centering\arraybackslash}X}@{}}
        \toprule
        \multirow{2}{*}{Tokenizer} & \multicolumn{2}{c}{RoboMimic} & \multicolumn{2}{c}{LIBERO} & \multicolumn{2}{c}{RoboTwin} \\
        \cmidrule(lr){2-3}\cmidrule(lr){4-5}\cmidrule(l){6-7}
        & R.O. & \methodname{} & R.O. & \methodname{} & R.O. & \methodname{} \\
        \midrule
        \tokenizerid{vqbase!60}{TF-VQ-S} & \textbf{0.02256} & 0.02558 & \textbf{0.01334} & 0.02327 & \textbf{0.00080} & 0.00088 \\
        \tokenizerid{vqbase!85}{TF-VQ-L} & \textbf{0.02329} & 0.04603 & \textbf{0.01567} & 0.03743 & \textbf{0.00077} & 0.00087 \\
        \addlinespace[2pt]
        \tokenizerid{rvqbase!40}{TF-RVQ-S} & \textbf{0.02896} & 0.03099 & \textbf{0.01941} & 0.02477 & \textbf{0.00085} & 0.00110 \\
        \tokenizerid{rvqbase!50}{TF-RVQ-L} & \textbf{0.02420} & 0.02746 & \textbf{0.01552} & 0.02101 & \textbf{0.00071} & 0.00103 \\
        \tokenizerid{rvqbase!60}{CT-RVQ-S} & \textbf{0.02690} & 0.02761 & 0.02052 & \textbf{0.01990} & \textbf{0.00068} & 0.00091 \\
        \tokenizerid{rvqbase!70}{CT-RVQ-L} & \textbf{0.02271} & 0.02560 & \textbf{0.01466} & 0.02155 & \textbf{0.00057} & 0.00075 \\
        \tokenizerid{rvqbase!80}{C2-RVQ-S} & \textbf{0.02657} & 0.02893 & \textbf{0.02025} & 0.02321 & \textbf{0.00090} & 0.00170 \\
        \tokenizerid{rvqbase!90}{C2-RVQ-L} & \textbf{0.02197} & 0.02866 & \textbf{0.01403} & 0.02429 & \textbf{0.00060} & 0.00075 \\
        \addlinespace[2pt]
        \tokenizerid{nrvqbase!40}{TF-NRVQ-S} & 0.03058 & \textbf{0.02984} & \textbf{0.01954} & 0.02385 & \textbf{0.00125} & 0.00157 \\
        \tokenizerid{nrvqbase!50}{TF-NRVQ-L} & \textbf{0.02422} & 0.02585 & \textbf{0.01488} & 0.01870 & \textbf{0.00098} & 0.00102 \\
        \tokenizerid{nrvqbase!60}{CT-NRVQ-S} & 0.26266 & \textbf{0.02977} & 0.28457 & \textbf{0.02093} & 0.06441 & \textbf{0.00136} \\
        \tokenizerid{nrvqbase!70}{CT-NRVQ-L} & 0.26257 & \textbf{0.03199} & 0.28403 & \textbf{0.02634} & 0.06438 & \textbf{0.00102} \\
        \tokenizerid{nrvqbase!80}{C2-NRVQ-S} & 0.03149 & \textbf{0.02789} & \textbf{0.01820} & 0.02096 & \textbf{0.00098} & 0.00111 \\
        \tokenizerid{nrvqbase!90}{C2-NRVQ-L} & \textbf{0.02250} & 0.02873 & \textbf{0.01504} & 0.02218 & \textbf{0.00068} & 0.00100 \\
        \addlinespace[2pt]
        \tokenizerid{fsqbase!60}{TF-FSQ-S} & \textbf{0.02688} & 0.03371 & \textbf{0.01937} & 0.02836 & \textbf{0.00092} & 0.00156 \\
        \tokenizerid{fsqbase!85}{TF-FSQ-L} & \textbf{0.02601} & 0.02891 & \textbf{0.01896} & 0.02381 & \textbf{0.00085} & 0.00143 \\
        \addlinespace[2pt]
        \tokenizerid{qincobase!40}{TF-QINCO-S} & \textbf{0.02475} & 0.03577 & \textbf{0.01555} & 0.03242 & \textbf{0.00081} & 0.00114 \\
        \tokenizerid{qincobase!50}{TF-QINCO-L} & \textbf{0.02281} & 0.03327 & \textbf{0.01395} & 0.02837 & \textbf{0.00061} & 0.00099 \\
        \tokenizerid{qincobase!60}{CT-QINCO-S} & \textbf{0.02306} & 0.03659 & \textbf{0.01430} & 0.03215 & \textbf{0.00066} & 0.00123 \\
        \tokenizerid{qincobase!70}{CT-QINCO-L} & \textbf{0.02129} & 0.03386 & \textbf{0.01402} & 0.02898 & \textbf{0.00059} & 0.00075 \\
        \tokenizerid{qincobase!80}{C2-QINCO-S} & \textbf{0.02150} & 0.03745 & \textbf{0.01455} & 0.03347 & \textbf{0.00077} & 0.00138 \\
        \tokenizerid{qincobase!90}{C2-QINCO-L} & \textbf{0.02177} & 0.03232 & \textbf{0.01408} & 0.03146 & \textbf{0.00049} & 0.00079 \\
        \bottomrule
    \end{tabularx}
    \endgroup
\end{table}

%% file: tables/experiments/simulation_results/appendix_robomimic_multitask_best.tex
\begin{table}[H]
    \centering
    \caption{Task-level multi-task RoboMimic success (\%) at the checkpoint with the best teacher-forced validation accuracy. Each tokenizer occupies two lines for $H_{\mathrm{exec}}=4$ and $8$ success rate $\pm$ 95\% confidence-interval half-width. R.O. denotes Reconstruction Only; \textbf{bold} marks the better objective per task and horizon.}
    \label{tab:robomimic-multitask-by-task}
    \tablebodyfont
    \setlength{\tabcolsep}{1.0pt}
    \renewcommand{\arraystretch}{0.96}
    \begin{tabularx}{\linewidth}{@{}>{\raggedright\arraybackslash}m{0.16\linewidth} *{6}{>{\centering\arraybackslash}X}@{}}
        \toprule
        \multirow{2}{*}{Tokenizer} & \multicolumn{2}{c}{Can} & \multicolumn{2}{c}{Lift} & \multicolumn{2}{c}{Square} \\
        \cmidrule(lr){2-3}\cmidrule(lr){4-5}\cmidrule(l){6-7}
        & R.O. & \methodname{} & R.O. & \methodname{} & R.O. & \methodname{} \\
        \midrule
         & $4$: \mbox{86.7 $\pm$ 3.8} & $4$: \mbox{\textbf{95.3 $\pm$ 2.4}} & $4$: \mbox{82.3 $\pm$ 4.3} & $4$: \mbox{\textbf{99.0 $\pm$ 1.1}} & $4$: \mbox{51.3 $\pm$ 5.7} & $4$: \mbox{\textbf{64.0 $\pm$ 5.5}} \\
        \multirow{-2}{*}{\tokenizerid{vqbase!60}{TF-VQ-S}} & $8$: \mbox{91.0 $\pm$ 3.2} & $8$: \mbox{\textbf{95.3 $\pm$ 2.4}} & $8$: \mbox{85.7 $\pm$ 4.0} & $8$: \mbox{\textbf{100.0 $\pm$ 0.0}} & $8$: \mbox{\textbf{64.0 $\pm$ 5.4}} & $8$: \mbox{63.3 $\pm$ 5.4} \\
        \rowcolor{black!4}  & $4$: \mbox{87.7 $\pm$ 3.7} & $4$: \mbox{\textbf{90.7 $\pm$ 3.3}} & $4$: \mbox{93.7 $\pm$ 2.7} & $4$: \mbox{\textbf{98.7 $\pm$ 1.3}} & $4$: \mbox{\textbf{46.3 $\pm$ 5.7}} & $4$: \mbox{43.3 $\pm$ 5.6} \\
        \rowcolor{black!4} \multirow{-2}{*}{\tokenizerid{vqbase!85}{TF-VQ-L}} & $8$: \mbox{86.0 $\pm$ 3.9} & $8$: \mbox{\textbf{86.3 $\pm$ 3.9}} & $8$: \mbox{87.0 $\pm$ 3.8} & $8$: \mbox{\textbf{98.3 $\pm$ 1.4}} & $8$: \mbox{\textbf{42.7 $\pm$ 5.6}} & $8$: \mbox{33.3 $\pm$ 5.3} \\
        \addlinespace[2pt]
         & $4$: \mbox{86.0 $\pm$ 3.9} & $4$: \mbox{\textbf{92.0 $\pm$ 3.1}} & $4$: \mbox{91.3 $\pm$ 3.2} & $4$: \mbox{\textbf{92.0 $\pm$ 3.1}} & $4$: \mbox{60.3 $\pm$ 5.5} & $4$: \mbox{\textbf{69.0 $\pm$ 5.3}} \\
        \multirow{-2}{*}{\tokenizerid{rvqbase!40}{TF-RVQ-S}} & $8$: \mbox{90.7 $\pm$ 3.3} & $8$: \mbox{\textbf{94.0 $\pm$ 2.7}} & $8$: \mbox{90.3 $\pm$ 3.3} & $8$: \mbox{\textbf{94.3 $\pm$ 2.6}} & $8$: \mbox{54.3 $\pm$ 5.7} & $8$: \mbox{\textbf{70.3 $\pm$ 5.1}} \\
        \rowcolor{black!4}  & $4$: \mbox{73.0 $\pm$ 5.1} & $4$: \mbox{\textbf{86.7 $\pm$ 3.9}} & $4$: \mbox{70.7 $\pm$ 5.2} & $4$: \mbox{\textbf{79.3 $\pm$ 4.6}} & $4$: \mbox{36.0 $\pm$ 5.4} & $4$: \mbox{\textbf{53.3 $\pm$ 5.7}} \\
        \rowcolor{black!4} \multirow{-2}{*}{\tokenizerid{rvqbase!50}{TF-RVQ-L}} & $8$: \mbox{63.7 $\pm$ 5.4} & $8$: \mbox{\textbf{80.7 $\pm$ 4.5}} & $8$: \mbox{77.0 $\pm$ 4.7} & $8$: \mbox{\textbf{81.0 $\pm$ 4.4}} & $8$: \mbox{26.0 $\pm$ 4.9} & $8$: \mbox{\textbf{43.3 $\pm$ 5.6}} \\
         & $4$: \mbox{91.0 $\pm$ 3.3} & $4$: \mbox{\textbf{95.0 $\pm$ 2.5}} & $4$: \mbox{\textbf{92.0 $\pm$ 3.1}} & $4$: \mbox{88.7 $\pm$ 3.6} & $4$: \mbox{\textbf{61.7 $\pm$ 5.5}} & $4$: \mbox{59.0 $\pm$ 5.6} \\
        \multirow{-2}{*}{\tokenizerid{rvqbase!60}{CT-RVQ-S}} & $8$: \mbox{91.0 $\pm$ 3.3} & $8$: \mbox{\textbf{96.0 $\pm$ 2.2}} & $8$: \mbox{92.3 $\pm$ 3.0} & $8$: \mbox{\textbf{95.0 $\pm$ 2.4}} & $8$: \mbox{55.3 $\pm$ 5.6} & $8$: \mbox{\textbf{64.3 $\pm$ 5.5}} \\
        \rowcolor{black!4}  & $4$: \mbox{75.3 $\pm$ 4.8} & $4$: \mbox{\textbf{80.0 $\pm$ 4.5}} & $4$: \mbox{74.7 $\pm$ 4.9} & $4$: \mbox{\textbf{80.0 $\pm$ 4.5}} & $4$: \mbox{36.0 $\pm$ 5.4} & $4$: \mbox{\textbf{40.7 $\pm$ 5.5}} \\
        \rowcolor{black!4} \multirow{-2}{*}{\tokenizerid{rvqbase!70}{CT-RVQ-L}} & $8$: \mbox{69.0 $\pm$ 5.2} & $8$: \mbox{\textbf{77.7 $\pm$ 4.7}} & $8$: \mbox{76.0 $\pm$ 4.9} & $8$: \mbox{\textbf{80.3 $\pm$ 4.5}} & $8$: \mbox{32.7 $\pm$ 5.3} & $8$: \mbox{\textbf{36.3 $\pm$ 5.4}} \\
         & $4$: \mbox{94.0 $\pm$ 2.7} & $4$: \mbox{\textbf{97.7 $\pm$ 1.7}} & $4$: \mbox{\textbf{97.3 $\pm$ 1.8}} & $4$: \mbox{90.7 $\pm$ 3.3} & $4$: \mbox{\textbf{64.0 $\pm$ 5.5}} & $4$: \mbox{57.7 $\pm$ 5.6} \\
        \multirow{-2}{*}{\tokenizerid{rvqbase!80}{C2-RVQ-S}} & $8$: \mbox{92.0 $\pm$ 3.1} & $8$: \mbox{\textbf{94.3 $\pm$ 2.6}} & $8$: \mbox{\textbf{96.0 $\pm$ 2.2}} & $8$: \mbox{93.0 $\pm$ 2.9} & $8$: \mbox{61.7 $\pm$ 5.5} & $8$: \mbox{\textbf{64.0 $\pm$ 5.5}} \\
        \rowcolor{black!4}  & $4$: \mbox{71.7 $\pm$ 5.1} & $4$: \mbox{\textbf{80.0 $\pm$ 4.5}} & $4$: \mbox{70.7 $\pm$ 5.2} & $4$: \mbox{\textbf{73.7 $\pm$ 5.0}} & $4$: \mbox{35.3 $\pm$ 5.4} & $4$: \mbox{\textbf{44.0 $\pm$ 5.6}} \\
        \rowcolor{black!4} \multirow{-2}{*}{\tokenizerid{rvqbase!90}{C2-RVQ-L}} & $8$: \mbox{66.0 $\pm$ 5.3} & $8$: \mbox{\textbf{74.0 $\pm$ 5.0}} & $8$: \mbox{\textbf{75.3 $\pm$ 4.9}} & $8$: \mbox{74.7 $\pm$ 4.9} & $8$: \mbox{32.3 $\pm$ 5.3} & $8$: \mbox{\textbf{35.7 $\pm$ 5.5}} \\
        \addlinespace[2pt]
         & $4$: \mbox{89.3 $\pm$ 3.5} & $4$: \mbox{\textbf{94.3 $\pm$ 2.6}} & $4$: \mbox{\textbf{89.0 $\pm$ 3.5}} & $4$: \mbox{88.7 $\pm$ 3.6} & $4$: \mbox{61.7 $\pm$ 5.5} & $4$: \mbox{\textbf{70.0 $\pm$ 5.2}} \\
        \multirow{-2}{*}{\tokenizerid{nrvqbase!40}{TF-NRVQ-S}} & $8$: \mbox{90.0 $\pm$ 3.4} & $8$: \mbox{\textbf{95.7 $\pm$ 2.3}} & $8$: \mbox{\textbf{88.3 $\pm$ 3.6}} & $8$: \mbox{88.0 $\pm$ 3.6} & $8$: \mbox{63.7 $\pm$ 5.5} & $8$: \mbox{\textbf{69.3 $\pm$ 5.2}} \\
        \rowcolor{black!4}  & $4$: \mbox{70.3 $\pm$ 5.1} & $4$: \mbox{\textbf{73.0 $\pm$ 5.0}} & $4$: \mbox{65.3 $\pm$ 5.3} & $4$: \mbox{\textbf{71.7 $\pm$ 5.1}} & $4$: \mbox{34.0 $\pm$ 5.3} & $4$: \mbox{\textbf{37.0 $\pm$ 5.5}} \\
        \rowcolor{black!4} \multirow{-2}{*}{\tokenizerid{nrvqbase!50}{TF-NRVQ-L}} & $8$: \mbox{60.0 $\pm$ 5.5} & $8$: \mbox{\textbf{65.3 $\pm$ 5.4}} & $8$: \mbox{64.0 $\pm$ 5.4} & $8$: \mbox{\textbf{75.0 $\pm$ 4.9}} & $8$: \mbox{21.0 $\pm$ 4.6} & $8$: \mbox{\textbf{31.0 $\pm$ 5.2}} \\
         & $4$: \mbox{0.0 $\pm$ 0.0} & $4$: \mbox{\textbf{91.0 $\pm$ 3.2}} & $4$: \mbox{0.0 $\pm$ 0.0} & $4$: \mbox{\textbf{88.3 $\pm$ 3.6}} & $4$: \mbox{0.0 $\pm$ 0.0} & $4$: \mbox{\textbf{64.0 $\pm$ 5.5}} \\
        \multirow{-2}{*}{\tokenizerid{nrvqbase!60}{CT-NRVQ-S}} & $8$: \mbox{0.0 $\pm$ 0.0} & $8$: \mbox{\textbf{92.3 $\pm$ 3.0}} & $8$: \mbox{0.0 $\pm$ 0.0} & $8$: \mbox{\textbf{96.7 $\pm$ 2.0}} & $8$: \mbox{0.0 $\pm$ 0.0} & $8$: \mbox{\textbf{70.7 $\pm$ 5.1}} \\
        \rowcolor{black!4}  & $4$: \mbox{0.0 $\pm$ 0.0} & $4$: \mbox{\textbf{73.0 $\pm$ 5.0}} & $4$: \mbox{0.0 $\pm$ 0.0} & $4$: \mbox{\textbf{79.3 $\pm$ 4.6}} & $4$: \mbox{0.0 $\pm$ 0.0} & $4$: \mbox{\textbf{32.7 $\pm$ 5.3}} \\
        \rowcolor{black!4} \multirow{-2}{*}{\tokenizerid{nrvqbase!70}{CT-NRVQ-L}} & $8$: \mbox{0.0 $\pm$ 0.0} & $8$: \mbox{\textbf{59.0 $\pm$ 5.5}} & $8$: \mbox{0.0 $\pm$ 0.0} & $8$: \mbox{\textbf{79.0 $\pm$ 4.6}} & $8$: \mbox{0.0 $\pm$ 0.0} & $8$: \mbox{\textbf{24.0 $\pm$ 4.8}} \\
         & $4$: \mbox{\textbf{91.7 $\pm$ 3.1}} & $4$: \mbox{89.3 $\pm$ 3.5} & $4$: \mbox{\textbf{90.0 $\pm$ 3.4}} & $4$: \mbox{85.0 $\pm$ 4.0} & $4$: \mbox{60.7 $\pm$ 5.5} & $4$: \mbox{\textbf{61.0 $\pm$ 5.6}} \\
        \multirow{-2}{*}{\tokenizerid{nrvqbase!80}{C2-NRVQ-S}} & $8$: \mbox{92.3 $\pm$ 3.0} & $8$: \mbox{\textbf{93.3 $\pm$ 2.8}} & $8$: \mbox{\textbf{95.3 $\pm$ 2.4}} & $8$: \mbox{87.7 $\pm$ 3.7} & $8$: \mbox{61.0 $\pm$ 5.5} & $8$: \mbox{\textbf{67.7 $\pm$ 5.3}} \\
        \rowcolor{black!4}  & $4$: \mbox{68.7 $\pm$ 5.3} & $4$: \mbox{\textbf{76.7 $\pm$ 4.8}} & $4$: \mbox{71.7 $\pm$ 5.1} & $4$: \mbox{\textbf{72.7 $\pm$ 5.1}} & $4$: \mbox{28.7 $\pm$ 5.1} & $4$: \mbox{\textbf{35.0 $\pm$ 5.4}} \\
        \rowcolor{black!4} \multirow{-2}{*}{\tokenizerid{nrvqbase!90}{C2-NRVQ-L}} & $8$: \mbox{54.0 $\pm$ 5.7} & $8$: \mbox{\textbf{62.3 $\pm$ 5.5}} & $8$: \mbox{68.7 $\pm$ 5.2} & $8$: \mbox{\textbf{69.0 $\pm$ 5.2}} & $8$: \mbox{21.3 $\pm$ 4.6} & $8$: \mbox{\textbf{24.7 $\pm$ 4.9}} \\
        \addlinespace[2pt]
         & $4$: \mbox{\textbf{91.0 $\pm$ 3.2}} & $4$: \mbox{89.0 $\pm$ 3.5} & $4$: \mbox{83.3 $\pm$ 4.2} & $4$: \mbox{\textbf{91.7 $\pm$ 3.1}} & $4$: \mbox{49.7 $\pm$ 5.6} & $4$: \mbox{\textbf{60.0 $\pm$ 5.5}} \\
        \multirow{-2}{*}{\tokenizerid{fsqbase!60}{TF-FSQ-S}} & $8$: \mbox{89.7 $\pm$ 3.5} & $8$: \mbox{\textbf{92.3 $\pm$ 3.0}} & $8$: \mbox{85.0 $\pm$ 4.0} & $8$: \mbox{\textbf{93.0 $\pm$ 2.9}} & $8$: \mbox{49.7 $\pm$ 5.6} & $8$: \mbox{\textbf{57.0 $\pm$ 5.6}} \\
        \rowcolor{black!4}  & $4$: \mbox{91.0 $\pm$ 3.2} & $4$: \mbox{\textbf{94.0 $\pm$ 2.7}} & $4$: \mbox{88.7 $\pm$ 3.6} & $4$: \mbox{\textbf{91.0 $\pm$ 3.3}} & $4$: \mbox{53.3 $\pm$ 5.7} & $4$: \mbox{\textbf{59.3 $\pm$ 5.6}} \\
        \rowcolor{black!4} \multirow{-2}{*}{\tokenizerid{fsqbase!85}{TF-FSQ-L}} & $8$: \mbox{87.3 $\pm$ 3.8} & $8$: \mbox{\textbf{92.7 $\pm$ 3.0}} & $8$: \mbox{84.7 $\pm$ 4.1} & $8$: \mbox{\textbf{94.0 $\pm$ 2.7}} & $8$: \mbox{50.0 $\pm$ 5.7} & $8$: \mbox{\textbf{63.7 $\pm$ 5.4}} \\
        \addlinespace[2pt]
         & $4$: \mbox{55.3 $\pm$ 5.7} & $4$: \mbox{\textbf{89.7 $\pm$ 3.4}} & $4$: \mbox{62.3 $\pm$ 5.5} & $4$: \mbox{\textbf{83.7 $\pm$ 4.2}} & $4$: \mbox{27.7 $\pm$ 5.0} & $4$: \mbox{\textbf{54.0 $\pm$ 5.6}} \\
        \multirow{-2}{*}{\tokenizerid{qincobase!40}{TF-QINCO-S}} & $8$: \mbox{47.3 $\pm$ 5.6} & $8$: \mbox{\textbf{86.3 $\pm$ 3.9}} & $8$: \mbox{61.3 $\pm$ 5.5} & $8$: \mbox{\textbf{84.0 $\pm$ 4.2}} & $8$: \mbox{12.7 $\pm$ 3.8} & $8$: \mbox{\textbf{51.0 $\pm$ 5.7}} \\
        \rowcolor{black!4}  & $4$: \mbox{51.0 $\pm$ 5.6} & $4$: \mbox{\textbf{86.3 $\pm$ 3.9}} & $4$: \mbox{64.7 $\pm$ 5.5} & $4$: \mbox{\textbf{82.7 $\pm$ 4.3}} & $4$: \mbox{19.0 $\pm$ 4.4} & $4$: \mbox{\textbf{53.0 $\pm$ 5.6}} \\
        \rowcolor{black!4} \multirow{-2}{*}{\tokenizerid{qincobase!50}{TF-QINCO-L}} & $8$: \mbox{32.7 $\pm$ 5.4} & $8$: \mbox{\textbf{85.7 $\pm$ 4.0}} & $8$: \mbox{56.3 $\pm$ 5.6} & $8$: \mbox{\textbf{82.0 $\pm$ 4.3}} & $8$: \mbox{8.7 $\pm$ 3.1} & $8$: \mbox{\textbf{48.7 $\pm$ 5.6}} \\
         & $4$: \mbox{67.0 $\pm$ 5.3} & $4$: \mbox{\textbf{86.0 $\pm$ 4.0}} & $4$: \mbox{74.0 $\pm$ 4.9} & $4$: \mbox{\textbf{83.7 $\pm$ 4.2}} & $4$: \mbox{34.0 $\pm$ 5.5} & $4$: \mbox{\textbf{43.0 $\pm$ 5.7}} \\
        \multirow{-2}{*}{\tokenizerid{qincobase!60}{CT-QINCO-S}} & $8$: \mbox{56.3 $\pm$ 5.7} & $8$: \mbox{\textbf{77.3 $\pm$ 4.7}} & $8$: \mbox{65.3 $\pm$ 5.5} & $8$: \mbox{\textbf{77.0 $\pm$ 4.8}} & $8$: \mbox{18.7 $\pm$ 4.4} & $8$: \mbox{\textbf{36.7 $\pm$ 5.5}} \\
        \rowcolor{black!4}  & $4$: \mbox{59.0 $\pm$ 5.6} & $4$: \mbox{\textbf{79.7 $\pm$ 4.6}} & $4$: \mbox{73.7 $\pm$ 5.0} & $4$: \mbox{\textbf{85.7 $\pm$ 4.0}} & $4$: \mbox{23.7 $\pm$ 4.8} & $4$: \mbox{\textbf{48.3 $\pm$ 5.7}} \\
        \rowcolor{black!4} \multirow{-2}{*}{\tokenizerid{qincobase!70}{CT-QINCO-L}} & $8$: \mbox{45.0 $\pm$ 5.6} & $8$: \mbox{\textbf{74.7 $\pm$ 4.9}} & $8$: \mbox{66.3 $\pm$ 5.3} & $8$: \mbox{\textbf{76.7 $\pm$ 4.8}} & $8$: \mbox{17.3 $\pm$ 4.3} & $8$: \mbox{\textbf{41.0 $\pm$ 5.7}} \\
         & $4$: \mbox{59.7 $\pm$ 5.5} & $4$: \mbox{\textbf{91.3 $\pm$ 3.2}} & $4$: \mbox{70.0 $\pm$ 5.2} & $4$: \mbox{\textbf{79.3 $\pm$ 4.5}} & $4$: \mbox{21.3 $\pm$ 4.7} & $4$: \mbox{\textbf{50.3 $\pm$ 5.6}} \\
        \multirow{-2}{*}{\tokenizerid{qincobase!80}{C2-QINCO-S}} & $8$: \mbox{50.0 $\pm$ 5.7} & $8$: \mbox{\textbf{87.7 $\pm$ 3.8}} & $8$: \mbox{64.3 $\pm$ 5.4} & $8$: \mbox{\textbf{78.0 $\pm$ 4.7}} & $8$: \mbox{11.3 $\pm$ 3.6} & $8$: \mbox{\textbf{48.0 $\pm$ 5.7}} \\
        \rowcolor{black!4}  & $4$: \mbox{61.7 $\pm$ 5.5} & $4$: \mbox{\textbf{73.3 $\pm$ 4.9}} & $4$: \mbox{72.3 $\pm$ 5.1} & $4$: \mbox{\textbf{80.0 $\pm$ 4.5}} & $4$: \mbox{17.7 $\pm$ 4.3} & $4$: \mbox{\textbf{29.0 $\pm$ 5.1}} \\
        \rowcolor{black!4} \multirow{-2}{*}{\tokenizerid{qincobase!90}{C2-QINCO-L}} & $8$: \mbox{47.0 $\pm$ 5.6} & $8$: \mbox{\textbf{61.7 $\pm$ 5.5}} & $8$: \mbox{63.3 $\pm$ 5.5} & $8$: \mbox{\textbf{74.0 $\pm$ 5.0}} & $8$: \mbox{9.7 $\pm$ 3.3} & $8$: \mbox{\textbf{21.3 $\pm$ 4.7}} \\
        \bottomrule
    \end{tabularx}
\end{table}

%% file: tables/experiments/simulation_results/appendix_libero_multitask_best.tex
\begin{table}[H]
    \centering
    \caption{Suite-level multi-task LIBERO success (\%) at the checkpoint with the best teacher-forced validation accuracy. Each tokenizer occupies two lines for $H_{\mathrm{exec}}=4$ and $8$ success rate $\pm$ the supplied 95\% confidence-interval half-width. Each suite contains 10 tasks and is evaluated for 500 episodes using one fixed evaluation seed. R.O. denotes Reconstruction Only; \textbf{bold} marks the better objective per suite and horizon.}
    \label{tab:libero-multitask-by-suite}
    \tablebodyfont
    \setlength{\tabcolsep}{0.55pt}
    \renewcommand{\arraystretch}{0.94}
    \begin{tabularx}{\linewidth}{@{}>{\raggedright\arraybackslash}m{0.145\linewidth} *{8}{>{\centering\arraybackslash}X}@{}}
        \toprule
        \multirow{2}{*}{Tokenizer} & \multicolumn{2}{c}{Spatial} & \multicolumn{2}{c}{Object} & \multicolumn{2}{c}{Goal} & \multicolumn{2}{c}{LIBERO-10} \\
        \cmidrule(lr){2-3}\cmidrule(lr){4-5}\cmidrule(lr){6-7}\cmidrule(l){8-9}
        & R.O. & \methodname{} & R.O. & \methodname{} & R.O. & \methodname{} & R.O. & \methodname{} \\
        \midrule
         & $4$: \mbox{54.6 $\pm$ 4.4} & $4$: \mbox{\textbf{61.6 $\pm$ 4.3}} & $4$: \mbox{71.8 $\pm$ 4.0} & $4$: \mbox{\textbf{74.0 $\pm$ 3.8}} & $4$: \mbox{28.2 $\pm$ 3.9} & $4$: \mbox{\textbf{38.2 $\pm$ 4.2}} & $4$: \mbox{47.8 $\pm$ 4.4} & $4$: \mbox{\textbf{52.8 $\pm$ 4.4}} \\
        \multirow{-2}{*}{\tokenizerid{vqbase!60}{TF-VQ-S}} & $8$: \mbox{42.6 $\pm$ 4.3} & $8$: \mbox{\textbf{52.4 $\pm$ 4.4}} & $8$: \mbox{67.0 $\pm$ 4.1} & $8$: \mbox{\textbf{71.0 $\pm$ 4.0}} & $8$: \mbox{23.0 $\pm$ 3.7} & $8$: \mbox{\textbf{40.6 $\pm$ 4.3}} & $8$: \mbox{\textbf{49.0 $\pm$ 4.4}} & $8$: \mbox{48.6 $\pm$ 4.3} \\
        \rowcolor{black!4}  & $4$: \mbox{57.6 $\pm$ 4.3} & $4$: \mbox{\textbf{59.4 $\pm$ 4.3}} & $4$: \mbox{71.2 $\pm$ 4.0} & $4$: \mbox{\textbf{83.8 $\pm$ 3.2}} & $4$: \mbox{29.0 $\pm$ 4.0} & $4$: \mbox{\textbf{45.8 $\pm$ 4.4}} & $4$: \mbox{42.2 $\pm$ 4.3} & $4$: \mbox{\textbf{45.4 $\pm$ 4.4}} \\
        \rowcolor{black!4} \multirow{-2}{*}{\tokenizerid{vqbase!85}{TF-VQ-L}} & $8$: \mbox{47.0 $\pm$ 4.4} & $8$: \mbox{\textbf{50.2 $\pm$ 4.4}} & $8$: \mbox{\textbf{76.4 $\pm$ 3.7}} & $8$: \mbox{76.0 $\pm$ 3.7} & $8$: \mbox{28.4 $\pm$ 4.0} & $8$: \mbox{\textbf{39.2 $\pm$ 4.3}} & $8$: \mbox{\textbf{45.6 $\pm$ 4.4}} & $8$: \mbox{45.4 $\pm$ 4.4} \\
        \addlinespace[1.5pt]
         & $4$: \mbox{67.6 $\pm$ 4.1} & $4$: \mbox{\textbf{71.6 $\pm$ 4.0}} & $4$: \mbox{80.0 $\pm$ 3.5} & $4$: \mbox{\textbf{85.8 $\pm$ 3.1}} & $4$: \mbox{65.2 $\pm$ 4.2} & $4$: \mbox{\textbf{68.2 $\pm$ 4.1}} & $4$: \mbox{59.6 $\pm$ 4.3} & $4$: \mbox{\textbf{63.0 $\pm$ 4.2}} \\
        \multirow{-2}{*}{\tokenizerid{rvqbase!40}{TF-RVQ-S}} & $8$: \mbox{59.0 $\pm$ 4.3} & $8$: \mbox{\textbf{61.0 $\pm$ 4.3}} & $8$: \mbox{76.4 $\pm$ 3.7} & $8$: \mbox{\textbf{85.6 $\pm$ 3.1}} & $8$: \mbox{61.2 $\pm$ 4.3} & $8$: \mbox{\textbf{68.2 $\pm$ 4.1}} & $8$: \mbox{61.2 $\pm$ 4.3} & $8$: \mbox{\textbf{63.4 $\pm$ 4.2}} \\
        \rowcolor{black!4}  & $4$: \mbox{60.0 $\pm$ 4.3} & $4$: \mbox{\textbf{66.8 $\pm$ 4.1}} & $4$: \mbox{75.8 $\pm$ 3.7} & $4$: \mbox{\textbf{85.0 $\pm$ 3.1}} & $4$: \mbox{51.4 $\pm$ 4.4} & $4$: \mbox{\textbf{63.0 $\pm$ 4.2}} & $4$: \mbox{42.8 $\pm$ 4.4} & $4$: \mbox{\textbf{56.6 $\pm$ 4.3}} \\
        \rowcolor{black!4} \multirow{-2}{*}{\tokenizerid{rvqbase!50}{TF-RVQ-L}} & $8$: \mbox{52.0 $\pm$ 4.4} & $8$: \mbox{\textbf{53.0 $\pm$ 4.4}} & $8$: \mbox{71.2 $\pm$ 4.0} & $8$: \mbox{\textbf{78.0 $\pm$ 3.6}} & $8$: \mbox{46.8 $\pm$ 4.4} & $8$: \mbox{\textbf{55.0 $\pm$ 4.3}} & $8$: \mbox{37.4 $\pm$ 4.2} & $8$: \mbox{\textbf{54.6 $\pm$ 4.4}} \\
         & $4$: \mbox{48.8 $\pm$ 4.4} & $4$: \mbox{\textbf{73.0 $\pm$ 3.9}} & $4$: \mbox{69.4 $\pm$ 4.1} & $4$: \mbox{\textbf{82.8 $\pm$ 3.3}} & $4$: \mbox{27.0 $\pm$ 3.9} & $4$: \mbox{\textbf{64.8 $\pm$ 4.2}} & $4$: \mbox{10.2 $\pm$ 2.7} & $4$: \mbox{\textbf{60.2 $\pm$ 4.3}} \\
        \multirow{-2}{*}{\tokenizerid{rvqbase!60}{CT-RVQ-S}} & $8$: \mbox{53.8 $\pm$ 4.4} & $8$: \mbox{\textbf{67.2 $\pm$ 4.1}} & $8$: \mbox{\textbf{81.8 $\pm$ 3.4}} & $8$: \mbox{80.2 $\pm$ 3.5} & $8$: \mbox{25.0 $\pm$ 3.8} & $8$: \mbox{\textbf{63.8 $\pm$ 4.2}} & $8$: \mbox{24.8 $\pm$ 3.8} & $8$: \mbox{\textbf{61.4 $\pm$ 4.3}} \\
        \rowcolor{black!4}  & $4$: \mbox{47.8 $\pm$ 4.3} & $4$: \mbox{\textbf{67.2 $\pm$ 4.1}} & $4$: \mbox{74.4 $\pm$ 3.8} & $4$: \mbox{\textbf{83.6 $\pm$ 3.3}} & $4$: \mbox{38.2 $\pm$ 4.2} & $4$: \mbox{\textbf{57.8 $\pm$ 4.3}} & $4$: \mbox{33.8 $\pm$ 4.1} & $4$: \mbox{\textbf{54.4 $\pm$ 4.4}} \\
        \rowcolor{black!4} \multirow{-2}{*}{\tokenizerid{rvqbase!70}{CT-RVQ-L}} & $8$: \mbox{44.8 $\pm$ 4.3} & $8$: \mbox{\textbf{58.0 $\pm$ 4.3}} & $8$: \mbox{66.0 $\pm$ 4.1} & $8$: \mbox{\textbf{77.2 $\pm$ 3.7}} & $8$: \mbox{38.4 $\pm$ 4.3} & $8$: \mbox{\textbf{53.4 $\pm$ 4.4}} & $8$: \mbox{35.2 $\pm$ 4.2} & $8$: \mbox{\textbf{47.2 $\pm$ 4.4}} \\
         & $4$: \mbox{75.0 $\pm$ 3.8} & $4$: \mbox{\textbf{75.2 $\pm$ 3.8}} & $4$: \mbox{81.8 $\pm$ 3.4} & $4$: \mbox{\textbf{83.0 $\pm$ 3.3}} & $4$: \mbox{59.8 $\pm$ 4.3} & $4$: \mbox{\textbf{68.6 $\pm$ 4.1}} & $4$: \mbox{\textbf{67.0 $\pm$ 4.1}} & $4$: \mbox{53.6 $\pm$ 4.3} \\
        \multirow{-2}{*}{\tokenizerid{rvqbase!80}{C2-RVQ-S}} & $8$: \mbox{\textbf{63.4 $\pm$ 4.2}} & $8$: \mbox{63.0 $\pm$ 4.2} & $8$: \mbox{84.0 $\pm$ 3.2} & $8$: \mbox{\textbf{84.6 $\pm$ 3.2}} & $8$: \mbox{64.4 $\pm$ 4.2} & $8$: \mbox{\textbf{67.6 $\pm$ 4.1}} & $8$: \mbox{\textbf{65.4 $\pm$ 4.1}} & $8$: \mbox{56.8 $\pm$ 4.4} \\
        \rowcolor{black!4}  & $4$: \mbox{55.6 $\pm$ 4.4} & $4$: \mbox{\textbf{63.0 $\pm$ 4.2}} & $4$: \mbox{74.6 $\pm$ 3.8} & $4$: \mbox{\textbf{83.4 $\pm$ 3.2}} & $4$: \mbox{50.4 $\pm$ 4.3} & $4$: \mbox{\textbf{58.4 $\pm$ 4.4}} & $4$: \mbox{36.8 $\pm$ 4.2} & $4$: \mbox{\textbf{54.0 $\pm$ 4.4}} \\
        \rowcolor{black!4} \multirow{-2}{*}{\tokenizerid{rvqbase!90}{C2-RVQ-L}} & $8$: \mbox{46.8 $\pm$ 4.3} & $8$: \mbox{\textbf{50.2 $\pm$ 4.3}} & $8$: \mbox{73.2 $\pm$ 3.9} & $8$: \mbox{\textbf{80.4 $\pm$ 3.5}} & $8$: \mbox{46.0 $\pm$ 4.4} & $8$: \mbox{\textbf{55.6 $\pm$ 4.3}} & $8$: \mbox{34.2 $\pm$ 4.2} & $8$: \mbox{\textbf{47.2 $\pm$ 4.4}} \\
        \addlinespace[1.5pt]
         & $4$: \mbox{44.4 $\pm$ 4.3} & $4$: \mbox{\textbf{54.0 $\pm$ 4.3}} & $4$: \mbox{65.6 $\pm$ 4.2} & $4$: \mbox{\textbf{73.4 $\pm$ 3.9}} & $4$: \mbox{28.8 $\pm$ 4.0} & $4$: \mbox{\textbf{31.2 $\pm$ 4.0}} & $4$: \mbox{6.0 $\pm$ 2.1} & $4$: \mbox{\textbf{7.2 $\pm$ 2.3}} \\
        \multirow{-2}{*}{\tokenizerid{nrvqbase!40}{TF-NRVQ-S}} & $8$: \mbox{37.0 $\pm$ 4.3} & $8$: \mbox{\textbf{57.2 $\pm$ 4.4}} & $8$: \mbox{69.8 $\pm$ 4.1} & $8$: \mbox{\textbf{81.0 $\pm$ 3.4}} & $8$: \mbox{30.8 $\pm$ 4.1} & $8$: \mbox{\textbf{32.8 $\pm$ 4.1}} & $8$: \mbox{17.2 $\pm$ 3.3} & $8$: \mbox{\textbf{19.0 $\pm$ 3.4}} \\
        \rowcolor{black!4}  & $4$: \mbox{61.6 $\pm$ 4.3} & $4$: \mbox{\textbf{67.2 $\pm$ 4.1}} & $4$: \mbox{71.0 $\pm$ 4.0} & $4$: \mbox{\textbf{80.6 $\pm$ 3.5}} & $4$: \mbox{50.8 $\pm$ 4.4} & $4$: \mbox{\textbf{54.2 $\pm$ 4.3}} & $4$: \mbox{37.6 $\pm$ 4.2} & $4$: \mbox{\textbf{44.8 $\pm$ 4.3}} \\
        \rowcolor{black!4} \multirow{-2}{*}{\tokenizerid{nrvqbase!50}{TF-NRVQ-L}} & $8$: \mbox{\textbf{55.6 $\pm$ 4.4}} & $8$: \mbox{53.4 $\pm$ 4.4} & $8$: \mbox{67.4 $\pm$ 4.1} & $8$: \mbox{\textbf{74.4 $\pm$ 3.8}} & $8$: \mbox{45.2 $\pm$ 4.4} & $8$: \mbox{\textbf{47.0 $\pm$ 4.4}} & $8$: \mbox{28.0 $\pm$ 4.0} & $8$: \mbox{\textbf{38.8 $\pm$ 4.3}} \\
         & $4$: \mbox{0.0 $\pm$ 0.0} & $4$: \mbox{\textbf{64.6 $\pm$ 4.1}} & $4$: \mbox{0.0 $\pm$ 0.0} & $4$: \mbox{\textbf{81.4 $\pm$ 3.4}} & $4$: \mbox{0.0 $\pm$ 0.0} & $4$: \mbox{\textbf{46.8 $\pm$ 4.4}} & $4$: \mbox{0.0 $\pm$ 0.0} & $4$: \mbox{\textbf{34.0 $\pm$ 4.1}} \\
        \multirow{-2}{*}{\tokenizerid{nrvqbase!60}{CT-NRVQ-S}} & $8$: \mbox{0.0 $\pm$ 0.0} & $8$: \mbox{\textbf{68.8 $\pm$ 4.1}} & $8$: \mbox{0.0 $\pm$ 0.0} & $8$: \mbox{\textbf{86.4 $\pm$ 3.0}} & $8$: \mbox{0.0 $\pm$ 0.0} & $8$: \mbox{\textbf{54.8 $\pm$ 4.4}} & $8$: \mbox{0.0 $\pm$ 0.0} & $8$: \mbox{\textbf{44.8 $\pm$ 4.4}} \\
        \rowcolor{black!4}  & $4$: \mbox{0.0 $\pm$ 0.0} & $4$: \mbox{\textbf{58.8 $\pm$ 4.3}} & $4$: \mbox{0.0 $\pm$ 0.0} & $4$: \mbox{\textbf{71.0 $\pm$ 4.0}} & $4$: \mbox{0.0 $\pm$ 0.0} & $4$: \mbox{\textbf{45.4 $\pm$ 4.4}} & $4$: \mbox{0.0 $\pm$ 0.0} & $4$: \mbox{\textbf{36.4 $\pm$ 4.3}} \\
        \rowcolor{black!4} \multirow{-2}{*}{\tokenizerid{nrvqbase!70}{CT-NRVQ-L}} & $8$: \mbox{0.0 $\pm$ 0.0} & $8$: \mbox{\textbf{46.4 $\pm$ 4.3}} & $8$: \mbox{0.0 $\pm$ 0.0} & $8$: \mbox{\textbf{64.2 $\pm$ 4.2}} & $8$: \mbox{0.0 $\pm$ 0.0} & $8$: \mbox{\textbf{42.0 $\pm$ 4.3}} & $8$: \mbox{0.0 $\pm$ 0.0} & $8$: \mbox{\textbf{26.2 $\pm$ 3.9}} \\
         & $4$: \mbox{76.6 $\pm$ 3.7} & $4$: \mbox{\textbf{78.4 $\pm$ 3.6}} & $4$: \mbox{\textbf{84.8 $\pm$ 3.1}} & $4$: \mbox{81.6 $\pm$ 3.4} & $4$: \mbox{\textbf{67.6 $\pm$ 4.1}} & $4$: \mbox{60.8 $\pm$ 4.3} & $4$: \mbox{\textbf{65.6 $\pm$ 4.2}} & $4$: \mbox{61.8 $\pm$ 4.2} \\
        \multirow{-2}{*}{\tokenizerid{nrvqbase!80}{C2-NRVQ-S}} & $8$: \mbox{\textbf{70.0 $\pm$ 4.0}} & $8$: \mbox{67.0 $\pm$ 4.1} & $8$: \mbox{\textbf{84.4 $\pm$ 3.2}} & $8$: \mbox{83.4 $\pm$ 3.2} & $8$: \mbox{\textbf{65.8 $\pm$ 4.2}} & $8$: \mbox{57.2 $\pm$ 4.3} & $8$: \mbox{\textbf{64.0 $\pm$ 4.2}} & $8$: \mbox{62.4 $\pm$ 4.2} \\
        \rowcolor{black!4}  & $4$: \mbox{\textbf{57.2 $\pm$ 4.4}} & $4$: \mbox{56.2 $\pm$ 4.3} & $4$: \mbox{75.8 $\pm$ 3.8} & $4$: \mbox{\textbf{76.4 $\pm$ 3.7}} & $4$: \mbox{47.4 $\pm$ 4.4} & $4$: \mbox{\textbf{49.4 $\pm$ 4.4}} & $4$: \mbox{35.6 $\pm$ 4.2} & $4$: \mbox{\textbf{37.6 $\pm$ 4.3}} \\
        \rowcolor{black!4} \multirow{-2}{*}{\tokenizerid{nrvqbase!90}{C2-NRVQ-L}} & $8$: \mbox{\textbf{43.4 $\pm$ 4.4}} & $8$: \mbox{42.2 $\pm$ 4.4} & $8$: \mbox{66.6 $\pm$ 4.1} & $8$: \mbox{\textbf{74.8 $\pm$ 3.8}} & $8$: \mbox{42.8 $\pm$ 4.4} & $8$: \mbox{\textbf{44.0 $\pm$ 4.3}} & $8$: \mbox{32.4 $\pm$ 4.1} & $8$: \mbox{\textbf{37.0 $\pm$ 4.2}} \\
        \addlinespace[1.5pt]
         & $4$: \mbox{36.2 $\pm$ 4.2} & $4$: \mbox{\textbf{63.2 $\pm$ 4.3}} & $4$: \mbox{67.6 $\pm$ 4.1} & $4$: \mbox{\textbf{79.0 $\pm$ 3.5}} & $4$: \mbox{11.0 $\pm$ 2.8} & $4$: \mbox{\textbf{47.2 $\pm$ 4.3}} & $4$: \mbox{8.0 $\pm$ 2.4} & $4$: \mbox{\textbf{54.4 $\pm$ 4.3}} \\
        \multirow{-2}{*}{\tokenizerid{fsqbase!60}{TF-FSQ-S}} & $8$: \mbox{33.6 $\pm$ 4.1} & $8$: \mbox{\textbf{51.2 $\pm$ 4.4}} & $8$: \mbox{72.4 $\pm$ 3.9} & $8$: \mbox{\textbf{82.4 $\pm$ 3.3}} & $8$: \mbox{10.8 $\pm$ 2.7} & $8$: \mbox{\textbf{43.4 $\pm$ 4.3}} & $8$: \mbox{12.0 $\pm$ 2.9} & $8$: \mbox{\textbf{56.6 $\pm$ 4.3}} \\
        \rowcolor{black!4}  & $4$: \mbox{15.4 $\pm$ 3.1} & $4$: \mbox{\textbf{64.8 $\pm$ 4.2}} & $4$: \mbox{36.4 $\pm$ 4.2} & $4$: \mbox{\textbf{83.2 $\pm$ 3.3}} & $4$: \mbox{9.6 $\pm$ 2.6} & $4$: \mbox{\textbf{45.0 $\pm$ 4.4}} & $4$: \mbox{2.0 $\pm$ 1.2} & $4$: \mbox{\textbf{48.6 $\pm$ 4.4}} \\
        \rowcolor{black!4} \multirow{-2}{*}{\tokenizerid{fsqbase!85}{TF-FSQ-L}} & $8$: \mbox{12.6 $\pm$ 2.9} & $8$: \mbox{\textbf{52.2 $\pm$ 4.3}} & $8$: \mbox{45.8 $\pm$ 4.3} & $8$: \mbox{\textbf{80.4 $\pm$ 3.4}} & $8$: \mbox{6.8 $\pm$ 2.2} & $8$: \mbox{\textbf{46.2 $\pm$ 4.4}} & $8$: \mbox{1.6 $\pm$ 1.1} & $8$: \mbox{\textbf{49.6 $\pm$ 4.3}} \\
        \addlinespace[1.5pt]
         & $4$: \mbox{\textbf{52.4 $\pm$ 4.4}} & $4$: \mbox{51.2 $\pm$ 4.4} & $4$: \mbox{58.8 $\pm$ 4.4} & $4$: \mbox{\textbf{66.4 $\pm$ 4.2}} & $4$: \mbox{\textbf{44.0 $\pm$ 4.4}} & $4$: \mbox{42.0 $\pm$ 4.3} & $4$: \mbox{\textbf{24.2 $\pm$ 3.7}} & $4$: \mbox{16.0 $\pm$ 3.2} \\
        \multirow{-2}{*}{\tokenizerid{qincobase!40}{TF-QINCO-S}} & $8$: \mbox{35.4 $\pm$ 4.2} & $8$: \mbox{\textbf{45.4 $\pm$ 4.4}} & $8$: \mbox{51.4 $\pm$ 4.4} & $8$: \mbox{\textbf{60.6 $\pm$ 4.3}} & $8$: \mbox{\textbf{35.6 $\pm$ 4.2}} & $8$: \mbox{34.6 $\pm$ 4.2} & $8$: \mbox{\textbf{15.2 $\pm$ 3.1}} & $8$: \mbox{15.0 $\pm$ 3.1} \\
        \rowcolor{black!4}  & $4$: \mbox{36.4 $\pm$ 4.2} & $4$: \mbox{\textbf{41.0 $\pm$ 4.3}} & $4$: \mbox{54.8 $\pm$ 4.3} & $4$: \mbox{\textbf{68.6 $\pm$ 4.1}} & $4$: \mbox{\textbf{34.8 $\pm$ 4.2}} & $4$: \mbox{32.2 $\pm$ 4.1} & $4$: \mbox{\textbf{14.8 $\pm$ 3.1}} & $4$: \mbox{8.0 $\pm$ 2.4} \\
        \rowcolor{black!4} \multirow{-2}{*}{\tokenizerid{qincobase!50}{TF-QINCO-L}} & $8$: \mbox{25.8 $\pm$ 3.8} & $8$: \mbox{\textbf{32.2 $\pm$ 4.1}} & $8$: \mbox{34.0 $\pm$ 4.2} & $8$: \mbox{\textbf{66.0 $\pm$ 4.1}} & $8$: \mbox{\textbf{28.2 $\pm$ 4.0}} & $8$: \mbox{27.4 $\pm$ 4.0} & $8$: \mbox{7.2 $\pm$ 2.3} & $8$: \mbox{\textbf{9.4 $\pm$ 2.6}} \\
         & $4$: \mbox{13.2 $\pm$ 2.9} & $4$: \mbox{\textbf{32.6 $\pm$ 4.1}} & $4$: \mbox{40.2 $\pm$ 4.3} & $4$: \mbox{\textbf{57.4 $\pm$ 4.3}} & $4$: \mbox{24.6 $\pm$ 3.8} & $4$: \mbox{\textbf{42.0 $\pm$ 4.3}} & $4$: \mbox{1.8 $\pm$ 1.2} & $4$: \mbox{\textbf{9.8 $\pm$ 2.6}} \\
        \multirow{-2}{*}{\tokenizerid{qincobase!60}{CT-QINCO-S}} & $8$: \mbox{16.4 $\pm$ 3.3} & $8$: \mbox{\textbf{32.2 $\pm$ 4.1}} & $8$: \mbox{35.8 $\pm$ 4.2} & $8$: \mbox{\textbf{51.4 $\pm$ 4.3}} & $8$: \mbox{21.6 $\pm$ 3.6} & $8$: \mbox{\textbf{36.6 $\pm$ 4.2}} & $8$: \mbox{2.2 $\pm$ 1.3} & $8$: \mbox{\textbf{12.2 $\pm$ 2.8}} \\
        \rowcolor{black!4}  & $4$: \mbox{26.6 $\pm$ 3.9} & $4$: \mbox{\textbf{53.0 $\pm$ 4.4}} & $4$: \mbox{53.4 $\pm$ 4.4} & $4$: \mbox{\textbf{64.4 $\pm$ 4.2}} & $4$: \mbox{23.6 $\pm$ 3.8} & $4$: \mbox{\textbf{37.4 $\pm$ 4.3}} & $4$: \mbox{9.8 $\pm$ 2.6} & $4$: \mbox{\textbf{15.2 $\pm$ 3.1}} \\
        \rowcolor{black!4} \multirow{-2}{*}{\tokenizerid{qincobase!70}{CT-QINCO-L}} & $8$: \mbox{21.2 $\pm$ 3.6} & $8$: \mbox{\textbf{42.6 $\pm$ 4.3}} & $8$: \mbox{35.0 $\pm$ 4.2} & $8$: \mbox{\textbf{62.2 $\pm$ 4.3}} & $8$: \mbox{21.2 $\pm$ 3.6} & $8$: \mbox{\textbf{37.8 $\pm$ 4.3}} & $8$: \mbox{5.2 $\pm$ 1.9} & $8$: \mbox{\textbf{15.2 $\pm$ 3.1}} \\
         & $4$: \mbox{5.0 $\pm$ 1.9} & $4$: \mbox{\textbf{53.2 $\pm$ 4.4}} & $4$: \mbox{15.8 $\pm$ 3.2} & $4$: \mbox{\textbf{62.8 $\pm$ 4.2}} & $4$: \mbox{7.4 $\pm$ 2.3} & $4$: \mbox{\textbf{27.0 $\pm$ 3.9}} & $4$: \mbox{0.8 $\pm$ 0.8} & $4$: \mbox{\textbf{9.0 $\pm$ 2.5}} \\
        \multirow{-2}{*}{\tokenizerid{qincobase!80}{C2-QINCO-S}} & $8$: \mbox{11.0 $\pm$ 2.7} & $8$: \mbox{\textbf{50.4 $\pm$ 4.3}} & $8$: \mbox{15.2 $\pm$ 3.1} & $8$: \mbox{\textbf{58.8 $\pm$ 4.3}} & $8$: \mbox{10.6 $\pm$ 2.7} & $8$: \mbox{\textbf{25.8 $\pm$ 3.8}} & $8$: \mbox{0.8 $\pm$ 0.8} & $8$: \mbox{\textbf{8.6 $\pm$ 2.5}} \\
        \rowcolor{black!4}  & $4$: \mbox{16.2 $\pm$ 3.2} & $4$: \mbox{\textbf{68.4 $\pm$ 4.1}} & $4$: \mbox{56.6 $\pm$ 4.4} & $4$: \mbox{\textbf{81.8 $\pm$ 3.4}} & $4$: \mbox{20.8 $\pm$ 3.6} & $4$: \mbox{\textbf{61.6 $\pm$ 4.3}} & $4$: \mbox{7.8 $\pm$ 2.4} & $4$: \mbox{\textbf{42.8 $\pm$ 4.4}} \\
        \rowcolor{black!4} \multirow{-2}{*}{\tokenizerid{qincobase!90}{C2-QINCO-L}} & $8$: \mbox{13.0 $\pm$ 2.9} & $8$: \mbox{\textbf{49.8 $\pm$ 4.4}} & $8$: \mbox{41.2 $\pm$ 4.3} & $8$: \mbox{\textbf{74.8 $\pm$ 3.8}} & $8$: \mbox{21.4 $\pm$ 3.6} & $8$: \mbox{\textbf{51.2 $\pm$ 4.4}} & $8$: \mbox{5.0 $\pm$ 1.9} & $8$: \mbox{\textbf{38.6 $\pm$ 4.3}} \\
        \bottomrule
    \end{tabularx}
\end{table}

%% file: tables/experiments/simulation_results/appendix_robotwin_multitask_best.tex
\begin{table}[H]
    \centering
    \caption{Task-level multi-task RoboTwin success (\%) at the checkpoint with the best teacher-forced validation accuracy. Each tokenizer occupies two lines for $H_{\mathrm{exec}}=4$ and $8$ success rate $\pm$ 95\% confidence-interval half-width (percentage points), computed by episode bootstrap from the success counts over 50 episodes per task and horizon. Place cans denotes placing cans in the plastic box. R.O. denotes Reconstruction Only; \textbf{bold} marks the better objective per task and horizon.}
    \label{tab:robotwin-multitask-by-task}
    \tablebodyfont
    \setlength{\tabcolsep}{1.0pt}
    \renewcommand{\arraystretch}{0.96}
    \begin{tabularx}{\linewidth}{@{}>{\raggedright\arraybackslash}m{0.16\linewidth} *{6}{>{\centering\arraybackslash}X}@{}}
        \toprule
        \multirow{2}{*}{Tokenizer} & \multicolumn{2}{c}{Grab roller} & \multicolumn{2}{c}{Place burger/fries} & \multicolumn{2}{c}{Place cans} \\
        \cmidrule(lr){2-3}\cmidrule(lr){4-5}\cmidrule(l){6-7}
        & R.O. & \methodname{} & R.O. & \methodname{} & R.O. & \methodname{} \\
        \midrule
         & $4$: \mbox{80.0 $\pm$ 10.9} & $4$: \mbox{\textbf{84.0 $\pm$ 10.1}} & $4$: \mbox{\textbf{76.0 $\pm$ 11.8}} & $4$: \mbox{74.0 $\pm$ 12.1} & $4$: \mbox{\textbf{64.0 $\pm$ 13.4}} & $4$: \mbox{30.0 $\pm$ 12.9} \\
        \multirow{-2}{*}{\tokenizerid{vqbase!60}{TF-VQ-S}} & $8$: \mbox{70.0 $\pm$ 12.8} & $8$: \mbox{\textbf{74.0 $\pm$ 12.1}} & $8$: \mbox{62.0 $\pm$ 13.5} & $8$: \mbox{\textbf{68.0 $\pm$ 13.0}} & $8$: \mbox{\textbf{78.0 $\pm$ 11.4}} & $8$: \mbox{50.0 $\pm$ 14.1} \\
        \rowcolor{black!4}  & $4$: \mbox{60.0 $\pm$ 13.6} & $4$: \mbox{\textbf{88.0 $\pm$ 9.0}} & $4$: \mbox{26.0 $\pm$ 12.4} & $4$: \mbox{\textbf{52.0 $\pm$ 14.0}} & $4$: \mbox{10.0 $\pm$ 8.4} & $4$: \mbox{\textbf{36.0 $\pm$ 13.4}} \\
        \rowcolor{black!4} \multirow{-2}{*}{\tokenizerid{vqbase!85}{TF-VQ-L}} & $8$: \mbox{56.0 $\pm$ 13.8} & $8$: \mbox{\textbf{78.0 $\pm$ 11.4}} & $8$: \mbox{34.0 $\pm$ 13.3} & $8$: \mbox{\textbf{74.0 $\pm$ 12.1}} & $8$: \mbox{34.0 $\pm$ 13.3} & $8$: \mbox{\textbf{58.0 $\pm$ 13.7}} \\
        \addlinespace[2pt]
         & $4$: \mbox{\textbf{94.0 $\pm$ 6.6}} & $4$: \mbox{92.0 $\pm$ 7.5} & $4$: \mbox{\textbf{76.0 $\pm$ 11.8}} & $4$: \mbox{58.0 $\pm$ 13.7} & $4$: \mbox{\textbf{46.0 $\pm$ 14.0}} & $4$: \mbox{42.0 $\pm$ 13.8} \\
        \multirow{-2}{*}{\tokenizerid{rvqbase!40}{TF-RVQ-S}} & $8$: \mbox{84.0 $\pm$ 10.1} & $8$: \mbox{\textbf{92.0 $\pm$ 7.5}} & $8$: \mbox{\textbf{60.0 $\pm$ 13.6}} & $8$: \mbox{56.0 $\pm$ 13.8} & $8$: \mbox{\textbf{64.0 $\pm$ 13.4}} & $8$: \mbox{60.0 $\pm$ 13.6} \\
        \rowcolor{black!4}  & $4$: \mbox{50.0 $\pm$ 14.1} & $4$: \mbox{\textbf{78.0 $\pm$ 11.4}} & $4$: \mbox{14.0 $\pm$ 9.6} & $4$: \mbox{\textbf{34.0 $\pm$ 13.3}} & $4$: \mbox{6.0 $\pm$ 6.6} & $4$: \mbox{\textbf{16.0 $\pm$ 10.2}} \\
        \rowcolor{black!4} \multirow{-2}{*}{\tokenizerid{rvqbase!50}{TF-RVQ-L}} & $8$: \mbox{40.0 $\pm$ 13.7} & $8$: \mbox{\textbf{66.0 $\pm$ 13.1}} & $8$: \mbox{12.0 $\pm$ 9.0} & $8$: \mbox{\textbf{28.0 $\pm$ 12.6}} & $8$: \mbox{16.0 $\pm$ 10.2} & $8$: \mbox{\textbf{22.0 $\pm$ 11.6}} \\
         & $4$: \mbox{82.0 $\pm$ 10.5} & $4$: \mbox{\textbf{90.0 $\pm$ 8.3}} & $4$: \mbox{60.0 $\pm$ 13.6} & $4$: \mbox{\textbf{72.0 $\pm$ 12.5}} & $4$: \mbox{\textbf{42.0 $\pm$ 13.8}} & $4$: \mbox{40.0 $\pm$ 13.7} \\
        \multirow{-2}{*}{\tokenizerid{rvqbase!60}{CT-RVQ-S}} & $8$: \mbox{\textbf{82.0 $\pm$ 10.5}} & $8$: \mbox{80.0 $\pm$ 10.9} & $8$: \mbox{62.0 $\pm$ 13.5} & $8$: \mbox{\textbf{72.0 $\pm$ 12.5}} & $8$: \mbox{48.0 $\pm$ 14.0} & $8$: \mbox{\textbf{50.0 $\pm$ 14.1}} \\
        \rowcolor{black!4}  & $4$: \mbox{46.0 $\pm$ 14.0} & $4$: \mbox{\textbf{56.0 $\pm$ 13.8}} & $4$: \mbox{6.0 $\pm$ 6.6} & $4$: \mbox{\textbf{28.0 $\pm$ 12.6}} & $4$: \mbox{0.0 $\pm$ 0.0} & $4$: \mbox{\textbf{16.0 $\pm$ 10.2}} \\
        \rowcolor{black!4} \multirow{-2}{*}{\tokenizerid{rvqbase!70}{CT-RVQ-L}} & $8$: \mbox{\textbf{48.0 $\pm$ 14.0}} & $8$: \mbox{\textbf{48.0 $\pm$ 14.0}} & $8$: \mbox{10.0 $\pm$ 8.4} & $8$: \mbox{\textbf{28.0 $\pm$ 12.6}} & $8$: \mbox{10.0 $\pm$ 8.4} & $8$: \mbox{\textbf{26.0 $\pm$ 12.4}} \\
         & $4$: \mbox{\textbf{78.0 $\pm$ 11.4}} & $4$: \mbox{54.0 $\pm$ 13.9} & $4$: \mbox{\textbf{46.0 $\pm$ 14.0}} & $4$: \mbox{16.0 $\pm$ 10.2} & $4$: \mbox{\textbf{20.0 $\pm$ 11.1}} & $4$: \mbox{0.0 $\pm$ 0.0} \\
        \multirow{-2}{*}{\tokenizerid{rvqbase!80}{C2-RVQ-S}} & $8$: \mbox{\textbf{82.0 $\pm$ 10.5}} & $8$: \mbox{54.0 $\pm$ 13.9} & $8$: \mbox{\textbf{42.0 $\pm$ 13.8}} & $8$: \mbox{12.0 $\pm$ 9.0} & $8$: \mbox{\textbf{40.0 $\pm$ 13.7}} & $8$: \mbox{0.0 $\pm$ 0.0} \\
        \rowcolor{black!4}  & $4$: \mbox{\textbf{72.0 $\pm$ 12.5}} & $4$: \mbox{54.0 $\pm$ 13.9} & $4$: \mbox{\textbf{18.0 $\pm$ 10.7}} & $4$: \mbox{\textbf{18.0 $\pm$ 10.7}} & $4$: \mbox{2.0 $\pm$ 3.9} & $4$: \mbox{\textbf{4.0 $\pm$ 5.4}} \\
        \rowcolor{black!4} \multirow{-2}{*}{\tokenizerid{rvqbase!90}{C2-RVQ-L}} & $8$: \mbox{\textbf{50.0 $\pm$ 14.1}} & $8$: \mbox{42.0 $\pm$ 13.8} & $8$: \mbox{10.0 $\pm$ 8.4} & $8$: \mbox{\textbf{28.0 $\pm$ 12.6}} & $8$: \mbox{8.0 $\pm$ 7.6} & $8$: \mbox{\textbf{22.0 $\pm$ 11.6}} \\
        \addlinespace[2pt]
         & $4$: \mbox{\textbf{88.0 $\pm$ 9.0}} & $4$: \mbox{86.0 $\pm$ 9.5} & $4$: \mbox{60.0 $\pm$ 13.6} & $4$: \mbox{\textbf{64.0 $\pm$ 13.4}} & $4$: \mbox{\textbf{40.0 $\pm$ 13.7}} & $4$: \mbox{34.0 $\pm$ 13.3} \\
        \multirow{-2}{*}{\tokenizerid{nrvqbase!40}{TF-NRVQ-S}} & $8$: \mbox{62.0 $\pm$ 13.5} & $8$: \mbox{\textbf{72.0 $\pm$ 12.5}} & $8$: \mbox{\textbf{52.0 $\pm$ 14.0}} & $8$: \mbox{\textbf{52.0 $\pm$ 14.0}} & $8$: \mbox{\textbf{56.0 $\pm$ 13.8}} & $8$: \mbox{50.0 $\pm$ 14.1} \\
        \rowcolor{black!4}  & $4$: \mbox{54.0 $\pm$ 13.9} & $4$: \mbox{\textbf{70.0 $\pm$ 12.8}} & $4$: \mbox{26.0 $\pm$ 12.4} & $4$: \mbox{\textbf{34.0 $\pm$ 13.3}} & $4$: \mbox{10.0 $\pm$ 8.4} & $4$: \mbox{\textbf{20.0 $\pm$ 11.1}} \\
        \rowcolor{black!4} \multirow{-2}{*}{\tokenizerid{nrvqbase!50}{TF-NRVQ-L}} & $8$: \mbox{60.0 $\pm$ 13.6} & $8$: \mbox{\textbf{66.0 $\pm$ 13.1}} & $8$: \mbox{28.0 $\pm$ 12.6} & $8$: \mbox{\textbf{30.0 $\pm$ 12.9}} & $8$: \mbox{20.0 $\pm$ 11.1} & $8$: \mbox{\textbf{22.0 $\pm$ 11.6}} \\
         & $4$: \mbox{0.0 $\pm$ 0.0} & $4$: \mbox{\textbf{76.0 $\pm$ 11.8}} & $4$: \mbox{0.0 $\pm$ 0.0} & $4$: \mbox{\textbf{40.0 $\pm$ 13.7}} & $4$: \mbox{0.0 $\pm$ 0.0} & $4$: \mbox{\textbf{28.0 $\pm$ 12.6}} \\
        \multirow{-2}{*}{\tokenizerid{nrvqbase!60}{CT-NRVQ-S}} & $8$: \mbox{0.0 $\pm$ 0.0} & $8$: \mbox{\textbf{88.0 $\pm$ 9.0}} & $8$: \mbox{0.0 $\pm$ 0.0} & $8$: \mbox{\textbf{44.0 $\pm$ 13.9}} & $8$: \mbox{0.0 $\pm$ 0.0} & $8$: \mbox{\textbf{72.0 $\pm$ 12.5}} \\
        \rowcolor{black!4}  & $4$: \mbox{0.0 $\pm$ 0.0} & $4$: \mbox{\textbf{76.0 $\pm$ 11.8}} & $4$: \mbox{0.0 $\pm$ 0.0} & $4$: \mbox{\textbf{36.0 $\pm$ 13.4}} & $4$: \mbox{0.0 $\pm$ 0.0} & $4$: \mbox{\textbf{14.0 $\pm$ 9.6}} \\
        \rowcolor{black!4} \multirow{-2}{*}{\tokenizerid{nrvqbase!70}{CT-NRVQ-L}} & $8$: \mbox{0.0 $\pm$ 0.0} & $8$: \mbox{\textbf{62.0 $\pm$ 13.5}} & $8$: \mbox{0.0 $\pm$ 0.0} & $8$: \mbox{\textbf{30.0 $\pm$ 12.9}} & $8$: \mbox{0.0 $\pm$ 0.0} & $8$: \mbox{\textbf{4.0 $\pm$ 5.4}} \\
         & $4$: \mbox{\textbf{74.0 $\pm$ 12.1}} & $4$: \mbox{70.0 $\pm$ 12.8} & $4$: \mbox{\textbf{60.0 $\pm$ 13.6}} & $4$: \mbox{58.0 $\pm$ 13.7} & $4$: \mbox{34.0 $\pm$ 13.3} & $4$: \mbox{\textbf{38.0 $\pm$ 13.6}} \\
        \multirow{-2}{*}{\tokenizerid{nrvqbase!80}{C2-NRVQ-S}} & $8$: \mbox{\textbf{76.0 $\pm$ 11.8}} & $8$: \mbox{\textbf{76.0 $\pm$ 11.8}} & $8$: \mbox{60.0 $\pm$ 13.6} & $8$: \mbox{\textbf{62.0 $\pm$ 13.5}} & $8$: \mbox{54.0 $\pm$ 13.9} & $8$: \mbox{\textbf{60.0 $\pm$ 13.6}} \\
        \rowcolor{black!4}  & $4$: \mbox{72.0 $\pm$ 12.5} & $4$: \mbox{\textbf{74.0 $\pm$ 12.1}} & $4$: \mbox{\textbf{46.0 $\pm$ 14.0}} & $4$: \mbox{44.0 $\pm$ 13.9} & $4$: \mbox{4.0 $\pm$ 5.4} & $4$: \mbox{\textbf{12.0 $\pm$ 9.0}} \\
        \rowcolor{black!4} \multirow{-2}{*}{\tokenizerid{nrvqbase!90}{C2-NRVQ-L}} & $8$: \mbox{68.0 $\pm$ 13.0} & $8$: \mbox{\textbf{70.0 $\pm$ 12.8}} & $8$: \mbox{\textbf{36.0 $\pm$ 13.4}} & $8$: \mbox{34.0 $\pm$ 13.3} & $8$: \mbox{\textbf{12.0 $\pm$ 9.0}} & $8$: \mbox{8.0 $\pm$ 7.6} \\
        \addlinespace[2pt]
         & $4$: \mbox{86.0 $\pm$ 9.5} & $4$: \mbox{\textbf{90.0 $\pm$ 8.3}} & $4$: \mbox{\textbf{70.0 $\pm$ 12.8}} & $4$: \mbox{\textbf{70.0 $\pm$ 12.8}} & $4$: \mbox{50.0 $\pm$ 14.1} & $4$: \mbox{\textbf{58.0 $\pm$ 13.7}} \\
        \multirow{-2}{*}{\tokenizerid{fsqbase!60}{TF-FSQ-S}} & $8$: \mbox{\textbf{90.0 $\pm$ 8.3}} & $8$: \mbox{88.0 $\pm$ 9.0} & $8$: \mbox{58.0 $\pm$ 13.7} & $8$: \mbox{\textbf{72.0 $\pm$ 12.5}} & $8$: \mbox{60.0 $\pm$ 13.6} & $8$: \mbox{\textbf{74.0 $\pm$ 12.1}} \\
        \rowcolor{black!4}  & $4$: \mbox{\textbf{90.0 $\pm$ 8.3}} & $4$: \mbox{88.0 $\pm$ 9.0} & $4$: \mbox{\textbf{66.0 $\pm$ 13.1}} & $4$: \mbox{60.0 $\pm$ 13.6} & $4$: \mbox{\textbf{46.0 $\pm$ 14.0}} & $4$: \mbox{28.0 $\pm$ 12.6} \\
        \rowcolor{black!4} \multirow{-2}{*}{\tokenizerid{fsqbase!85}{TF-FSQ-L}} & $8$: \mbox{\textbf{86.0 $\pm$ 9.5}} & $8$: \mbox{\textbf{86.0 $\pm$ 9.5}} & $8$: \mbox{66.0 $\pm$ 13.1} & $8$: \mbox{\textbf{70.0 $\pm$ 12.8}} & $8$: \mbox{54.0 $\pm$ 13.9} & $8$: \mbox{\textbf{66.0 $\pm$ 13.1}} \\
        \addlinespace[2pt]
         & $4$: \mbox{40.0 $\pm$ 13.7} & $4$: \mbox{\textbf{82.0 $\pm$ 10.5}} & $4$: \mbox{2.0 $\pm$ 3.9} & $4$: \mbox{\textbf{52.0 $\pm$ 14.0}} & $4$: \mbox{2.0 $\pm$ 3.9} & $4$: \mbox{\textbf{24.0 $\pm$ 12.0}} \\
        \multirow{-2}{*}{\tokenizerid{qincobase!40}{TF-QINCO-S}} & $8$: \mbox{30.0 $\pm$ 12.9} & $8$: \mbox{\textbf{74.0 $\pm$ 12.1}} & $8$: \mbox{6.0 $\pm$ 6.6} & $8$: \mbox{\textbf{52.0 $\pm$ 14.0}} & $8$: \mbox{6.0 $\pm$ 6.6} & $8$: \mbox{\textbf{20.0 $\pm$ 11.1}} \\
        \rowcolor{black!4}  & $4$: \mbox{42.0 $\pm$ 13.8} & $4$: \mbox{\textbf{70.0 $\pm$ 12.8}} & $4$: \mbox{10.0 $\pm$ 8.4} & $4$: \mbox{\textbf{42.0 $\pm$ 13.8}} & $4$: \mbox{0.0 $\pm$ 0.0} & $4$: \mbox{\textbf{26.0 $\pm$ 12.4}} \\
        \rowcolor{black!4} \multirow{-2}{*}{\tokenizerid{qincobase!50}{TF-QINCO-L}} & $8$: \mbox{52.0 $\pm$ 14.0} & $8$: \mbox{\textbf{68.0 $\pm$ 13.0}} & $8$: \mbox{2.0 $\pm$ 3.9} & $8$: \mbox{\textbf{48.0 $\pm$ 14.0}} & $8$: \mbox{0.0 $\pm$ 0.0} & $8$: \mbox{\textbf{22.0 $\pm$ 11.6}} \\
         & $4$: \mbox{38.0 $\pm$ 13.6} & $4$: \mbox{\textbf{88.0 $\pm$ 9.0}} & $4$: \mbox{18.0 $\pm$ 10.7} & $4$: \mbox{\textbf{42.0 $\pm$ 13.8}} & $4$: \mbox{0.0 $\pm$ 0.0} & $4$: \mbox{\textbf{24.0 $\pm$ 12.0}} \\
        \multirow{-2}{*}{\tokenizerid{qincobase!60}{CT-QINCO-S}} & $8$: \mbox{40.0 $\pm$ 13.7} & $8$: \mbox{\textbf{84.0 $\pm$ 10.1}} & $8$: \mbox{12.0 $\pm$ 9.0} & $8$: \mbox{\textbf{42.0 $\pm$ 13.8}} & $8$: \mbox{6.0 $\pm$ 6.6} & $8$: \mbox{\textbf{22.0 $\pm$ 11.6}} \\
        \rowcolor{black!4}  & $4$: \mbox{36.0 $\pm$ 13.4} & $4$: \mbox{\textbf{66.0 $\pm$ 13.1}} & $4$: \mbox{28.0 $\pm$ 12.6} & $4$: \mbox{\textbf{60.0 $\pm$ 13.6}} & $4$: \mbox{6.0 $\pm$ 6.6} & $4$: \mbox{\textbf{16.0 $\pm$ 10.2}} \\
        \rowcolor{black!4} \multirow{-2}{*}{\tokenizerid{qincobase!70}{CT-QINCO-L}} & $8$: \mbox{32.0 $\pm$ 13.1} & $8$: \mbox{\textbf{54.0 $\pm$ 13.9}} & $8$: \mbox{18.0 $\pm$ 10.7} & $8$: \mbox{\textbf{58.0 $\pm$ 13.7}} & $8$: \mbox{8.0 $\pm$ 7.6} & $8$: \mbox{\textbf{20.0 $\pm$ 11.1}} \\
         & $4$: \mbox{62.0 $\pm$ 13.5} & $4$: \mbox{\textbf{88.0 $\pm$ 9.0}} & $4$: \mbox{40.0 $\pm$ 13.7} & $4$: \mbox{\textbf{56.0 $\pm$ 13.8}} & $4$: \mbox{\textbf{26.0 $\pm$ 12.4}} & $4$: \mbox{\textbf{26.0 $\pm$ 12.4}} \\
        \multirow{-2}{*}{\tokenizerid{qincobase!80}{C2-QINCO-S}} & $8$: \mbox{44.0 $\pm$ 13.9} & $8$: \mbox{\textbf{78.0 $\pm$ 11.4}} & $8$: \mbox{24.0 $\pm$ 12.0} & $8$: \mbox{\textbf{44.0 $\pm$ 13.9}} & $8$: \mbox{14.0 $\pm$ 9.6} & $8$: \mbox{\textbf{20.0 $\pm$ 11.1}} \\
        \rowcolor{black!4}  & $4$: \mbox{\textbf{54.0 $\pm$ 13.9}} & $4$: \mbox{34.0 $\pm$ 13.3} & $4$: \mbox{8.0 $\pm$ 7.6} & $4$: \mbox{\textbf{22.0 $\pm$ 11.6}} & $4$: \mbox{0.0 $\pm$ 0.0} & $4$: \mbox{\textbf{20.0 $\pm$ 11.1}} \\
        \rowcolor{black!4} \multirow{-2}{*}{\tokenizerid{qincobase!90}{C2-QINCO-L}} & $8$: \mbox{\textbf{38.0 $\pm$ 13.6}} & $8$: \mbox{\textbf{38.0 $\pm$ 13.6}} & $8$: \mbox{2.0 $\pm$ 3.9} & $8$: \mbox{\textbf{16.0 $\pm$ 10.2}} & $8$: \mbox{6.0 $\pm$ 6.6} & $8$: \mbox{\textbf{20.0 $\pm$ 11.1}} \\
        \bottomrule
    \end{tabularx}
\end{table}

%% file: tables/experiments/simulation_results/appendix_robomimic_multitask_last.tex
\begin{table}[H]
    \centering
    \caption{Task-level multi-task RoboMimic results at the last checkpoint. Each tokenizer occupies two lines for $H_{\mathrm{exec}}=4$ and $8$ success rate $\pm$ 95\% confidence-interval half-width (percentage points). R.O. denotes Reconstruction Only; \textbf{bold} marks the better objective per task and horizon.}
    \label{tab:robomimic-multitask-last}
    \tablebodyfont
    \setlength{\tabcolsep}{1.0pt}
    \renewcommand{\arraystretch}{0.96}
    \begin{tabularx}{\linewidth}{@{}>{\raggedright\arraybackslash}m{0.16\linewidth} *{6}{>{\centering\arraybackslash}X}@{}}
        \toprule
        \multirow{2}{*}{Tokenizer} & \multicolumn{2}{c}{Can} & \multicolumn{2}{c}{Lift} & \multicolumn{2}{c}{Square} \\
        \cmidrule(lr){2-3}\cmidrule(lr){4-5}\cmidrule(l){6-7}
        & R.O. & \methodname{} & R.O. & \methodname{} & R.O. & \methodname{} \\
        \midrule
         & $4$: \mbox{88.3 $\pm$ 3.6} & $4$: \mbox{\textbf{93.3 $\pm$ 2.9}} & $4$: \mbox{81.3 $\pm$ 4.4} & $4$: \mbox{\textbf{99.0 $\pm$ 1.1}} & $4$: \mbox{55.3 $\pm$ 5.6} & $4$: \mbox{\textbf{61.0 $\pm$ 5.5}} \\
        \multirow{-2}{*}{\tokenizerid{vqbase!60}{TF-VQ-S}} & $8$: \mbox{87.3 $\pm$ 3.8} & $8$: \mbox{\textbf{93.3 $\pm$ 2.8}} & $8$: \mbox{87.7 $\pm$ 3.7} & $8$: \mbox{\textbf{99.0 $\pm$ 1.1}} & $8$: \mbox{56.0 $\pm$ 5.6} & $8$: \mbox{\textbf{60.0 $\pm$ 5.6}} \\
        \rowcolor{black!4}  & $4$: \mbox{85.7 $\pm$ 4.0} & $4$: \mbox{\textbf{87.7 $\pm$ 3.8}} & $4$: \mbox{91.7 $\pm$ 3.1} & $4$: \mbox{\textbf{97.7 $\pm$ 1.7}} & $4$: \mbox{42.7 $\pm$ 5.6} & $4$: \mbox{\textbf{44.7 $\pm$ 5.6}} \\
        \rowcolor{black!4} \multirow{-2}{*}{\tokenizerid{vqbase!85}{TF-VQ-L}} & $8$: \mbox{\textbf{84.7 $\pm$ 4.1}} & $8$: \mbox{83.7 $\pm$ 4.2} & $8$: \mbox{88.7 $\pm$ 3.6} & $8$: \mbox{\textbf{96.7 $\pm$ 2.0}} & $8$: \mbox{\textbf{45.3 $\pm$ 5.6}} & $8$: \mbox{33.7 $\pm$ 5.4} \\
        \addlinespace[2pt]
         & $4$: \mbox{89.3 $\pm$ 3.5} & $4$: \mbox{\textbf{91.0 $\pm$ 3.2}} & $4$: \mbox{\textbf{91.7 $\pm$ 3.2}} & $4$: \mbox{90.7 $\pm$ 3.3} & $4$: \mbox{60.3 $\pm$ 5.5} & $4$: \mbox{\textbf{69.7 $\pm$ 5.3}} \\
        \multirow{-2}{*}{\tokenizerid{rvqbase!40}{TF-RVQ-S}} & $8$: \mbox{92.3 $\pm$ 3.0} & $8$: \mbox{\textbf{96.0 $\pm$ 2.2}} & $8$: \mbox{91.3 $\pm$ 3.2} & $8$: \mbox{\textbf{92.3 $\pm$ 3.0}} & $8$: \mbox{60.7 $\pm$ 5.6} & $8$: \mbox{\textbf{65.0 $\pm$ 5.4}} \\
        \rowcolor{black!4}  & $4$: \mbox{72.7 $\pm$ 5.0} & $4$: \mbox{\textbf{87.7 $\pm$ 3.8}} & $4$: \mbox{\textbf{81.3 $\pm$ 4.4}} & $4$: \mbox{81.0 $\pm$ 4.5} & $4$: \mbox{38.0 $\pm$ 5.5} & $4$: \mbox{\textbf{45.3 $\pm$ 5.7}} \\
        \rowcolor{black!4} \multirow{-2}{*}{\tokenizerid{rvqbase!50}{TF-RVQ-L}} & $8$: \mbox{66.3 $\pm$ 5.4} & $8$: \mbox{\textbf{83.7 $\pm$ 4.2}} & $8$: \mbox{73.0 $\pm$ 4.9} & $8$: \mbox{\textbf{81.7 $\pm$ 4.4}} & $8$: \mbox{35.3 $\pm$ 5.4} & $8$: \mbox{\textbf{47.7 $\pm$ 5.7}} \\
         & $4$: \mbox{89.7 $\pm$ 3.5} & $4$: \mbox{\textbf{93.7 $\pm$ 2.8}} & $4$: \mbox{\textbf{91.3 $\pm$ 3.2}} & $4$: \mbox{89.0 $\pm$ 3.5} & $4$: \mbox{\textbf{61.3 $\pm$ 5.5}} & $4$: \mbox{59.0 $\pm$ 5.6} \\
        \multirow{-2}{*}{\tokenizerid{rvqbase!60}{CT-RVQ-S}} & $8$: \mbox{91.7 $\pm$ 3.1} & $8$: \mbox{\textbf{97.0 $\pm$ 1.9}} & $8$: \mbox{93.7 $\pm$ 2.8} & $8$: \mbox{\textbf{94.3 $\pm$ 2.6}} & $8$: \mbox{59.7 $\pm$ 5.6} & $8$: \mbox{\textbf{63.3 $\pm$ 5.4}} \\
        \rowcolor{black!4}  & $4$: \mbox{75.3 $\pm$ 4.9} & $4$: \mbox{\textbf{82.0 $\pm$ 4.4}} & $4$: \mbox{71.0 $\pm$ 5.1} & $4$: \mbox{\textbf{82.7 $\pm$ 4.3}} & $4$: \mbox{32.7 $\pm$ 5.3} & $4$: \mbox{\textbf{45.0 $\pm$ 5.6}} \\
        \rowcolor{black!4} \multirow{-2}{*}{\tokenizerid{rvqbase!70}{CT-RVQ-L}} & $8$: \mbox{69.7 $\pm$ 5.3} & $8$: \mbox{\textbf{73.0 $\pm$ 5.0}} & $8$: \mbox{76.0 $\pm$ 4.8} & $8$: \mbox{\textbf{78.7 $\pm$ 4.7}} & $8$: \mbox{30.0 $\pm$ 5.2} & $8$: \mbox{\textbf{32.3 $\pm$ 5.3}} \\
         & $4$: \mbox{93.7 $\pm$ 2.8} & $4$: \mbox{\textbf{96.7 $\pm$ 2.1}} & $4$: \mbox{\textbf{95.7 $\pm$ 2.3}} & $4$: \mbox{88.3 $\pm$ 3.7} & $4$: \mbox{\textbf{61.3 $\pm$ 5.5}} & $4$: \mbox{58.3 $\pm$ 5.6} \\
        \multirow{-2}{*}{\tokenizerid{rvqbase!80}{C2-RVQ-S}} & $8$: \mbox{90.7 $\pm$ 3.3} & $8$: \mbox{\textbf{95.3 $\pm$ 2.4}} & $8$: \mbox{\textbf{95.7 $\pm$ 2.3}} & $8$: \mbox{91.7 $\pm$ 3.1} & $8$: \mbox{\textbf{68.0 $\pm$ 5.3}} & $8$: \mbox{59.7 $\pm$ 5.6} \\
        \rowcolor{black!4}  & $4$: \mbox{74.3 $\pm$ 4.9} & $4$: \mbox{\textbf{78.3 $\pm$ 4.6}} & $4$: \mbox{\textbf{77.0 $\pm$ 4.8}} & $4$: \mbox{74.7 $\pm$ 5.0} & $4$: \mbox{38.3 $\pm$ 5.6} & $4$: \mbox{\textbf{41.0 $\pm$ 5.6}} \\
        \rowcolor{black!4} \multirow{-2}{*}{\tokenizerid{rvqbase!90}{C2-RVQ-L}} & $8$: \mbox{69.0 $\pm$ 5.2} & $8$: \mbox{\textbf{76.0 $\pm$ 4.9}} & $8$: \mbox{\textbf{73.0 $\pm$ 5.0}} & $8$: \mbox{72.7 $\pm$ 5.1} & $8$: \mbox{26.3 $\pm$ 5.0} & $8$: \mbox{\textbf{32.0 $\pm$ 5.3}} \\
        \addlinespace[2pt]
         & $4$: \mbox{89.0 $\pm$ 3.5} & $4$: \mbox{\textbf{93.0 $\pm$ 2.9}} & $4$: \mbox{85.3 $\pm$ 4.0} & $4$: \mbox{\textbf{88.3 $\pm$ 3.6}} & $4$: \mbox{64.0 $\pm$ 5.4} & $4$: \mbox{\textbf{71.0 $\pm$ 5.1}} \\
        \multirow{-2}{*}{\tokenizerid{nrvqbase!40}{TF-NRVQ-S}} & $8$: \mbox{85.7 $\pm$ 4.0} & $8$: \mbox{\textbf{95.3 $\pm$ 2.4}} & $8$: \mbox{\textbf{92.3 $\pm$ 3.0}} & $8$: \mbox{88.7 $\pm$ 3.6} & $8$: \mbox{60.0 $\pm$ 5.5} & $8$: \mbox{\textbf{67.7 $\pm$ 5.3}} \\
        \rowcolor{black!4}  & $4$: \mbox{63.7 $\pm$ 5.4} & $4$: \mbox{\textbf{71.0 $\pm$ 5.2}} & $4$: \mbox{64.3 $\pm$ 5.4} & $4$: \mbox{\textbf{80.0 $\pm$ 4.5}} & $4$: \mbox{29.3 $\pm$ 5.1} & $4$: \mbox{\textbf{42.0 $\pm$ 5.6}} \\
        \rowcolor{black!4} \multirow{-2}{*}{\tokenizerid{nrvqbase!50}{TF-NRVQ-L}} & $8$: \mbox{54.7 $\pm$ 5.6} & $8$: \mbox{\textbf{63.3 $\pm$ 5.4}} & $8$: \mbox{63.3 $\pm$ 5.5} & $8$: \mbox{\textbf{77.0 $\pm$ 4.8}} & $8$: \mbox{19.0 $\pm$ 4.4} & $8$: \mbox{\textbf{28.3 $\pm$ 5.1}} \\
         & $4$: \mbox{0.0 $\pm$ 0.0} & $4$: \mbox{\textbf{93.3 $\pm$ 2.8}} & $4$: \mbox{0.0 $\pm$ 0.0} & $4$: \mbox{\textbf{82.0 $\pm$ 4.3}} & $4$: \mbox{0.0 $\pm$ 0.0} & $4$: \mbox{\textbf{57.3 $\pm$ 5.6}} \\
        \multirow{-2}{*}{\tokenizerid{nrvqbase!60}{CT-NRVQ-S}} & $8$: \mbox{0.0 $\pm$ 0.0} & $8$: \mbox{\textbf{92.3 $\pm$ 3.0}} & $8$: \mbox{0.0 $\pm$ 0.0} & $8$: \mbox{\textbf{95.3 $\pm$ 2.4}} & $8$: \mbox{0.0 $\pm$ 0.0} & $8$: \mbox{\textbf{64.3 $\pm$ 5.5}} \\
        \rowcolor{black!4}  & $4$: \mbox{0.0 $\pm$ 0.0} & $4$: \mbox{\textbf{69.7 $\pm$ 5.2}} & $4$: \mbox{0.0 $\pm$ 0.0} & $4$: \mbox{\textbf{77.3 $\pm$ 4.7}} & $4$: \mbox{0.0 $\pm$ 0.0} & $4$: \mbox{\textbf{30.0 $\pm$ 5.2}} \\
        \rowcolor{black!4} \multirow{-2}{*}{\tokenizerid{nrvqbase!70}{CT-NRVQ-L}} & $8$: \mbox{0.0 $\pm$ 0.0} & $8$: \mbox{\textbf{59.3 $\pm$ 5.6}} & $8$: \mbox{0.0 $\pm$ 0.0} & $8$: \mbox{\textbf{78.0 $\pm$ 4.7}} & $8$: \mbox{0.0 $\pm$ 0.0} & $8$: \mbox{\textbf{25.7 $\pm$ 4.9}} \\
         & $4$: \mbox{89.0 $\pm$ 3.5} & $4$: \mbox{\textbf{93.3 $\pm$ 2.8}} & $4$: \mbox{\textbf{93.3 $\pm$ 2.8}} & $4$: \mbox{83.7 $\pm$ 4.2} & $4$: \mbox{\textbf{57.3 $\pm$ 5.6}} & $4$: \mbox{\textbf{57.3 $\pm$ 5.6}} \\
        \multirow{-2}{*}{\tokenizerid{nrvqbase!80}{C2-NRVQ-S}} & $8$: \mbox{91.0 $\pm$ 3.2} & $8$: \mbox{\textbf{93.0 $\pm$ 2.9}} & $8$: \mbox{\textbf{93.3 $\pm$ 2.8}} & $8$: \mbox{89.7 $\pm$ 3.5} & $8$: \mbox{58.7 $\pm$ 5.6} & $8$: \mbox{\textbf{66.7 $\pm$ 5.4}} \\
        \rowcolor{black!4}  & $4$: \mbox{68.7 $\pm$ 5.3} & $4$: \mbox{\textbf{74.7 $\pm$ 4.9}} & $4$: \mbox{\textbf{71.7 $\pm$ 5.1}} & $4$: \mbox{71.0 $\pm$ 5.1} & $4$: \mbox{27.7 $\pm$ 5.1} & $4$: \mbox{\textbf{34.7 $\pm$ 5.4}} \\
        \rowcolor{black!4} \multirow{-2}{*}{\tokenizerid{nrvqbase!90}{C2-NRVQ-L}} & $8$: \mbox{51.7 $\pm$ 5.7} & $8$: \mbox{\textbf{65.3 $\pm$ 5.3}} & $8$: \mbox{64.7 $\pm$ 5.4} & $8$: \mbox{\textbf{67.0 $\pm$ 5.3}} & $8$: \mbox{24.7 $\pm$ 4.9} & $8$: \mbox{\textbf{29.0 $\pm$ 5.1}} \\
        \addlinespace[2pt]
         & $4$: \mbox{89.3 $\pm$ 3.5} & $4$: \mbox{\textbf{91.3 $\pm$ 3.2}} & $4$: \mbox{81.3 $\pm$ 4.4} & $4$: \mbox{\textbf{91.0 $\pm$ 3.2}} & $4$: \mbox{50.3 $\pm$ 5.7} & $4$: \mbox{\textbf{57.0 $\pm$ 5.6}} \\
        \multirow{-2}{*}{\tokenizerid{fsqbase!60}{TF-FSQ-S}} & $8$: \mbox{87.3 $\pm$ 3.7} & $8$: \mbox{\textbf{91.7 $\pm$ 3.1}} & $8$: \mbox{83.0 $\pm$ 4.2} & $8$: \mbox{\textbf{90.3 $\pm$ 3.4}} & $8$: \mbox{46.7 $\pm$ 5.7} & $8$: \mbox{\textbf{55.7 $\pm$ 5.7}} \\
        \rowcolor{black!4}  & $4$: \mbox{87.3 $\pm$ 3.7} & $4$: \mbox{\textbf{94.0 $\pm$ 2.7}} & $4$: \mbox{92.0 $\pm$ 3.1} & $4$: \mbox{\textbf{93.7 $\pm$ 2.8}} & $4$: \mbox{47.3 $\pm$ 5.6} & $4$: \mbox{\textbf{56.7 $\pm$ 5.6}} \\
        \rowcolor{black!4} \multirow{-2}{*}{\tokenizerid{fsqbase!85}{TF-FSQ-L}} & $8$: \mbox{87.0 $\pm$ 3.8} & $8$: \mbox{\textbf{92.3 $\pm$ 3.0}} & $8$: \mbox{89.7 $\pm$ 3.5} & $8$: \mbox{\textbf{93.3 $\pm$ 2.9}} & $8$: \mbox{43.3 $\pm$ 5.6} & $8$: \mbox{\textbf{56.0 $\pm$ 5.6}} \\
        \addlinespace[2pt]
         & $4$: \mbox{54.3 $\pm$ 5.6} & $4$: \mbox{\textbf{93.0 $\pm$ 2.9}} & $4$: \mbox{70.7 $\pm$ 5.1} & $4$: \mbox{\textbf{84.0 $\pm$ 4.1}} & $4$: \mbox{18.3 $\pm$ 4.4} & $4$: \mbox{\textbf{58.3 $\pm$ 5.7}} \\
        \multirow{-2}{*}{\tokenizerid{qincobase!40}{TF-QINCO-S}} & $8$: \mbox{50.7 $\pm$ 5.7} & $8$: \mbox{\textbf{87.0 $\pm$ 3.8}} & $8$: \mbox{64.0 $\pm$ 5.4} & $8$: \mbox{\textbf{82.3 $\pm$ 4.3}} & $8$: \mbox{12.0 $\pm$ 3.7} & $8$: \mbox{\textbf{50.0 $\pm$ 5.6}} \\
        \rowcolor{black!4}  & $4$: \mbox{44.7 $\pm$ 5.6} & $4$: \mbox{\textbf{88.7 $\pm$ 3.6}} & $4$: \mbox{66.0 $\pm$ 5.4} & $4$: \mbox{\textbf{82.7 $\pm$ 4.3}} & $4$: \mbox{13.7 $\pm$ 3.9} & $4$: \mbox{\textbf{58.0 $\pm$ 5.6}} \\
        \rowcolor{black!4} \multirow{-2}{*}{\tokenizerid{qincobase!50}{TF-QINCO-L}} & $8$: \mbox{36.3 $\pm$ 5.5} & $8$: \mbox{\textbf{85.7 $\pm$ 3.9}} & $8$: \mbox{55.7 $\pm$ 5.6} & $8$: \mbox{\textbf{85.0 $\pm$ 4.0}} & $8$: \mbox{6.3 $\pm$ 2.7} & $8$: \mbox{\textbf{47.3 $\pm$ 5.6}} \\
         & $4$: \mbox{61.0 $\pm$ 5.5} & $4$: \mbox{\textbf{85.0 $\pm$ 4.0}} & $4$: \mbox{74.0 $\pm$ 5.0} & $4$: \mbox{\textbf{82.0 $\pm$ 4.3}} & $4$: \mbox{28.0 $\pm$ 5.1} & $4$: \mbox{\textbf{50.3 $\pm$ 5.6}} \\
        \multirow{-2}{*}{\tokenizerid{qincobase!60}{CT-QINCO-S}} & $8$: \mbox{52.7 $\pm$ 5.7} & $8$: \mbox{\textbf{77.7 $\pm$ 4.7}} & $8$: \mbox{70.7 $\pm$ 5.2} & $8$: \mbox{\textbf{76.3 $\pm$ 4.8}} & $8$: \mbox{20.3 $\pm$ 4.6} & $8$: \mbox{\textbf{32.0 $\pm$ 5.3}} \\
        \rowcolor{black!4}  & $4$: \mbox{55.3 $\pm$ 5.7} & $4$: \mbox{\textbf{78.7 $\pm$ 4.7}} & $4$: \mbox{71.7 $\pm$ 5.1} & $4$: \mbox{\textbf{82.3 $\pm$ 4.4}} & $4$: \mbox{21.0 $\pm$ 4.6} & $4$: \mbox{\textbf{47.0 $\pm$ 5.6}} \\
        \rowcolor{black!4} \multirow{-2}{*}{\tokenizerid{qincobase!70}{CT-QINCO-L}} & $8$: \mbox{42.0 $\pm$ 5.6} & $8$: \mbox{\textbf{75.7 $\pm$ 4.8}} & $8$: \mbox{68.3 $\pm$ 5.3} & $8$: \mbox{\textbf{80.0 $\pm$ 4.5}} & $8$: \mbox{12.0 $\pm$ 3.7} & $8$: \mbox{\textbf{35.3 $\pm$ 5.4}} \\
         & $4$: \mbox{62.0 $\pm$ 5.5} & $4$: \mbox{\textbf{86.7 $\pm$ 3.8}} & $4$: \mbox{67.0 $\pm$ 5.4} & $4$: \mbox{\textbf{79.3 $\pm$ 4.6}} & $4$: \mbox{15.7 $\pm$ 4.1} & $4$: \mbox{\textbf{52.0 $\pm$ 5.7}} \\
        \multirow{-2}{*}{\tokenizerid{qincobase!80}{C2-QINCO-S}} & $8$: \mbox{46.0 $\pm$ 5.6} & $8$: \mbox{\textbf{83.3 $\pm$ 4.2}} & $8$: \mbox{64.3 $\pm$ 5.4} & $8$: \mbox{\textbf{79.0 $\pm$ 4.6}} & $8$: \mbox{12.7 $\pm$ 3.8} & $8$: \mbox{\textbf{49.7 $\pm$ 5.7}} \\
        \rowcolor{black!4}  & $4$: \mbox{62.3 $\pm$ 5.5} & $4$: \mbox{\textbf{65.7 $\pm$ 5.4}} & $4$: \mbox{64.0 $\pm$ 5.5} & $4$: \mbox{\textbf{77.3 $\pm$ 4.7}} & $4$: \mbox{19.0 $\pm$ 4.4} & $4$: \mbox{\textbf{33.3 $\pm$ 5.3}} \\
        \rowcolor{black!4} \multirow{-2}{*}{\tokenizerid{qincobase!90}{C2-QINCO-L}} & $8$: \mbox{52.3 $\pm$ 5.6} & $8$: \mbox{\textbf{64.0 $\pm$ 5.4}} & $8$: \mbox{60.3 $\pm$ 5.5} & $8$: \mbox{\textbf{72.0 $\pm$ 5.1}} & $8$: \mbox{7.7 $\pm$ 3.0} & $8$: \mbox{\textbf{20.7 $\pm$ 4.6}} \\
        \bottomrule
    \end{tabularx}
\end{table}

%% file: tables/experiments/simulation_results/appendix_robomimic_single_task.tex
\begin{table}[H]
    \centering
    \caption{RoboMimic results for policies trained on the corresponding single task, at the checkpoint with the best teacher-forced validation accuracy. Each tokenizer occupies two lines for $H_{\mathrm{exec}}=4$ and $8$ success rate $\pm$ 95\% confidence-interval half-width (percentage points). R.O. denotes Reconstruction Only; \textbf{bold} marks the better objective per task and horizon.}
    \label{tab:robomimic-single-task}
    \tablebodyfont
    \setlength{\tabcolsep}{1.0pt}
    \renewcommand{\arraystretch}{0.96}
    \begin{tabularx}{\linewidth}{@{}>{\raggedright\arraybackslash}m{0.16\linewidth} *{6}{>{\centering\arraybackslash}X}@{}}
        \toprule
        \multirow{2}{*}{Tokenizer} & \multicolumn{2}{c}{Can} & \multicolumn{2}{c}{Lift} & \multicolumn{2}{c}{Square} \\
        \cmidrule(lr){2-3}\cmidrule(lr){4-5}\cmidrule(l){6-7}
        & R.O. & \methodname{} & R.O. & \methodname{} & R.O. & \methodname{} \\
        \midrule
         & $4$: \mbox{90.3 $\pm$ 3.3} & $4$: \mbox{\textbf{90.7 $\pm$ 3.3}} & $4$: \mbox{93.0 $\pm$ 2.9} & $4$: \mbox{\textbf{99.0 $\pm$ 1.1}} & $4$: \mbox{64.3 $\pm$ 5.4} & $4$: \mbox{\textbf{74.7 $\pm$ 4.9}} \\
        \multirow{-2}{*}{\tokenizerid{vqbase!60}{TF-VQ-S}} & $8$: \mbox{\textbf{93.3 $\pm$ 2.8}} & $8$: \mbox{92.0 $\pm$ 3.1} & $8$: \mbox{93.0 $\pm$ 2.9} & $8$: \mbox{\textbf{99.3 $\pm$ 0.9}} & $8$: \mbox{\textbf{69.3 $\pm$ 5.2}} & $8$: \mbox{\textbf{69.3 $\pm$ 5.2}} \\
        \rowcolor{black!4}  & $4$: \mbox{93.3 $\pm$ 2.8} & $4$: \mbox{\textbf{93.7 $\pm$ 2.8}} & $4$: \mbox{96.3 $\pm$ 2.1} & $4$: \mbox{\textbf{98.7 $\pm$ 1.3}} & $4$: \mbox{50.7 $\pm$ 5.7} & $4$: \mbox{\textbf{63.7 $\pm$ 5.5}} \\
        \rowcolor{black!4} \multirow{-2}{*}{\tokenizerid{vqbase!85}{TF-VQ-L}} & $8$: \mbox{93.0 $\pm$ 2.9} & $8$: \mbox{\textbf{94.3 $\pm$ 2.6}} & $8$: \mbox{95.0 $\pm$ 2.5} & $8$: \mbox{\textbf{98.7 $\pm$ 1.3}} & $8$: \mbox{61.0 $\pm$ 5.5} & $8$: \mbox{\textbf{62.0 $\pm$ 5.4}} \\
        \addlinespace[2pt]
         & $4$: \mbox{\textbf{94.0 $\pm$ 2.7}} & $4$: \mbox{92.7 $\pm$ 2.9} & $4$: \mbox{84.7 $\pm$ 4.1} & $4$: \mbox{\textbf{92.3 $\pm$ 3.1}} & $4$: \mbox{63.7 $\pm$ 5.4} & $4$: \mbox{\textbf{66.7 $\pm$ 5.3}} \\
        \multirow{-2}{*}{\tokenizerid{rvqbase!40}{TF-RVQ-S}} & $8$: \mbox{93.0 $\pm$ 2.9} & $8$: \mbox{\textbf{93.7 $\pm$ 2.8}} & $8$: \mbox{87.7 $\pm$ 3.7} & $8$: \mbox{\textbf{96.7 $\pm$ 2.0}} & $8$: \mbox{\textbf{71.7 $\pm$ 5.1}} & $8$: \mbox{69.7 $\pm$ 5.2} \\
        \rowcolor{black!4}  & $4$: \mbox{58.3 $\pm$ 5.6} & $4$: \mbox{\textbf{83.3 $\pm$ 4.2}} & $4$: \mbox{86.3 $\pm$ 3.9} & $4$: \mbox{\textbf{91.0 $\pm$ 3.2}} & $4$: \mbox{51.0 $\pm$ 5.6} & $4$: \mbox{\textbf{53.3 $\pm$ 5.6}} \\
        \rowcolor{black!4} \multirow{-2}{*}{\tokenizerid{rvqbase!50}{TF-RVQ-L}} & $8$: \mbox{46.7 $\pm$ 5.7} & $8$: \mbox{\textbf{73.7 $\pm$ 4.9}} & $8$: \mbox{74.7 $\pm$ 4.9} & $8$: \mbox{\textbf{91.3 $\pm$ 3.1}} & $8$: \mbox{27.0 $\pm$ 4.9} & $8$: \mbox{\textbf{49.0 $\pm$ 5.7}} \\
         & $4$: \mbox{85.0 $\pm$ 4.1} & $4$: \mbox{\textbf{89.7 $\pm$ 3.4}} & $4$: \mbox{85.7 $\pm$ 4.0} & $4$: \mbox{\textbf{93.3 $\pm$ 2.8}} & $4$: \mbox{\textbf{69.0 $\pm$ 5.2}} & $4$: \mbox{65.0 $\pm$ 5.4} \\
        \multirow{-2}{*}{\tokenizerid{rvqbase!60}{CT-RVQ-S}} & $8$: \mbox{89.3 $\pm$ 3.5} & $8$: \mbox{\textbf{95.7 $\pm$ 2.3}} & $8$: \mbox{92.7 $\pm$ 3.0} & $8$: \mbox{\textbf{99.7 $\pm$ 0.7}} & $8$: \mbox{\textbf{73.0 $\pm$ 5.1}} & $8$: \mbox{71.3 $\pm$ 5.1} \\
        \rowcolor{black!4}  & $4$: \mbox{73.7 $\pm$ 5.0} & $4$: \mbox{\textbf{81.7 $\pm$ 4.4}} & $4$: \mbox{\textbf{93.0 $\pm$ 2.9}} & $4$: \mbox{89.0 $\pm$ 3.5} & $4$: \mbox{47.0 $\pm$ 5.6} & $4$: \mbox{\textbf{47.7 $\pm$ 5.7}} \\
        \rowcolor{black!4} \multirow{-2}{*}{\tokenizerid{rvqbase!70}{CT-RVQ-L}} & $8$: \mbox{60.0 $\pm$ 5.5} & $8$: \mbox{\textbf{73.3 $\pm$ 5.0}} & $8$: \mbox{88.3 $\pm$ 3.6} & $8$: \mbox{\textbf{90.3 $\pm$ 3.3}} & $8$: \mbox{33.0 $\pm$ 5.2} & $8$: \mbox{\textbf{41.3 $\pm$ 5.6}} \\
         & $4$: \mbox{87.0 $\pm$ 3.8} & $4$: \mbox{\textbf{94.7 $\pm$ 2.6}} & $4$: \mbox{\textbf{93.0 $\pm$ 2.9}} & $4$: \mbox{90.0 $\pm$ 3.4} & $4$: \mbox{52.0 $\pm$ 5.6} & $4$: \mbox{\textbf{58.3 $\pm$ 5.6}} \\
        \multirow{-2}{*}{\tokenizerid{rvqbase!80}{C2-RVQ-S}} & $8$: \mbox{87.0 $\pm$ 3.9} & $8$: \mbox{\textbf{94.3 $\pm$ 2.6}} & $8$: \mbox{96.0 $\pm$ 2.2} & $8$: \mbox{\textbf{97.0 $\pm$ 1.9}} & $8$: \mbox{\textbf{70.3 $\pm$ 5.2}} & $8$: \mbox{63.7 $\pm$ 5.4} \\
        \rowcolor{black!4}  & $4$: \mbox{70.0 $\pm$ 5.2} & $4$: \mbox{\textbf{80.3 $\pm$ 4.5}} & $4$: \mbox{\textbf{90.3 $\pm$ 3.4}} & $4$: \mbox{84.7 $\pm$ 4.1} & $4$: \mbox{48.0 $\pm$ 5.6} & $4$: \mbox{\textbf{52.0 $\pm$ 5.7}} \\
        \rowcolor{black!4} \multirow{-2}{*}{\tokenizerid{rvqbase!90}{C2-RVQ-L}} & $8$: \mbox{64.7 $\pm$ 5.4} & $8$: \mbox{\textbf{77.0 $\pm$ 4.8}} & $8$: \mbox{\textbf{89.7 $\pm$ 3.5}} & $8$: \mbox{85.3 $\pm$ 4.0} & $8$: \mbox{34.7 $\pm$ 5.4} & $8$: \mbox{\textbf{49.0 $\pm$ 5.7}} \\
        \addlinespace[2pt]
         & $4$: \mbox{94.3 $\pm$ 2.6} & $4$: \mbox{\textbf{94.7 $\pm$ 2.5}} & $4$: \mbox{94.7 $\pm$ 2.6} & $4$: \mbox{\textbf{95.7 $\pm$ 2.3}} & $4$: \mbox{68.0 $\pm$ 5.2} & $4$: \mbox{\textbf{76.0 $\pm$ 4.8}} \\
        \multirow{-2}{*}{\tokenizerid{nrvqbase!40}{TF-NRVQ-S}} & $8$: \mbox{92.0 $\pm$ 3.1} & $8$: \mbox{\textbf{95.3 $\pm$ 2.4}} & $8$: \mbox{\textbf{97.7 $\pm$ 1.7}} & $8$: \mbox{96.7 $\pm$ 2.0} & $8$: \mbox{67.7 $\pm$ 5.4} & $8$: \mbox{\textbf{77.0 $\pm$ 4.8}} \\
        \rowcolor{black!4}  & $4$: \mbox{64.7 $\pm$ 5.4} & $4$: \mbox{\textbf{74.0 $\pm$ 5.0}} & $4$: \mbox{68.7 $\pm$ 5.3} & $4$: \mbox{\textbf{86.0 $\pm$ 3.9}} & $4$: \mbox{33.3 $\pm$ 5.3} & $4$: \mbox{\textbf{34.3 $\pm$ 5.4}} \\
        \rowcolor{black!4} \multirow{-2}{*}{\tokenizerid{nrvqbase!50}{TF-NRVQ-L}} & $8$: \mbox{48.3 $\pm$ 5.6} & $8$: \mbox{\textbf{63.3 $\pm$ 5.5}} & $8$: \mbox{70.0 $\pm$ 5.3} & $8$: \mbox{\textbf{83.3 $\pm$ 4.2}} & $8$: \mbox{19.3 $\pm$ 4.5} & $8$: \mbox{\textbf{24.0 $\pm$ 4.8}} \\
         & $4$: \mbox{0.0 $\pm$ 0.0} & $4$: \mbox{\textbf{93.3 $\pm$ 2.8}} & $4$: \mbox{90.0 $\pm$ 3.4} & $4$: \mbox{\textbf{93.3 $\pm$ 2.8}} & $4$: \mbox{0.0 $\pm$ 0.0} & $4$: \mbox{\textbf{47.0 $\pm$ 5.7}} \\
        \multirow{-2}{*}{\tokenizerid{nrvqbase!60}{CT-NRVQ-S}} & $8$: \mbox{0.0 $\pm$ 0.0} & $8$: \mbox{\textbf{93.3 $\pm$ 2.8}} & $8$: \mbox{\textbf{94.7 $\pm$ 2.5}} & $8$: \mbox{89.3 $\pm$ 3.5} & $8$: \mbox{0.0 $\pm$ 0.0} & $8$: \mbox{\textbf{65.7 $\pm$ 5.4}} \\
        \rowcolor{black!4}  & $4$: \mbox{56.3 $\pm$ 5.7} & $4$: \mbox{\textbf{62.7 $\pm$ 5.5}} & $4$: \mbox{77.0 $\pm$ 4.8} & $4$: \mbox{\textbf{77.7 $\pm$ 4.7}} & $4$: \mbox{24.7 $\pm$ 5.0} & $4$: \mbox{\textbf{28.7 $\pm$ 5.1}} \\
        \rowcolor{black!4} \multirow{-2}{*}{\tokenizerid{nrvqbase!70}{CT-NRVQ-L}} & $8$: \mbox{37.3 $\pm$ 5.5} & $8$: \mbox{\textbf{46.3 $\pm$ 5.7}} & $8$: \mbox{\textbf{74.3 $\pm$ 4.9}} & $8$: \mbox{67.3 $\pm$ 5.2} & $8$: \mbox{14.3 $\pm$ 4.0} & $8$: \mbox{\textbf{21.0 $\pm$ 4.6}} \\
         & $4$: \mbox{\textbf{94.7 $\pm$ 2.5}} & $4$: \mbox{89.3 $\pm$ 3.5} & $4$: \mbox{93.0 $\pm$ 2.9} & $4$: \mbox{\textbf{94.7 $\pm$ 2.5}} & $4$: \mbox{61.0 $\pm$ 5.5} & $4$: \mbox{\textbf{65.7 $\pm$ 5.4}} \\
        \multirow{-2}{*}{\tokenizerid{nrvqbase!80}{C2-NRVQ-S}} & $8$: \mbox{87.7 $\pm$ 3.7} & $8$: \mbox{\textbf{91.3 $\pm$ 3.2}} & $8$: \mbox{\textbf{98.7 $\pm$ 1.3}} & $8$: \mbox{95.3 $\pm$ 2.4} & $8$: \mbox{59.3 $\pm$ 5.5} & $8$: \mbox{\textbf{63.3 $\pm$ 5.5}} \\
        \rowcolor{black!4}  & $4$: \mbox{67.0 $\pm$ 5.4} & $4$: \mbox{\textbf{70.7 $\pm$ 5.1}} & $4$: \mbox{\textbf{79.7 $\pm$ 4.5}} & $4$: \mbox{78.0 $\pm$ 4.7} & $4$: \mbox{\textbf{34.0 $\pm$ 5.3}} & $4$: \mbox{31.0 $\pm$ 5.2} \\
        \rowcolor{black!4} \multirow{-2}{*}{\tokenizerid{nrvqbase!90}{C2-NRVQ-L}} & $8$: \mbox{50.3 $\pm$ 5.6} & $8$: \mbox{\textbf{61.7 $\pm$ 5.5}} & $8$: \mbox{\textbf{76.0 $\pm$ 4.8}} & $8$: \mbox{74.7 $\pm$ 4.9} & $8$: \mbox{25.0 $\pm$ 4.8} & $8$: \mbox{\textbf{28.7 $\pm$ 5.1}} \\
        \addlinespace[2pt]
         & $4$: \mbox{94.0 $\pm$ 2.7} & $4$: \mbox{\textbf{95.3 $\pm$ 2.4}} & $4$: \mbox{93.7 $\pm$ 2.8} & $4$: \mbox{\textbf{98.0 $\pm$ 1.6}} & $4$: \mbox{\textbf{74.7 $\pm$ 5.0}} & $4$: \mbox{72.0 $\pm$ 5.1} \\
        \multirow{-2}{*}{\tokenizerid{fsqbase!60}{TF-FSQ-S}} & $8$: \mbox{\textbf{95.0 $\pm$ 2.5}} & $8$: \mbox{93.0 $\pm$ 2.9} & $8$: \mbox{97.3 $\pm$ 1.8} & $8$: \mbox{\textbf{98.3 $\pm$ 1.4}} & $8$: \mbox{\textbf{71.7 $\pm$ 5.1}} & $8$: \mbox{71.3 $\pm$ 5.1} \\
        \rowcolor{black!4}  & $4$: \mbox{93.7 $\pm$ 2.7} & $4$: \mbox{\textbf{94.7 $\pm$ 2.5}} & $4$: \mbox{96.3 $\pm$ 2.1} & $4$: \mbox{\textbf{97.7 $\pm$ 1.7}} & $4$: \mbox{63.3 $\pm$ 5.4} & $4$: \mbox{\textbf{65.0 $\pm$ 5.4}} \\
        \rowcolor{black!4} \multirow{-2}{*}{\tokenizerid{fsqbase!85}{TF-FSQ-L}} & $8$: \mbox{\textbf{91.7 $\pm$ 3.1}} & $8$: \mbox{91.0 $\pm$ 3.3} & $8$: \mbox{\textbf{99.3 $\pm$ 0.9}} & $8$: \mbox{98.3 $\pm$ 1.4} & $8$: \mbox{\textbf{73.3 $\pm$ 5.0}} & $8$: \mbox{66.3 $\pm$ 5.3} \\
        \addlinespace[2pt]
         & $4$: \mbox{70.0 $\pm$ 5.2} & $4$: \mbox{\textbf{80.7 $\pm$ 4.5}} & $4$: \mbox{77.7 $\pm$ 4.7} & $4$: \mbox{\textbf{92.0 $\pm$ 3.1}} & $4$: \mbox{36.3 $\pm$ 5.4} & $4$: \mbox{\textbf{57.3 $\pm$ 5.6}} \\
        \multirow{-2}{*}{\tokenizerid{qincobase!40}{TF-QINCO-S}} & $8$: \mbox{52.3 $\pm$ 5.7} & $8$: \mbox{\textbf{76.7 $\pm$ 4.7}} & $8$: \mbox{74.0 $\pm$ 4.9} & $8$: \mbox{\textbf{91.3 $\pm$ 3.2}} & $8$: \mbox{18.0 $\pm$ 4.3} & $8$: \mbox{\textbf{54.3 $\pm$ 5.6}} \\
        \rowcolor{black!4}  & $4$: \mbox{58.7 $\pm$ 5.6} & $4$: \mbox{\textbf{87.3 $\pm$ 3.8}} & $4$: \mbox{79.3 $\pm$ 4.6} & $4$: \mbox{\textbf{91.3 $\pm$ 3.2}} & $4$: \mbox{19.3 $\pm$ 4.5} & $4$: \mbox{\textbf{60.7 $\pm$ 5.5}} \\
        \rowcolor{black!4} \multirow{-2}{*}{\tokenizerid{qincobase!50}{TF-QINCO-L}} & $8$: \mbox{41.7 $\pm$ 5.6} & $8$: \mbox{\textbf{82.0 $\pm$ 4.3}} & $8$: \mbox{72.0 $\pm$ 5.1} & $8$: \mbox{\textbf{92.0 $\pm$ 3.1}} & $8$: \mbox{11.0 $\pm$ 3.5} & $8$: \mbox{\textbf{66.3 $\pm$ 5.4}} \\
         & $4$: \mbox{73.0 $\pm$ 5.0} & $4$: \mbox{\textbf{86.0 $\pm$ 3.9}} & $4$: \mbox{89.0 $\pm$ 3.6} & $4$: \mbox{\textbf{95.7 $\pm$ 2.3}} & $4$: \mbox{41.7 $\pm$ 5.6} & $4$: \mbox{\textbf{60.3 $\pm$ 5.6}} \\
        \multirow{-2}{*}{\tokenizerid{qincobase!60}{CT-QINCO-S}} & $8$: \mbox{57.0 $\pm$ 5.6} & $8$: \mbox{\textbf{87.7 $\pm$ 3.7}} & $8$: \mbox{88.3 $\pm$ 3.6} & $8$: \mbox{\textbf{97.0 $\pm$ 1.9}} & $8$: \mbox{28.0 $\pm$ 5.1} & $8$: \mbox{\textbf{53.7 $\pm$ 5.6}} \\
        \rowcolor{black!4}  & $4$: \mbox{62.0 $\pm$ 5.4} & $4$: \mbox{\textbf{87.7 $\pm$ 3.8}} & $4$: \mbox{83.0 $\pm$ 4.2} & $4$: \mbox{\textbf{88.3 $\pm$ 3.7}} & $4$: \mbox{24.3 $\pm$ 4.9} & $4$: \mbox{\textbf{58.0 $\pm$ 5.6}} \\
        \rowcolor{black!4} \multirow{-2}{*}{\tokenizerid{qincobase!70}{CT-QINCO-L}} & $8$: \mbox{51.3 $\pm$ 5.7} & $8$: \mbox{\textbf{84.7 $\pm$ 4.1}} & $8$: \mbox{79.7 $\pm$ 4.5} & $8$: \mbox{\textbf{86.7 $\pm$ 3.9}} & $8$: \mbox{14.7 $\pm$ 4.0} & $8$: \mbox{\textbf{55.0 $\pm$ 5.6}} \\
         & $4$: \mbox{67.7 $\pm$ 5.3} & $4$: \mbox{\textbf{92.3 $\pm$ 3.0}} & $4$: \mbox{\textbf{90.7 $\pm$ 3.3}} & $4$: \mbox{89.7 $\pm$ 3.5} & $4$: \mbox{14.0 $\pm$ 3.9} & $4$: \mbox{\textbf{51.7 $\pm$ 5.6}} \\
        \multirow{-2}{*}{\tokenizerid{qincobase!80}{C2-QINCO-S}} & $8$: \mbox{54.0 $\pm$ 5.7} & $8$: \mbox{\textbf{88.0 $\pm$ 3.6}} & $8$: \mbox{80.0 $\pm$ 4.6} & $8$: \mbox{\textbf{96.0 $\pm$ 2.2}} & $8$: \mbox{12.0 $\pm$ 3.7} & $8$: \mbox{\textbf{53.7 $\pm$ 5.6}} \\
        \rowcolor{black!4}  & $4$: \mbox{65.0 $\pm$ 5.4} & $4$: \mbox{\textbf{87.3 $\pm$ 3.7}} & $4$: \mbox{84.0 $\pm$ 4.2} & $4$: \mbox{\textbf{87.0 $\pm$ 3.8}} & $4$: \mbox{26.7 $\pm$ 5.1} & $4$: \mbox{\textbf{51.3 $\pm$ 5.7}} \\
        \rowcolor{black!4} \multirow{-2}{*}{\tokenizerid{qincobase!90}{C2-QINCO-L}} & $8$: \mbox{54.0 $\pm$ 5.7} & $8$: \mbox{\textbf{80.7 $\pm$ 4.4}} & $8$: \mbox{81.7 $\pm$ 4.4} & $8$: \mbox{\textbf{87.3 $\pm$ 3.7}} & $8$: \mbox{18.7 $\pm$ 4.4} & $8$: \mbox{\textbf{48.3 $\pm$ 5.7}} \\
        \bottomrule
    \end{tabularx}
\end{table}

%% file: tables/experiments/simulation_results/appendix_robomimic_family_ablation.tex
\begin{table}[H]
    \centering
    \caption{RoboMimic component ablation on the 20-configuration cohort. Entries give success (\%) at the checkpoint with the best teacher-forced validation accuracy at $H_{\mathrm{exec}}=4\,/\,8$, averaged equally over tasks and configurations within each family. \textbf{Bold} marks the best objective at each horizon. \texttt{CT-NRVQ-S/L} collapsed under Reconstruction-only training; these configurations are excluded from the aggregate calculation.}
    \label{tab:robomimic-family-ablation}
    \tablebodyfont
    \setlength{\tabcolsep}{4.0pt}
    \renewcommand{\arraystretch}{1.0}
    \begin{tabularx}{0.94\linewidth}{@{}l *{4}{>{\centering\arraybackslash}X}@{}}
        \toprule
        Quantizer & R.O. & \wotbr{} & \wotp{} & \methodname{} \\
        \midrule
        \tokenizerid{vqbase!80}{VQ} & 74.7\,/\,76.1 & 79.1\,/\,74.6 & \textbf{84.3}\,/\,\textbf{83.8} & 81.8\,/\,79.4 \\
        \tokenizerid{rvqbase!80}{RVQ} & 71.2\,/\,69.0 & 75.3\,/\,73.6 & \textbf{77.3}\,/\,74.1 & 75.5\,/\,\textbf{74.9} \\
        \tokenizerid{nrvqbase!80}{NRVQ} & 68.4\,/\,65.0 & \textbf{73.1}\,/\,68.0 & 71.3\,/\,67.9 & 71.2\,/\,\textbf{69.1} \\
        \tokenizerid{fsqbase!80}{FSQ} & 76.2\,/\,74.4 & 77.9\,/\,76.0 & \textbf{83.3}\,/\,\textbf{83.1} & 80.8\,/\,82.1 \\
        \tokenizerid{qincobase!80}{QinCo} & 50.8\,/\,40.8 & 63.7\,/\,57.2 & 65.2\,/\,55.7 & \textbf{71.1}\,/\,\textbf{66.2} \\
        \midrule
        \textbf{Overall} & 65.4\,/\,61.0 & 72.0\,/\,67.9 & 73.8\,/\,69.2 & \textbf{74.5}\,/\,\textbf{72.3} \\
        \bottomrule
    \end{tabularx}
\end{table}

%% file: tables/experiments/simulation_results/appendix_robomimic_multitask_best_can.tex
\begin{table}[H]
    \centering
    \caption{Multi-task RoboMimic ablation on Can for all 22 configurations, using the checkpoint with the best teacher-forced validation accuracy. The two rows per tokenizer report $H_{\mathrm{exec}}=4$ and $8$ success (\%) $\pm$ 95\% confidence-interval half-width. TP and TBR denote Token Perturbation and Token Balance Regularization; \textbf{bold} marks the best treatment per horizon.}
    \label{tab:robomimic-multitask-best-can}
    \tablebodyfont
    \setlength{\tabcolsep}{2.2pt}
    \renewcommand{\arraystretch}{0.96}
    \begin{tabularx}{\linewidth}{@{}>{\raggedright\arraybackslash}m{0.18\linewidth} *{4}{>{\centering\arraybackslash}X}@{}}
        \toprule
        Tokenizer & Reconstruction-only & \wotbr{} & \wotp{} & \methodname{} \\
        \midrule
         & $4$: \mbox{86.7 $\pm$ 3.8} & $4$: \mbox{88.0 $\pm$ 3.7} & $4$: \mbox{94.3 $\pm$ 2.6} & $4$: \mbox{\textbf{95.3 $\pm$ 2.4}} \\
        \multirow{-2}{*}{\tokenizerid{vqbase!60}{TF-VQ-S}} & $8$: \mbox{91.0 $\pm$ 3.2} & $8$: \mbox{89.3 $\pm$ 3.5} & $8$: \mbox{89.0 $\pm$ 3.6} & $8$: \mbox{\textbf{95.3 $\pm$ 2.4}} \\
        \rowcolor{black!4}  & $4$: \mbox{87.7 $\pm$ 3.7} & $4$: \mbox{89.3 $\pm$ 3.5} & $4$: \mbox{\textbf{94.0 $\pm$ 2.7}} & $4$: \mbox{90.7 $\pm$ 3.3} \\
        \rowcolor{black!4} \multirow{-2}{*}{\tokenizerid{vqbase!85}{TF-VQ-L}} & $8$: \mbox{86.0 $\pm$ 3.9} & $8$: \mbox{80.0 $\pm$ 4.5} & $8$: \mbox{\textbf{94.7 $\pm$ 2.5}} & $8$: \mbox{86.3 $\pm$ 3.9} \\
        \addlinespace[2pt]
         & $4$: \mbox{86.0 $\pm$ 3.9} & $4$: \mbox{97.0 $\pm$ 1.9} & $4$: \mbox{\textbf{98.3 $\pm$ 1.4}} & $4$: \mbox{92.0 $\pm$ 3.1} \\
        \multirow{-2}{*}{\tokenizerid{rvqbase!40}{TF-RVQ-S}} & $8$: \mbox{90.7 $\pm$ 3.3} & $8$: \mbox{95.0 $\pm$ 2.5} & $8$: \mbox{\textbf{95.3 $\pm$ 2.4}} & $8$: \mbox{94.0 $\pm$ 2.7} \\
        \rowcolor{black!4}  & $4$: \mbox{73.0 $\pm$ 5.1} & $4$: \mbox{80.7 $\pm$ 4.5} & $4$: \mbox{82.3 $\pm$ 4.4} & $4$: \mbox{\textbf{86.7 $\pm$ 3.9}} \\
        \rowcolor{black!4} \multirow{-2}{*}{\tokenizerid{rvqbase!50}{TF-RVQ-L}} & $8$: \mbox{63.7 $\pm$ 5.4} & $8$: \mbox{74.7 $\pm$ 4.9} & $8$: \mbox{74.7 $\pm$ 4.9} & $8$: \mbox{\textbf{80.7 $\pm$ 4.5}} \\
         & $4$: \mbox{91.0 $\pm$ 3.3} & $4$: \mbox{94.3 $\pm$ 2.6} & $4$: \mbox{\textbf{96.7 $\pm$ 2.0}} & $4$: \mbox{95.0 $\pm$ 2.5} \\
        \multirow{-2}{*}{\tokenizerid{rvqbase!60}{CT-RVQ-S}} & $8$: \mbox{91.0 $\pm$ 3.3} & $8$: \mbox{95.7 $\pm$ 2.3} & $8$: \mbox{\textbf{96.0 $\pm$ 2.2}} & $8$: \mbox{\textbf{96.0 $\pm$ 2.2}} \\
        \rowcolor{black!4}  & $4$: \mbox{75.3 $\pm$ 4.8} & $4$: \mbox{78.7 $\pm$ 4.6} & $4$: \mbox{\textbf{85.7 $\pm$ 4.0}} & $4$: \mbox{80.0 $\pm$ 4.5} \\
        \rowcolor{black!4} \multirow{-2}{*}{\tokenizerid{rvqbase!70}{CT-RVQ-L}} & $8$: \mbox{69.0 $\pm$ 5.2} & $8$: \mbox{74.7 $\pm$ 5.0} & $8$: \mbox{72.3 $\pm$ 5.0} & $8$: \mbox{\textbf{77.7 $\pm$ 4.7}} \\
         & $4$: \mbox{94.0 $\pm$ 2.7} & $4$: \mbox{92.7 $\pm$ 2.9} & $4$: \mbox{94.3 $\pm$ 2.6} & $4$: \mbox{\textbf{97.7 $\pm$ 1.7}} \\
        \multirow{-2}{*}{\tokenizerid{rvqbase!80}{C2-RVQ-S}} & $8$: \mbox{92.0 $\pm$ 3.1} & $8$: \mbox{\textbf{95.7 $\pm$ 2.3}} & $8$: \mbox{93.7 $\pm$ 2.8} & $8$: \mbox{94.3 $\pm$ 2.6} \\
        \rowcolor{black!4}  & $4$: \mbox{71.7 $\pm$ 5.1} & $4$: \mbox{82.3 $\pm$ 4.4} & $4$: \mbox{\textbf{82.7 $\pm$ 4.3}} & $4$: \mbox{80.0 $\pm$ 4.5} \\
        \rowcolor{black!4} \multirow{-2}{*}{\tokenizerid{rvqbase!90}{C2-RVQ-L}} & $8$: \mbox{66.0 $\pm$ 5.3} & $8$: \mbox{71.3 $\pm$ 5.1} & $8$: \mbox{\textbf{74.0 $\pm$ 5.0}} & $8$: \mbox{\textbf{74.0 $\pm$ 5.0}} \\
        \addlinespace[2pt]
         & $4$: \mbox{89.3 $\pm$ 3.5} & $4$: \mbox{\textbf{95.0 $\pm$ 2.5}} & $4$: \mbox{94.0 $\pm$ 2.7} & $4$: \mbox{94.3 $\pm$ 2.6} \\
        \multirow{-2}{*}{\tokenizerid{nrvqbase!40}{TF-NRVQ-S}} & $8$: \mbox{90.0 $\pm$ 3.4} & $8$: \mbox{94.0 $\pm$ 2.7} & $8$: \mbox{\textbf{95.7 $\pm$ 2.3}} & $8$: \mbox{\textbf{95.7 $\pm$ 2.3}} \\
        \rowcolor{black!4}  & $4$: \mbox{70.3 $\pm$ 5.1} & $4$: \mbox{\textbf{81.0 $\pm$ 4.4}} & $4$: \mbox{60.7 $\pm$ 5.5} & $4$: \mbox{73.0 $\pm$ 5.0} \\
        \rowcolor{black!4} \multirow{-2}{*}{\tokenizerid{nrvqbase!50}{TF-NRVQ-L}} & $8$: \mbox{60.0 $\pm$ 5.5} & $8$: \mbox{\textbf{67.7 $\pm$ 5.2}} & $8$: \mbox{45.3 $\pm$ 5.6} & $8$: \mbox{65.3 $\pm$ 5.4} \\
         & $4$: \mbox{0.0 $\pm$ 0.0} & $4$: \mbox{91.3 $\pm$ 3.2} & $4$: \mbox{\textbf{94.3 $\pm$ 2.6}} & $4$: \mbox{91.0 $\pm$ 3.2} \\
        \multirow{-2}{*}{\tokenizerid{nrvqbase!60}{CT-NRVQ-S}} & $8$: \mbox{0.0 $\pm$ 0.0} & $8$: \mbox{\textbf{94.3 $\pm$ 2.6}} & $8$: \mbox{94.0 $\pm$ 2.7} & $8$: \mbox{92.3 $\pm$ 3.0} \\
        \rowcolor{black!4}  & $4$: \mbox{0.0 $\pm$ 0.0} & $4$: \mbox{\textbf{75.3 $\pm$ 4.9}} & $4$: \mbox{72.3 $\pm$ 5.1} & $4$: \mbox{73.0 $\pm$ 5.0} \\
        \rowcolor{black!4} \multirow{-2}{*}{\tokenizerid{nrvqbase!70}{CT-NRVQ-L}} & $8$: \mbox{0.0 $\pm$ 0.0} & $8$: \mbox{\textbf{64.0 $\pm$ 5.6}} & $8$: \mbox{61.7 $\pm$ 5.5} & $8$: \mbox{59.0 $\pm$ 5.5} \\
         & $4$: \mbox{91.7 $\pm$ 3.1} & $4$: \mbox{89.3 $\pm$ 3.5} & $4$: \mbox{\textbf{95.3 $\pm$ 2.4}} & $4$: \mbox{89.3 $\pm$ 3.5} \\
        \multirow{-2}{*}{\tokenizerid{nrvqbase!80}{C2-NRVQ-S}} & $8$: \mbox{92.3 $\pm$ 3.0} & $8$: \mbox{91.7 $\pm$ 3.2} & $8$: \mbox{\textbf{96.3 $\pm$ 2.1}} & $8$: \mbox{93.3 $\pm$ 2.8} \\
        \rowcolor{black!4}  & $4$: \mbox{68.7 $\pm$ 5.3} & $4$: \mbox{72.3 $\pm$ 5.0} & $4$: \mbox{\textbf{76.7 $\pm$ 4.8}} & $4$: \mbox{\textbf{76.7 $\pm$ 4.8}} \\
        \rowcolor{black!4} \multirow{-2}{*}{\tokenizerid{nrvqbase!90}{C2-NRVQ-L}} & $8$: \mbox{54.0 $\pm$ 5.7} & $8$: \mbox{\textbf{69.3 $\pm$ 5.3}} & $8$: \mbox{65.0 $\pm$ 5.4} & $8$: \mbox{62.3 $\pm$ 5.5} \\
        \addlinespace[2pt]
         & $4$: \mbox{91.0 $\pm$ 3.2} & $4$: \mbox{90.7 $\pm$ 3.3} & $4$: \mbox{\textbf{95.3 $\pm$ 2.4}} & $4$: \mbox{89.0 $\pm$ 3.5} \\
        \multirow{-2}{*}{\tokenizerid{fsqbase!60}{TF-FSQ-S}} & $8$: \mbox{89.7 $\pm$ 3.5} & $8$: \mbox{90.3 $\pm$ 3.3} & $8$: \mbox{\textbf{94.3 $\pm$ 2.6}} & $8$: \mbox{92.3 $\pm$ 3.0} \\
        \rowcolor{black!4}  & $4$: \mbox{91.0 $\pm$ 3.2} & $4$: \mbox{93.3 $\pm$ 2.8} & $4$: \mbox{92.7 $\pm$ 2.9} & $4$: \mbox{\textbf{94.0 $\pm$ 2.7}} \\
        \rowcolor{black!4} \multirow{-2}{*}{\tokenizerid{fsqbase!85}{TF-FSQ-L}} & $8$: \mbox{87.3 $\pm$ 3.8} & $8$: \mbox{91.0 $\pm$ 3.2} & $8$: \mbox{92.0 $\pm$ 3.1} & $8$: \mbox{\textbf{92.7 $\pm$ 3.0}} \\
        \addlinespace[2pt]
         & $4$: \mbox{55.3 $\pm$ 5.7} & $4$: \mbox{75.7 $\pm$ 4.9} & $4$: \mbox{85.3 $\pm$ 3.9} & $4$: \mbox{\textbf{89.7 $\pm$ 3.4}} \\
        \multirow{-2}{*}{\tokenizerid{qincobase!40}{TF-QINCO-S}} & $8$: \mbox{47.3 $\pm$ 5.6} & $8$: \mbox{73.3 $\pm$ 5.0} & $8$: \mbox{66.0 $\pm$ 5.3} & $8$: \mbox{\textbf{86.3 $\pm$ 3.9}} \\
        \rowcolor{black!4}  & $4$: \mbox{51.0 $\pm$ 5.6} & $4$: \mbox{68.0 $\pm$ 5.2} & $4$: \mbox{64.0 $\pm$ 5.4} & $4$: \mbox{\textbf{86.3 $\pm$ 3.9}} \\
        \rowcolor{black!4} \multirow{-2}{*}{\tokenizerid{qincobase!50}{TF-QINCO-L}} & $8$: \mbox{32.7 $\pm$ 5.4} & $8$: \mbox{51.3 $\pm$ 5.6} & $8$: \mbox{47.7 $\pm$ 5.6} & $8$: \mbox{\textbf{85.7 $\pm$ 4.0}} \\
         & $4$: \mbox{67.0 $\pm$ 5.3} & $4$: \mbox{80.7 $\pm$ 4.4} & $4$: \mbox{75.7 $\pm$ 4.8} & $4$: \mbox{\textbf{86.0 $\pm$ 4.0}} \\
        \multirow{-2}{*}{\tokenizerid{qincobase!60}{CT-QINCO-S}} & $8$: \mbox{56.3 $\pm$ 5.7} & $8$: \mbox{\textbf{77.7 $\pm$ 4.7}} & $8$: \mbox{58.7 $\pm$ 5.6} & $8$: \mbox{77.3 $\pm$ 4.7} \\
        \rowcolor{black!4}  & $4$: \mbox{59.0 $\pm$ 5.6} & $4$: \mbox{69.3 $\pm$ 5.2} & $4$: \mbox{77.3 $\pm$ 4.7} & $4$: \mbox{\textbf{79.7 $\pm$ 4.6}} \\
        \rowcolor{black!4} \multirow{-2}{*}{\tokenizerid{qincobase!70}{CT-QINCO-L}} & $8$: \mbox{45.0 $\pm$ 5.6} & $8$: \mbox{62.0 $\pm$ 5.4} & $8$: \mbox{60.0 $\pm$ 5.6} & $8$: \mbox{\textbf{74.7 $\pm$ 4.9}} \\
         & $4$: \mbox{59.7 $\pm$ 5.5} & $4$: \mbox{79.3 $\pm$ 4.5} & $4$: \mbox{85.3 $\pm$ 4.0} & $4$: \mbox{\textbf{91.3 $\pm$ 3.2}} \\
        \multirow{-2}{*}{\tokenizerid{qincobase!80}{C2-QINCO-S}} & $8$: \mbox{50.0 $\pm$ 5.7} & $8$: \mbox{73.3 $\pm$ 5.0} & $8$: \mbox{75.7 $\pm$ 4.9} & $8$: \mbox{\textbf{87.7 $\pm$ 3.8}} \\
        \rowcolor{black!4}  & $4$: \mbox{61.7 $\pm$ 5.5} & $4$: \mbox{73.0 $\pm$ 5.0} & $4$: \mbox{\textbf{79.3 $\pm$ 4.6}} & $4$: \mbox{73.3 $\pm$ 4.9} \\
        \rowcolor{black!4} \multirow{-2}{*}{\tokenizerid{qincobase!90}{C2-QINCO-L}} & $8$: \mbox{47.0 $\pm$ 5.6} & $8$: \mbox{66.7 $\pm$ 5.4} & $8$: \mbox{\textbf{68.3 $\pm$ 5.2}} & $8$: \mbox{61.7 $\pm$ 5.5} \\
        \bottomrule
    \end{tabularx}
\end{table}

%% file: tables/experiments/simulation_results/appendix_robomimic_multitask_best_lift.tex
\begin{table}[H]
    \centering
    \caption{Multi-task RoboMimic ablation on Lift for all 22 configurations, using the checkpoint with the best teacher-forced validation accuracy. The two rows per tokenizer report $H_{\mathrm{exec}}=4$ and $8$ success (\%) $\pm$ 95\% confidence-interval half-width. TP and TBR denote Token Perturbation and Token Balance Regularization; \textbf{bold} marks the best treatment per horizon.}
    \label{tab:robomimic-multitask-best-lift}
    \tablebodyfont
    \setlength{\tabcolsep}{2.2pt}
    \renewcommand{\arraystretch}{0.96}
    \begin{tabularx}{\linewidth}{@{}>{\raggedright\arraybackslash}m{0.18\linewidth} *{4}{>{\centering\arraybackslash}X}@{}}
        \toprule
        Tokenizer & Reconstruction-only & \wotbr{} & \wotp{} & \methodname{} \\
        \midrule
         & $4$: \mbox{82.3 $\pm$ 4.3} & $4$: \mbox{90.7 $\pm$ 3.3} & $4$: \mbox{97.3 $\pm$ 1.8} & $4$: \mbox{\textbf{99.0 $\pm$ 1.1}} \\
        \multirow{-2}{*}{\tokenizerid{vqbase!60}{TF-VQ-S}} & $8$: \mbox{85.7 $\pm$ 4.0} & $8$: \mbox{86.0 $\pm$ 3.9} & $8$: \mbox{98.7 $\pm$ 1.3} & $8$: \mbox{\textbf{100.0 $\pm$ 0.0}} \\
        \rowcolor{black!4}  & $4$: \mbox{93.7 $\pm$ 2.7} & $4$: \mbox{85.3 $\pm$ 4.0} & $4$: \mbox{\textbf{99.7 $\pm$ 0.7}} & $4$: \mbox{98.7 $\pm$ 1.3} \\
        \rowcolor{black!4} \multirow{-2}{*}{\tokenizerid{vqbase!85}{TF-VQ-L}} & $8$: \mbox{87.0 $\pm$ 3.8} & $8$: \mbox{82.7 $\pm$ 4.3} & $8$: \mbox{\textbf{100.0 $\pm$ 0.0}} & $8$: \mbox{98.3 $\pm$ 1.4} \\
        \addlinespace[2pt]
         & $4$: \mbox{91.3 $\pm$ 3.2} & $4$: \mbox{92.0 $\pm$ 3.1} & $4$: \mbox{\textbf{98.3 $\pm$ 1.4}} & $4$: \mbox{92.0 $\pm$ 3.1} \\
        \multirow{-2}{*}{\tokenizerid{rvqbase!40}{TF-RVQ-S}} & $8$: \mbox{90.3 $\pm$ 3.3} & $8$: \mbox{95.7 $\pm$ 2.3} & $8$: \mbox{\textbf{98.3 $\pm$ 1.5}} & $8$: \mbox{94.3 $\pm$ 2.6} \\
        \rowcolor{black!4}  & $4$: \mbox{70.7 $\pm$ 5.2} & $4$: \mbox{81.7 $\pm$ 4.4} & $4$: \mbox{\textbf{89.0 $\pm$ 3.5}} & $4$: \mbox{79.3 $\pm$ 4.6} \\
        \rowcolor{black!4} \multirow{-2}{*}{\tokenizerid{rvqbase!50}{TF-RVQ-L}} & $8$: \mbox{77.0 $\pm$ 4.7} & $8$: \mbox{79.7 $\pm$ 4.6} & $8$: \mbox{\textbf{88.0 $\pm$ 3.7}} & $8$: \mbox{81.0 $\pm$ 4.4} \\
         & $4$: \mbox{92.0 $\pm$ 3.1} & $4$: \mbox{93.7 $\pm$ 2.7} & $4$: \mbox{\textbf{100.0 $\pm$ 0.0}} & $4$: \mbox{88.7 $\pm$ 3.6} \\
        \multirow{-2}{*}{\tokenizerid{rvqbase!60}{CT-RVQ-S}} & $8$: \mbox{92.3 $\pm$ 3.0} & $8$: \mbox{93.3 $\pm$ 2.8} & $8$: \mbox{\textbf{99.7 $\pm$ 0.7}} & $8$: \mbox{95.0 $\pm$ 2.4} \\
        \rowcolor{black!4}  & $4$: \mbox{74.7 $\pm$ 4.9} & $4$: \mbox{80.3 $\pm$ 4.5} & $4$: \mbox{\textbf{95.3 $\pm$ 2.4}} & $4$: \mbox{80.0 $\pm$ 4.5} \\
        \rowcolor{black!4} \multirow{-2}{*}{\tokenizerid{rvqbase!70}{CT-RVQ-L}} & $8$: \mbox{76.0 $\pm$ 4.9} & $8$: \mbox{80.3 $\pm$ 4.5} & $8$: \mbox{\textbf{94.3 $\pm$ 2.6}} & $8$: \mbox{80.3 $\pm$ 4.5} \\
         & $4$: \mbox{97.3 $\pm$ 1.8} & $4$: \mbox{89.7 $\pm$ 3.4} & $4$: \mbox{\textbf{98.3 $\pm$ 1.4}} & $4$: \mbox{90.7 $\pm$ 3.3} \\
        \multirow{-2}{*}{\tokenizerid{rvqbase!80}{C2-RVQ-S}} & $8$: \mbox{96.0 $\pm$ 2.2} & $8$: \mbox{91.0 $\pm$ 3.2} & $8$: \mbox{\textbf{100.0 $\pm$ 0.0}} & $8$: \mbox{93.0 $\pm$ 2.9} \\
        \rowcolor{black!4}  & $4$: \mbox{70.7 $\pm$ 5.2} & $4$: \mbox{79.0 $\pm$ 4.6} & $4$: \mbox{\textbf{93.7 $\pm$ 2.8}} & $4$: \mbox{73.7 $\pm$ 5.0} \\
        \rowcolor{black!4} \multirow{-2}{*}{\tokenizerid{rvqbase!90}{C2-RVQ-L}} & $8$: \mbox{75.3 $\pm$ 4.9} & $8$: \mbox{83.0 $\pm$ 4.3} & $8$: \mbox{\textbf{89.7 $\pm$ 3.5}} & $8$: \mbox{74.7 $\pm$ 4.9} \\
        \addlinespace[2pt]
         & $4$: \mbox{89.0 $\pm$ 3.5} & $4$: \mbox{97.3 $\pm$ 1.8} & $4$: \mbox{\textbf{98.3 $\pm$ 1.4}} & $4$: \mbox{88.7 $\pm$ 3.6} \\
        \multirow{-2}{*}{\tokenizerid{nrvqbase!40}{TF-NRVQ-S}} & $8$: \mbox{88.3 $\pm$ 3.6} & $8$: \mbox{98.0 $\pm$ 1.6} & $8$: \mbox{\textbf{100.0 $\pm$ 0.0}} & $8$: \mbox{88.0 $\pm$ 3.6} \\
        \rowcolor{black!4}  & $4$: \mbox{65.3 $\pm$ 5.3} & $4$: \mbox{\textbf{77.0 $\pm$ 4.8}} & $4$: \mbox{\textbf{77.0 $\pm$ 4.8}} & $4$: \mbox{71.7 $\pm$ 5.1} \\
        \rowcolor{black!4} \multirow{-2}{*}{\tokenizerid{nrvqbase!50}{TF-NRVQ-L}} & $8$: \mbox{64.0 $\pm$ 5.4} & $8$: \mbox{66.3 $\pm$ 5.4} & $8$: \mbox{\textbf{80.0 $\pm$ 4.6}} & $8$: \mbox{75.0 $\pm$ 4.9} \\
         & $4$: \mbox{0.0 $\pm$ 0.0} & $4$: \mbox{87.7 $\pm$ 3.7} & $4$: \mbox{\textbf{98.0 $\pm$ 1.6}} & $4$: \mbox{88.3 $\pm$ 3.6} \\
        \multirow{-2}{*}{\tokenizerid{nrvqbase!60}{CT-NRVQ-S}} & $8$: \mbox{0.0 $\pm$ 0.0} & $8$: \mbox{94.7 $\pm$ 2.5} & $8$: \mbox{\textbf{100.0 $\pm$ 0.0}} & $8$: \mbox{96.7 $\pm$ 2.0} \\
        \rowcolor{black!4}  & $4$: \mbox{0.0 $\pm$ 0.0} & $4$: \mbox{74.3 $\pm$ 4.9} & $4$: \mbox{\textbf{83.7 $\pm$ 4.1}} & $4$: \mbox{79.3 $\pm$ 4.6} \\
        \rowcolor{black!4} \multirow{-2}{*}{\tokenizerid{nrvqbase!70}{CT-NRVQ-L}} & $8$: \mbox{0.0 $\pm$ 0.0} & $8$: \mbox{76.0 $\pm$ 4.8} & $8$: \mbox{\textbf{86.7 $\pm$ 3.9}} & $8$: \mbox{79.0 $\pm$ 4.6} \\
         & $4$: \mbox{90.0 $\pm$ 3.4} & $4$: \mbox{89.3 $\pm$ 3.4} & $4$: \mbox{\textbf{99.3 $\pm$ 0.9}} & $4$: \mbox{85.0 $\pm$ 4.0} \\
        \multirow{-2}{*}{\tokenizerid{nrvqbase!80}{C2-NRVQ-S}} & $8$: \mbox{95.3 $\pm$ 2.4} & $8$: \mbox{86.3 $\pm$ 3.9} & $8$: \mbox{\textbf{99.7 $\pm$ 0.7}} & $8$: \mbox{87.7 $\pm$ 3.7} \\
        \rowcolor{black!4}  & $4$: \mbox{71.7 $\pm$ 5.1} & $4$: \mbox{80.0 $\pm$ 4.5} & $4$: \mbox{\textbf{88.7 $\pm$ 3.6}} & $4$: \mbox{72.7 $\pm$ 5.1} \\
        \rowcolor{black!4} \multirow{-2}{*}{\tokenizerid{nrvqbase!90}{C2-NRVQ-L}} & $8$: \mbox{68.7 $\pm$ 5.2} & $8$: \mbox{70.3 $\pm$ 5.1} & $8$: \mbox{\textbf{86.7 $\pm$ 3.8}} & $8$: \mbox{69.0 $\pm$ 5.2} \\
        \addlinespace[2pt]
         & $4$: \mbox{83.3 $\pm$ 4.2} & $4$: \mbox{94.0 $\pm$ 2.7} & $4$: \mbox{\textbf{96.7 $\pm$ 2.1}} & $4$: \mbox{91.7 $\pm$ 3.1} \\
        \multirow{-2}{*}{\tokenizerid{fsqbase!60}{TF-FSQ-S}} & $8$: \mbox{85.0 $\pm$ 4.0} & $8$: \mbox{91.0 $\pm$ 3.2} & $8$: \mbox{\textbf{98.7 $\pm$ 1.3}} & $8$: \mbox{93.0 $\pm$ 2.9} \\
        \rowcolor{black!4}  & $4$: \mbox{88.7 $\pm$ 3.6} & $4$: \mbox{87.7 $\pm$ 3.7} & $4$: \mbox{\textbf{96.7 $\pm$ 2.0}} & $4$: \mbox{91.0 $\pm$ 3.3} \\
        \rowcolor{black!4} \multirow{-2}{*}{\tokenizerid{fsqbase!85}{TF-FSQ-L}} & $8$: \mbox{84.7 $\pm$ 4.1} & $8$: \mbox{88.7 $\pm$ 3.6} & $8$: \mbox{\textbf{98.0 $\pm$ 1.6}} & $8$: \mbox{94.0 $\pm$ 2.7} \\
        \addlinespace[2pt]
         & $4$: \mbox{62.3 $\pm$ 5.5} & $4$: \mbox{76.7 $\pm$ 4.8} & $4$: \mbox{\textbf{91.7 $\pm$ 3.1}} & $4$: \mbox{83.7 $\pm$ 4.2} \\
        \multirow{-2}{*}{\tokenizerid{qincobase!40}{TF-QINCO-S}} & $8$: \mbox{61.3 $\pm$ 5.5} & $8$: \mbox{72.3 $\pm$ 5.1} & $8$: \mbox{\textbf{89.3 $\pm$ 3.5}} & $8$: \mbox{84.0 $\pm$ 4.2} \\
        \rowcolor{black!4}  & $4$: \mbox{64.7 $\pm$ 5.5} & $4$: \mbox{72.0 $\pm$ 5.1} & $4$: \mbox{80.7 $\pm$ 4.5} & $4$: \mbox{\textbf{82.7 $\pm$ 4.3}} \\
        \rowcolor{black!4} \multirow{-2}{*}{\tokenizerid{qincobase!50}{TF-QINCO-L}} & $8$: \mbox{56.3 $\pm$ 5.6} & $8$: \mbox{70.0 $\pm$ 5.2} & $8$: \mbox{73.0 $\pm$ 5.0} & $8$: \mbox{\textbf{82.0 $\pm$ 4.3}} \\
         & $4$: \mbox{74.0 $\pm$ 4.9} & $4$: \mbox{82.3 $\pm$ 4.3} & $4$: \mbox{\textbf{93.3 $\pm$ 2.8}} & $4$: \mbox{83.7 $\pm$ 4.2} \\
        \multirow{-2}{*}{\tokenizerid{qincobase!60}{CT-QINCO-S}} & $8$: \mbox{65.3 $\pm$ 5.5} & $8$: \mbox{78.0 $\pm$ 4.7} & $8$: \mbox{\textbf{89.3 $\pm$ 3.5}} & $8$: \mbox{77.0 $\pm$ 4.8} \\
        \rowcolor{black!4}  & $4$: \mbox{73.7 $\pm$ 5.0} & $4$: \mbox{74.3 $\pm$ 4.9} & $4$: \mbox{\textbf{86.7 $\pm$ 3.8}} & $4$: \mbox{85.7 $\pm$ 4.0} \\
        \rowcolor{black!4} \multirow{-2}{*}{\tokenizerid{qincobase!70}{CT-QINCO-L}} & $8$: \mbox{66.3 $\pm$ 5.3} & $8$: \mbox{67.3 $\pm$ 5.4} & $8$: \mbox{\textbf{86.7 $\pm$ 3.9}} & $8$: \mbox{76.7 $\pm$ 4.8} \\
         & $4$: \mbox{70.0 $\pm$ 5.2} & $4$: \mbox{84.7 $\pm$ 4.2} & $4$: \mbox{\textbf{96.7 $\pm$ 2.0}} & $4$: \mbox{79.3 $\pm$ 4.5} \\
        \multirow{-2}{*}{\tokenizerid{qincobase!80}{C2-QINCO-S}} & $8$: \mbox{64.3 $\pm$ 5.4} & $8$: \mbox{78.7 $\pm$ 4.6} & $8$: \mbox{\textbf{96.7 $\pm$ 2.0}} & $8$: \mbox{78.0 $\pm$ 4.7} \\
        \rowcolor{black!4}  & $4$: \mbox{72.3 $\pm$ 5.1} & $4$: \mbox{74.0 $\pm$ 4.9} & $4$: \mbox{\textbf{93.3 $\pm$ 2.8}} & $4$: \mbox{80.0 $\pm$ 4.5} \\
        \rowcolor{black!4} \multirow{-2}{*}{\tokenizerid{qincobase!90}{C2-QINCO-L}} & $8$: \mbox{63.3 $\pm$ 5.5} & $8$: \mbox{73.0 $\pm$ 5.0} & $8$: \mbox{\textbf{91.0 $\pm$ 3.3}} & $8$: \mbox{74.0 $\pm$ 5.0} \\
        \bottomrule
    \end{tabularx}
\end{table}

%% file: tables/experiments/simulation_results/appendix_robomimic_multitask_best_square.tex
\begin{table}[H]
    \centering
    \caption{Multi-task RoboMimic ablation on Square for all 22 configurations, using the checkpoint with the best teacher-forced validation accuracy. The two rows per tokenizer report $H_{\mathrm{exec}}=4$ and $8$ success (\%) $\pm$ 95\% confidence-interval half-width. TP and TBR denote Token Perturbation and Token Balance Regularization; \textbf{bold} marks the best treatment per horizon.}
    \label{tab:robomimic-multitask-best-square}
    \tablebodyfont
    \setlength{\tabcolsep}{2.2pt}
    \renewcommand{\arraystretch}{0.96}
    \begin{tabularx}{\linewidth}{@{}>{\raggedright\arraybackslash}m{0.18\linewidth} *{4}{>{\centering\arraybackslash}X}@{}}
        \toprule
        Tokenizer & Reconstruction-only & \wotbr{} & \wotp{} & \methodname{} \\
        \midrule
         & $4$: \mbox{51.3 $\pm$ 5.7} & $4$: \mbox{\textbf{65.3 $\pm$ 5.4}} & $4$: \mbox{59.7 $\pm$ 5.6} & $4$: \mbox{64.0 $\pm$ 5.5} \\
        \multirow{-2}{*}{\tokenizerid{vqbase!60}{TF-VQ-S}} & $8$: \mbox{\textbf{64.0 $\pm$ 5.4}} & $8$: \mbox{58.0 $\pm$ 5.6} & $8$: \mbox{58.0 $\pm$ 5.5} & $8$: \mbox{63.3 $\pm$ 5.4} \\
        \rowcolor{black!4}  & $4$: \mbox{46.3 $\pm$ 5.7} & $4$: \mbox{55.7 $\pm$ 5.6} & $4$: \mbox{\textbf{61.0 $\pm$ 5.6}} & $4$: \mbox{43.3 $\pm$ 5.6} \\
        \rowcolor{black!4} \multirow{-2}{*}{\tokenizerid{vqbase!85}{TF-VQ-L}} & $8$: \mbox{42.7 $\pm$ 5.6} & $8$: \mbox{51.7 $\pm$ 5.7} & $8$: \mbox{\textbf{62.3 $\pm$ 5.5}} & $8$: \mbox{33.3 $\pm$ 5.3} \\
        \addlinespace[2pt]
         & $4$: \mbox{60.3 $\pm$ 5.5} & $4$: \mbox{63.3 $\pm$ 5.5} & $4$: \mbox{60.0 $\pm$ 5.6} & $4$: \mbox{\textbf{69.0 $\pm$ 5.3}} \\
        \multirow{-2}{*}{\tokenizerid{rvqbase!40}{TF-RVQ-S}} & $8$: \mbox{54.3 $\pm$ 5.7} & $8$: \mbox{68.0 $\pm$ 5.3} & $8$: \mbox{65.7 $\pm$ 5.4} & $8$: \mbox{\textbf{70.3 $\pm$ 5.1}} \\
        \rowcolor{black!4}  & $4$: \mbox{36.0 $\pm$ 5.4} & $4$: \mbox{44.0 $\pm$ 5.6} & $4$: \mbox{32.0 $\pm$ 5.3} & $4$: \mbox{\textbf{53.3 $\pm$ 5.7}} \\
        \rowcolor{black!4} \multirow{-2}{*}{\tokenizerid{rvqbase!50}{TF-RVQ-L}} & $8$: \mbox{26.0 $\pm$ 4.9} & $8$: \mbox{34.0 $\pm$ 5.4} & $8$: \mbox{23.0 $\pm$ 4.8} & $8$: \mbox{\textbf{43.3 $\pm$ 5.6}} \\
         & $4$: \mbox{61.7 $\pm$ 5.5} & $4$: \mbox{\textbf{67.0 $\pm$ 5.3}} & $4$: \mbox{61.0 $\pm$ 5.5} & $4$: \mbox{59.0 $\pm$ 5.6} \\
        \multirow{-2}{*}{\tokenizerid{rvqbase!60}{CT-RVQ-S}} & $8$: \mbox{55.3 $\pm$ 5.6} & $8$: \mbox{\textbf{65.0 $\pm$ 5.4}} & $8$: \mbox{63.7 $\pm$ 5.5} & $8$: \mbox{64.3 $\pm$ 5.5} \\
        \rowcolor{black!4}  & $4$: \mbox{36.0 $\pm$ 5.4} & $4$: \mbox{39.0 $\pm$ 5.5} & $4$: \mbox{37.0 $\pm$ 5.5} & $4$: \mbox{\textbf{40.7 $\pm$ 5.5}} \\
        \rowcolor{black!4} \multirow{-2}{*}{\tokenizerid{rvqbase!70}{CT-RVQ-L}} & $8$: \mbox{32.7 $\pm$ 5.3} & $8$: \mbox{33.7 $\pm$ 5.4} & $8$: \mbox{27.3 $\pm$ 5.0} & $8$: \mbox{\textbf{36.3 $\pm$ 5.4}} \\
         & $4$: \mbox{\textbf{64.0 $\pm$ 5.5}} & $4$: \mbox{57.3 $\pm$ 5.6} & $4$: \mbox{60.3 $\pm$ 5.6} & $4$: \mbox{57.7 $\pm$ 5.6} \\
        \multirow{-2}{*}{\tokenizerid{rvqbase!80}{C2-RVQ-S}} & $8$: \mbox{61.7 $\pm$ 5.5} & $8$: \mbox{60.7 $\pm$ 5.4} & $8$: \mbox{57.0 $\pm$ 5.5} & $8$: \mbox{\textbf{64.0 $\pm$ 5.5}} \\
        \rowcolor{black!4}  & $4$: \mbox{35.3 $\pm$ 5.4} & $4$: \mbox{42.0 $\pm$ 5.6} & $4$: \mbox{26.7 $\pm$ 5.0} & $4$: \mbox{\textbf{44.0 $\pm$ 5.6}} \\
        \rowcolor{black!4} \multirow{-2}{*}{\tokenizerid{rvqbase!90}{C2-RVQ-L}} & $8$: \mbox{32.3 $\pm$ 5.3} & $8$: \mbox{34.3 $\pm$ 5.4} & $8$: \mbox{21.0 $\pm$ 4.6} & $8$: \mbox{\textbf{35.7 $\pm$ 5.5}} \\
        \addlinespace[2pt]
         & $4$: \mbox{61.7 $\pm$ 5.5} & $4$: \mbox{68.3 $\pm$ 5.2} & $4$: \mbox{64.3 $\pm$ 5.5} & $4$: \mbox{\textbf{70.0 $\pm$ 5.2}} \\
        \multirow{-2}{*}{\tokenizerid{nrvqbase!40}{TF-NRVQ-S}} & $8$: \mbox{63.7 $\pm$ 5.5} & $8$: \mbox{65.3 $\pm$ 5.4} & $8$: \mbox{59.3 $\pm$ 5.6} & $8$: \mbox{\textbf{69.3 $\pm$ 5.2}} \\
        \rowcolor{black!4}  & $4$: \mbox{34.0 $\pm$ 5.3} & $4$: \mbox{31.0 $\pm$ 5.2} & $4$: \mbox{11.0 $\pm$ 3.6} & $4$: \mbox{\textbf{37.0 $\pm$ 5.5}} \\
        \rowcolor{black!4} \multirow{-2}{*}{\tokenizerid{nrvqbase!50}{TF-NRVQ-L}} & $8$: \mbox{21.0 $\pm$ 4.6} & $8$: \mbox{22.0 $\pm$ 4.7} & $8$: \mbox{9.3 $\pm$ 3.3} & $8$: \mbox{\textbf{31.0 $\pm$ 5.2}} \\
         & $4$: \mbox{0.0 $\pm$ 0.0} & $4$: \mbox{61.7 $\pm$ 5.6} & $4$: \mbox{56.3 $\pm$ 5.7} & $4$: \mbox{\textbf{64.0 $\pm$ 5.5}} \\
        \multirow{-2}{*}{\tokenizerid{nrvqbase!60}{CT-NRVQ-S}} & $8$: \mbox{0.0 $\pm$ 0.0} & $8$: \mbox{67.0 $\pm$ 5.3} & $8$: \mbox{55.0 $\pm$ 5.6} & $8$: \mbox{\textbf{70.7 $\pm$ 5.1}} \\
        \rowcolor{black!4}  & $4$: \mbox{0.0 $\pm$ 0.0} & $4$: \mbox{\textbf{41.0 $\pm$ 5.5}} & $4$: \mbox{26.7 $\pm$ 5.0} & $4$: \mbox{32.7 $\pm$ 5.3} \\
        \rowcolor{black!4} \multirow{-2}{*}{\tokenizerid{nrvqbase!70}{CT-NRVQ-L}} & $8$: \mbox{0.0 $\pm$ 0.0} & $8$: \mbox{\textbf{29.0 $\pm$ 5.2}} & $8$: \mbox{20.0 $\pm$ 4.5} & $8$: \mbox{24.0 $\pm$ 4.8} \\
         & $4$: \mbox{60.7 $\pm$ 5.5} & $4$: \mbox{62.0 $\pm$ 5.5} & $4$: \mbox{\textbf{64.3 $\pm$ 5.5}} & $4$: \mbox{61.0 $\pm$ 5.6} \\
        \multirow{-2}{*}{\tokenizerid{nrvqbase!80}{C2-NRVQ-S}} & $8$: \mbox{61.0 $\pm$ 5.5} & $8$: \mbox{57.3 $\pm$ 5.6} & $8$: \mbox{64.3 $\pm$ 5.4} & $8$: \mbox{\textbf{67.7 $\pm$ 5.3}} \\
        \rowcolor{black!4}  & $4$: \mbox{28.7 $\pm$ 5.1} & $4$: \mbox{\textbf{35.0 $\pm$ 5.4}} & $4$: \mbox{25.7 $\pm$ 4.9} & $4$: \mbox{\textbf{35.0 $\pm$ 5.4}} \\
        \rowcolor{black!4} \multirow{-2}{*}{\tokenizerid{nrvqbase!90}{C2-NRVQ-L}} & $8$: \mbox{21.3 $\pm$ 4.6} & $8$: \mbox{\textbf{28.0 $\pm$ 5.1}} & $8$: \mbox{13.0 $\pm$ 3.8} & $8$: \mbox{24.7 $\pm$ 4.9} \\
        \addlinespace[2pt]
         & $4$: \mbox{49.7 $\pm$ 5.6} & $4$: \mbox{53.7 $\pm$ 5.6} & $4$: \mbox{\textbf{61.7 $\pm$ 5.5}} & $4$: \mbox{60.0 $\pm$ 5.5} \\
        \multirow{-2}{*}{\tokenizerid{fsqbase!60}{TF-FSQ-S}} & $8$: \mbox{49.7 $\pm$ 5.6} & $8$: \mbox{49.0 $\pm$ 5.6} & $8$: \mbox{\textbf{58.0 $\pm$ 5.6}} & $8$: \mbox{57.0 $\pm$ 5.6} \\
        \rowcolor{black!4}  & $4$: \mbox{53.3 $\pm$ 5.7} & $4$: \mbox{48.3 $\pm$ 5.6} & $4$: \mbox{56.7 $\pm$ 5.6} & $4$: \mbox{\textbf{59.3 $\pm$ 5.6}} \\
        \rowcolor{black!4} \multirow{-2}{*}{\tokenizerid{fsqbase!85}{TF-FSQ-L}} & $8$: \mbox{50.0 $\pm$ 5.7} & $8$: \mbox{46.0 $\pm$ 5.7} & $8$: \mbox{57.3 $\pm$ 5.5} & $8$: \mbox{\textbf{63.7 $\pm$ 5.4}} \\
        \addlinespace[2pt]
         & $4$: \mbox{27.7 $\pm$ 5.0} & $4$: \mbox{40.0 $\pm$ 5.5} & $4$: \mbox{28.3 $\pm$ 5.1} & $4$: \mbox{\textbf{54.0 $\pm$ 5.6}} \\
        \multirow{-2}{*}{\tokenizerid{qincobase!40}{TF-QINCO-S}} & $8$: \mbox{12.7 $\pm$ 3.8} & $8$: \mbox{32.3 $\pm$ 5.3} & $8$: \mbox{20.0 $\pm$ 4.5} & $8$: \mbox{\textbf{51.0 $\pm$ 5.7}} \\
        \rowcolor{black!4}  & $4$: \mbox{19.0 $\pm$ 4.4} & $4$: \mbox{29.7 $\pm$ 5.2} & $4$: \mbox{15.0 $\pm$ 4.0} & $4$: \mbox{\textbf{53.0 $\pm$ 5.6}} \\
        \rowcolor{black!4} \multirow{-2}{*}{\tokenizerid{qincobase!50}{TF-QINCO-L}} & $8$: \mbox{8.7 $\pm$ 3.1} & $8$: \mbox{21.7 $\pm$ 4.6} & $8$: \mbox{5.7 $\pm$ 2.6} & $8$: \mbox{\textbf{48.7 $\pm$ 5.6}} \\
         & $4$: \mbox{34.0 $\pm$ 5.5} & $4$: \mbox{\textbf{46.7 $\pm$ 5.6}} & $4$: \mbox{30.0 $\pm$ 5.2} & $4$: \mbox{43.0 $\pm$ 5.7} \\
        \multirow{-2}{*}{\tokenizerid{qincobase!60}{CT-QINCO-S}} & $8$: \mbox{18.7 $\pm$ 4.4} & $8$: \mbox{\textbf{44.3 $\pm$ 5.7}} & $8$: \mbox{17.3 $\pm$ 4.3} & $8$: \mbox{36.7 $\pm$ 5.5} \\
        \rowcolor{black!4}  & $4$: \mbox{23.7 $\pm$ 4.8} & $4$: \mbox{31.0 $\pm$ 5.2} & $4$: \mbox{22.3 $\pm$ 4.7} & $4$: \mbox{\textbf{48.3 $\pm$ 5.7}} \\
        \rowcolor{black!4} \multirow{-2}{*}{\tokenizerid{qincobase!70}{CT-QINCO-L}} & $8$: \mbox{17.3 $\pm$ 4.3} & $8$: \mbox{21.7 $\pm$ 4.6} & $8$: \mbox{12.7 $\pm$ 3.8} & $8$: \mbox{\textbf{41.0 $\pm$ 5.7}} \\
         & $4$: \mbox{21.3 $\pm$ 4.7} & $4$: \mbox{\textbf{52.0 $\pm$ 5.6}} & $4$: \mbox{42.7 $\pm$ 5.6} & $4$: \mbox{50.3 $\pm$ 5.6} \\
        \multirow{-2}{*}{\tokenizerid{qincobase!80}{C2-QINCO-S}} & $8$: \mbox{11.3 $\pm$ 3.6} & $8$: \mbox{34.7 $\pm$ 5.4} & $8$: \mbox{30.0 $\pm$ 5.1} & $8$: \mbox{\textbf{48.0 $\pm$ 5.7}} \\
        \rowcolor{black!4}  & $4$: \mbox{17.7 $\pm$ 4.3} & $4$: \mbox{\textbf{37.7 $\pm$ 5.5}} & $4$: \mbox{26.3 $\pm$ 5.0} & $4$: \mbox{29.0 $\pm$ 5.1} \\
        \rowcolor{black!4} \multirow{-2}{*}{\tokenizerid{qincobase!90}{C2-QINCO-L}} & $8$: \mbox{9.7 $\pm$ 3.3} & $8$: \mbox{\textbf{32.0 $\pm$ 5.2}} & $8$: \mbox{15.0 $\pm$ 4.0} & $8$: \mbox{21.3 $\pm$ 4.7} \\
        \bottomrule
    \end{tabularx}
\end{table}

%% file: figures/experiments/prediction_error/appendix_validation_loss_by_family.tex
\begin{tikzpicture}
\begin{axis}[
  width=4.05cm,height=3.1cm,scale only axis,
  xmin=-4.0,xmax=1.0,xtick={-4,-3,-2,-1,0,1},
  ymin=0.45,ymax=5.55,ytick={1,2,3,4,5},
  yticklabels={\tokenizerid{qincobase}{QinCo},\tokenizerid{fsqbase}{FSQ},\tokenizerid{nrvqbase}{NRVQ},\tokenizerid{rvqbase}{RVQ},\tokenizerid{vqbase}{VQ}},
  scaled ticks=false,
  tick label style={font=\fontsize{7}{8}\selectfont,/pgf/number format/fixed},
  yticklabel style={anchor=east,inner sep=1pt,font=\fontsize{7}{8}\selectfont},
  label style={font=\fontsize{7}{8}\selectfont},
  xlabel={$\Delta\mathcal L_{\mathrm{val}}^{\mathrm{TBR}}$},
  tick align=outside,tick pos=left,
  axis line style={black!55,line width=0.35pt},
  xmajorgrids,grid style={black!10,line width=0.3pt},
]
\addplot[black!65,line width=0.45pt,forget plot] coordinates {(0,0.45) (0,5.55)};
\addplot+[xbar,bar width=5pt,mark=none,draw=vqbase,fill=vqbase,line width=0.5pt,forget plot,error bars/.cd,x dir=both,x explicit,error bar style={draw=vqbase!82,line width=0.65pt},error mark=none] coordinates {(-0.992271074,5.00) += (0.377454806,0) -= (0.377454806,0)};
\addplot+[xbar,bar width=5pt,mark=none,draw=rvqbase,fill=rvqbase,line width=0.5pt,forget plot,error bars/.cd,x dir=both,x explicit,error bar style={draw=rvqbase!82,line width=0.65pt},error mark=none] coordinates {(0.152342809,4.00) += (0.603361023,0) -= (1.308949377,0)};
\addplot+[xbar,bar width=5pt,mark=none,draw=nrvqbase,fill=nrvqbase,line width=0.5pt,forget plot,error bars/.cd,x dir=both,x explicit,error bar style={draw=nrvqbase!82,line width=0.65pt},error mark=none] coordinates {(-0.047527516,3.00) += (0.723400494,0) -= (0.583526500,0)};
\addplot+[xbar,bar width=5pt,mark=none,draw=fsqbase,fill=fsqbase,line width=0.5pt,forget plot,error bars/.cd,x dir=both,x explicit,error bar style={draw=fsqbase!82,line width=0.65pt},error mark=none] coordinates {(-0.834347878,2.00) += (0.128797836,0) -= (0.128797836,0)};
\addplot+[xbar,bar width=5pt,mark=none,draw=qincobase,fill=qincobase,line width=0.5pt,forget plot,error bars/.cd,x dir=both,x explicit,error bar style={draw=qincobase!82,line width=0.65pt},error mark=none] coordinates {(-1.896879653,1.00) += (0.917410104,0) -= (1.719193795,0)};
\end{axis}
\end{tikzpicture}

%% file: tables/experiments/prediction_error/appendix_validation_metrics.tex
\begin{table}[H]
    \centering
    \caption{Policy validation metrics for all 22 multi-task RoboMimic tokenizers at the checkpoint with the best teacher-forced validation accuracy. Lower loss and higher accuracy are better; \textbf{bold} marks the better value in each pair. A green down-arrow marks \methodname{} values whose tokenizer also has lower validation reconstruction error.}
    \label{tab:policy-validation-metrics}
    \begingroup
    \fontsize{8}{9}\selectfont
    \setlength{\tabcolsep}{4pt}
    \renewcommand{\arraystretch}{1.05}
    \begin{tabularx}{\linewidth}{@{}l *{4}{>{\centering\arraybackslash}X}@{}}
        \toprule
        \multirow{2}{*}{Tokenizer} & \multicolumn{2}{c}{Validation loss} & \multicolumn{2}{c}{Validation accuracy (\%)} \\
        \cmidrule(lr){2-3}\cmidrule(l){4-5}
        & \wotbr{} & \methodname{} & \wotbr{} & \methodname{} \\
        \midrule
        \tokenizerid{vqbase!60}{TF-VQ-S} & 4.646 & \textbf{3.276} & 37.7 & \textbf{44.3} \\
        \rowcolor{black!4} \tokenizerid{vqbase!85}{TF-VQ-L} & 2.064 & \textbf{1.449} & 68.3 & \textbf{74.7} \\
        \addlinespace[2pt]
        \tokenizerid{rvqbase!40}{TF-RVQ-S} & \textbf{3.773} & 4.529 & \textbf{44.7} & 35.8 \\
        \rowcolor{black!4} \tokenizerid{rvqbase!50}{TF-RVQ-L} & \textbf{6.531} & \reconbetter{6.793} & \textbf{40.9} & \reconbetter{33.7} \\
        \tokenizerid{rvqbase!60}{CT-RVQ-S} & \textbf{4.443} & \reconbetter{4.736} & \textbf{43.9} & \reconbetter{42.8} \\
        \rowcolor{black!4} \tokenizerid{rvqbase!70}{CT-RVQ-L} & \textbf{6.630} & \reconbetter{7.336} & \textbf{40.8} & \reconbetter{35.2} \\
        \tokenizerid{rvqbase!80}{C2-RVQ-S} & 4.704 & \textbf{3.547} & 43.9 & \textbf{45.4} \\
        \rowcolor{black!4} \tokenizerid{rvqbase!90}{C2-RVQ-L} & \textbf{6.351} & \reconbetter{6.405} & \textbf{41.8} & \reconbetter{41.5} \\
        \addlinespace[2pt]
        \tokenizerid{nrvqbase!40}{TF-NRVQ-S} & 5.343 & \reconbetter{\textbf{5.191}} & \textbf{39.3} & \reconbetter{36.6} \\
        \rowcolor{black!4} \tokenizerid{nrvqbase!50}{TF-NRVQ-L} & 9.577 & \reconbetter{\textbf{9.209}} & 20.2 & \reconbetter{\textbf{23.3}} \\
        \tokenizerid{nrvqbase!60}{CT-NRVQ-S} & 4.675 & \textbf{4.044} & \textbf{42.5} & 42.1 \\
        \rowcolor{black!4} \tokenizerid{nrvqbase!70}{CT-NRVQ-L} & \textbf{9.514} & 9.605 & 19.3 & \textbf{20.9} \\
        \tokenizerid{nrvqbase!80}{C2-NRVQ-S} & \textbf{4.000} & \reconbetter{4.676} & \textbf{44.5} & \reconbetter{43.1} \\
        \rowcolor{black!4} \tokenizerid{nrvqbase!90}{C2-NRVQ-L} & \textbf{9.196} & 9.296 & 20.2 & \textbf{23.3} \\
        \addlinespace[2pt]
        \tokenizerid{fsqbase!60}{TF-FSQ-S} & 3.481 & \textbf{2.776} & 41.9 & \textbf{54.7} \\
        \rowcolor{black!4} \tokenizerid{fsqbase!85}{TF-FSQ-L} & 4.013 & \textbf{3.050} & 43.7 & \textbf{48.5} \\
        \addlinespace[2pt]
        \tokenizerid{qincobase!40}{TF-QINCO-S} & 5.575 & \textbf{4.311} & 40.1 & \textbf{44.7} \\
        \rowcolor{black!4} \tokenizerid{qincobase!50}{TF-QINCO-L} & 7.770 & \textbf{4.154} & 35.1 & \textbf{45.8} \\
        \tokenizerid{qincobase!60}{CT-QINCO-S} & 5.899 & \textbf{4.347} & 40.6 & \textbf{43.9} \\
        \rowcolor{black!4} \tokenizerid{qincobase!70}{CT-QINCO-L} & 6.812 & \textbf{5.022} & 40.8 & \textbf{42.0} \\
        \tokenizerid{qincobase!80}{C2-QINCO-S} & 5.438 & \textbf{4.458} & 42.0 & \textbf{44.6} \\
        \rowcolor{black!4} \tokenizerid{qincobase!90}{C2-QINCO-L} & 6.957 & \textbf{4.778} & 37.9 & \textbf{44.6} \\
        \bottomrule
    \end{tabularx}
    \endgroup
\end{table}

%% file: figures/experiments/reconstruction_error/appendix_decoded_robustness_pooled.tex
\begin{tikzpicture}
\definecolor{robustshared}{RGB}{184,121,99}
\definecolor{robustlevel}{RGB}{104,151,187}
\begin{groupplot}[group style={group size=3 by 1,horizontal sep=0.55cm},
  width=3.55cm,height=2.65cm,scale only axis,
  xmode=log,log basis x=2,xmin=1,xmax=2048,xtick={1,8,64,512,2048},xticklabels={1,8,64,512,2048},
  ymin=-0.0400,ymax=0.0200,ytick={-0.04,-0.02,0.00,0.02},scaled ticks=false,
  tick label style={font=\fontsize{8}{9}\selectfont,/pgf/number format/fixed,/pgf/number format/precision=2},
  label style={font=\fontsize{8}{9}\selectfont},
  tick align=outside,tick pos=left,
  axis line style={black!55,line width=0.35pt},xmajorgrids,ymajorgrids,grid style={black!10,line width=0.2pt}]
\nextgroupplot[extra description/.code={
  \node[anchor=north east,fill=none,draw=none,text opacity=1,inner sep=0.6pt,font=\fontsize{8}{9}\selectfont]
    at (axis description cs:0.988,0.965) {Can};
  \node[anchor=south west,fill=none,draw=none,text opacity=1,inner sep=0pt,align=left,font=\fontsize{8}{9}\selectfont]
    at (axis description cs:0.04,0.04) {\textcolor{robustshared}{$\bullet$} All tokenizers};
},ylabel={Decoded-error change (MAE)}]
\addplot[draw=none,fill=robustshared,fill opacity=0.08,forget plot] coordinates {(1,-0.0025412793) (2048,-0.0025412793) (2048,0.0161207386) (1,0.0161207386)} \closedcycle;
\addplot[robustshared,dashed,line width=0.65pt,forget plot] coordinates {(1,0.0042034615) (2048,0.0042034615)};
\addplot[name path=robust00Low,draw=none,forget plot] coordinates {(1,-0.0059852198) (2,-0.0075604722) (4,-0.0090269885) (8,-0.0108819870) (16,-0.0121408905) (32,-0.0130168487) (64,-0.0141948263) (128,-0.0156738105) (256,-0.0168108643) (512,-0.0177406107) (1024,-0.0185368231) (2048,-0.0192750480)};
\addplot[name path=robust00High,draw=none,forget plot] coordinates {(1,-0.0045185508) (2,-0.0058631173) (4,-0.0069038768) (8,-0.0085213299) (16,-0.0095362696) (32,-0.0105317509) (64,-0.0112102332) (128,-0.0120583074) (256,-0.0128026400) (512,-0.0131700971) (1024,-0.0134405718) (2048,-0.0136488859)};
\addplot[fill=robustshared,fill opacity=0.25,draw=none,forget plot] fill between[of=robust00Low and robust00High];
\addplot[robustshared,solid,line width=0.85pt,mark=*,mark size=1.25pt,mark options={solid,fill=robustshared},forget plot] coordinates {(1,-0.0052394504) (2,-0.0066982579) (4,-0.0079348945) (8,-0.0096738840) (16,-0.0108654254) (32,-0.0117869142) (64,-0.0127190204) (128,-0.0137633769) (256,-0.0147140540) (512,-0.0153895683) (1024,-0.0159207129) (2048,-0.0164502796)};
\addplot[black!55,line width=0.45pt,forget plot] coordinates {(1,0) (2048,0)};
\nextgroupplot[extra description/.code={
  \node[anchor=north east,fill=none,draw=none,text opacity=1,inner sep=0.6pt,font=\fontsize{8}{9}\selectfont]
    at (axis description cs:0.988,0.965) {Lift};
},yticklabels={}]
\addplot[draw=none,fill=robustshared,fill opacity=0.08,forget plot] coordinates {(1,-0.0025412793) (2048,-0.0025412793) (2048,0.0161207386) (1,0.0161207386)} \closedcycle;
\addplot[robustshared,dashed,line width=0.65pt,forget plot] coordinates {(1,0.0042034615) (2048,0.0042034615)};
\addplot[name path=robust10Low,draw=none,forget plot] coordinates {(1,-0.0045514997) (2,-0.0054908928) (4,-0.0068583128) (8,-0.0077968586) (16,-0.0086579525) (32,-0.0099109259) (64,-0.0112562389) (128,-0.0120976231) (256,-0.0121849221) (512,-0.0122852766) (1024,-0.0121909548) (2048,-0.0127209179)};
\addplot[name path=robust10High,draw=none,forget plot] coordinates {(1,-0.0031040325) (2,-0.0037883906) (4,-0.0047787694) (8,-0.0058856602) (16,-0.0066559031) (32,-0.0075433717) (64,-0.0086583180) (128,-0.0093339743) (256,-0.0090447670) (512,-0.0091427606) (1024,-0.0084005361) (2048,-0.0083972187)};
\addplot[fill=robustshared,fill opacity=0.25,draw=none,forget plot] fill between[of=robust10Low and robust10High];
\addplot[robustshared,solid,line width=0.85pt,mark=*,mark size=1.25pt,mark options={solid,fill=robustshared},forget plot] coordinates {(1,-0.0037934896) (2,-0.0046090236) (4,-0.0057471596) (8,-0.0068157769) (16,-0.0076508844) (32,-0.0087183120) (64,-0.0099354048) (128,-0.0107450719) (256,-0.0106786793) (512,-0.0107989720) (1024,-0.0104080625) (2048,-0.0106950959)};
\addplot[black!55,line width=0.45pt,forget plot] coordinates {(1,0) (2048,0)};
\nextgroupplot[extra description/.code={
  \node[anchor=north east,fill=none,draw=none,text opacity=1,inner sep=0.6pt,font=\fontsize{8}{9}\selectfont]
    at (axis description cs:0.988,0.965) {Square};
},yticklabels={}]
\addplot[draw=none,fill=robustshared,fill opacity=0.08,forget plot] coordinates {(1,-0.0025412793) (2048,-0.0025412793) (2048,0.0161207386) (1,0.0161207386)} \closedcycle;
\addplot[robustshared,dashed,line width=0.65pt,forget plot] coordinates {(1,0.0042034615) (2048,0.0042034615)};
\addplot[name path=robust20Low,draw=none,forget plot] coordinates {(1,-0.0056198479) (2,-0.0071283631) (4,-0.0084844426) (8,-0.0102583287) (16,-0.0120410221) (32,-0.0139788514) (64,-0.0161926138) (128,-0.0187644727) (256,-0.0208762297) (512,-0.0225143372) (1024,-0.0240293874) (2048,-0.0260212929)};
\addplot[name path=robust20High,draw=none,forget plot] coordinates {(1,-0.0045691099) (2,-0.0060326611) (4,-0.0072508523) (8,-0.0088010135) (16,-0.0101387598) (32,-0.0116670566) (64,-0.0134550599) (128,-0.0153586615) (256,-0.0169778856) (512,-0.0183951721) (1024,-0.0196153144) (2048,-0.0209458412)};
\addplot[fill=robustshared,fill opacity=0.25,draw=none,forget plot] fill between[of=robust20Low and robust20High];
\addplot[robustshared,solid,line width=0.85pt,mark=*,mark size=1.25pt,mark options={solid,fill=robustshared},forget plot] coordinates {(1,-0.0050945221) (2,-0.0065957879) (4,-0.0078989850) (8,-0.0095432502) (16,-0.0111194181) (32,-0.0128340894) (64,-0.0148294783) (128,-0.0170786429) (256,-0.0189763461) (512,-0.0205266791) (1024,-0.0218997131) (2048,-0.0236659776)};
\addplot[black!55,line width=0.45pt,forget plot] coordinates {(1,0) (2048,0)};
\end{groupplot}
\node[anchor=north,font=\fontsize{8}{9}\selectfont,inner sep=0pt] at ([yshift=-0.65cm]group c2r1.south) {$M$ policy samples};
\end{tikzpicture}